\documentclass[sigconf,pbalance,screen]{acmart}
\AtBeginDocument{%
  }

\copyrightyear{2025}
\setcopyright{cc}
\setcctype{by}
\acmConference[SIGMOD '26]{2026 International
Conference on Management of Data}{May 31--June 5, 2026}{Bengaluru, India}
\acmBooktitle{Proceedings of the 2026 International Conference on Management
of Data (SIGMOD '26), May 31--June 5, 2026, Bengaluru, India}
\acmYear{2025} \acmVolume{3} \acmNumber{4 (SIGMOD)} \acmArticle{238} \acmMonth{9} \acmPrice{}\acmDOI{10.1145/3749156}
\usepackage[utf8]{inputenc}
\usepackage{xspace}
\usepackage{xpatch}
\usepackage{xstring}
\usepackage{nicefrac}
\usepackage{soul}
\usepackage{setspace}
\usepackage{pgfmath}
\usepackage[capitalise]{cleveref}
\usepackage{amsmath}
\usepackage{amsfonts}
\usepackage{pifont}
\usepackage{bm}
\usepackage{siunitx}
\usepackage{stmaryrd}
\usepackage{titlesec}
\usepackage[inline]{enumitem}
\usepackage[labelformat=simple]{subcaption}
\usepackage{stfloats}
\usepackage[most]{tcolorbox}
\usepackage{mdframed}
\usepackage{array}
\usepackage{makecell}
\usepackage{tabu}
\usepackage{tabulary}
\usepackage{multirow}
\usepackage{multicol}
\usepackage{booktabs}
\usepackage{adjustbox}
\usepackage[flushleft,para]{threeparttable}
\usepackage{rotating}
\usepackage{collcell}
\usepackage{tikz}
\usepackage{algorithm}
\usepackage{algpseudocodex}

\newcounter{rqcnt}

\makeatletter
\newcommand*{\tran}{^{\mathpalette\@transpose{}}}
\newcommand*{\@transpose}[2]{\raisebox{\depth}{$\m@th#1\intercal$}}
  {}%
  {}%
\newenvironment{rquestionul}
  {\noindent\refstepcounter{rqcnt}\ul{\textbf{RQ\therqcnt:}}\itshape}%
  {}%
\newcommand{\crefnames}[3]{%
   \@for\next:=#1\do{%
     \expandafter\crefformat\expandafter{\next}{#2##2##1##3}
     \expandafter\crefmultiformat\expandafter{\next}
     {#3##2##1##3}{\&##2##1##3}{, ##2##1##3}{, \&##2##1##3}
    }%
 }
\makeatother
\graphicspath{{figs/}}
\DeclareGraphicsExtensions{.pdf,.eps,.png,.jpg,.jpeg}

\crefname{figure}{Figure}{Figures}

\crefnames{rqcnt}{RQ}{RQ}
\newlist{wenumerate}{enumerate}{3}
\setlist[wenumerate,1]{wide,labelwidth=!,labelindent=0pt,nosep,topsep=2pt,label=\arabic*.}
\newlist{witemize}{itemize}{3}
\setlist[witemize,1]{wide,labelwidth=!,labelindent=0pt,nosep,topsep=2pt,label=$\bullet$}
\setlist[itemize,2]{label=$\circ$}
\mdfsetup{%
innerleftmargin=0pt,
innerrightmargin=0pt,
innertopmargin=0pt,
innerbottommargin=0pt,}
\theoremstyle{plain}

\DeclareMathOperator{\diag}{diag}

\newcommand{\RR}[1]{\mathbb{R}^{#1}}

\newcommand{\binos}[2]{\bigl( \begin{smallmatrix}{#1}\\{#2}\end{smallmatrix}\bigr)}
\newcommand{\bmu}[1]{\boldsymbol{\mathbf{#1}}}
\newcommand{\ccmark}[1]{\textcolor{ForestGreen}{\ding{51}}}
\newcommand{\xxmark}[1]{\textcolor{red}{\ding{55}}}
\newcommand{\ft}[1]{\textls[-60]{\textsf{#1}}}
\newcommand{\ds}[1]{\textls[-60]{\textsc{\MakeLowercase{#1}}}}
\newcommand{\tpm}[1]{{\scriptsize{\textpm#1}}}
\newcommand{\dshh}[1]{\fontsize{8pt}{6pt}\selectfont{#1}}
\newcommand{\rkoo}[1]{\colorbox[HTML]{cce5d7}{#1}}
\newcommand{\rko}[1]{#1}
\newcommand{\rka}[1]{#1}
\newcommand{\rkat}[1]{#1}
\newcommand{\rkb}[1]{#1}
\newcommand{\rkbt}[1]{#1}
\newcommand{\rkc}[1]{#1}
\newcommand{\rkct}[1]{#1}
\newcommand{\tz}[2]{\cellcolor[HTML]{#1}{#2}}

\newcommand{\rra}[1]{#1}
\newcommand{\rrb}[1]{#1}
\newcommand{\rrc}[1]{#1}
\newcommand{\blue}[1]{\textcolor{black}{{#1}}}

\begin{document}

\begin{abstract}
  With recent advancements in graph neural networks (GNNs), spectral GNNs have received increasing popularity by virtue of their ability to retrieve graph signals in the spectral domain. These models feature uniqueness in efficient computation as well as rich expressiveness, which stems from advanced management and profound understanding of graph data. However, few systematic studies have been conducted to assess spectral GNNs, particularly in benchmarking their efficiency, memory consumption, and effectiveness in a unified and fair manner. There is also a pressing need to select spectral models suitable for learning specific graph data and deploying them to massive web-scale graphs, which is currently constrained by the varied model designs and training settings.

  In this work, we extensively benchmark spectral GNNs with a focus on the spectral perspective, demystifying them as spectral graph filters. We analyze and categorize 35 GNNs with 27 corresponding filters, spanning diverse formulations and utilizations of the graph data. Then, we implement the filters within a unified spectral-oriented framework with dedicated graph computations and efficient training schemes. In particular, our implementation enables the deployment of spectral GNNs over million-scale graphs and various tasks with comparable performance and less overhead. Thorough experiments are conducted on the graph filters with comprehensive metrics on effectiveness and efficiency, offering novel observations and practical guidelines that are only available from our evaluations across graph scales. Different from the prevailing belief, our benchmark reveals an intricate landscape regarding the effectiveness and efficiency of spectral graph filters, demonstrating the potential to achieve desirable performance through tailored spectral manipulation of graph data.
  Our code and evaluation is available at: \url{https://github.com/gdmnl/Spectral-GNN-Benchmark}.
\end{abstract}

\setlist[enumerate,1]{wide,labelwidth=!,labelindent=0pt,nosep,topsep=2pt,label=\arabic*.}
\setlist[itemize,1]{wide,labelwidth=!,labelindent=0pt,nosep,topsep=2pt}

\titlespacing*{\section}{0pt}{.45\baselineskip plus 2pt minus .2pt}{.25\baselineskip}
\titlespacing*{\subsection}{0pt}{.4\baselineskip plus .6pt minus .2pt}{.25\baselineskip}
\titlespacing*{\subsubsection}{0pt}{.15\baselineskip plus .2pt minus .2pt}{.15\baselineskip}
\titlespacing*{\noindentparagraph}{0pt}{.15\baselineskip minus .2pt}{.125\baselineskip}

\title[A Comprehensive Benchmark on Spectral GNNs: The Impact on Efficiency, Memory, and Effectiveness]
{A Comprehensive Benchmark on Spectral GNNs:\\
The Impact on Efficiency, Memory, and Effectiveness}
\subtitle{[Technical Report]}

\author{Ningyi Liao}
\orcid{0000-0003-3176-4401}
\affiliation{%
  \department{College of Computing and Data Science}
  \institution{Nanyang Technological University}
  \country{Singapore}
}
\email{liao0090@e.ntu.edu.sg}

\author{Haoyu Liu}
\orcid{0000-0002-0839-5460}
\affiliation{%
  \department{College of Computing and Data Science}
  \institution{Nanyang Technological University}
  \country{Singapore}
}
\email{haoyu.liu@ntu.edu.sg}

\author{Zulun Zhu}
\orcid{0000-0002-5176-6378}
\affiliation{%
  \department{College of Computing and Data Science}
  \institution{Nanyang Technological University}
  \country{Singapore}
}
\email{zulun001@ntu.edu.sg}

\author{Siqiang Luo}
\orcid{0000-0001-8197-0903}
\affiliation{%
  \department{College of Computing and Data Science}
  \institution{Nanyang Technological University}
  \country{Singapore}
}
\email{siqiang.luo@ntu.edu.sg}

\author{Laks V.S. Lakshmanan}
\orcid{0000-0002-9775-4241}
\affiliation{%
  \department{Department of Computer Science}
  \institution{The University of British Columbia}
  \country{Canada}
}
\email{laks@cs.ubc.ca}

\renewcommand{\shortauthors}{N. Liao et al.}



\received{January 2025}
\received[revised]{April 2025}
\received[accepted]{May 2025}

\maketitle

\setlength{\textfloatsep}{4pt plus 0.6pt minus 4.0pt}
\setlength{\dbltextfloatsep}{4pt plus 0.6pt minus 4.0pt}
\setlength{\floatsep}{4pt plus 0.8pt minus 2.2pt}
\setlength{\intextsep}{4pt plus 0.8pt minus 3.2pt}
\setlength{\abovedisplayskip}{1.5pt plus 0.2pt minus 2.0pt}
\setlength{\abovedisplayshortskip}{1.0pt plus 0.2pt minus 2.0pt}
\setlength{\belowdisplayskip}{1.5pt plus 0.2pt minus 2.0pt}
\setlength{\belowdisplayshortskip}{1.0pt plus 0.2pt minus 2.0pt}
\captionsetup{font={stretch=0.96}}

\vspace{-1ex}%
\section{Introduction}
\label{sec:introduction}
Graph neural networks (GNNs) have emerged by integrating neural network architectures with graph data processing techniques, serving as powerful tools for a wide range of graph understanding tasks \cite{kipf2016semi,velivckovic2017graph,zhang2018link,xu2019}. Among them, a notable branch is spectral GNNs, which are characterized by understanding the graph data through leveraging the \blue{graph spectrum, i.e., eigenvalues} \cite{bruna2014spectral,kipf2016semi,atwood2016,defferrard2016convolutional}. In the graph, such spectral information is powerful for characterizing various patterns, from local edge connections to global clustering structures \cite{wang2022b,gong2024}. Hence, spectral GNNs have garnered exceptional utility for real-world applications encompassing specialized graph topologies, such as multivariate time-series forecasting~\cite{cao2020spectral,liu2022multivariate} and point cloud analysis~\cite{li2021towards,hu2021graph}.

\begin{figure}[!t]
\centering
    \includegraphics[width=0.9\linewidth]{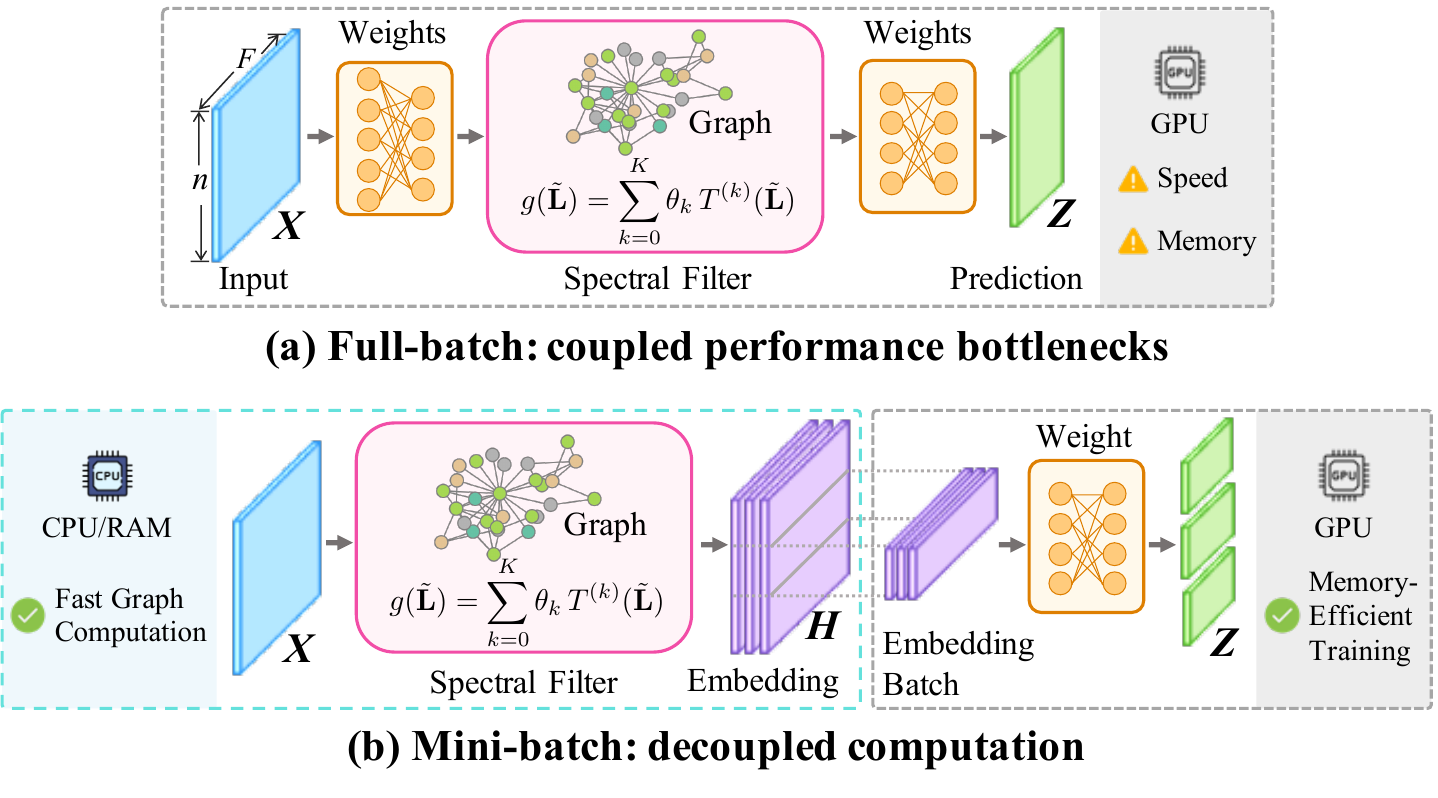}
    \caption{Model composition and training pipeline of \textbf{(a)}~full-batch scheme for common GNN training and inference, where model weights and graph processing are integrated; and \textbf{(b)}~mini-batch learning, which is specialized for spectral GNNs with separated graph computations. Among these two settings, the graph operations remain consistent and are extracted as spectral filters.}
    \label{plot:schemes}
\vspace{-.8em}
\end{figure}

Particularly, spectral models are superior in addressing the notorious scalability issue of common GNNs \cite{Ying2018graph,sahu2018,shen2024}. This merit stems from the unique graph data management as illustrated in \cref{plot:schemes}. Conventionally, learning GNNs is achieved through the full-batch scheme, where graph data and neural network weights are integrated, rendering iterative computation and high GPU overhead. In comparison, the mini-batch setup is uniquely available for spectral GNNs by dividing learning data into small batches, thereby alleviating both efficiency bottlenecks and memory footprints \cite{duan2022comprehensive,ward2022}. The graph data and weights can be processed separately with tailored computations, enabling spectral GNNs to excel in learning web-scale graphs with hundreds of millions of nodes \cite{frasca2020,maurya2022,liao2024scalable}.

\noindentparagraph{Challenges: Benchmarking Spectral GNNs at Scale.}
Previous surveys on spectral GNNs \cite{chen2023,bo2023b} primarily focus on deriving spectral models and evaluating their accuracy on small datasets. Although the spectral design has been shown to be promising in scaling up GNNs, there is a lack of comprehensive study, especially on their efficiency and scalability performance. We note that implementing and evaluating these models for large-scale graphs remains a demanding task considering the following major challenges:

\par\noindent$\bullet$\;\textbf{Under-explored practical performance on large-scale graphs.}
For the interest of fast and useful GNNs in a broader scope, it is crucial to understand the performance of advanced solutions, extensively considering their time efficiency, memory overhead, effectiveness, and the relationships among these aspects. However, it remains unknown of such systematic insights regarding the feasibility and performance characteristics of spectral GNNs at various data scales, rendering it difficult to decide the appropriate model.

\par\noindent$\bullet$\;\textbf{Limited availability of efficient and scalable computation techniques.}
Most existing scalable computations based on spectral GNNs, such as graph operation accelerations \cite{Chen2020a,wang2021b,feng2022a,liao2022} and mini-batch training techniques \cite{he2022a,feng2022,liao2023ld2} are exclusively available for only a few models. The majority of spectral GNNs have never been implemented or evaluated with these techniques, consequently constraining their practical application in large-scale scenarios.

\par\noindent$\bullet$\;\textbf{Varied implementations and inconsistent evaluations.}
Although a broad scope of models can be regarded as spectral GNNs, their computational designs vary greatly in terms of graph managements, network architectures, and training schemes. The distinct settings and implementations lead to varied time and memory overheads and complicate fair comparison among spectral models.

\noindentparagraph{Our Contributions.}
In this research, we present a comprehensive benchmark on spectral GNNs, especially targeting the above challenges. We extensively review existing GNN designs with spectral interpretations, ranging from traditional to cutting-edge studies, and reach a total of 35 models.
To facilitate fair and versatile evaluations, we encapsulate the \textit{filters}, i.e., the specialized graph management process, among these models. A novel taxonomy (\cref{tab:summary}) is proposed to identify and categorize these filters into three categories, each exhibiting distinct capabilities and performances.

We implement the spectral filters in a unified manner applicable to a variety of settings and tasks. Our framework embraces scalable graph processing techniques, especially the dedicated mini-batch training scheme, and allows for \textbf{widely applicable spectral operations on million-scale graphs}, which have not been achieved before.
On top of our taxonomy and implementation, we perform evaluation on efficiency, memory consumption, and effectiveness of spectral filters. Our findings aim to provide new assessment criteria, insightful observations, and useful guidelines, which are of pragmatic interest for developing and deploying spectral GNNs.

\noindentparagraph{Our Findings.}
Our benchmark evaluations cover an inclusive range of spectral GNN components, graph filter designs, and training schemes. The observations feature diverse graph properties and impact factors on the performance of spectral models, and reveal the intricate relationship between efficiency and effectiveness across different scales of graphs, which is \textbf{only available by our unified framework highlighting scalability}. In specific, we outline the novel findings from benchmarking GNNs at scale as follows:
\begin{itemize}
\item \blue{GNN efficiency bottleneck shifts across data scales. Graph operations only dominate time and memory overheads on graphs above the million scale. Additional variable weights further intensify the computation.}
\item \blue{Mini-batch training is capable of enhancing efficiency and scalability. Similarly, the improvement is only significant on larger graphs, where the graph computation overhead is considerable.}
\item The effectiveness and efficiency of spectral GNNs are not mutually exclusive, challenging the prevailing belief. With adequate configurations, simple filters can excel in high accuracy and fast computation on most graphs.
\item The effectiveness of spectral filters primarily roots in their alignment with the graph information; that is, preserving useful patterns while attenuating noise. Contrarily, sophisticated designs usually benefit generality but do not guarantee improved accuracy.
\item \blue{Our taxonomy can effectively characterize filter efficiency.} As the approach to harmonizing effectiveness and efficiency in spectral GNNs, it is recommended to employ simple but suitable filters through a better understanding of the graph data.
\end{itemize}

\section{Preliminaries}
\label{sec:related}
In this section, we introduce the concepts in spectral graph processing and GNNs, followed by configurations of filter operations, model architectures and learning pipelines. A comparative analysis with existing studies on GNN evaluations is provided to highlight the novelty of this benchmark.

\begin{table*}[!t]
\caption{Taxonomy of spectral GNN filters with notable corresponding models included in this paper. 
``Param.'' represents learnable filter parameters acquired during model training, while ``HP'' represents tunable filter hyperparameters acquired through hyperparameter search. 
``Time'' and ``Memory'' are time and space complexity for filter computation, respectively.}
\label{tab:summary}
\centering 
\fontsize{7.5pt}{9.5pt}\selectfont
\rowcolors{2}{white}{gray!5}
\setlength{\tabcolsep}{3.5pt}
\renewcommand{\arraystretch}{1.2}
\renewcommand\cellset{\renewcommand\arraystretch{0.6}\setlength\extrarowheight{0pt}} 
\begin{adjustbox}{max width=\textwidth}
\begin{threeparttable}
\begin{tabular}{>{\cellcolor{white}}c@{\hspace{2ex}}c|c@{\hspace{1pt}}c@{\hspace{1pt}}c|c@{\hspace{2pt}}c|>{\fontsize{6pt}{9.5pt}\selectfont}c}
\toprule 
  \textbf{Type} & \textbf{Filter}$^\ast$ & \textbf{Filter Function} $g(\Tilde{\bmu{L}})$ & \textbf{Param.} & \textbf{HP} & \textbf{Time} & \textbf{Memory} & \textbf{\fontsize{7pt}{9pt}\selectfont Model}$^\dagger$ \\
\midrule 
  & \ft{Identity}    & $\bmu{I}$ 
    & / & / & $O(KnF)$ & $O(nF)$ 
    & MLP \\
  & \ft{Linear}  & $2\bmu{I}-\Tilde{\bmu{L}}$ 
    & / & / & $O(KmF)$ & $O(nF)$ 
    & $\mathbb{I}$: GCN~\cite{kipf2016semi} \\
  & \ft{Impulse} & $(\bmu{I}-\Tilde{\bmu{L}})^K$ 
    & / & / & $O(KmF)$ & $O(nF)$ 
    & $\mathbb{D}$: SGC~\cite{wu19sim}, gfNN~\cite{nt2019revisiting}, GZoom~\cite{deng2020}, GRAND+~\cite{feng2022} \\
  & \ft{Monomial}& $\frac{1}{K+1}\sum_{k=0}^{K} (\bmu{I} - \Tilde{\bmu{L}})^k$ 
    & / & / & $O(KmF)$ & $O(nF)$ 
    & $\mathbb{D}$: S$^2$GC~\cite{zhu2021a}, AGP~\cite{wang2021b}, GRAND+~\cite{feng2022} \\
  & \ft{PPR}     & $\sum_{k=0}^K \alpha(1-\alpha)^k (\bmu{I} - \Tilde{\bmu{L}})^k$ 
    & / & $\alpha$ & $O(KmF)$ & $O(nF)$ 
    & \makecell{$\mathbb{I}$: GLP~\cite{li2019e}, GCNII~\cite{ming2020};\\ $\mathbb{D}$: APPNP~\cite{Klicpera2019}, GDC~\cite{gasteiger2019}, AGP~\cite{wang2021b}, GRAND+~\cite{feng2022}} \\
  & \ft{HK}      & $\sum_{k=0}^K \frac{e^{-\alpha}\alpha^k}{k!} (\bmu{I} - \Tilde{\bmu{L}})^k$ 
    & / & $\alpha$ & $O(KmF)$ & $O(nF)$ 
    & $\mathbb{D}$: GDC~\cite{gasteiger2019}, AGP~\cite{wang2021b}, DGC~\cite{NEURIPS2021_2d95666e} \\
 \multirow{-7}{*}{Fixed}
  & \ft{Gaussian} & $\sum_{k=0}^K \frac{\alpha^k}{k!} (2\bmu{I} - \Tilde{\bmu{L}})^{k}$ 
    & / & $\alpha$ & $O(KmF)$ & $O(nF)$ 
    & $\mathbb{D}$: G$^2$CN~\cite{li2022e} \\
\midrule 
  & \ft{Linear}  & $(1 + \theta)\bmu{I} - \Tilde{\bmu{L}}$ 
    & $\theta_k$ & / & $O(KmF)$ & $O(nF)$ 
    & $\mathbb{I}$: GIN~\cite{xu2019}, AKGNN~\cite{AKGNN} \\
  & \ft{Monomial}& $\sum_{k=0}^{K} \theta_k (\bmu{I} - \Tilde{\bmu{L}})^k$ 
    & $\theta_k$ & / & $O(KmF)$ & $O(nF)$ 
    & $\mathbb{D}$: DAGNN~\cite{liu2020b}, GPRGNN~\cite{chien2021} \\
  & \ft{Horner}  & $\sum_{k=0}^{K} \theta_k (\bmu{I} - \Tilde{\bmu{L}})^k$ 
    & $\theta_k$ & / & $O(KmF)$ & $O(2nF)$ 
    & $\mathbb{I}$: ARMAGNN~\cite{arma}, HornerGCN~\cite{guo2023a} \\
  & \ft{Chebyshev}& $\sum_{k=0}^K \theta_k T_{\text{Cheb}}^{(k)}(\Tilde{\bmu{L}})$ 
    & $\theta_k$ & / & $O(KmF)$ & $O(2nF)$ 
    & $\mathbb{I}$: ChebNet~\cite{defferrard2016convolutional}; $\mathbb{D}$: ChebBase~\cite{he2022a} \\ 
  & \ft{ChebInterp}& $\frac{2}{K+1} \sum_{k=0}^{K} \sum_{\kappa=0}^{K} \theta_\kappa T_{\text{Cheb}}^{(k)}(x_\kappa) T_{\text{Cheb}}^{(k)}(\Tilde{\bmu{L}})$ 
    & $\theta_k$ & / & $O(KmF+K^2nF)$ & $O(2nF)$ 
    & $\mathbb{D}$: ChebNetII~\cite{he2022a} \\
  & \ft{Clenshaw}& $\sum_{k=0}^K \theta_k T_{\text{Cheb2}}^{(k)}(\Tilde{\bmu{L}})$ 
    & $\theta_k$ & / & $O(KmF)$ & $O(3nF)$ 
    & $\mathbb{I}$: ClenshawGCN~\cite{guo2023a} \\ 
  & \ft{Bernstein}& $\sum_{k=0}^{K} \frac{\theta_k}{2^{K}} \binos{K}{k} (2\bmu{I} - \Tilde{\bmu{L}})^{K-k} \Tilde{\bmu{L}}^k$ 
    & $\theta_k$ & / & $O(K^2mF)$ & $O(nF)$ 
    & $\mathbb{D}$: BernNet~\cite{he2021} \\
  & \ft{Legendre}& $\sum_{k=0}^K \theta_k \frac{(-1)^k}{k!\binos{2k}{k}}\, (2\bmu{I} - \Tilde{\bmu{L}})^{k} \Tilde{\bmu{L}}^k$ 
    & $\theta_k$ & / & $O(KmF)$ & $O(2nF)$ 
    & $\mathbb{D}$: LegendreNet~\cite{Chen2023Improved} \\
  & \ft{Jacobi}  & $\sum_{k=0}^K \theta_k T_{\text{Jacobi}}^{(k)}(\Tilde{\bmu{L}})$ 
    & $\theta_k$ & $\alpha, \beta$ & $O(KmF)$ & $O(2nF)$ 
    & $\mathbb{D}$: JacobiConv~\cite{wang2022b} \\ 
  & \ft{Favard}  & $\sum_{k=0}^K \theta_k T_{\text{Favard}}^{(k)}(\Tilde{\bmu{L}})$ 
    & $\theta_k$ & / & $O(KmF+KnF)$ & $O(2nF)$ 
    & $\mathbb{D}$: FavardGNN~\cite{guo2023} \\ 
 \multirow{-11}{*}{Variable}
  & \ft{OptBasis}  & $\sum_{k=0}^K \theta_k T_{\text{OptBasis}}^{(k)}(\Tilde{\bmu{L}})$ 
    & $\theta_k$ & / & $O(KmF+KnF^2)$ & $O(2nF)$ 
    & $\mathbb{D}$: OptBasisGNN~\cite{guo2023} \\ 
\midrule 
  & \ft{Linear} (channel-wise) & $\bigparallel_{q=1}^{Q} (\bmu{I} - \gamma_q\Tilde{\bmu{L}})$ 
    & $\gamma_q$ & / & $O(KmF)$ & $O(nF)$ 
    & $\mathbb{D}$: AdaGNN~\cite{dong2021} \\
  & \ft{Linear} (LP, HP)     & $\gamma_1 (\bmu{I} - \Tilde{\bmu{L}}) + \gamma_2\Tilde{\bmu{L}}$ 
    & $\gamma_q$ & / & $O(QKmF+QKnF)$ & $O(QnF)$ 
    & $\mathbb{I}$: FBGCN (I/II)~\cite{luan2022a} \\
  & \ft{Linear} (LP, HP, ID) & $\gamma_1 (\bmu{I} - \Tilde{\bmu{L}}) + \gamma_2\Tilde{\bmu{L}} + \gamma_3\bmu{I}$ 
    & $\gamma_q$ & / & $O(QKmF+QKnF)$ & $O(QnF)$ 
    & $\mathbb{I}$: ACMGNN (I/II)~\cite{luan2022} \\
  & \ft{Linear} (LP, HP)     & $\gamma_1 ((\beta+1)\bmu{I}-\Tilde{\bmu{L}}) + \gamma_2 ((\beta-1)\bmu{I} + \Tilde{\bmu{L}})$ 
    & $\gamma_q$ & $\beta$ & $O(QKmF)$ & $O(QnF)$ 
    & $\mathbb{D}$: FAGNN~\cite{bo2021} \\
  & \ft{Gaussian} (LP, HP)   & $\sum_{q=1}^{Q}\sum_{k=0}^{\lfloor K/2\rfloor} \frac{\alpha_q^k}{k!} ((1+\beta_q)\bmu{I} - \Tilde{\bmu{L}})^{2k}$ 
    & $\gamma_q$ & $\alpha_q, \beta_q$ & $O(QKmF)$ & $O(QnF)$ 
    & $\mathbb{D}$: G$^2$CN~\cite{li2022e} \\
  & \ft{PPR} (LP, HP)  & $\sum_{q=1}^{Q}\sum_{k=0}^{K} \alpha_q(1-\alpha_q)^k (\bmu{I} + \beta_q\Tilde{\bmu{L}})(\bmu{I} - \Tilde{\bmu{L}})^k$ 
    & $\gamma_q$ & $\alpha_q, \beta_q$ & $O(QKmF)$ & $O(QnF)$ 
    & $\mathbb{D}$: GNN-LF/HF~\cite{zhu2021} \\
 \multirow{-7}{*}{Bank}
  & \ft{Mono, Cheb, Bern, ID} & $\sum_{q=1}^{\mathit{Q}}\sum_{k=0}^{K} \gamma_q \theta_{q,k} T_q^{(k)}(\Tilde{\bmu{L}})$ 
    & $\gamma_q, \theta_{q,k}$ & / & $O(QKmF)$ & $O(QnF)$ 
    & $\mathbb{D}$: FiGURe~\cite{ekbote2023figure} \\
\bottomrule 
\end{tabular}
\begin{tablenotes}
    \item [$\ast$] \textbf{Filter}: LP = Low-pass, HP = High-pass, ID = Identity. 
    \item [$^\dagger$] \textbf{Model}: $\mathbb{I}$ = Iterative architecture, $\mathbb{D}$ = Decoupled architecture.
\end{tablenotes}
\end{threeparttable}
\end{adjustbox}
\end{table*}

\subsection{Formulation of Spectral GNN Paradigm}
\label{ssec:paradigm}
\noindentparagraph{Graph Notation.}
\label{ssec:spatial}
An undirected graph is denoted as $\mathcal{G} = \langle\mathcal{V}, \mathcal{E}\rangle$ with $n = |\mathcal{V}|$ nodes and $m = |\mathcal{E}|$ edges.
The adjacency matrix without and with self-loops are $\bmu{A}$ and $\Bar{\bmu{A}}=\bmu{A}+\bmu{I}$, respectively. The normalized adjacency matrix is $\Tilde{\bmu{A}} = \Bar{\bmu{D}}^{\rho-1} \Bar{\bmu{A}} \Bar{\bmu{D}}^{-\rho}$, where $\Bar{\bmu{D}}$ is the self-looped diagonal degree matrix and $\rho\in[0,1]$ is the normalization coefficient \cite{wang2021b,yang2022a}.
The attributed graph is associated with a node attribute matrix $\bmu{X} \in \RR{n \times F}$, carrying an $F$-dimensional attribute vector for each node.

\noindentparagraph{Spectral Graph Theory.}
\label{ssec:spectral}
Spectral graph theory \cite{dong2020graph} relate the graph with signal processing techniques, which utilizes the graph Laplacian matrix $\bmu{L}=\bmu{D}-\bmu{A}$ or its normalized counterpart $\Tilde{\bmu{L}}=\bmu{I}-\Tilde{\bmu{A}}$. Graph Laplacian leads to its eigen-decomposition $\Tilde{\bmu{L}} = \bmu{U} \bmu{\Lambda} \bmu{U}\tran$, where the graph \textit{spectrum} $\bmu{\Lambda} = \diag(\lambda_1, \cdots, \lambda_n)$ is the diagonal matrix of eigenvalues $0=\lambda_1 \le \lambda_2 \le\cdots\le \lambda_n \le 2$, and $\bmu{U}$ is the matrix consisting of eigenvectors. The eigenvalues are also known as the signal \textit{frequencies}, where small and large values correspond to low and high frequencies in the graph, respectively.

The spectral graph property is highly related to the heterophily pattern, which depicts the spatial distribution of similar graph nodes and motivates a number of spectral GNN designs \cite{luan2024}. A \textit{homophilous} graph implies that nodes with the same labels tend to connect to each other, while a \textit{heterophilous} graph does not. Graphs of high heterophily usually signify high-frequency spectral signals.
\blue{An exemplar measure of the heterophily property is} the node homophily score \cite{HongbinPei2020GeomGCNGG}: {$\mathcal{H} = \frac{1}{n}\sum_{u \in V} |\{v | v \in \mathcal{N}(u), y(v) = y(u) \}|/|\mathcal{N}(u)|$, where $y(v)$ is the label of node $v$. A score closer to $0$ indicates higher homophily with more similar neighbors.

\noindentparagraph{Spatial Graph Neural Network.}
\label{ssec:spatial_gnn}
General GNNs are based on the \textit{spatial} interpretation of the graph structure $\Tilde{\bmu{A}}$, which is exploited to perform the \textit{propagation} operation $f(\Tilde{\bmu{A}})$. The model aims to iteratively update the node-wise representation $\bmu{H}$ through its $j$-th layer: $\bmu{H}^{(j+1)} = \varphi( f^{(j)}(\Tilde{\bmu{A}}) \cdot \bmu{H}^{(j)})$, where $\varphi$ denotes the \textit{transformation} consisting of learnable weights and non-linear activations. The representation can be initialized by node attributes $\bmu{H}^{(0)} = \bmu{X}$.

Typically, one GNN layer corresponds to one hop of propagation through graph edges, and spatial GNNs stack $J$ layers to receive more information from a larger field. The layer combination derives two variants of the GNN architecture, since the two types of operations can be either integrated or separated: An \textit{iterative} model performs propagation $f$ and transformation $\varphi$ alternatively across its layers, while the \textit{decoupled} architecture executes all propagations together, with weight transformations only applied on the propagation results.
Despite the different combinations of operations, these two architectures are considered to posses the same expressiveness in propagating graph information \cite{zhang2021,wang2022b}.

\noindentparagraph{Spectral Graph Neural Network.}
\label{ssec:spectral_gnn}
Spectral GNNs view the graph information as a matrix to perform signal analysis techniques such as eigen-decomposition. Although it is based on the distinct interpretation with spatial GNNs, these two variants can be derived from each other, as revealed by previous research \cite{kipf2016semi,chen2023}.
To demonstrate, we start from the spatial GNN formulation by focusing on the graph-related components $f^{(j)}(\Tilde{\bmu{A}})$ in GNN and omitting the non-linear transformation, i.e., let $\varphi(\bmu{x}) = \bmu{x}$. We also utilize the graph Laplacian $\Tilde{\bmu{L}}$ instead of adjacency to denote graph information, and use the $F=1$ attribute vector $\bmu{x}$ as node-wise signal instead of the attribute matrix.
In this case, the $K$-hop graph computation can be characterized by a polynomial $g(\Tilde{\bmu{L}})$:
\begin{equation}
\label{eq:sgnn}
    g(\Tilde{\bmu{L}}) \cdot \bmu{x} = \sum_{k=0}^{K} \theta_k\, T^{(k)}(\Tilde{\bmu{L}}) \,\bmu{x},
\end{equation}
where $T^{(k)}(\Tilde{\bmu{L}})$ is the $k$-th term of the polynomial basis, and $\theta_k$ is the corresponding parameter. Combination of these two components determines the actual management of graph signals \cite{wang2022b}.

In spectral domain, a \textit{filter} $\hat{g}: [0,2] \rightarrow \mathbb{R}$ processes the input \textit{signal} into \textit{response} by amplifying or attenuating certain frequencies. Denote the filtering operation on signal $\bmu{x}$ as $\hat{g} \ast \bmu{x}$, its relationship with spatial operations is given by the Fourier transform:
\begin{equation}
\label{eq:sconv}
    \hat{g} \ast \bmu{x}
    = \bmu{U} \hat{g}(\bmu{\Lambda}) \bmu{U}\tran \bmu{x}
    = \sum_{i=0}^{n} \hat{g}(\lambda_i)\bmu{u}_i\bmu{u}_i\tran \bmu{x}.
\end{equation}
\blue{\cref{eq:sconv} can be understood by three consecutive stages: the Fourier transform $\bmu{y}_1 = \bmu{U}\tran \bmu{x}$ which transfers the input signal to the spectral domain; the modulation $\bmu{y}_2 = \hat{g}(\bmu{\Lambda}) \bmu{y}_1$ which applies graph information through the filter; and lastly the inverse Fourier transform $\bmu{x}^\star = \bmu{U} \bmu{y}_2$ which transfers the output back from the spectral domain. }

\cref{eq:sconv} implies that the spectral filtering operation can be achieved by iterative spatial multiplications with the signal. Therefore, \cref{eq:sgnn} provides a feasible approximation by only considering $K \ll n$ orders of those relatively \blue{significant} graph spectrum. In practice, this polynomial form dominates the actual implementations in spectral GNNs, bypassing the prohibitive eigen-decomposition through the well-established propagation based on graph adjacency. As a consequence, spectral and spatial GNNs share the same elementary operations, rendering the possibility to derive and evaluate a large family of GNNs under the interpretation of spectral filters.

\subsection{GNN Learning Workflow}
\noindentparagraph{Complexity Analysis.}
\label{ssec:model}
As indicated by \cref{eq:sconv}, spectral and spatial GNNs share the common elementary operations.
\textit{Propagation} describes the process of applying graph matrix to the node representation. Each application of the basis $T^{(k)}(\Tilde{\bmu{L}})$ to the graph signal in \cref{eq:sgnn} is regarded as one propagation step.
The operation is conducted by the matrix multiplication between the $n \times n$ sparse graph matrix and the $n \times F$ dense representation, resulting in $O(mF)$ time. The memory overhead of graph matrix is $O(m)$.
The \textit{transformation} operation updates the representation with learnable weights. Typical transformations include scalar and matrix multiplication, corresponding to time complexity $O(nF)$ and $O(nF^2)$, respectively. Employing a representation matrix to be iteratively updated for all nodes throughout learning consumes $O(nF)$ memory space, while advanced transformation architectures such as residual connection and concatenation cause time and memory expense to rise in accordance.

\noindentparagraph{Learning Schemes.}
\label{ssec:batch}
The learning scheme describes how the model and data are deployed to CPU and GPU for training and inference. \textit{Full-batch} (FB) training loads all input data onto the GPU, which suffices the need on small-scale graphs and is the de facto scheme for most GNNs. However, for larger graphs, the critical scalability issue emerges, as the full graph topology exceeds the GPU memory capacity. To mitigate the overhead, a common model-agnostic solution is to employ \textit{graph partition} (GP) algorithms to divide and train the graph data in batches. However, this process impacts the graph topology and undermines GNN expressiveness \cite{duan2022comprehensive,ma2024}.

In contrast, the decoupled \textit{mini-batch} (MB) scheme is uniquely available for the spectral paradigm. It chooses to perform graph-related operations to generate one or more node-wise representation matrices in a precomputation stage on CPU. Then, only the intermediate representations are loaded onto the GPU in batches during training, which prevents the memory footprint from being coupled with the graph size and enjoys better scalability. As showcased in \cref{plot:schemes}, the filter $g(\Tilde{\bmu{L}})$ remains invariant across different model architectures and training schemes with varied efficiency and scalability scenarios. \blue{This invariance allows for a shared} formulation and implementation across these scenarios, which can be further enhanced by dedicated graph data management techniques.

\begin{table}[!t]
\begin{adjustbox}{max width=\columnwidth}
\begin{threeparttable}
\caption{Review of GNN frameworks and benchmarks literature with empirical evaluations. ``Scheme'': available learning schemes. ``Eff.'': whether efficiency is evaluated. ``Scale'': largest dataset used in terms of the number of nodes. 
}
\label{tab:related}
\centering
\rowcolors{2}{white}{gray!5}
\setlength{\tabcolsep}{3.5pt}
\renewcommand{\arraystretch}{1.05}
\begin{tabular}{@{}>{\cellcolor{white}\small}c@{\vspace{2pt}}>{\small}l|>{\small}p{0.32\columnwidth}ccc@{}} \toprule
\textbf{Type} & \textbf{\normalsize{Study}} & \textbf{\normalsize{Spectral Models}} & \textbf{Scheme}$^\ast$ & \textbf{Eff.} & \textbf{Scale} \\ \midrule
    & PyG~\cite{pyg} & GCN, ChebNet, SGC, APPNP, ARMA & FB \& GP & \xxmark{} & / \\  
    & \citet{olegplatonov2023} & GCN, CPGNN, FAGCN, GPRGNN, JacobiConv & FB & \xxmark{} & 49K \\ 
    & OGB \cite{Hu2020} & GCN, SGC & FB & \xxmark{} & 100M \\ 
    \multirow{-4}{*}{General} & LINKX \cite{lim2021} & GCN, SGC, APPNP, GPRGNN & FB \& MB & \xxmark{} & 3M \\ 
\midrule[0.5pt]
    & \citet{maekawa2022beyond} & GCN, ChebNet, SGC, GPRGNN & FB & \ccmark{} & 15K \\
    & \citet{dwivedi2022} & GCN & FB \& GP & \ccmark{} & 235K \\ 
    & \citet{duan2022comprehensive} & SGC & FB \& MB & \ccmark{} & 3M \\
    \multirow{-4}{*}{Efficiency} & SGL \cite{ma2024} & SGC, S$^2$GC, APPNP & FB \& MB & \ccmark{} & 3M \\
\midrule[0.5pt]
    & GAMMA-Spec \cite{bo2023b} & 14 & FB & \xxmark{} & 20K \\ 
    \multirow{-2}{*}{Spectral} & \textbf{Ours} & \ul{35} & FB \& MB & \ccmark{} & 3M \\ 
\bottomrule
\end{tabular}
\begin{tablenotes}
    \item [$\ast$] \textbf{Scheme}: FB = Full-Batch, GP = Graph Partition, MB = Mini-Batch.
\end{tablenotes}
\end{threeparttable}
\end{adjustbox}
\vspace{-.6em}
\end{table}

\subsection{Related Works}
\label{ssec:related}
\noindentparagraph{GNNs Surveys and Benchmarks.}
The broad area of graph neural networks has been extensively reviewed from different aspects, while the spectral paradigm is a relatively unprecedented topic. We respectively discuss the related benchmark studies:

\noindent$\bullet$ Surveys for \textit{general} GNNs examine the algorithmic designs including architectural selections, convolutional and sequential operations, and overall graph learning pipelines \cite{zhang2020b,zhou2020,wu2021Comprehensive}. Software frameworks \cite{pyg,wang2019dgl} and datasets \cite{olegplatonov2023,Hu2020,lim2021} have also been developed.
Although a few classic spectral models are covered, they are not distinguished from other spatial models.

\noindent$\bullet$ The GNN \textit{efficiency} is especially surveyed in \cite{abadal2022,liu2022,zhang2023d} and evaluated in \cite{dwivedi2022,maekawa2022beyond,duan2022comprehensive,ma2024}, focusing on the empirical performance of algorithmic designs in GNN architectures.
However, most of these techniques are limited to spatial interpretations, which are orthogonal to the spectral perspective in our work.

\noindent$\bullet$ Two recent surveys target \textit{spectral} GNNs: \cite{chen2023} derives the general connection between typical spatial and spectral operators, advancing the unified interpretation. Our definition of spectral GNNs is primarily based on this work.
\cite{bo2023b} reviews 14 spectral models identified by the eigen-decomposition. However, both frameworks contain computation-intensive operations, which are prohibitive for large graphs. We further elaborate on the derivation between our taxonomy and these works in \cref{seca:model}.

\cref{tab:related} presents the literature containing empirical evaluations. Overall, we note a lack of research encompassing a broad range of up-to-date spectral GNNs, particularly in benchmarking graph operations and scalable implementations. This study hence contributes by offering efficiency comparisons, performance on large-scale datasets, and a comprehensive coverage of spectral filters.

\noindentparagraph{Efficient GNN Computation.}
To mitigate the computational bottleneck, a series of studies \cite{liao2022,liao2024scalable,li2023,zheng2023,feng2022a,gao2024,liu2025sigma} accelerate graph propagation in specific existing \textit{spectral} filters using dedicated approximate algorithms. \textit{Spatial} techniques such as sparsification~\cite{li2021e,Wickman2022,chen2023d,liao2025unifews}, sampling~\cite{luo2018TOAIN,yoon2021,song2023b}, and compression~\cite{zhou2021,ma2024a} are also widely explored to reduce the overhead of model operations. 
Although we do not directly evaluate these works on compute optimization, some efficient data processing techniques from them have been incorporated into our pipeline.

\noindentparagraph{GNN Training Systems.}
Instead of tailoring particular models, system-level research aims to enhance efficiency and scalability through data management techniques on graph topology \cite{li2023e,ai2024,zhu2019a,zhang2020c,wan2023} as well as structured embeddings \cite{huang2024,park2022,sheng2024,miao2021a}. Additionally, a line of studies advances GNNs training in distributed environments \cite{Liu2020,gandhi2021,zheng2022,guliyev2024,zhang2023h}.
These systems typically partition the graph for data parallelism, which differs from our straightforward mini-batch scheme in \cref{ssec:batch}. We refer readers to \cite{shen2024,yuan2024} for more comprehensive reviews on this topic.

\section{Taxonomy of Spectral Filters}
\label{sec:taxonomy}
In this study, we propose a novel taxonomy for spectral GNNs based on their bases $T(\Tilde{\bmu{L}})$ and parameters $\theta$, and categorize them into three types in \cref{tab:summary}.
We respectively introduce the filter types and their representative models, while detailed derivations of filters within each type are in \cref{seca:summary}.

\noindentparagraph{Selection and Naming of Filters.}
\label{ssec:naming}
We extensively survey existing GNN designs that can be represented in the polynomial form \cref{eq:sgnn}. In specific, we focus on works encompassing graph data and propagation operations, which signify the spectral formulations $g(\Tilde{\bmu{L}}; \theta)$ of the filter.
The naming of the filters generally follows the cited original works in \cref{tab:summary}, since most of the GNN models are inspired by traditional polynomial functions or signal processing filters. Exceptional cases include the novel filters \ft{ChebInterp} and \ft{OptBasis} as well as filter bank models, which are named after the respective works in \cref{tab:summary}.

\subsection{Fixed Filter GNN}
For the first type of spectral GNNs, the basis and parameters are both constant during learning, resulting in fixed filters $g(\Tilde{\bmu{L}})$. In practice, these filters are usually well-studied graph data processing schemes such as monomial summation or PPR propagation.
Below we give an example of the classic APPNP model and its filter:

\noindentparagraph{\ft{Personalized PageRank (PPR)}.}
APPNP~\cite{Klicpera2019} is a pioneering decoupled GNN that iteratively applies a decaying graph propagation $\bmu{H}^{(j+1)} = \varphi((1-\alpha)\Tilde{\bmu{A}} \bmu{H}^{(j)} + \alpha \bmu{H}^{(0)})$, where $\alpha \in [0,1]$ denotes the decay coefficient. It is revealed to be equivalent to the well-studied PPR calculation~\cite{Page1999,liu2024bird}, i.e., iteratively multiplying the signal with $\Tilde{\bmu{A}}^k$, and accumulating with a decaying coefficient $\alpha(1-\alpha)^k$. Recall that $\Tilde{\bmu{L}}=\bmu{I}-\Tilde{\bmu{A}}$, the spectral basis for each iteration can be written as $T^{(k)}(\Tilde{\bmu{L}}) = (\bmu{I} - \Tilde{\bmu{L}})^k$, and the corresponding parameter is $\theta_k = \alpha(1-\alpha)^k$. We name it as the \ft{PPR} filter:
\begin{equation*}
    g(\Tilde{\bmu{L}})
    = \sum_{k=0}^K \alpha(1-\alpha)^k (\bmu{I} - \Tilde{\bmu{L}})^k,\quad
    \theta_k = \alpha(1-\alpha)^k.
\end{equation*}
Regarding computational complexity, calculating the $K$-order filter requires $O(KmF)$ time for the graph computation and $O(nF)$ space for storing the representation matrix.

\subsection{Variable Filter GNN}
The second type in our taxonomy also features a predetermined basis $T^{(k)}(\Tilde{\bmu{L}})$ in \cref{eq:sgnn}. The series of parameters $\theta_k$ is variable, usually acquired through gradient descent during model training, leading to a variable filter denoted as $g(\Tilde{\bmu{L}}; \theta)$.
Compared to spectral GNNs with fixed filters, these models are claimed to enjoy better capability, especially for fitting high-frequency signals and capturing non-local topology.
Below we present the \ft{Chebyshev} filter:

\noindentparagraph{\ft{Chebyshev}.}
ChebNet~\cite{defferrard2016convolutional} is a representative spectral GNN that explicitly utilizes the classical Chebyshev polynomial \cite{HAMMOND2011129} as basis, which possesses orthogonal terms with straightforward computation. Each term $T^{(k)}(\Tilde{\bmu{L}})$ is expressed in the recursive form based on the computation results $T^{(k-1)}(\Tilde{\bmu{L}})$ and $T^{(k-2)}(\Tilde{\bmu{L}})$ of previous hops. The \ft{Chebyshev} filter can be formulated as:
\begin{align*}
    g(\Tilde{\bmu{L}}; \theta)
    = \sum_{k=0}^K \theta_k T^{(k)}(\Tilde{\bmu{L}}),\quad
    &T^{(k)}(\Tilde{\bmu{L}}) = 2 \Tilde{\bmu{L}} T^{(k-1)}(\Tilde{\bmu{L}}) - T^{(k-2)}(\Tilde{\bmu{L}}),\\
    &T^{(1)}(\Tilde{\bmu{L}}) = \Tilde{\bmu{L}},\, T^{(0)}(\Tilde{\bmu{L}}) = \bmu{I}.
\end{align*}
The computation scheme of ChebNet can be directly inferred from the filter formulation. Initially, it employs $\bmu{X}$ and $\Tilde{\bmu{L}}\bmu{X}$ as the $0$- and $1$-order terms. Then, the recursive equation is followed to apply $T^{(k)}(\Tilde{\bmu{L}})$ and results are accumulated by multiplying $\theta_k$.
Its complexity for performing graph operations is $O(KmF)$ We denote its memory footprint as $O(2nF)$ for maintaining two $n\times F$ matrices during the iterative filter computation.

\subsection{Filter Bank GNN}
Some studies argue that a single filter is limited in leveraging complex graph signals. Hence, spectral GNNs with filter bank arise as an advantageous approach to provide abundant information for learning. In this study, we innovatively incorporate these models into the spectral GNN taxonomy as a mixture of $\mathit{Q}$ fixed or variable filters $g_q(\Tilde{\bmu{L}}; \theta)$, each assigned a filter weight parameter $\gamma_q$:
\begin{equation}
\label{eq:fbank}
    \mathscr{g}(\Tilde{\bmu{L}}; \gamma, \theta) = \bigoplus_{q=1}^{\mathit{Q}} \gamma_q \cdot g_q(\Tilde{\bmu{L}}; \theta),\quad
    g_q(\Tilde{\bmu{L}}; \theta) = \sum_{k=0}^{K} \theta_{q,k}\, T_q^{(k)}(\Tilde{\bmu{L}}),
\end{equation}
where $\bigoplus$ denotes an arbitrary fusion function such as summation or concatenation.
By this means, the filter bank is able to cover different \textit{channels}, i.e., frequency ranges, to generate more comprehensive embeddings of the graph. The filter parameter $\gamma$ can be either learned separately or along with GNN training, depending on the specific model implementation.

\noindentparagraph{\ft{FiGURe}.}
FiGURe~\cite{ekbote2023figure} is an exemplar filter bank GNN following exactly \cref{eq:fbank} with summation fusion $\mathscr{g}(\Tilde{\bmu{L}}; \gamma, \theta) = \sum_{q=1}^{\mathit{Q}} \gamma_q \cdot g_q(\Tilde{\bmu{L}}; \theta)$, where filter-level parameters $\gamma_q$ are learned to control the channel strength.
Common variable filters including \ft{Monomial}, \ft{Chebyshev}, and \ft{Bernstein} bases are utilized to compose the filter bank.
With a straightforward implementation, the time and memory complexity of the combination are $Q$ times that of an individual filter.

\section{Benchmark Design}
\label{sec:design}

\noindentparagraph{Tasks and Metrics.}
In the main experiment, we focus on semi-supervised learning for node-classification task on a single graph, while other tasks are also analyzed in \cref{sec:experiment-extra}. Particularly, the evaluation features some of the largest datasets available for graph learning, rendering it the mainstream task for spectral GNNs and is ideal for assessing scalability.
We follow the dataset settings for using \textit{efficacy} metrics including accuracy and ROC AUC.
\textit{Efficiency} with respect to both running speed and CPU/GPU memory footprint is separately measured for each existing learning stage, including precomputation, training, inference.
We conduct 10 runs with different random seeds by default and report the average metrics with standard deviation. Evaluations are conducted on a single machine with 32 Intel Xeon CPUs (2.4GHz), an Nvidia A30 GPU (24GB memory), and 512GB RAM.

\begin{table}[!b]
\captionsetup{skip=6pt,font={large,stretch=0.9}}
\begin{adjustbox}{max width=\columnwidth}
\begin{threeparttable}
\caption{Dataset statistics. $\mathcal{H}$, $F_{i}$, and $F_{o}$ are homophily score, input attribute dimension, and number of class labels, respectively.}
\label{taba:dataset}
  \centering
  \renewcommand{\arraystretch}{0.95}
  \setlength{\tabcolsep}{2.2pt}
\begin{tabular}{cc@{}c|rrrrrc@{}} \toprule
    \multicolumn{2}{c@{}}{\,\textbf{Category}$^\ast$} & \textbf{Dataset} 
        & \textbf{Nodes} $n$ & \textbf{Edges} $m$ & $\mathcal{H}$ & $F_{i}$ & $F_{o}$ & \textbf{Metric} \\ 
\midrule
    \multirow{11}{*}{S} & \multirow{6}{*}{Homo.} 
    & \ds{cora} \cite{Sen2008}
        & 2,708 & 10,556 & 0.83 & 1,433 & 7 & Accuracy  \\
    & & \ds{citeseer} \cite{Sen2008}
        & 3,327 & 9,104 & 0.72 & 3,703 & 6 & Accuracy  \\
    & & \ds{pubmed} \cite{Sen2008}
        & 19,717 & 88,648 & 0.79 & 500 & 3 & Accuracy  \\
    & & \ds{minesweeper} \cite{olegplatonov2023} 
        & 10,000 & 78,804 & 0.68 & 7 & 2 & ROC AUC  \\
    & & \ds{questions} \cite{olegplatonov2023} 
        & 48,921 & 307,080 & 0.90 & 301 & 2 & ROC AUC  \\
    & & \ds{tolokers} \cite{olegplatonov2023} 
        & 11,758 & 1,038,000 & 0.63 & 10 & 2 & ROC AUC  \\
\cmidrule{2-9}
    & \multirow{5}{*}{Hetero.} 
    & \ds{chameleon} \cite{HongbinPei2020GeomGCNGG} 
        & 890 & 17,708 & 0.24 & 2,325 & 5 & Accuracy  \\
    & & \ds{squirrel} \cite{HongbinPei2020GeomGCNGG} 
        & 2,223 & 93,996 & 0.19 & 2,089 & 5 & Accuracy  \\
    & & \ds{actor} \cite{HongbinPei2020GeomGCNGG} 
        & 7,600 & 30,019 & 0.22 & 932 & 5 & Accuracy  \\
    & & \ds{roman} \cite{olegplatonov2023} 
        & 22,662 & 65,854 & 0.05 & 300 & 18 & Accuracy  \\
    & & \ds{ratings} \cite{olegplatonov2023} 
        & 24,492 & 186,100 & 0.38 & 300 & 5 & Accuracy  \\
\hline
    \multirow{6}{*}{M} & \multirow{2}{*}{Homo.} 
    & \ds{flickr} \cite{zeng2019graphsaint}
        & 89,250 & 899,756 & 0.32 & 500 & 7 & Accuracy  \\
    & & \ds{ogbn-arxiv} \cite{Hu2020} 
        & 169,343 & 1,166,243 & 0.63 & 128 & 40 & Accuracy  \\
\cmidrule{2-9}
    & \multirow{4}{*}{Hetero.} 
    &  \ds{arxiv-year} \cite{lim2021} 
        & 169,343 & 1,166,243 & 0.31 & 128 & 5 & Accuracy  \\
    & & \ds{penn94} \cite{lim2021} 
        & 41,554 & 2,724,458 & 0.48 & 4,814 & 2 & Accuracy  \\
    & & \ds{genius} \cite{lim2021} 
        & 421,961 & 984,979 & 0.08 & 12 & 2 & ROC AUC  \\
    & & \ds{twitch-gamer} \cite{lim2021} 
        & 168,114 & 6,797,557 & 0.10 & 7 & 2 & Accuracy  \\
\hline
    \multirow{5}{*}{L} & \multirow{2}{*}{Homo.} 
    & \ds{ogbn-mag} \cite{Hu2020} 
        & 736,389 & 5,416,271 & 0.31 & 128 & 349 & Accuracy    \\
    & & \ds{ogbn-products} \cite{Hu2020} 
        & 2,449,029 & 123,718,280 & 0.83 & 100 & 47 & Accuracy    \\
\cmidrule{2-9}
    & \multirow{3}{*}{Hetero.} 
    & \ds{pokec} \cite{lim2021} 
       & 1,632,803 & 30,622,564 & 0.43 & 65 & 2 & Accuracy  \\
    & & \ds{snap-patents} \cite{lim2021} 
       & 2,923,922 & 13,972,555 & 0.22 & 269 & 5 & Accuracy  \\
    & & \ds{wiki} \cite{lim2021} 
       & 1,925,342 & 303,434,860 & 0.28 & 600 & 5 & Accuracy  \\
\bottomrule
  \end{tabular}
\begin{tablenotes}
    \item [$\ast$] \textbf{Category}: S = Small-scale, M = Medium-scale, L = Large-scale;\quad Homo. = Homophily, Hetero. = Heterophily.
\end{tablenotes}
\end{threeparttable}
\end{adjustbox}
\end{table}

\noindentparagraph{Datasets.}
We comprehensively involve 22 datasets that are widely used in GNN node classification detailed in \cref{taba:dataset}. In the table, we incorporate self-loop edges and count undirected edges twice to better reflect the graph computation overhead. We mainly categorize the datasets from two aspects.
In efficacy evaluations, we separately investigate \textit{homophilous} and \textit{heterophilous} datasets considering they exhibit different patterns of graph signals.
Based on the graph scale, we alternatively distinguish them into \textit{small}, \textit{medium}, and \textit{large} categories. We focus on medium and large datasets for efficiency analysis, with large graphs identified by empirical evaluation that at least some models are prohibitive due to the critical scalability issue. We follow the unified dataset processing protocol for common node classification task including edge direction, degree normalization, and feature transformation in line with the original literature. Random 60\%/20\%/20\% splits for train/validation/test sets are used for graphs without predefined splits.

\noindentparagraph{Model Architectures and Hyperparameters.}
Our selection of model architectures and hyperparameters intends to provide fair and comparable results of both effectiveness and efficiency while minimizing the effect of non-spectral aspects. Hence, we focus on the decoupled architecture with a same number of transformation layers.
Referring to \cref{tab:summary}, the number of hops $K$ and hidden dimension $F$ are the critical model-level hyperparameters affecting computational complexity. A \textit{universal} choice is therefore searched and then
fixed for experiments across all models and datasets: the number of training epochs is kept constant at 500, while batch sizes are set to 4,096 and 200,000 for mini-batch learning on small/medium- and large-scale datasets, respectively. For the other hyperparameters, including filter-level ones in \cref{tab:summary}, we perform \textit{individual} tuning for each model and dataset to pursue satisfying accuracy. The ranges and used values of hyperparameter search are listed in \cref{tab:param}.

\begin{table}[t]
\captionsetup{font={small,stretch=0.9}}
\caption{Configuration search scheme. Hyperparameters are explored based on the combination of listed ranges, and \ul{underlined} values are the universal settings used across main experiments.}
\label{tab:param}
\centering
\setlength{\tabcolsep}{3pt}
\begin{adjustbox}{max width=\columnwidth}
\begin{tabular}{cc|ccc} \toprule
    \textbf{Scheme} & \textbf{Hyperparameter} & \textbf{Full-batch} & \textbf{Mini-batch} \\ \midrule
    & Propagation hop $K$ & \{2, 4, $\cdots$, \ul{10}, $\cdots$, 30\} & \{2, 4, $\cdots$, \ul{10}, $\cdots$, 30\} \\
    & Hidden width $F$ & \{16, 32, 64, \ul{128}, 256\} & \{16, 32, 64, \ul{128}, 256\} \\
    & Layer of linear $\varphi_0$ & \{\ul{1}, 2, 3\} & \{\ul{0}\} \\
    \multirow{-4}{*}{Universal} & Layer of linear $\varphi_1$ & \{\ul{1}, 2, 3\} & \{1, \ul{2}, 3\} \\
\midrule
    & Graph normalization $\rho$ & $[0, 1]$ & $[0, 1]$ \\
    & Learning rate of $\varphi_0, \varphi_1$ & $[10^{-5}, 0.5]$ & $[10^{-5}, 0.5]$ \\
    & Learning rate of $\theta, \gamma$ & $[10^{-5}, 0.5]$ & $[10^{-5}, 0.5]$ \\
    & Weight decay of $\varphi_0, \varphi_1$ & $[10^{-7}, 10^{-3}]$ & $[10^{-7}, 10^{-3}]$ \\
    \multirow{-5}{*}{Individual} & Weight decay of $\theta, \gamma$ & $[10^{-7}, 10^{-3}]$ & $[10^{-7}, 10^{-3}]$ \\
\bottomrule
\end{tabular}
\end{adjustbox}
\end{table}

\vspace{-0.8em}
\section{Results: Efficiency and Effectiveness}
\label{sec:experiment-main}
In this section, we present primary research questions (RQs) and analysis on the performance comparison among a total of 27 filters across three types, benchmarking their effectiveness, efficiency, and memory overhead under different datasets and training schemes. Full results can be found in \cref{seca:exp}.

\begin{figure*}[tp]
\centering
\setlength{\tabcolsep}{0pt}
\begin{tabular}{cccc}
    \includegraphics[height=1.45in]{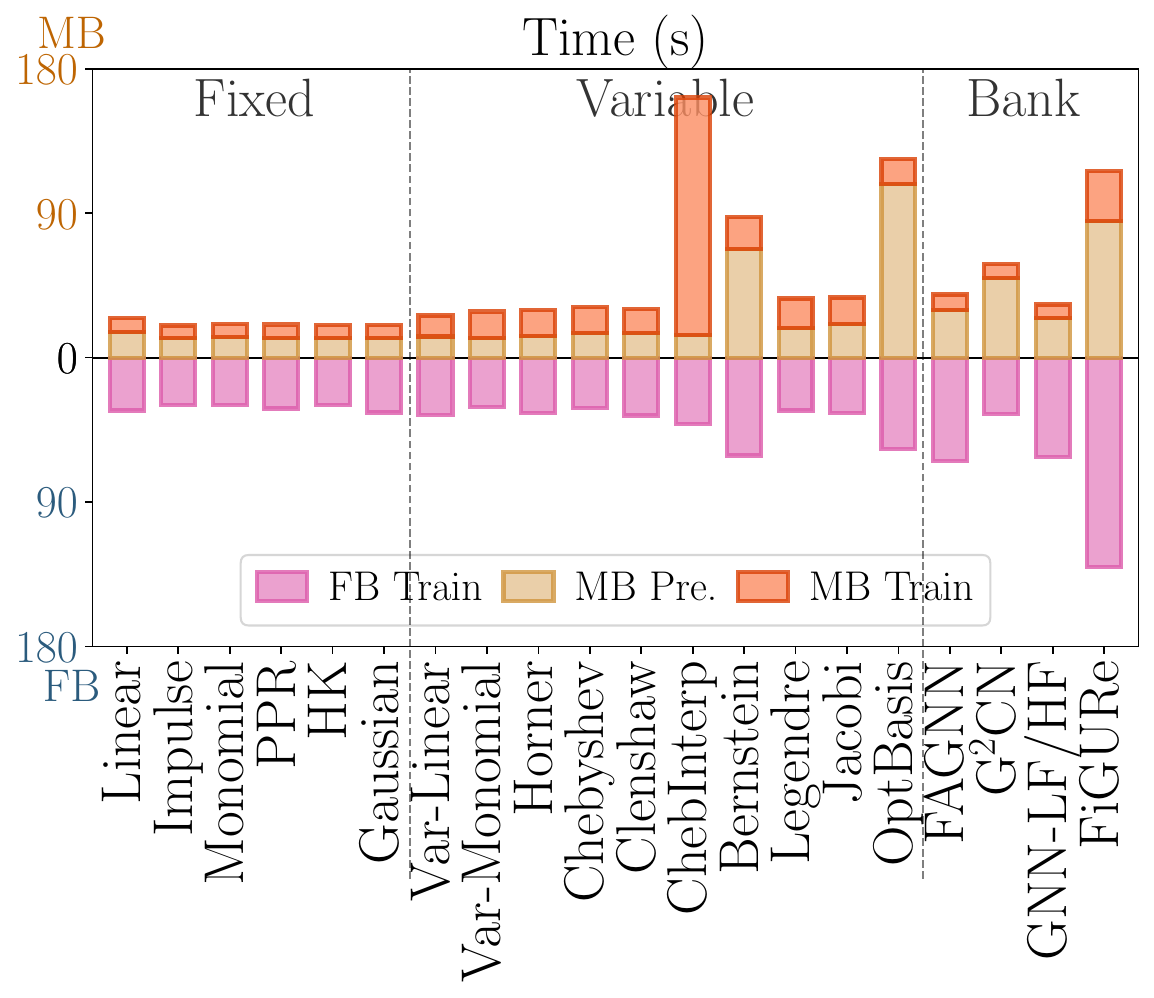} & \includegraphics[height=1.45in]{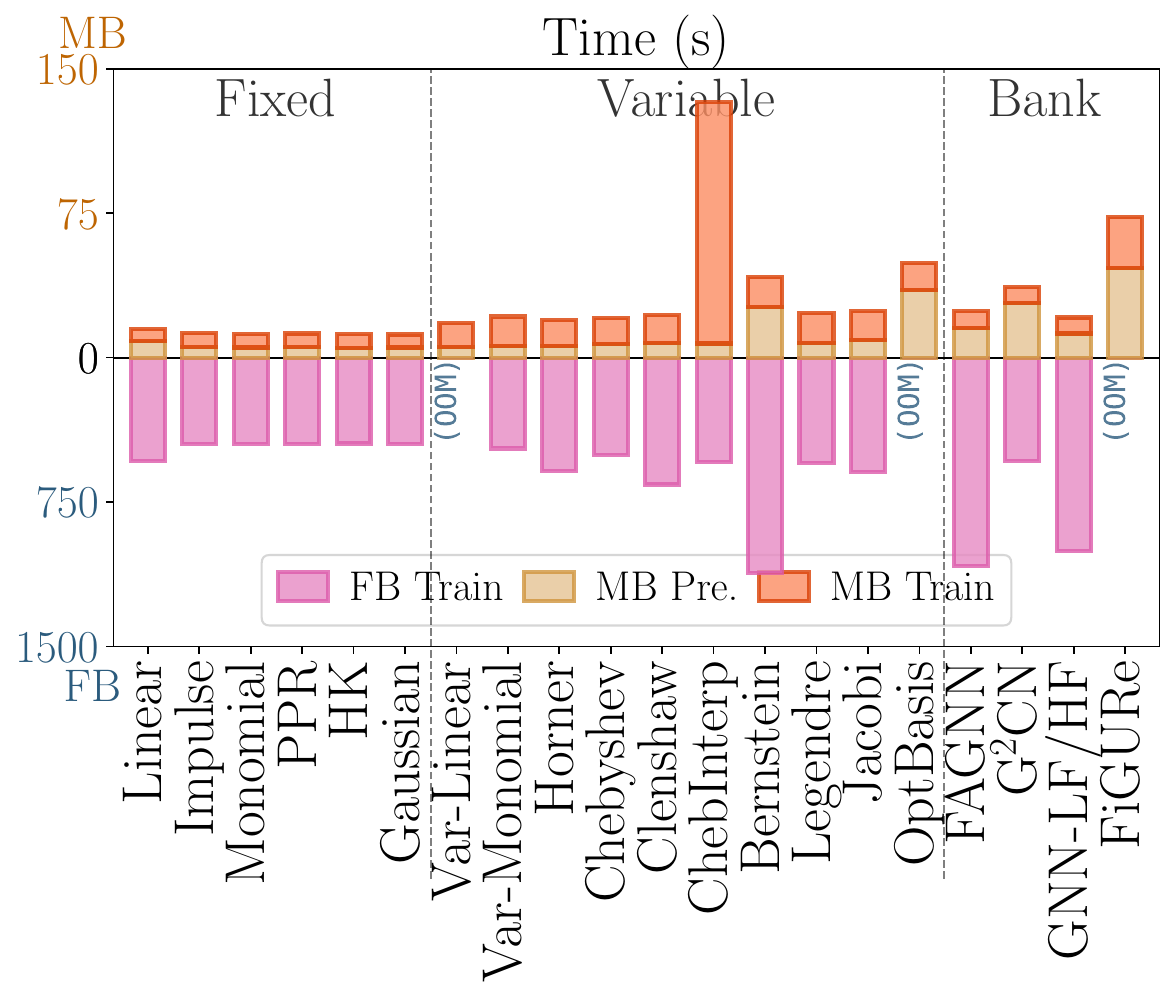} & \includegraphics[height=1.45in]{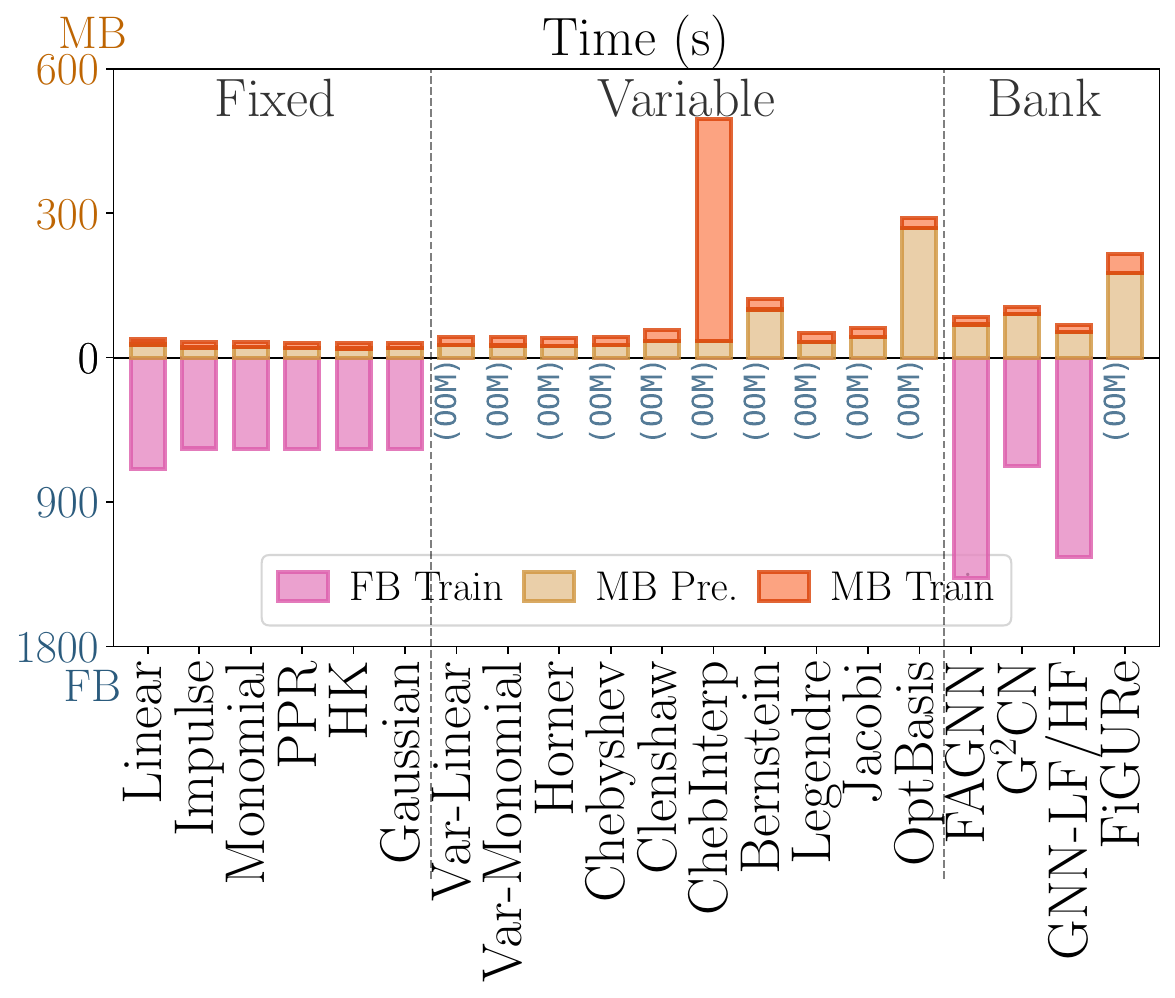} & \includegraphics[height=1.45in]{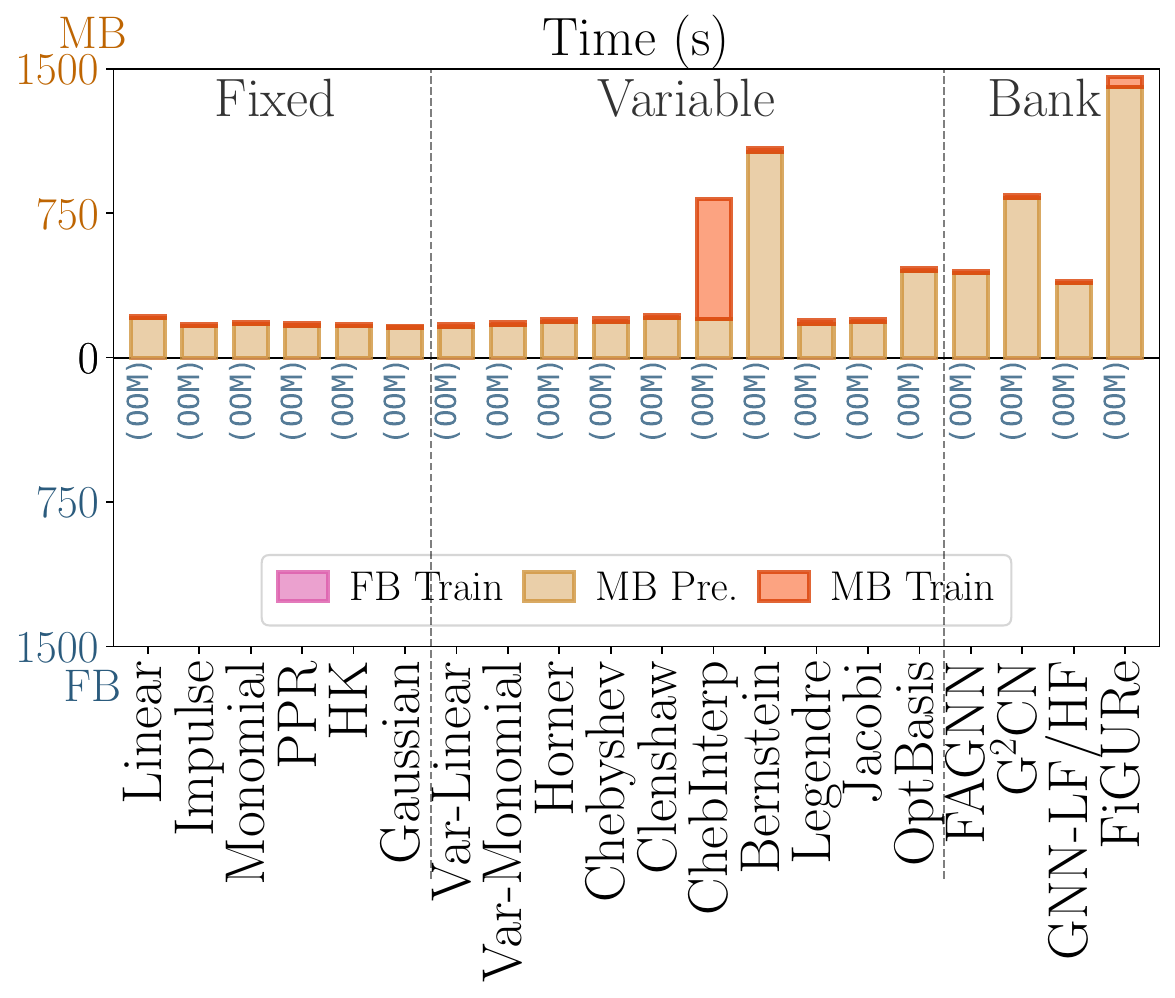} \\
    \includegraphics[height=1.45in]{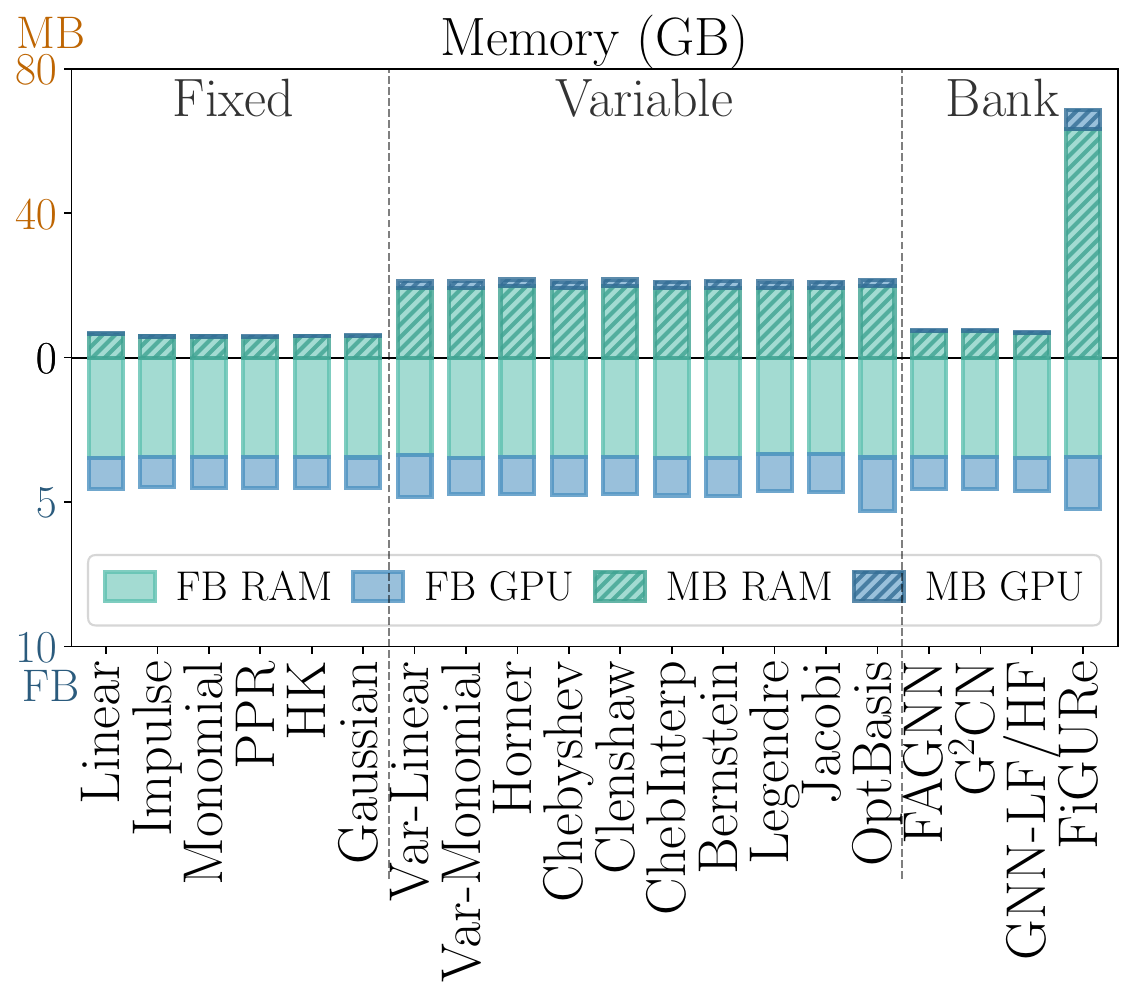} & \includegraphics[height=1.45in]{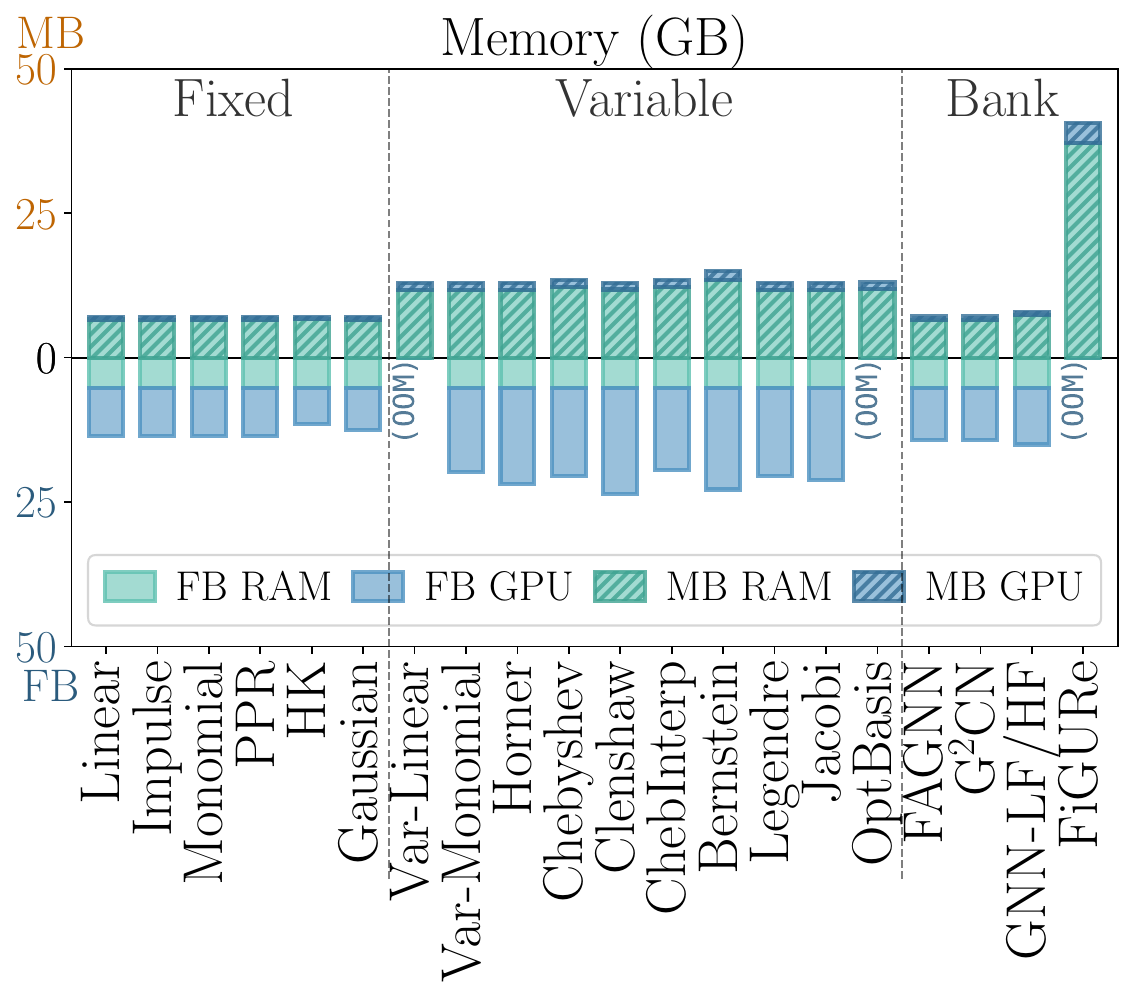} & \includegraphics[height=1.45in]{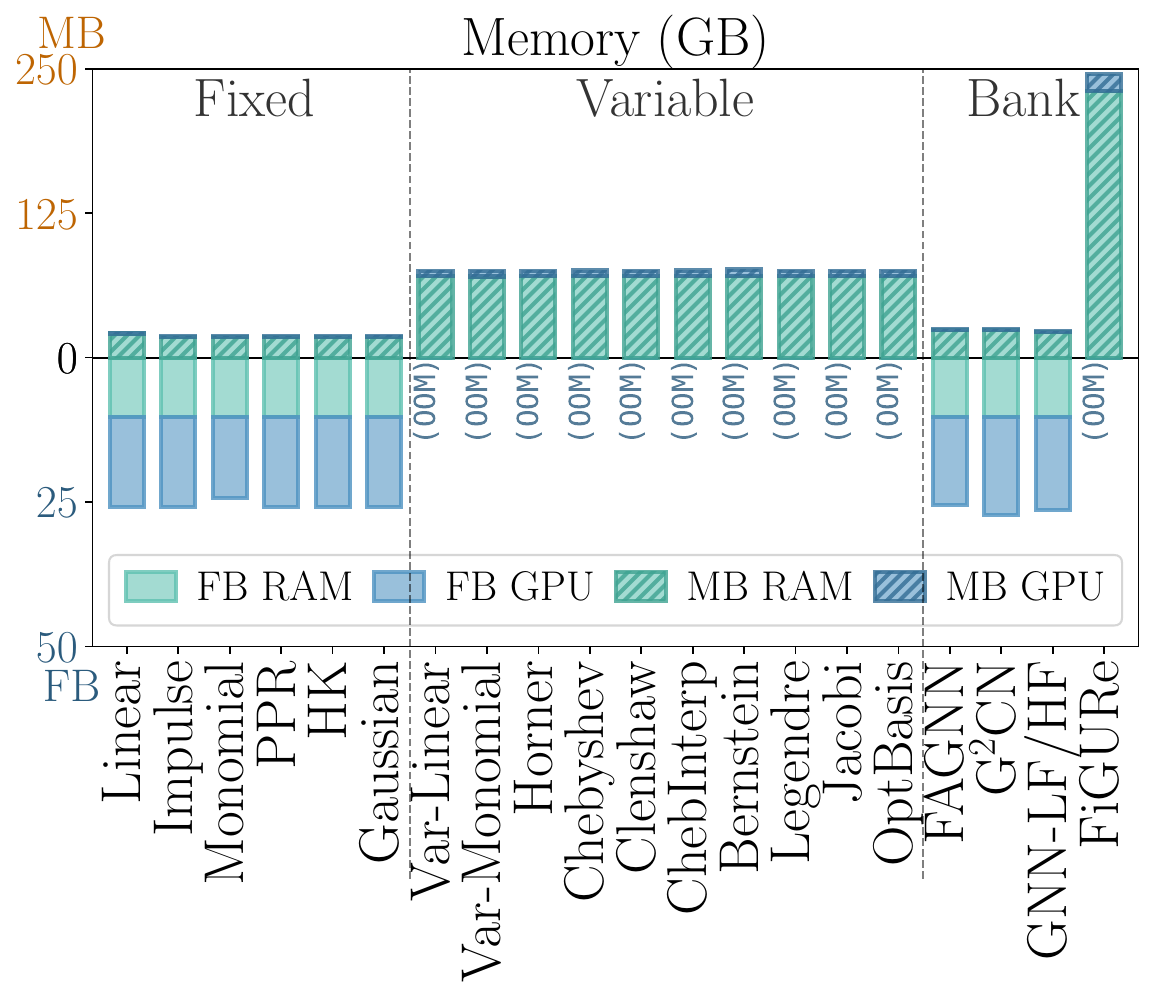} & \includegraphics[height=1.45in]{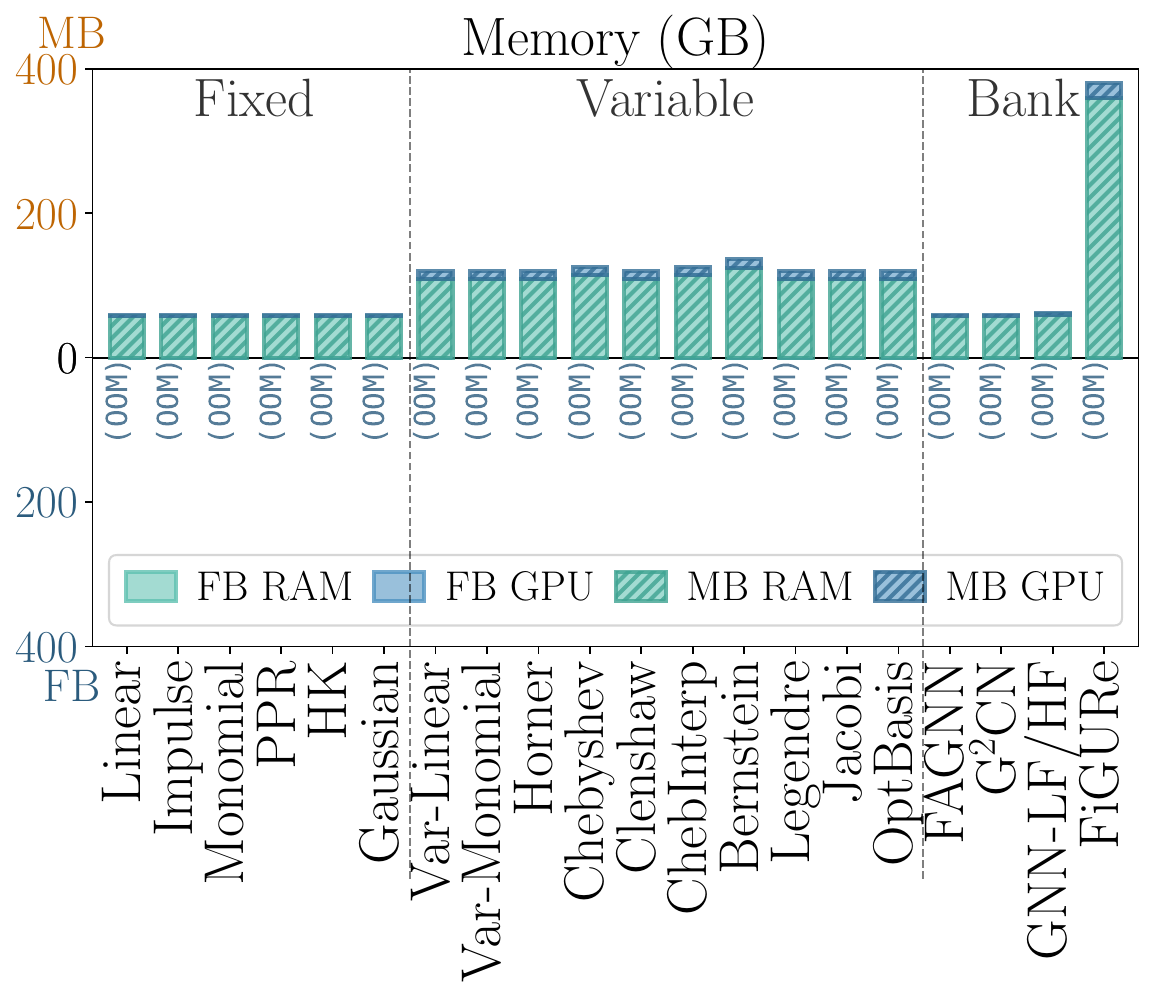} \\
    \textbf{(a) \ds{penn94}} & \textbf{(b) \ds{pokec}} & \textbf{(c) \ds{snap}} & \textbf{(d) \ds{wiki}} \\
\end{tabular}
\captionsetup{font={small}}
\caption{Comparison of filter time and memory efficiency under mini-batch (MB, upper axis) and full-batch (FB, lower axis) training on four medium- to large-scale datasets. Note that FB and MB axes may be on different scales. Empty bars are models with OOM error.}
\label{fig:eff_tm}
\end{figure*}

\vspace{-0.8em}
\subsection{Efficiency}
\label{ssec:main-efficiency}
Thanks to our unified implementation framework, the time and memory efficiency of spectral GNNs can be compared across different graph scales under fair criteria.
\cref{fig:eff_tm} provides a detailed breakdown of full- and mini-batch schemes, separating learning stages and devices for the time and memory evaluations, respectively. We respectively assess the performance from multiple perspectives of filter operations and learning schemes.



\subsubsection{Model Operations.}
\label{ob:efficiency}
~\begin{rquestionul}\label{rq:efficiency-filter}
What are the efficiency bottlenecks of filters under different graph scales?
\end{rquestionul}
As analyzed in \cref{ssec:model}, the key operations of graph propagation and weight transformation exhibit diverse computational overheads. Overall, it can be observed from \cref{fig:eff_tm} that filter efficiency can be effectively characterized by our taxonomy \cref{tab:summary}. The three types of filters exhibit distinct performances, and the empirical time and memory efficiency are in line with our complexity analysis.

Regarding specific operations, \textbf{variable \textit{transformations} come at the cost of additional computational resources}. In the FB scheme, this leads to reduced speed and increased GPU memory footprint. While some simple variable filters (\ft{Monomial}, \ft{Chebyshev}) and filter banks consisting of fixed filters (\ft{G$^2$CN}, \ft{GNN-LF/HF}) attain favorable performance when the graph scale is relatively small, models with complex transformations (\ft{Favard}, \ft{OptBasis}) experience prolonged training times and poor GPU memory scalability, and even result in out-of-memory (OOM) errors on large graphs. For MB training, the cost is reflected in increased RAM usage. Compared to fixed filters, memory footprint of variable designs is $K$ times larger for separately storing results in each hop. Similarly, a filter bank design with $Q$ variable filters demands up to $Q$ times more memory.

On the other hand, \textbf{\textit{propagation} becomes more computationally intensive on larger graphs}. In \cref{fig:eff_tm}, network transformation represented by MB training entails a large portion of time overhead on medium-scale graphs (\ds{penn94}), while propagation in precomputation dominates the MB training on larger graphs (\ds{pokec}, \ds{snap}).
Additional propagation further increases learning time but barely affects memory scalability, which is evident from variable filters with more graph-related propagations (\ft{ChebInterp}, \ft{Bernstein}). Nonetheless, thanks to our scalable implementation, their memory footprints are on par with other variable filters and are applicable to large-scale data even in memory-constrained environments.

\subsubsection{Learning Schemes.}~
\begin{rquestionul}\label{rq:efficiency-model}
What are the efficiency and scalability improvements brought by mini-batch training?
\end{rquestionul}
\cref{fig:eff_tm} implies that filter \textit{scalability}, i.e., whether filters can be applied to larger graphs, is mainly constrained by GPU memory capacity.
With FB, models typically encounter the OOM issue on datasets larger than \ds{ogbn-mag} and \ds{pokec} due to the representation storage proportional to node size $n$. In comparison, \textbf{MB training benefits scalability by shifting the memory overhead from GPU to RAM}, enabling the employment of memory-intensive filters (\ft{Favard}, \ft{OptBasis}, \ft{FiGURe}) on \ds{pokec} and even larger graphs. To the best of our knowledge, our implementation is the first attempt to successfully scale up an array of advanced spectral GNNs to these million-scale datasets and comprehensively characterize the performance.

Another advantage of MB lie in \textit{time efficiency}. Even though FB for some filters is applicable to some million-scale graphs (\ds{pokec}, \ds{snap}), the training time is excessively long. \textbf{MB achieves a significant $10-50\times$ speedup} by decoupling the computation-intensive propagation operations and simplifying the iterative training.
Moreover, \textbf{the speedup is mainly observable on large-scale datasets}, while FB and MB present the same level of overall time on medium-sized graphs (\ds{penn94}). This can be explained by \cref{rq:efficiency-filter}, as the increase in efficiency stems from faster propagation and is only effective when propagation is the bottleneck.


\begin{table*}[!t]
\captionsetup{font={small}}
\caption{Effectiveness results (\%) and standard deviations of spectral filters with full-batch training on available datasets. \rrc{For each dataset, results are highlighted based on the relative effectiveness among filters, where \rkoo{green} results are better.}
}
\label{res:acc}
    \setlength{\tabcolsep}{1.45pt}
    \renewcommand{\arraystretch}{0.95}
    \centering
\begin{adjustbox}{max width=\linewidth}
\begin{tabular}{@{}c@{}c|ccccc cccc| ccccc ccccc@{}}
\toprule
    Type & Filter & \ds{cora} & \ds{citeseer} & \ds{pubmed} & \ds{mine} & \ds{questions} & \ds{tolokers} & \ds{flickr} & \ds{arxiv} & \ds{mag} & \ds{chameleon} & \ds{squirrel} & \ds{actor} & \ds{roman} & \ds{ratings} & \ds{year} & \ds{penn94} & \ds{genius} & \ds{twitch} & \ds{pokec}
\\ \midrule
  \multirow{7}{*}{Fixed} 
	& \ft{Identity} & \tz{fee7d7}{72.37\tpm{2.02}} & \tz{fffceb}{72.24\tpm{1.51}} & \tz{fffdec}{87.74\tpm{0.62}} & \tz{fde4d5}{50.52\tpm{2.42}} & \tz{cee6d7}{71.29\tpm{1.81}} & \tz{fffbe9}{72.30\tpm{1.09}} & \tz{fff6e2}{46.83\tpm{0.30}} & \tz{fff8e5}{53.46\tpm{0.49}} & \tz{fffcea}{25.43\tpm{1.21}} & \tz{feebd9}{31.60\tpm{2.28}} & \tz{fff5e1}{29.18\tpm{6.07}} & \tz{f5fae6}{35.32\tpm{1.31}} & \tz{eff7e2}{64.64\tpm{0.72}} & \tz{f3f9e5}{41.87\tpm{0.54}} & \tz{fde4d5}{35.14\tpm{0.03}} & \tz{fffeee}{74.22\tpm{0.73}} & \tz{fffceb}{86.21\tpm{0.11}} & \tz{d0e8d8}{97.73\tpm{1.45}} & \tz{f8fce9}{62.21\tpm{0.07}} \\
	& \ft{Linear} & \tz{e9f4df}{86.14\tpm{1.15}} & \tz{fafdec}{74.07\tpm{1.91}} & \tz{fff0dd}{85.15\tpm{0.75}} & \tz{fff7e3}{68.50\tpm{1.49}} & \tz{cce5d7}{71.47\tpm{0.70}} & \tz{fffbe9}{72.74\tpm{1.55}} & \tz{fefef1}{50.55\tpm{0.46}} & \tz{e9f5df}{65.56\tpm{1.86}} & \tz{d9ecdb}{31.52\tpm{0.97}} & \tz{cce5d7}{40.66\tpm{2.70}} & \tz{f4fae5}{36.43\tpm{2.42}} & \tz{fee9d8}{26.88\tpm{1.43}} & \tz{feefdc}{37.60\tpm{0.68}} & \tz{f0f8e2}{42.17\tpm{0.42}} & \tz{fffeee}{44.52\tpm{2.04}} & \tz{fff5e1}{63.57\tpm{9.48}} & \tz{fffbe9}{85.71\tpm{3.76}} & \tz{fde5d6}{96.93\tpm{0.06}} & \tz{fde4d5}{51.59\tpm{1.54}} \\
	& \ft{Impulse} & \tz{fdfef0}{84.43\tpm{0.82}} & \tz{f8fce9}{74.24\tpm{2.19}} & \tz{fde4d5}{83.38\tpm{0.54}} & \tz{fff3df}{64.06\tpm{1.61}} & \tz{e2f1dd}{69.24\tpm{1.37}} & \tz{fde4d5}{56.75\tpm{2.13}} & \tz{fffeee}{49.92\tpm{0.23}} & \tz{f3f9e5}{63.84\tpm{1.58}} & \tz{f3f9e5}{29.32\tpm{1.57}} & \tz{f8fbe8}{37.57\tpm{3.71}} & \tz{fdfef0}{35.48\tpm{2.12}} & \tz{fde4d5}{25.67\tpm{1.13}} & \tz{fde4d5}{27.37\tpm{3.27}} & \tz{feefdc}{31.67\tpm{3.94}} & \tz{fff6e2}{40.68\tpm{0.46}} & \tz{fde4d5}{51.00\tpm{2.85}} & \tz{e1f0dd}{88.02\tpm{0.15}} & \tz{fde5d6}{96.93\tpm{0.06}} & \tz{fde4d5}{51.40\tpm{0.81}} \\
	& \ft{Monomial} & \tz{e6f3de}{86.28\tpm{1.72}} & \tz{e5f3de}{75.21\tpm{1.62}} & \tz{e9f5df}{89.07\tpm{0.49}} & \tz{fff8e5}{70.21\tpm{1.46}} & \tz{f5fae6}{66.61\tpm{6.57}} & \tz{ffffef}{76.51\tpm{1.36}} & \tz{ffffef}{50.27\tpm{0.73}} & \tz{fffded}{59.41\tpm{0.55}} & \tz{cce5d7}{32.41\tpm{0.57}} & \tz{daeddc}{39.82\tpm{3.38}} & \tz{dbeddc}{38.31\tpm{2.64}} & \tz{cce5d7}{37.84\tpm{1.65}} & \tz{f3f9e4}{63.44\tpm{1.48}} & \tz{cce5d7}{44.57\tpm{1.12}} & \tz{ffffef}{44.74\tpm{2.55}} & \tz{ffffef}{75.27\tpm{0.29}} & \tz{d3e9d9}{88.30\tpm{0.07}} & \tz{feeedb}{96.99\tpm{0.10}} & \tz{fefff1}{60.06\tpm{0.22}} \\
	& \ft{PPR} & \tz{e1f0dd}{86.58\tpm{1.96}} & \tz{cee7d8}{76.09\tpm{2.30}} & \tz{e8f4df}{89.12\tpm{0.39}} & \tz{fff7e2}{67.97\tpm{1.95}} & \tz{fffeee}{64.19\tpm{3.85}} & \tz{ffffef}{76.82\tpm{2.37}} & \tz{fffdec}{49.45\tpm{0.72}} & \tz{ecf6e0}{65.14\tpm{1.06}} & \tz{fffdec}{26.18\tpm{0.96}} & \tz{f1f8e3}{38.20\tpm{3.82}} & \tz{fbfded}{35.73\tpm{1.45}} & \tz{e0f0dd}{36.77\tpm{1.29}} & \tz{f1f8e3}{63.78\tpm{1.72}} & \tz{fffded}{39.38\tpm{0.49}} & \tz{fff8e4}{41.21\tpm{0.48}} & \tz{fffeee}{74.14\tpm{0.24}} & \tz{d2e9d9}{88.31\tpm{0.16}} & \tz{fffeee}{97.15\tpm{0.35}} & \tz{fffff2}{59.91\tpm{0.22}} \\
	& \ft{HK} & \tz{eaf5df}{86.02\tpm{1.53}} & \tz{f1f8e3}{74.62\tpm{1.96}} & \tz{e7f3de}{89.15\tpm{0.66}} & \tz{fffae8}{72.95\tpm{1.15}} & \tz{d4e9d9}{70.71\tpm{1.64}} & \tz{ffffef}{76.72\tpm{1.91}} & \tz{fff7e3}{47.15\tpm{0.06}} & \tz{cde6d7}{69.26\tpm{0.37}} & \tz{fff9e6}{23.53\tpm{1.04}} & \tz{d6ebda}{40.10\tpm{4.77}} & \tz{e2f1dd}{37.89\tpm{2.27}} & \tz{d1e8d9}{37.57\tpm{1.04}} & \tz{edf6e1}{65.08\tpm{1.08}} & \tz{f8fce9}{41.40\tpm{0.91}} & \tz{fffeee}{44.62\tpm{0.98}} & \tz{fffeee}{74.59\tpm{0.34}} & \tz{d3e9d9}{88.31\tpm{0.21}} & \tz{fee7d7}{96.94\tpm{0.08}} & \tz{f9fcea}{62.10\tpm{1.09}} \\
	& \ft{Gaussian} & \tz{e4f2de}{86.44\tpm{1.02}} & \tz{e2f1dd}{75.36\tpm{1.64}} & \tz{ecf6e0}{88.97\tpm{0.73}} & \tz{fffae7}{72.61\tpm{1.09}} & \tz{f7fbe8}{66.25\tpm{3.71}} & \tz{f9fcea}{78.32\tpm{1.69}} & \tz{fff6e2}{46.67\tpm{0.30}} & \tz{f4fae5}{63.72\tpm{0.83}} & \tz{fffff2}{27.87\tpm{3.45}} & \tz{fffeee}{36.38\tpm{4.57}} & \tz{f5fae6}{36.35\tpm{1.36}} & \tz{d5eada}{37.35\tpm{1.22}} & \tz{f0f8e3}{64.04\tpm{0.90}} & \tz{f5fae6}{41.68\tpm{0.40}} & \tz{f6fbe6}{46.11\tpm{1.94}} & \tz{ffffef}{74.64\tpm{0.17}} & \tz{e5f2de}{87.93\tpm{0.21}} & \tz{dceedc}{97.63\tpm{0.06}} & \tz{fffff2}{59.95\tpm{0.06}} \\
\midrule
  \multirow{11}{*}{Variable} 
	& \ft{Linear} & \tz{eff7e2}{85.72\tpm{1.46}} & \tz{e6f3de}{75.17\tpm{1.95}} & \tz{fff2df}{85.50\tpm{0.69}} & \tz{fffae7}{72.41\tpm{1.21}} & \tz{e4f2de}{68.97\tpm{3.02}} & \tz{fffae8}{72.04\tpm{6.40}} & \tz{fefef1}{50.53\tpm{0.10}} & \tz{f4fae5}{63.69\tpm{1.22}} & \tz{dfefdd}{31.11\tpm{0.60}} & \tz{cce5d7}{40.66\tpm{3.57}} & \tz{e5f3de}{37.61\tpm{2.27}} & \tz{fee9d8}{26.81\tpm{1.71}} & \tz{fee8d7}{31.16\tpm{0.86}} & \tz{d8ecdb}{43.83\tpm{1.36}} & \tz{f0f8e3}{46.58\tpm{0.63}} & \tz{fffcea}{70.81\tpm{0.09}} & \tz{cde6d7}{88.42\tpm{0.11}} & \tz{fffff2}{97.18\tpm{0.49}} & (OOM) \\
	& \ft{Monomial} & \tz{dceedc}{86.85\tpm{0.65}} & \tz{cce5d7}{76.18\tpm{1.86}} & \tz{eaf5df}{89.05\tpm{0.77}} & \tz{d1e8d9}{89.26\tpm{0.80}} & \tz{fcfeee}{65.29\tpm{4.63}} & \tz{f5fae6}{78.95\tpm{0.72}} & \tz{f0f8e3}{52.23\tpm{1.30}} & \tz{eaf5df}{65.47\tpm{1.37}} & \tz{d2e8d9}{32.03\tpm{1.38}} & \tz{d6ebda}{40.10\tpm{3.84}} & \tz{eef7e1}{36.94\tpm{1.97}} & \tz{f9fcea}{34.95\tpm{1.64}} & \tz{f0f8e3}{64.16\tpm{1.68}} & \tz{fffded}{39.50\tpm{1.38}} & \tz{f5fae6}{46.19\tpm{0.50}} & \tz{ffffef}{75.23\tpm{0.99}} & \tz{cde6d7}{88.42\tpm{0.06}} & \tz{fde4d5}{96.91\tpm{0.07}} & \tz{fffded}{59.10\tpm{1.49}} \\
	& \ft{Horner} & \tz{e4f2de}{86.39\tpm{2.06}} & \tz{e0f0dd}{75.44\tpm{1.83}} & \tz{d1e8d9}{89.71\tpm{0.48}} & \tz{d6ebda}{88.53\tpm{1.06}} & \tz{fffeee}{64.01\tpm{1.32}} & \tz{fdfeef}{77.40\tpm{0.69}} & \tz{e9f4df}{52.94\tpm{0.46}} & \tz{f8fce9}{62.94\tpm{2.40}} & \tz{eff7e2}{29.71\tpm{0.55}} & \tz{cde6d7}{40.59\tpm{3.09}} & \tz{fffdec}{33.74\tpm{1.24}} & \tz{cfe7d8}{37.67\tpm{1.43}} & \tz{ffffef}{58.27\tpm{2.38}} & \tz{f3f9e5}{41.82\tpm{1.81}} & \tz{d6ebda}{48.43\tpm{0.48}} & \tz{fffeee}{74.56\tpm{0.39}} & \tz{cce5d7}{88.43\tpm{0.99}} & \tz{fde5d6}{96.93\tpm{0.06}} & \tz{cce5d7}{72.02\tpm{0.82}} \\
	& \ft{Chebyshev} & \tz{ecf6df}{85.95\tpm{1.13}} & \tz{e2f1dd}{75.38\tpm{2.13}} & \tz{e7f3de}{89.14\tpm{0.61}} & \tz{cce5d7}{89.92\tpm{0.78}} & \tz{fffcea}{62.65\tpm{4.21}} & \tz{f6fbe7}{78.85\tpm{0.83}} & \tz{cce5d7}{55.11\tpm{0.14}} & \tz{e4f2de}{66.31\tpm{0.61}} & \tz{e2f1dd}{30.90\tpm{1.42}} & \tz{fff5e1}{33.71\tpm{3.30}} & \tz{ecf6e0}{37.11\tpm{2.39}} & \tz{eff7e2}{35.76\tpm{1.03}} & \tz{e2f1dd}{67.92\tpm{2.22}} & \tz{fffeed}{39.70\tpm{1.17}} & \tz{fffbe9}{42.91\tpm{1.42}} & \tz{d0e7d8}{84.31\tpm{0.27}} & \tz{f7fbe7}{87.47\tpm{0.03}} & \tz{fde5d6}{96.93\tpm{0.06}} & \tz{feefdc}{54.26\tpm{0.43}} \\
	& \ft{Clenshaw} & \tz{ffffef}{83.90\tpm{1.98}} & \tz{ffffef}{73.63\tpm{1.83}} & \tz{f9fceb}{88.49\tpm{2.85}} & \tz{d5eada}{88.70\tpm{1.33}} & \tz{fffcea}{62.49\tpm{5.24}} & \tz{f1f8e3}{79.66\tpm{0.54}} & \tz{eef7e1}{52.46\tpm{0.38}} & \tz{fff0dd}{47.25\tpm{10.56}} & \tz{fffae7}{24.06\tpm{5.66}} & \tz{fde5d6}{30.62\tpm{2.91}} & \tz{fffeee}{34.66\tpm{0.56}} & \tz{eaf5df}{36.11\tpm{1.95}} & \tz{d8ecdb}{70.37\tpm{3.27}} & \tz{ffffef}{40.44\tpm{3.20}} & \tz{fffceb}{43.72\tpm{2.00}} & \tz{dceedc}{82.55\tpm{0.33}} & \tz{f9fcea}{87.40\tpm{0.12}} & \tz{feedda}{96.98\tpm{0.04}} & \tz{feebd9}{53.02\tpm{0.29}} \\
	& \ft{ChebInterp} & \tz{fffeee}{83.52\tpm{5.59}} & \tz{fff2de}{66.96\tpm{6.76}} & \tz{d0e8d8}{89.72\tpm{0.87}} & \tz{d0e7d8}{89.43\tpm{0.86}} & \tz{eaf5df}{68.15\tpm{5.93}} & \tz{eef7e1}{80.27\tpm{0.91}} & \tz{cee6d7}{54.97\tpm{2.55}} & \tz{f9fcea}{62.81\tpm{1.84}} & \tz{fff9e6}{23.25\tpm{3.75}} & \tz{fde4d5}{30.41\tpm{4.08}} & \tz{fcfeee}{35.59\tpm{1.74}} & \tz{eff7e2}{35.78\tpm{2.29}} & \tz{dceedc}{69.38\tpm{1.98}} & \tz{fffae7}{36.70\tpm{0.14}} & \tz{f1f8e3}{46.50\tpm{0.06}} & \tz{deefdd}{82.26\tpm{1.37}} & \tz{fde4d5}{79.65\tpm{7.86}} & \tz{fde5d6}{96.93\tpm{0.07}} & \tz{dceedc}{69.15\tpm{0.10}} \\
	& \ft{Bernstein} & \tz{fde4d5}{71.03\tpm{2.49}} & \tz{fde4d5}{62.21\tpm{2.39}} & \tz{feead9}{84.24\tpm{0.97}} & \tz{dfefdd}{87.38\tpm{1.12}} & \tz{fffcea}{62.51\tpm{2.89}} & \tz{ffffef}{76.36\tpm{1.24}} & \tz{fff9e6}{47.95\tpm{0.56}} & \tz{fff3df}{48.86\tpm{0.73}} & \tz{fde4d5}{13.83\tpm{3.31}} & \tz{fffae7}{34.97\tpm{3.41}} & \tz{fafdeb}{35.84\tpm{1.81}} & \tz{f9fceb}{34.90\tpm{0.97}} & \tz{fff8e4}{46.77\tpm{2.18}} & \tz{fde4d5}{27.56\tpm{2.78}} & \tz{fffbe9}{43.08\tpm{0.21}} & \tz{e3f2de}{81.36\tpm{0.56}} & \tz{fbfdec}{87.32\tpm{0.10}} & \tz{fde4d5}{96.92\tpm{0.06}} & \tz{cce5d7}{72.03\tpm{0.21}} \\
	& \ft{Legendre} & \tz{cce5d7}{87.68\tpm{0.96}} & \tz{eff7e2}{74.75\tpm{2.13}} & \tz{d0e8d8}{89.72\tpm{0.36}} & \tz{cce5d7}{89.86\tpm{0.76}} & \tz{f7fbe7}{66.36\tpm{2.00}} & \tz{f5fae6}{79.04\tpm{2.09}} & \tz{fffbe9}{48.86\tpm{0.92}} & \tz{fffdec}{59.04\tpm{1.84}} & \tz{fffbe9}{25.00\tpm{2.69}} & \tz{fffded}{36.17\tpm{2.48}} & \tz{eef7e1}{36.94\tpm{2.39}} & \tz{cfe7d8}{37.69\tpm{1.43}} & \tz{fefff1}{59.00\tpm{6.53}} & \tz{e7f3de}{42.90\tpm{0.22}} & \tz{fffeee}{44.64\tpm{3.46}} & \tz{cee7d8}{84.43\tpm{0.51}} & \tz{ecf6df}{87.79\tpm{0.71}} & \tz{fee9d8}{96.95\tpm{0.04}} & \tz{fffae8}{57.76\tpm{0.21}} \\
	& \ft{Jacobi} & \tz{d9ecdb}{87.02\tpm{0.87}} & \tz{e4f2de}{75.25\tpm{2.28}} & \tz{dff0dd}{89.35\tpm{0.95}} & \tz{cfe7d8}{89.48\tpm{0.80}} & \tz{ebf5df}{68.08\tpm{1.57}} & \tz{f9fceb}{78.13\tpm{2.62}} & \tz{fffcea}{49.09\tpm{0.55}} & \tz{fde4d5}{38.92\tpm{2.73}} & \tz{f9fceb}{28.58\tpm{1.78}} & \tz{f9fcea}{37.43\tpm{4.48}} & \tz{fffeee}{34.86\tpm{1.74}} & \tz{fff6e2}{30.28\tpm{1.91}} & \tz{eef7e1}{64.77\tpm{1.74}} & \tz{ddeedc}{43.58\tpm{0.98}} & \tz{d6ebda}{48.38\tpm{0.07}} & \tz{cce5d7}{84.77\tpm{0.48}} & \tz{f0f8e2}{87.67\tpm{0.27}} & \tz{ffffef}{97.16\tpm{0.35}} & \tz{fff7e2}{56.15\tpm{3.20}} \\
	& \ft{Favard} & \tz{e9f4df}{86.14\tpm{1.60}} & \tz{fffeee}{73.29\tpm{2.77}} & \tz{e5f3de}{89.18\tpm{1.00}} & \tz{cee6d7}{89.74\tpm{0.79}} & \tz{fff5e1}{58.86\tpm{4.25}} & \tz{fffdec}{74.51\tpm{4.90}} & \tz{fde4d5}{42.26\tpm{0.00}} & \tz{f0f8e2}{64.53\tpm{0.35}} & \tz{dceedc}{31.29\tpm{0.95}} & \tz{fee6d6}{30.83\tpm{5.11}} & \tz{f3f9e5}{36.54\tpm{2.69}} & \tz{ecf6e0}{35.97\tpm{1.58}} & \tz{d5eada}{70.82\tpm{0.86}} & \tz{fbfded}{40.98\tpm{3.44}} & \tz{dceedc}{48.05\tpm{0.20}} & \tz{eef7e1}{79.59\tpm{9.46}} & \tz{fffeee}{86.82\tpm{0.24}} & \tz{d7ebdb}{97.67\tpm{0.66}} & (OOM) \\
	& \ft{OptBasis} & \tz{fffae7}{80.74\tpm{2.03}} & \tz{fff8e5}{69.65\tpm{2.84}} & \tz{e2f1dd}{89.28\tpm{0.75}} & \tz{d5eada}{88.73\tpm{0.51}} & \tz{feeedb}{56.31\tpm{7.84}} & \tz{f0f8e2}{79.93\tpm{1.25}} & \tz{feefdc}{44.81\tpm{1.21}} & \tz{fffdec}{58.67\tpm{1.92}} & (OOM) & \tz{fffcea}{35.53\tpm{2.54}} & \tz{fffded}{34.13\tpm{1.09}} & \tz{deefdd}{36.88\tpm{1.13}} & \tz{cce5d7}{72.76\tpm{1.06}} & \tz{f5fae6}{41.65\tpm{0.73}} & \tz{cce5d7}{48.98\tpm{0.08}} & \tz{e7f3de}{80.72\tpm{4.70}} & \tz{ebf5df}{87.80\tpm{0.69}} & \tz{f2f9e4}{97.39\tpm{0.44}} & (OOM) \\
\midrule
  \multirow{9}{*}{Bank} 
	& \ft{AdaGNN} & \tz{f2f9e4}{85.45\tpm{1.25}} & \tz{f3f9e5}{74.52\tpm{2.09}} & \tz{cce5d7}{89.82\tpm{0.65}} & \tz{e8f4df}{85.83\tpm{4.37}} & \tz{eff7e2}{67.48\tpm{1.86}} & \tz{f1f8e3}{79.65\tpm{1.29}} & \tz{e2f1dd}{53.55\tpm{0.75}} & \tz{fffae7}{54.62\tpm{2.20}} & \tz{fdfef0}{28.06\tpm{5.25}} & \tz{fffdec}{36.10\tpm{4.05}} & \tz{d5eada}{38.62\tpm{2.19}} & \tz{e0f0dd}{36.77\tpm{1.23}} & \tz{eff7e2}{64.62\tpm{0.89}} & \tz{fffded}{39.39\tpm{1.48}} & \tz{f2f9e4}{46.41\tpm{0.58}} & \tz{f9fceb}{77.02\tpm{1.94}} & \tz{eef7e1}{87.73\tpm{0.36}} & \tz{ffffef}{97.17\tpm{0.42}} & \tz{f2f9e4}{63.99\tpm{1.00}} \\
	& \ft{FBGNNI} & \tz{fffbe9}{81.43\tpm{5.82}} & \tz{ffffef}{73.56\tpm{1.36}} & \tz{fffae8}{87.06\tpm{0.85}} & \tz{cee7d8}{89.60\tpm{3.13}} & \tz{fff9e6}{61.14\tpm{3.56}} & \tz{ffffef}{76.63\tpm{2.09}} & \tz{fafdec}{51.08\tpm{0.66}} & \tz{f4fae5}{63.73\tpm{3.91}} & (OOM) & \tz{fffceb}{35.88\tpm{3.40}} & \tz{fffcea}{33.20\tpm{2.22}} & \tz{fcfdee}{34.66\tpm{1.72}} & \tz{fbfded}{60.35\tpm{1.52}} & \tz{ffffef}{40.36\tpm{1.71}} & \tz{e2f1dd}{47.70\tpm{0.48}} & \tz{fffae7}{68.67\tpm{2.35}} & \tz{f1f8e3}{87.62\tpm{0.31}} & \tz{dceedc}{97.63\tpm{0.14}} & (OOM) \\
	& \ft{FBGNNII} & \tz{fffbe9}{81.18\tpm{4.04}} & \tz{fffeee}{73.08\tpm{2.08}} & \tz{ffffef}{88.12\tpm{0.91}} & \tz{f0f8e3}{84.37\tpm{3.70}} & \tz{fffcea}{62.72\tpm{5.60}} & \tz{fdfeef}{77.31\tpm{1.48}} & \tz{eaf5df}{52.84\tpm{1.12}} & \tz{e1f0dd}{66.76\tpm{3.93}} & (OOM) & \tz{d8ecdb}{39.97\tpm{4.30}} & \tz{fffdec}{34.05\tpm{1.66}} & \tz{ecf6e0}{36.02\tpm{1.18}} & \tz{fffff2}{58.82\tpm{2.97}} & \tz{e6f3de}{42.91\tpm{0.41}} & \tz{fefff1}{45.14\tpm{2.53}} & \tz{fffceb}{72.22\tpm{0.35}} & \tz{feefdc}{82.11\tpm{0.36}} & \tz{cce5d7}{97.77\tpm{0.24}} & (OOM) \\
	& \ft{ACMGNNI} & \tz{f7fbe7}{85.12\tpm{1.98}} & \tz{f1f8e3}{74.62\tpm{1.90}} & \tz{f4fae5}{88.70\tpm{0.94}} & \tz{e6f3de}{86.21\tpm{4.93}} & \tz{feeedb}{56.29\tpm{7.46}} & \tz{edf6e1}{80.30\tpm{1.85}} & \tz{fbfdec}{51.04\tpm{1.07}} & \tz{cce5d7}{69.41\tpm{0.51}} & (OOM) & \tz{e9f5df}{38.83\tpm{4.15}} & \tz{fde4d5}{22.78\tpm{6.72}} & \tz{f0f8e2}{35.73\tpm{0.81}} & \tz{fffded}{55.79\tpm{2.70}} & \tz{f0f8e2}{42.13\tpm{0.25}} & \tz{fffcea}{43.46\tpm{2.03}} & \tz{fffdec}{72.50\tpm{0.27}} & \tz{fffbe9}{85.73\tpm{1.86}} & \tz{fde5d6}{96.93\tpm{0.06}} & (OOM) \\
	& \ft{ACMGNNII} & \tz{ffffef}{84.16\tpm{4.09}} & \tz{fcfeee}{73.95\tpm{2.11}} & \tz{e6f3de}{89.16\tpm{0.54}} & \tz{e6f3de}{86.23\tpm{3.50}} & \tz{fde4d5}{52.93\tpm{5.17}} & \tz{cce5d7}{84.51\tpm{1.09}} & \tz{eaf5df}{52.83\tpm{1.65}} & \tz{fff9e6}{54.49\tpm{2.39}} & (OOM) & \tz{fffae8}{35.04\tpm{4.34}} & \tz{fffeee}{34.69\tpm{1.42}} & \tz{fffbe9}{32.50\tpm{3.91}} & \tz{fffdec}{55.20\tpm{2.08}} & \tz{fffeee}{40.04\tpm{2.83}} & \tz{ecf6e0}{46.94\tpm{0.37}} & \tz{fffeee}{73.96\tpm{0.30}} & \tz{f8fbe8}{87.44\tpm{0.37}} & \tz{fff8e5}{97.08\tpm{0.25}} & (OOM) \\
	& \ft{FAGNN} & \tz{f2f9e4}{85.45\tpm{0.85}} & \tz{deefdd}{75.53\tpm{1.58}} & \tz{fffbe9}{87.19\tpm{1.36}} & \tz{fffbe9}{73.98\tpm{2.10}} & \tz{f3f9e4}{66.95\tpm{4.62}} & \tz{fffded}{74.99\tpm{1.20}} & \tz{fffeee}{50.06\tpm{1.75}} & \tz{dceedc}{67.51\tpm{0.40}} & \tz{e3f2de}{30.78\tpm{1.89}} & \tz{f7fbe7}{37.71\tpm{2.79}} & \tz{fefff1}{35.34\tpm{1.62}} & \tz{fee7d7}{26.50\tpm{1.77}} & \tz{fff1de}{39.50\tpm{0.36}} & \tz{e1f0dd}{43.28\tpm{0.77}} & \tz{d9ecdb}{48.25\tpm{0.28}} & \tz{fff9e5}{67.15\tpm{2.28}} & \tz{dceedc}{88.13\tpm{0.16}} & \tz{fff7e2}{97.06\tpm{0.21}} & \tz{feedda}{53.55\tpm{0.28}} \\
	& \ft{G$^2$CN} & \tz{f7fbe7}{85.12\tpm{3.32}} & \tz{dfefdd}{75.50\tpm{1.44}} & \tz{f5fae6}{88.67\tpm{0.69}} & \tz{fff7e2}{67.98\tpm{1.24}} & \tz{e4f2de}{68.95\tpm{2.30}} & \tz{fafdec}{77.91\tpm{1.95}} & \tz{f4fae5}{51.83\tpm{0.21}} & \tz{d0e7d8}{68.96\tpm{0.92}} & \tz{eff7e2}{29.72\tpm{2.17}} & \tz{d0e7d8}{40.45\tpm{2.75}} & \tz{fffeee}{34.63\tpm{1.12}} & \tz{eff7e2}{35.76\tpm{1.22}} & \tz{f8fce9}{61.45\tpm{1.89}} & \tz{d0e7d8}{44.35\tpm{1.35}} & \tz{d8ecdb}{48.29\tpm{0.46}} & \tz{f8fce9}{77.26\tpm{0.65}} & \tz{d0e7d8}{88.37\tpm{0.27}} & \tz{eef7e1}{97.43\tpm{0.15}} & \tz{feefdc}{54.13\tpm{0.04}} \\
	& \ft{GNN-LF/HF} & \tz{ddeedc}{86.83\tpm{0.98}} & \tz{deefdd}{75.53\tpm{1.67}} & \tz{eaf5df}{89.03\tpm{0.66}} & \tz{dceedc}{87.80\tpm{0.77}} & \tz{fff8e4}{60.29\tpm{3.55}} & \tz{f9fcea}{78.27\tpm{2.02}} & \tz{eff7e2}{52.34\tpm{0.19}} & \tz{ffffef}{60.97\tpm{0.32}} & \tz{ffffef}{27.63\tpm{0.48}} & \tz{fefff1}{36.80\tpm{2.40}} & \tz{f8fce9}{36.04\tpm{2.04}} & \tz{fffff2}{34.29\tpm{2.07}} & \tz{f5fae6}{62.68\tpm{3.63}} & \tz{fffceb}{38.72\tpm{0.04}} & \tz{fff8e4}{41.28\tpm{0.26}} & \tz{fffeee}{74.54\tpm{0.68}} & \tz{eaf5df}{87.81\tpm{0.11}} & \tz{fff2de}{97.02\tpm{0.14}} & \tz{f6fbe6}{63.06\tpm{0.23}} \\
	& \ft{FiGURe} & \tz{d9ecdb}{87.04\tpm{0.99}} & \tz{f4fae5}{74.49\tpm{1.71}} & \tz{ffffef}{88.22\tpm{0.95}} & \tz{cee6d7}{89.71\tpm{0.56}} & \tz{fcfdee}{65.38\tpm{8.16}} & \tz{f1f8e3}{79.76\tpm{1.39}} & \tz{e8f4df}{53.02\tpm{0.65}} & \tz{dbeddc}{67.57\tpm{0.40}} & \tz{f2f9e4}{29.43\tpm{2.18}} & \tz{fffae7}{34.90\tpm{1.63}} & \tz{cce5d7}{39.16\tpm{1.24}} & \tz{fffdec}{33.45\tpm{1.09}} & \tz{f4fae5}{62.84\tpm{1.45}} & \tz{f5fae6}{41.69\tpm{0.34}} & \tz{fffded}{44.14\tpm{3.99}} & \tz{d6ebda}{83.30\tpm{0.31}} & \tz{d9ecdb}{88.19\tpm{0.83}} & \tz{fffeee}{97.15\tpm{0.38}} & (OOM) \\
\bottomrule
\end{tabular}
\end{adjustbox}
\end{table*}

By summarizing \cref{rq:efficiency-filter,rq:efficiency-model}, we conclude our \textbf{guidelines for employing mini-batch training}: Firstly, MB is particularly suitable for filters without variable parameters, showcasing faster computation and a smaller memory footprint thanks to better device utilization.
Another ideal use case arises when FB for complex filters is prohibitive due to memory constraints. MB is capable of conducting graph filtering operations on CPU and storing full-scale representations in RAM at the cost of increased CPU/RAM usage.
Even so, since GPU memory often has stricter constraints on practical platforms, MB with GPU memory complexity independent of the graph scale is favorable for deploying spectral GNNs across a wider range of environments.

\subsection{Effectiveness}
\label{ssec:main-effect}
Our benchmark specifically features the filter effectiveness from the varied ancillary settings, aiming to demystify their capability in processing graph signals with fair comparison. \cref{res:acc} displays the filter accuracy with decoupled architecture and full-batch training. In particular, \ft{Identity} represents the baseline without graph information. Our results align with previous evaluations involving spectral GNNs such as \cite{lim2021,olegplatonov2023,bo2023b} and are mostly comparable to those in the original papers listed in \cref{tab:summary}.

\subsubsection{Filter-level Comparison.}
\label{ob:effective}
As an overview, our results suggest that filter efficacy depends on graph patterns, and no single filter consistently achieves dominant accuracy across all scenarios. In this section, we analyze the effectiveness among the filter variants, particularly focusing on two aspects: filters under varying degrees of heterophily and filters across different categories.

\begin{rquestionul}\label{rq:effective-hetero}
How do the graph pattern of homophily/heterophily affect filter effectiveness?
\end{rquestionul}
As introduced in \cref{ssec:spectral}, under \textit{homophily}, simple filters such as \ft{Linear} can leverage graph inductive bias in low-frequency components and benefit effectiveness over \ft{Identity} \cite{gao2023}.
On conventional homophilous graphs (\ds{cora} and \ds{pubmed}), it is common that a large part of filters can achieve top-tier accuracy within the error interval. This observation indicates that the graph signal is relatively easy to learn, and is in line with other recent GNN evaluations \cite{luo2024,sancak2024}. With proper settings, \textbf{even conventional fixed filters such as \ft{PPR} and \ft{HK} can deliver satisfying performance}.
Contrarily, variable filters and filter banks (\ft{AdaGNN} and \ft{FiGURe}) are superior on \ds{tolokers} and \ds{minesweeper}. Although these graphs are also classified as homophilous, we deduce their graph patterns are more complex than simple low-frequency clustering, and thus demands more adaptive filters.

The scenario of \textit{heterophily} is more challenging, as the local graph structure does not assist graph learning and entails advanced filters featuring high-frequency spectral signals, i.e., filters emphasizing $\theta_k$ for larger $k$-s under our polynomial formulation \cref{eq:sgnn}. Empirical results in \cref{res:acc} support the design intuitive, as low-pass filters (\ft{Linear}, \ft{Impulse}) fail on heterophilous graphs, sometimes performing even worse than \ft{Identity}. \textbf{Filters incorporating high-pass components are effective in learning under heterophily}. This guideline applies to all filter categories: for fixed filters, this can be achieved by increasing the hop number $K$ (\ft{Monomial}) or reducing the decay factor $\alpha$ (\ft{PPR}, \ft{Gaussian}). For variable filters, it is usually beneficial to select bases that facilitate learning for high-frequency parameters (\ft{Horner}, \ft{Bernstein}) \cite{luan2024}.
Meanwhile, there are also critiques that some variable bases, including \ft{Monomial} and \ft{Bernstein}, prioritize heterophilous accuracy and sacrifice their capability under homophily \cite{luan2024}.

\begin{rquestionul}\label{rq:effective-taxa}
What is the impact of variable and filter bank designs on filter effectiveness?
\end{rquestionul}
In our taxonomy \cref{tab:summary}, we identify the different designs of filter parameterization, typically discriminating fixed, variable, and filter bank designs. Our observation in \cref{res:acc} suggests that, \textbf{different designs do not guarantee improved accuracy on \textit{certain} graphs, but may benefit generality across \textit{various} datasets}.
In specific, the simple and fast \textit{fixed} filters sufficiently achieve top-tier accuracy on a number of datasets, as demonstrated in \cref{rq:effective-hetero}.
\textit{Variable} filters offer a more flexible approach for approximating a wider range of the graph spectrum. This allows for dynamically learning weight parameters from input signals, which is advantageous for capturing and leveraging richer information in the frequency domain. In cases where graph information is useful, these filters empirically produce better effectiveness, although the level of improvement varies with different data distributions.
The \textit{filter bank} design alternatively enhances model capacity by adopting multiple filters, usually spanning diverse frequency ranges. This approach is effective in mitigating the failure of a single filter and ensures reasonable accuracy in broad scenarios. On the other hand, this design does not necessarily outperform a single filter since there is no inherent improvement in filter expressiveness.

\subsubsection{Relation to Efficiency.}
Then, we relate filter effectiveness and efficiency to deliver a comprehensive discussion on choosing proper filters. We first investigate the impact of learning schemes in \cref{rq:effective-model}, then discuss the relationship between effectiveness and efficiency in our core research question \cref{rq:balance}.

\begin{figure}[!t]
\captionsetup{font={stretch=0.98,small}}
\centering
    \includegraphics[width=0.9\linewidth]{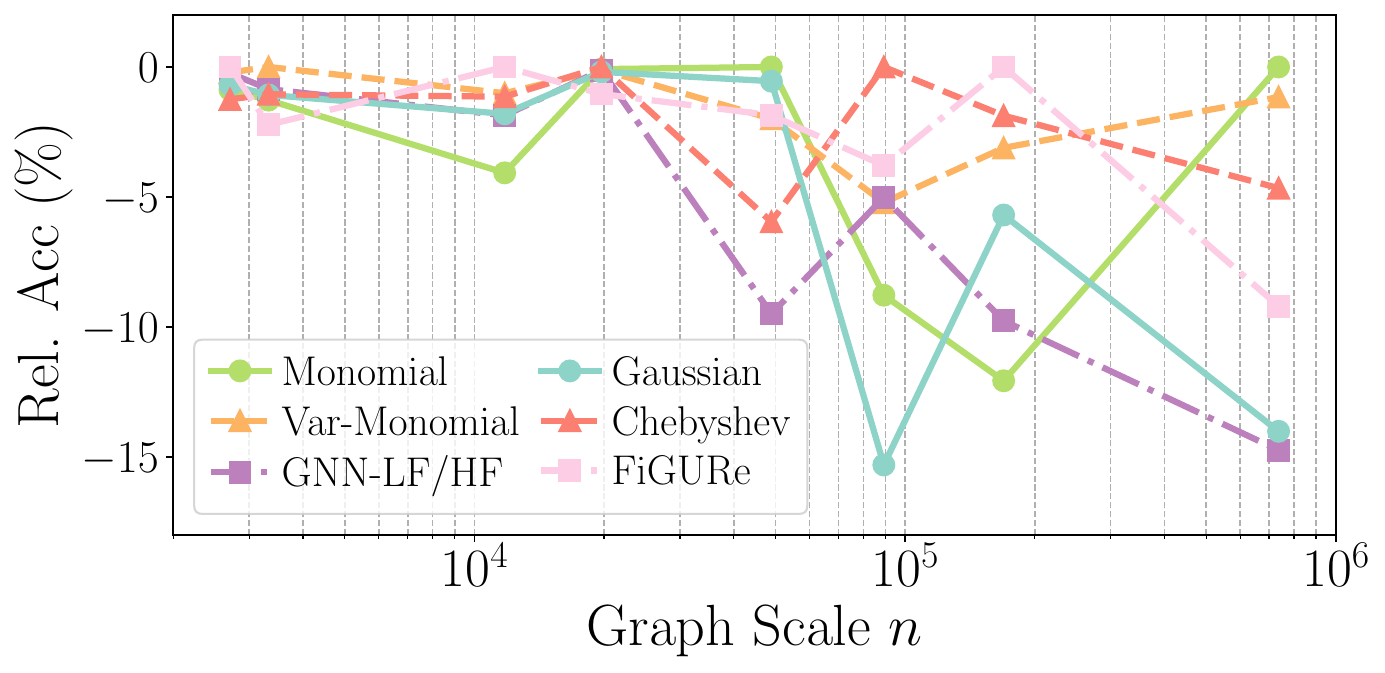}
    \caption{Shift of filter effectiveness on homophilous datasets across different scales. Effectiveness ($y$-axis) is presented by the relative accuracy to the highest filter in each dataset. The graph scale ($x$-axis) is presented by node size $n$ in log scale.}
    \label{fig:balance}
\end{figure}

\begin{rquestionul}\label{rq:effective-model}
How do full-batch and mini-batch training schemes affect model efficacy?
\end{rquestionul}
\textit{Theoretically}, mini-batch computation is identical to the full-batch scheme from the aspect of spectral operations, and the only difference roots in the feature transformation procedure, as MB models do not perform the pre-transformation on input attributes before applying graph filters.
\textit{Empirically}, the majority results in \cref{resa:acc_mini} confirm that \textbf{MB training delivers comparable accuracy to the corresponding FB results} in \cref{res:acc}, and our key observations \cref{rq:effective-hetero,rq:effective-taxa} on filter effectiveness still hold. It also supports our motivation for extracting and benchmarking spectral filters, as they are generally applicable to different training schemes without affecting the capability of GNN learning.

Meanwhile, accuracy drops of MB can be observed on graphs with a small attribute dimension $F_{i}$ (\ds{minesweeper}, \ds{tolokers}, \ds{ratings}). This defect can be explained by the over-squashing phenomenon \cite{alon2021}, which stems from the information loss when encoding the comprehensive graph topology into a small $F_{i}$ during the separate filtering process. Variable filters are relatively prone to this issue since their filter parameters are not sufficiently trained under these circumstances.
In contrast, MB on heterophilous datasets such as \ds{chameleon} and \ds{roman} achieves higher accuracy than full-batch models. This is precisely the opposite outcome of over-squashing, where malignant graph information is implicitly alleviated by the spectral filtering process on raw attributes.

\begin{rquestionul}\label{rq:balance}
What is the relationship between filter effectiveness and efficiency? Moreover, how to choose spectral filters that are both effective and efficient?
\end{rquestionul}
A prevailing trend in spectral GNN studies favors more sophisticated filter designs for better effectiveness, which, as indicated by \cref{rq:efficiency-filter}, compromise time and memory efficiency. However, \cref{rq:effective-hetero} demonstrates that these two techniques lead to distinctive outcomes compared with fixed filters: Improving filter variability through \textit{learnable parameters} can expand model capability in some cases, though the impact largely depends on the usefulness of graph signals. The design also incurs additional efficiency overhead related to the feature transformation.
The \textit{filter bank} approach usually contributes to overall filter adaptability across various graph signals, rather than improving accuracy of existing filters. This comes at the cost of multiplying the computation time and memory by the number of filters involved.

In other words, different from the common view of a straightforward trade-off between efficacy and efficiency, \textbf{our benchmark study uncovers a more intricate nature: these two aspects are not mutually exclusive}. The filter effectiveness is determined by its \textit{inherent frequency response}. Even with the same variability and complexity, different filters yield varied accuracy, and filters appropriate to the graph signals results in better accuracy.
Alternatively, time and memory efficiency are relevant to the \textit{external designs} of graph computation and feature transformations, which can be explicitly inferred from their complexity. For instance, when processing heterophilous graph topology, a fixed high-order filter is more likely to achieve superior performance compared to a filter with a large parameter size focusing on low-frequency components.

\begin{figure}[!b]
\captionsetup{font={stretch=0.96,small}}
    \centering
    \subcaptionbox{\ds{cora}\label{ffig:box_cora}}%
    [0.49\columnwidth]{\includegraphics[height=1.46in]{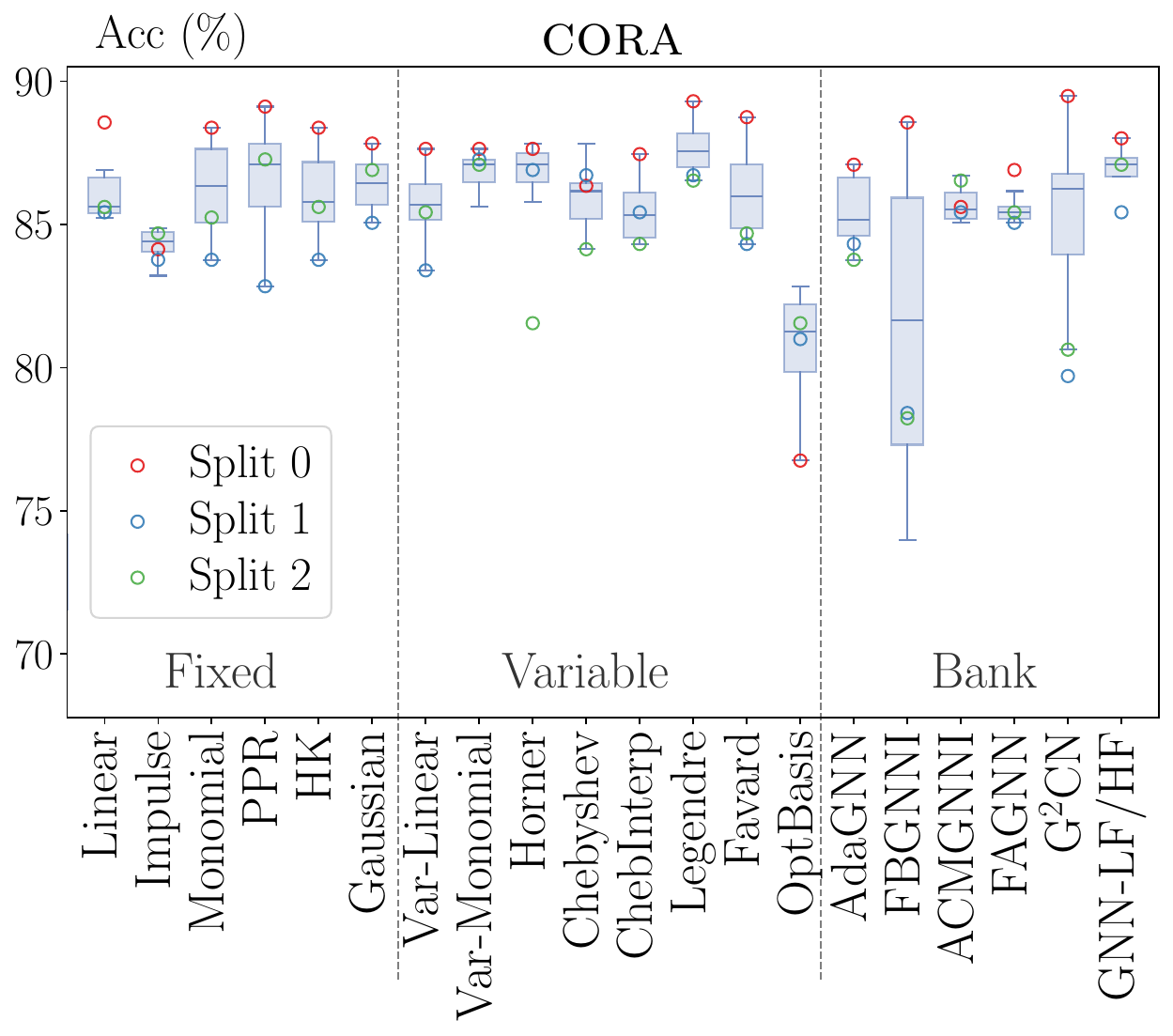}}
    \hfil
    \subcaptionbox{\ds{arixv}\label{ffig:box_ogbn-arxiv}}%
    [0.49\columnwidth]{\includegraphics[height=1.46in]{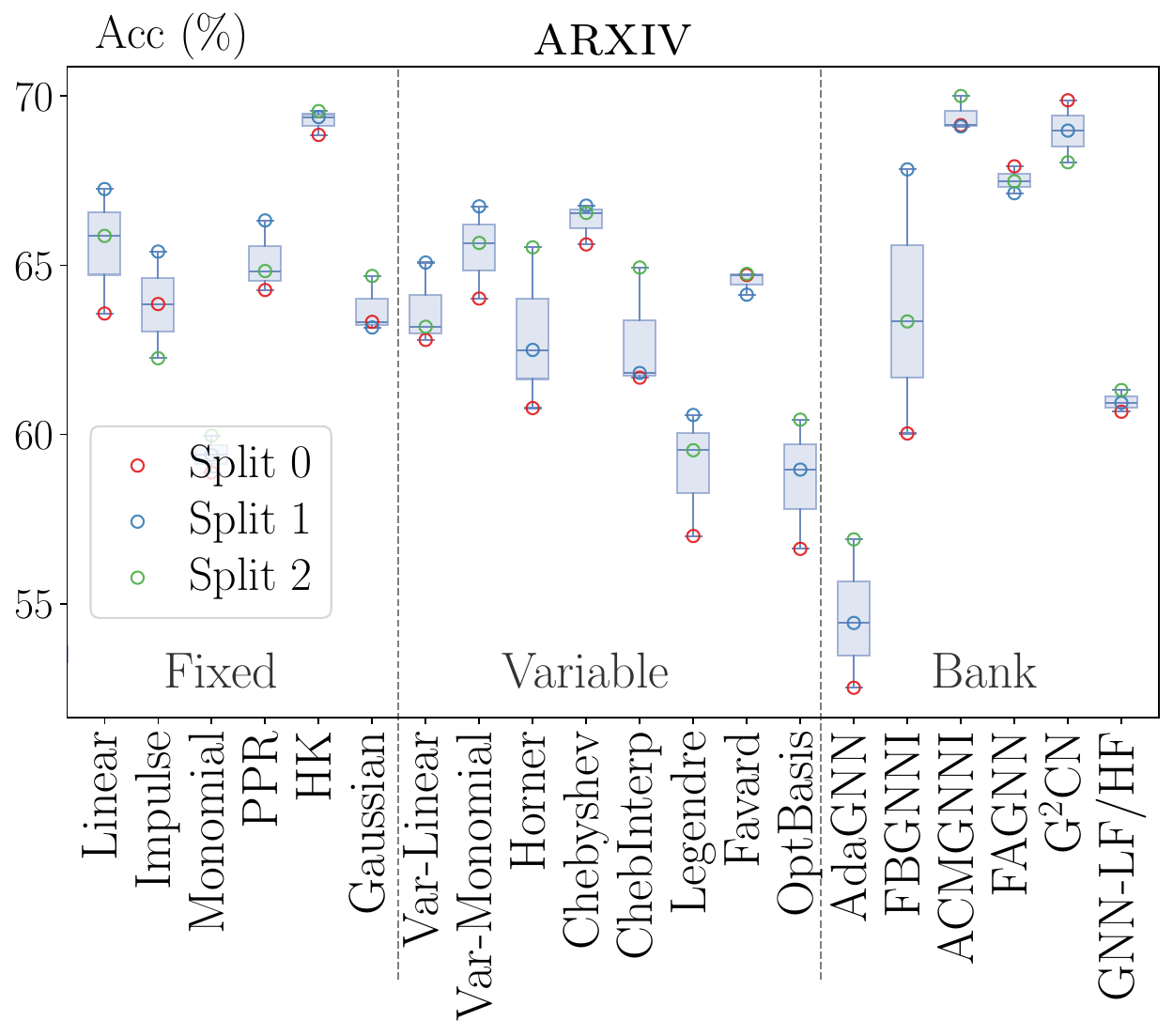}}
    \caption{Statistical significance of filter effectiveness. Parameter initialization and dataset split for all filters follow the same set of random seeds. Results from three seeds are marked in the figure.}
  \label{fig:box}
\end{figure}

Thus, balancing effectiveness and efficiency preferably requires a comprehensive consideration of graph knowledge and available environments. It is more important to find a suitable spectral expression by examining the particular graph input, instead of simply exploiting sophisticated designs. Significant factors affecting learning efficacy and spectral expressiveness are thus explored more specifically in \cref{sec:experiment-extra}.
\textbf{We suggest the following practice as an attempt toward effective and efficient spectral GNNs}: When learning on simple and homophilous graphs, one can stick to fixed filters for the best efficiency and comparably superb precision. For tasks with complex graph signals, or when fixed filters are inadequate, it is recommended to carefully choose an appropriate model that fits the graph spectrum and producing satisfactory performance, while ensuring that the time and memory overheads remain feasible in the given environment.

\subsection{Result Stability}
\label{ssec:main-robust}
In this section, we especially investigate the statistical significance of our evaluation on filter effectiveness and efficiency, ensuring that our conclusions are widely applicable to general circumstances.

\subsubsection{Efficacy Variance and Divergence.}
\label{ssec:main-var}
We further visualize the confidence intervals of filter accuracy in \blue{\cref{res:acc,resa:acc_mini}} by box plot in \cref{fig:box} with selected seeds. Particularly, \ds{cora} and \ds{arxiv} represent the datasets with random and attribute-based splits, respectively, and all filters learn on the same split for the same seed. \cref{ffig:box_cora} implies that the variance in datasets like \ds{cora} is largely caused by the split difference, as some seeds lead to high accuracy for most filters while others greatly impede efficacy. This is a common phenomenon in semi-supervised learning, where random splits may not be representative of containing sufficient information, such as lacking minor label groups, which results in large accuracy deviation. On \ds{arxiv}, where the splits are more stable, the filter accuracy is more concentrated. Nonetheless, in both cases, \textbf{the relative effectiveness among filters can be effectively depicted by average accuracy} \blue{while certain outliers are caused by less representative data splits}, which supports our \cref{rq:effective-hetero,rq:effective-taxa} based on \cref{res:acc}. \blue{Results between FB and MB are also supports the observation in \cref{rq:effective-model} that MB largely maintains accuracy, while being more unstable on graphs with low-dimensional attributes such as \ds{minesweeper}.}

Our scalable implementation offer an unprecedented opportunity to study filter efficacy across graphs with different scales, which is showcased in \cref{fig:balance} with representatives in three categories. Together with \cref{fig:box}, we observe that \textbf{the relative difference among filters is more significant on larger graphs}. On small-scale graphs (\cref{ffig:box_cora}), the accuracy difference among filters are marginal, and a broad range of filters can achieve high accuracy. However, filter effectiveness becomes more divergent on larger graphs where $n>10^5$ (\cref{ffig:box_ogbn-arxiv}). While filters with suitable frequency responses, e.g. the fixed \ft{Monomial}, can still achieve leading performance on corresponding graphs, inappropriate ones exhibit larger accuracy gaps from the best filters. The finding further underscores the importance of our conclusion \cref{rq:effective-hetero} especially on large-scale graphs, that choosing filters that fit the graph spectrum is critical for effectiveness.

\begin{figure}[!t]
    \centering
    \includegraphics[width=0.88\columnwidth]{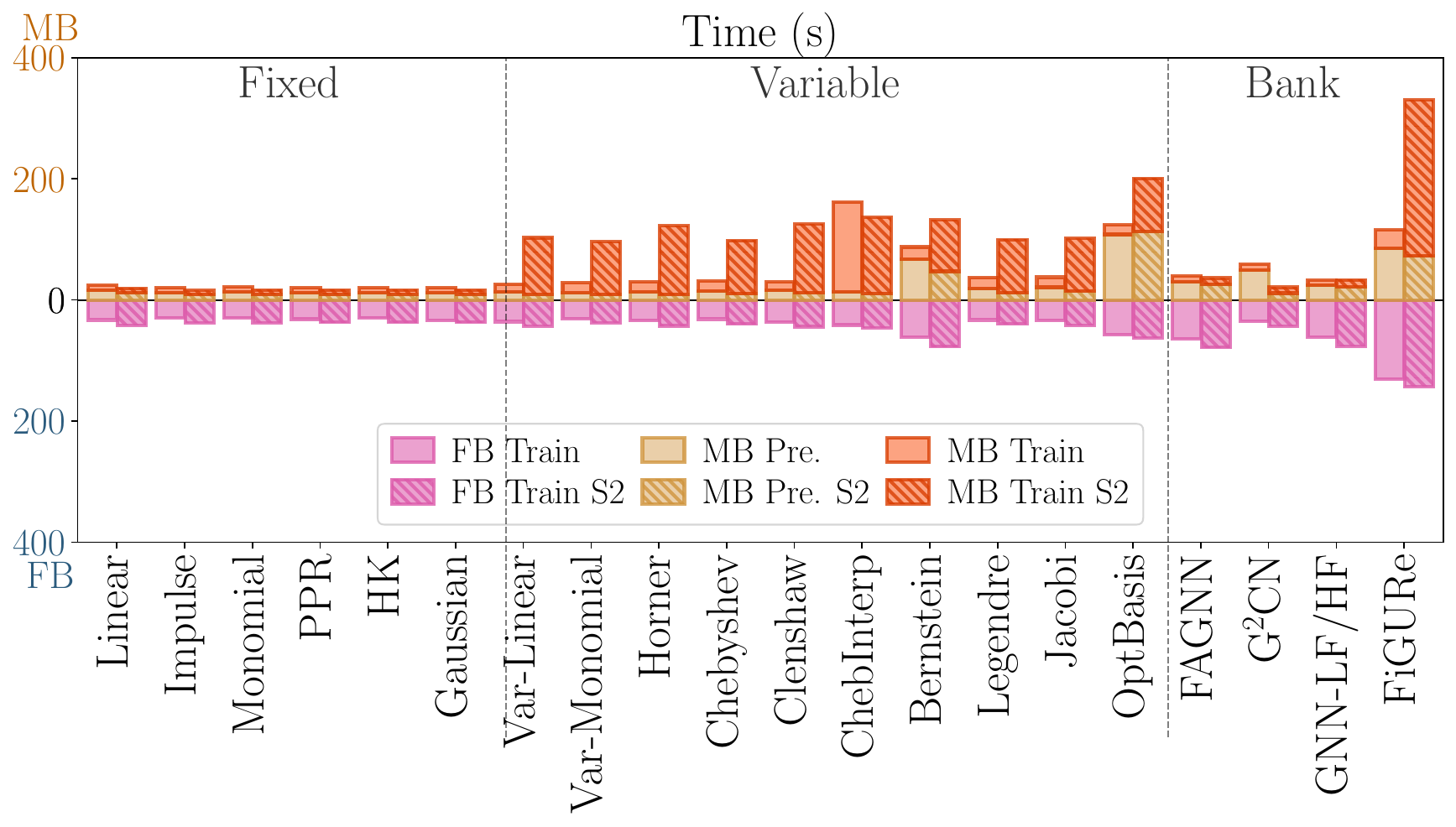}
\captionsetup{font={small}}
    \caption{Time efficiency comparison of FB (lower axis) and MB (upper axis) training on \ds{penn94} with different hardware.}
  \label{fig:plat}
\vspace{-0.6em}
\end{figure}

\subsubsection{Efficiency on Different Hardware.}
\label{ssec:main-hardware}
To validate our efficiency observations on diverse hardware platforms, we evaluate the filter efficiency on another server marked as S2, which is with slower CPUs (Intel Xeon, 2.2GHz) and a faster GPU (NVIDIA RTX A5000). The time breakdown on the typical dataset \ds{penn94} is in \cref{fig:plat}.
Notably, the figure demonstrates \textbf{varied bottlenecks for different filter types} due to the hardware difference. For MB fixed filters with transformation being the bottleneck, the overall learning time on S2 is shorter thanks to faster GPU computation. In comparison, as graph propagation dominates the efficiency of FB training and MB variable filters, the empirical speed is relatively slower. The slowdown is more significant on datasets larger than \ds{penn94}. The observation aligns with \cref{rq:efficiency-filter,rq:efficiency-model} regarding the efficiency of model operations and learning schemes, verifying that they are generally applicable irrespective of hardware settings.

\subsection{Key Conclusions}
\label{ssec:main-ob}
\begin{enumerate}[label=C\arabic*.]
    \item Model time and memory efficiency are respectively dominated by graph propagation and weight transformation operations as indicated by our taxonomy. On graphs above million-scale, propagation turns into the bottleneck of GNN training. (\cref{rq:efficiency-filter})
    \item Mini-batch training is a unique learning scheme of spectral GNN models, offering comparable efficacy and better scalability, especially on large graphs. In turn, it falls short in flexibility and is more sensitive to raw graph attributes. (\cref{rq:effective-model,rq:efficiency-model})
    \item Model effectiveness mainly stems from the inherit filter property of frequency response, i.e., the emphasis on low- and high-frequency signals. With proper configurations, simple filters can also excel on suitable graphs. (\cref{rq:effective-hetero})
    \item While sophisticated filter designs are favored in common consensus, they are less related to accuracy improvement, but potentially benefit generality across datasets. Variable filters increase the capacity to deal complex input signals, while filter banks can assemble the performance of individual filters. (\cref{rq:effective-taxa})
    \item Contrary to the prevailing belief, we reveal that efficacy and efficiency are not mutually exclusive. We provide our practical guidelines for analyzing the graph spectrum and choosing appropriate filters to achieve both efficacy and efficiency. (\cref{rq:balance})
\end{enumerate}

\begin{table}[!b]
\captionsetup{font={stretch=0.9}}
\begin{adjustbox}{max width=\columnwidth}
\begin{threeparttable}
\caption{Effectiveness and efficiency of models outside our framework on representative datasets. ``Train''  and ``Infer'' respectively refer to average training time per epoch and inference time ($\si{s}$), while precomputation time is separated in applicable models.
}
\label{res:pyg}
\centering
\newcommand{\dsh}[2]{\multicolumn{4}{c#1}{#2}}
\renewcommand{\arraystretch}{0.9}
\setlength{\tabcolsep}{3pt}
\begin{tabular}{c|cccc|cccc} \toprule
  \multirow{2}{*}{Model (Backend$^\ast$)}
    & \dsh{|}{\ds{arxiv}} & \dsh{}{\ds{mag}} \\
   ~ & \dshh{Acc} & \dshh{Train} & \dshh{Infer} & \dshh{GPU}
  & \dshh{Acc} & \dshh{Train} & \dshh{Infer} & \dshh{GPU}
\\\midrule
    GCN (SP) & 53.2 & 0.05 & 0.03 & 1.1 & 27.4 & 0.31 & 0.11 & 7.8 \\
    GraphSAGE (SP) & 50.3 & 0.04 & 0.005 & 1.2 & 24.8 & 0.25 & 0.01 & 7.7 \\
    GCN (EI) & 53.0 & 0.06 & 0.06 & 3.2 & \dsh{}{(OOM)} \\
    GraphSAGE (EI) & 54.1 & 0.09 & 0.03 & 2.7 & 28.2 & 0.33 & 0.18 & 10.8 \\
    ChebNet (EI) & 53.4 & 0.11 & 0.05 & 3.0 & \dsh{}{(OOM)} \\
    NAGphormer (EI) & 67.8 & 280+3.2 & 2.1 & 2.3 & 33.2 & 89+10.3 & 2.2 & 3.8 \\
    ANS-GT (EI) & 71.1 & 16205+109 & 2.7 & 11.3 & \dsh{}{(OOM)} \\
\toprule
    Model (Backend) & \dsh{|}{\ds{penn94}} & \dsh{}{\ds{pokec}} \\\midrule
    GCN (SP) & 73.1 & 0.04 & 0.03 & 1.3 & 60.4 & 663.3 & 0.35 & 0.01 \\
    GraphSAGE (SP) & 74.2 & 0.08 & 0.003 & 2.3 & 63.4 & 0.45 & 0.009 & 9.6 \\
    GCN (EI) & 67.6 & 0.06 & 0.06 & 3.7 & \dsh{}{(OOM)} \\
    GraphSAGE (EI) & \dsh{|}{(OOM)} & \dsh{}{(OOM)} \\
    ChebNet (EI) & \dsh{|}{(OOM)} & \dsh{}{(OOM)} \\
    NAGphormer (EI) & 74.4 & 237+6.1 & 2.1 & 2.3 & 73.1 & 70+16.1 & 3.1 & 8.9 \\
    ANS-GT (EI) & 67.8 & 34092+37 & 5.0 & 8.7 & \dsh{}{(OOM)} \\
\bottomrule
\end{tabular}
\begin{tablenotes}
    \item [$\ast$] \textbf{Backend}: SP = \texttt{torch.sparse}, EI = \texttt{torch\_geometric.EdgeIndex} \cite{pyg}.
\end{tablenotes}
\end{threeparttable}
\end{adjustbox}
\end{table}

\section{Results: Specific Evaluations}
\label{sec:experiment-extra}
In this section, we conduct in-depth evaluations regarding specific properties of individual spectral filters, offering ancillary view on obtaining desired filter effectiveness and efficiency. Further research questions are raised for evaluations leading to new findings.

\subsection{Extended Tasks}
\label{ssec:extra-extend}
In this section, we highlight the generalizability of our implementation and evaluation by comparing to other graph processing methods and performing tasks other than node classification.

\subsubsection{Implementation Comparison.}
\label{ssec:extra-model}
To demonstrate the performance of our implementation, especially regarding GNN efficiency and scalability, \cref{res:pyg} presents the additional evaluation for typical GNN models deployed in other popular frameworks. The baselines include spatial message-passing GNNs (GraphSAGE \cite{hamilton2017inductive}), spectral message-passing GNNs (GCN \cite{kipf2016semi}, ChebNet \cite{defferrard2016convolutional}), as well as scalable Graph Transformers (NAGphormer \cite{chen2023c}, ANS-GT \cite{zhang2022ansgt}). The models are implemented with available graph backends from the PyG framework \cite{pyg}. Especially, the EI backend is the most common backend used in PyG by default entailing $O(mF)$ space footprint for graph propagation. The SP backend is more efficient and used as the default backend in our main experiments, while only a handful of PyG models are compatible with it.

It can be observed that most accuracy are in line with our results in \cref{res:acc}. Regarding efficiency and scalability, message-passing models with the SP backend is more superior than EI for faster training and less GPU memory footprint, while the most common EI backend encounters OOM on large datasets. Graph Transformers are more computational intensive for demanding significantly long precomputation, slower training speed, and more memory overhead due to the complicated model architecture.


\subsubsection{Link Prediction.}
\label{ssec:extra-link}
Link prediction is another popular task that presents substantial challenges regarding GNN scalability \cite{hu2021ogblsc,li2023c}. Compared to the node classification task in \cref{sec:experiment-main}, link prediction focuses on transformation operations, as identifying node-pair correlation through neural networks is critical for prediction accuracy. Graph information retrieved by the filters only serve as fundamental knowledge for the downstream transformation. Different from \cref{ssec:model}, the typical complexity for transformation is $O(\kappa mF^2)$. This is because for a graph with $m$ ground-truth edges, the model needs to processes a total of $\kappa m$ positive and negative samples, where $\kappa$ is usually $2-10$. Hence, performing the \textbf{link prediction task inevitably entails mini-batch training}, as the full-scale memory overhead is prohibitive.


As the effectiveness of link prediction is mainly determined by transformation networks which are orthogonal to this study, we mainly stick to a simple MLP network and focus on efficiency evaluations shown in \cref{fig:lp_eff}, while more sophisticated downstream modules can be integrated in a plug-and-play manner as introduced in \cref{sec:design}.
Compared to \cref{rq:efficiency-filter}, link prediction \textbf{efficiency is dominated by the transformation operation on larger graphs} due to the considerable amount of iterative edge-wise computations, while GPU memory can be controlled by the batch size. Nonetheless, such computational bottleneck is less related to the topic of graph processing, and a range of dedicated accelerations are available \cite{zhang2018link,SUREL,yu2024}.

\begin{figure}[!t]
    \centering
    \subcaptionbox{Time\label{ffig:timelp_ogbl-ppa}}%
    [0.49\columnwidth]{\includegraphics[height=1.25in]{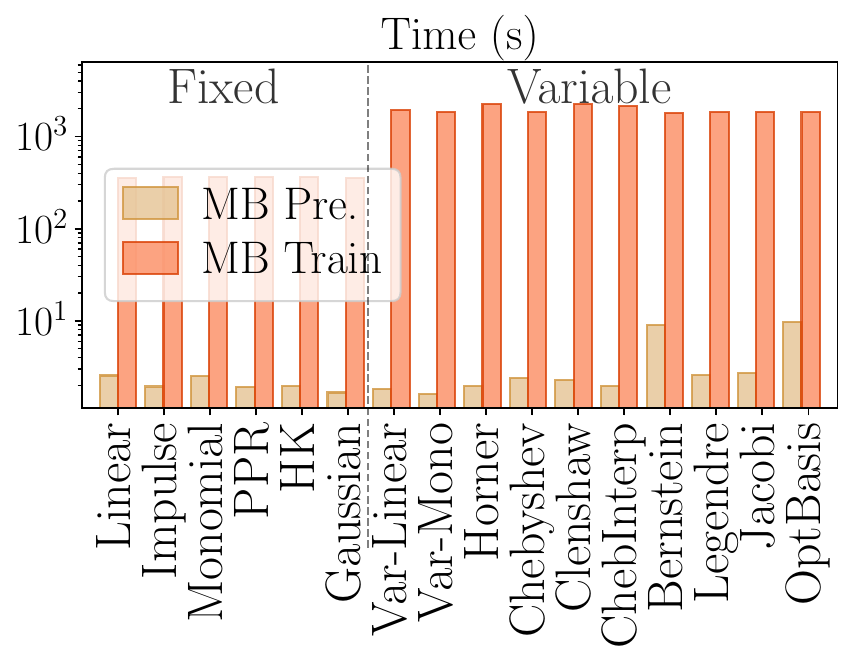}}
    \hfil
    \subcaptionbox{Memory\label{ffig:memlp_ogbl-ppa}}%
    [0.49\columnwidth]{\includegraphics[height=1.25in]{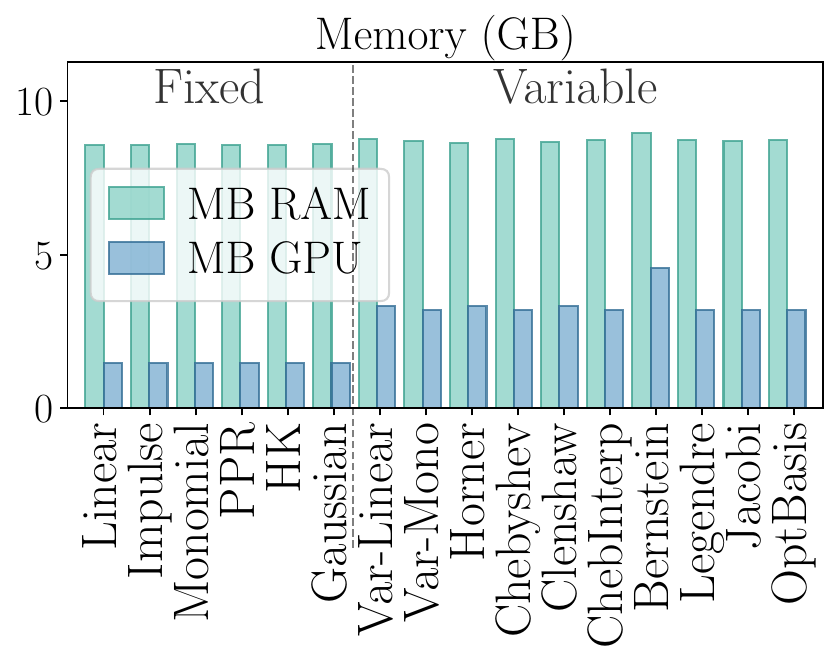}}
\captionsetup{font={small,stretch=0.9}}
    \caption{Filter time and memory efficiency of MB link prediction on \ds{ppa}. Note that the time axis is on log scale.}
  \label{fig:lp_eff}
\vspace{-0.8em}
\end{figure}

\subsubsection{Signal Regression.}
Spectral filtering also offers a regression task for fitting a given signal encoded in the graph, which is intuitive in characterizing the frequency response to signals spanning diverse frequency components.
We define the fully-supervised graph regression task \cite{he2021} by learning from the predefined pair of input $\mathbf{x}$ and the objective $\bmu{z} = g^\star \ast \bmu{x}$, as an approach to approximate the given simple spectral filter function $g^\star$. \cref{res:filter} presents the average $R^2$ score of five typical signal functions, where larger scores indicate better regression precision.

\begin{table}[!b]
\captionsetup{font={small,stretch=0.9}}
\caption{Average $R^2$ scores of graph regression on five signal functions. For each dataset, results are highlighted based on the relative effectiveness among filters, where \rkoo{green} results are better.}
\label{res:filter}
\small
\setlength{\tabcolsep}{2.5pt}
\renewcommand{\arraystretch}{0.9}
\centering
\begin{adjustbox}{max width=\columnwidth}
\begin{tabular}{@{}c@{}c|ccccc@{}}
\toprule
\multirow{2}{*}{Type} & Dataset & \ds{band} & \ds{combine} & \ds{high} & \ds{low} & \ds{reject} \\
& Signal & \tiny{$e^{-10(\lambda-1)^2}$}&\tiny{$|\sin(\pi\lambda)|$}&\tiny{$1-e^{-10\lambda^2}$}&\tiny{$e^{-10\lambda^2}$}&\tiny{$1-e^{-10(\lambda-1)^2}$}\\
\midrule
\multirow{6}{*}{Fixed}
    & \ft{PPR} & \tz{fdfeef}{21.40} & \tz{f9fcea}{32.13} & \tz{f9fcea}{35.42} & \tz{fff3e0}{77.58} & \tz{ebf5df}{91.24} \\
	& \ft{Linear} & \tz{fee9d8}{6.74} & \tz{feead8}{10.12} & \tz{fee9d8}{8.87} & \tz{e8f4df}{94.90} & \tz{ffffef}{83.22} \\
	& \ft{Impulse} & \tz{fee9d8}{6.73} & \tz{fee9d8}{9.99} & \tz{fee8d7}{8.79} & \tz{f0f8e3}{93.27} & \tz{fffeed}{82.25} \\
	& \ft{Monomial} & \tz{fdfeef}{21.40} & \tz{f9fcea}{32.13} & \tz{f9fcea}{35.42} & \tz{fff3e0}{77.58} & \tz{ebf5df}{91.24} \\
	& \ft{HK} & \tz{feecd9}{7.57} & \tz{feeedb}{11.77} & \tz{feecd9}{10.23} & \tz{dceedc}{96.93} & \tz{f7fbe8}{86.94} \\
	& \ft{Gaussian} & \tz{feecd9}{7.62} & \tz{feeedb}{11.96} & \tz{feecd9}{10.26} & \tz{dceedc}{96.92} & \tz{f7fbe8}{86.93} \\
\midrule
\multirow{10}{*}{Variable}
    & \ft{Monomial} & \tz{fee9d8}{6.87} & \tz{feead8}{10.21} & \tz{fee9d8}{8.87} & \tz{e9f4df}{94.80} & \tz{fffbe9}{79.75} \\
	& \ft{Horner} & \tz{ecf6df}{48.98} & \tz{d8ecdb}{69.10} & \tz{daeddc}{78.87} & \tz{ffffef}{89.14} & \tz{fffae7}{78.96} \\
	& \ft{Chebyshev} & \tz{fee9d8}{6.74} & \tz{fde4d5}{8.40} & \tz{fde4d5}{6.87} & \tz{f9fceb}{91.15} & \tz{fffceb}{81.12} \\
	& \ft{Clenshaw} & \tz{fee9d8}{6.77} & \tz{fde4d5}{8.06} & \tz{fee7d7}{8.14} & \tz{fff2de}{76.68} & \tz{fffdec}{81.49} \\
	& \ft{ChebInterp} & \tz{fee9d8}{6.72} & \tz{fee8d7}{9.58} & \tz{fee9d8}{8.88} & \tz{fcfeee}{90.42} & \tz{fffceb}{80.95} \\
	& \ft{Bernstein} & \tz{fff9e6}{13.03} & \tz{fffae8}{17.79} & \tz{ffffef}{22.10} & \tz{d8ecdb}{97.56} & \tz{f9fcea}{86.08} \\
	& \ft{Legendre} & \tz{fee9d8}{6.68} & \tz{fee9d8}{9.82} & \tz{fee8d7}{8.59} & \tz{edf6e0}{94.00} & \tz{feefdc}{72.86} \\
	& \ft{Jacobi} & \tz{fee9d8}{6.69} & \tz{fee9d8}{9.73} & \tz{fee8d7}{8.62} & \tz{edf6e0}{93.99} & \tz{fffeed}{82.14} \\
	& \ft{Favard} & \tz{fde4d5}{5.23} & \tz{fee6d6}{8.89} & \tz{fde5d6}{7.46} & \tz{fde4d5}{67.57} & \tz{fde4d5}{67.68} \\
	& \ft{OptBasis} & \tz{cce5d7}{82.88} & \tz{cce5d7}{79.73} & \tz{cce5d7}{93.69} & \tz{cce5d7}{99.19} & \tz{cce5d7}{99.06} \\
\bottomrule
\end{tabular}
\end{adjustbox}
\end{table}

The observation implies that most filters still highly focus on low-frequency components, demonstrating higher precision on \ds{low} and \ds{band reject}. This is in line with our findings in \cref{rq:effective-taxa} that introducing variable bases does not necessarily strengthen filter capability, as the performance is principally determined by its spectral formulation.
On the contrary, \textbf{filters enforcing higher attention on high-frequency domains}, including \ft{Monomial}, \ft{Horner}, and \ft{OptBasis}, \textbf{achieve strong performance on high-frequency signals} (\ds{BAND}, \ds{COMBINE}, and \ds{HIGH}), although \ft{Monomial} and \ft{Horner} sacrifice low-frequency ability to certain degrees. \ft{OptBasis} outperforms on all signals thanks to its flexible parameter acquisition. However, it should be noted that the regression precision of simple signals does not guarantee node classification effectiveness in our main experiment, as better utilization of graph signals is not always beneficial due to the complex grounding of the realistic task.

\subsection{Spectral Capabilities}
\label{ssec:extra-spectral}
As analyzed in \cref{rq:effective-hetero}, the effectiveness of graph learning with different filters is related to the compatibility between their spectral expressions and the graph data. Therefore, we delve into the spectral properties to provide a thorough understanding of the spectral capabilities concerning different filter formulations. These experiments orient the following question:
\begin{rquestionul}\label{rq:extra-spectral}
What inherent spectral properties are impactful in determining filter effectiveness?
\end{rquestionul}

\subsubsection{Effect of Propagation Hop.}
\label{sseca:extra_hopk}
The number of propagation hops $K$ is a critical hyperparameter governing both spectral expressiveness and empirical efficiency. Deciding the value of $K$ requires careful consideration in the spectral domain. A small $K$ indicates a limited number of polynomial terms in \cref{eq:sgnn}, leading to restricted filter capability. Conversely, a large number of propagations may introduce excessive graph information, which also implies linearly increased computational overhead.

The value of $K$ in existing literature varies significantly due to non-spectral designs, complicating the assessment of the filter capability. Our framework offers a unified architecture for spectral GNNs, enabling a more thorough examination of the impact of propagation hops on graph filters. \cref{fig:hop} illustrates the model accuracy patterns when varying the number of hops in common the range $K \in [2, 20]$ on homophilous and heterophilous graphs. \textbf{Generally, a limited $K$ is more favorable, especially for homophilous graphs.} Overall, our selection of $K=10$ in the main experiments is reasonable for maintaining the performance of most models across various datasets.

For low-pass filters (\ft{Linear} and \ft{Impulse}), the efficacy gradually decreases when the hop number increases for both homophilous and heterophilous graphs. This corresponds to the over-smoothing issue \cite{li2018,yang2021c,yan2022}, where the signal is overwhelm by non-local noise and loses useful information.
Filters such as \ft{PPR} and \ft{Gaussian} can alleviate this issue by tuning the decaying factor, while filters with high-frequency components, especially those with orthogonal basis, are also more stable by manipulating those signals.

\begin{figure}[!t]
    \centering
    \includegraphics[width=0.8\columnwidth]{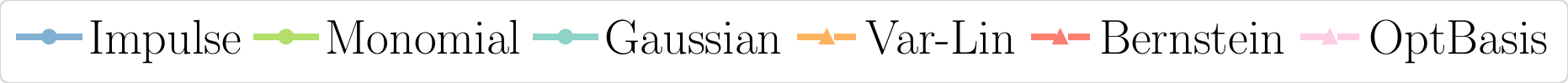}
    \subcaptionbox{\ds{citeseer}\label{ffig:hop_fix_citeseer}}%
    [0.49\columnwidth]{\includegraphics[height=1.2in]{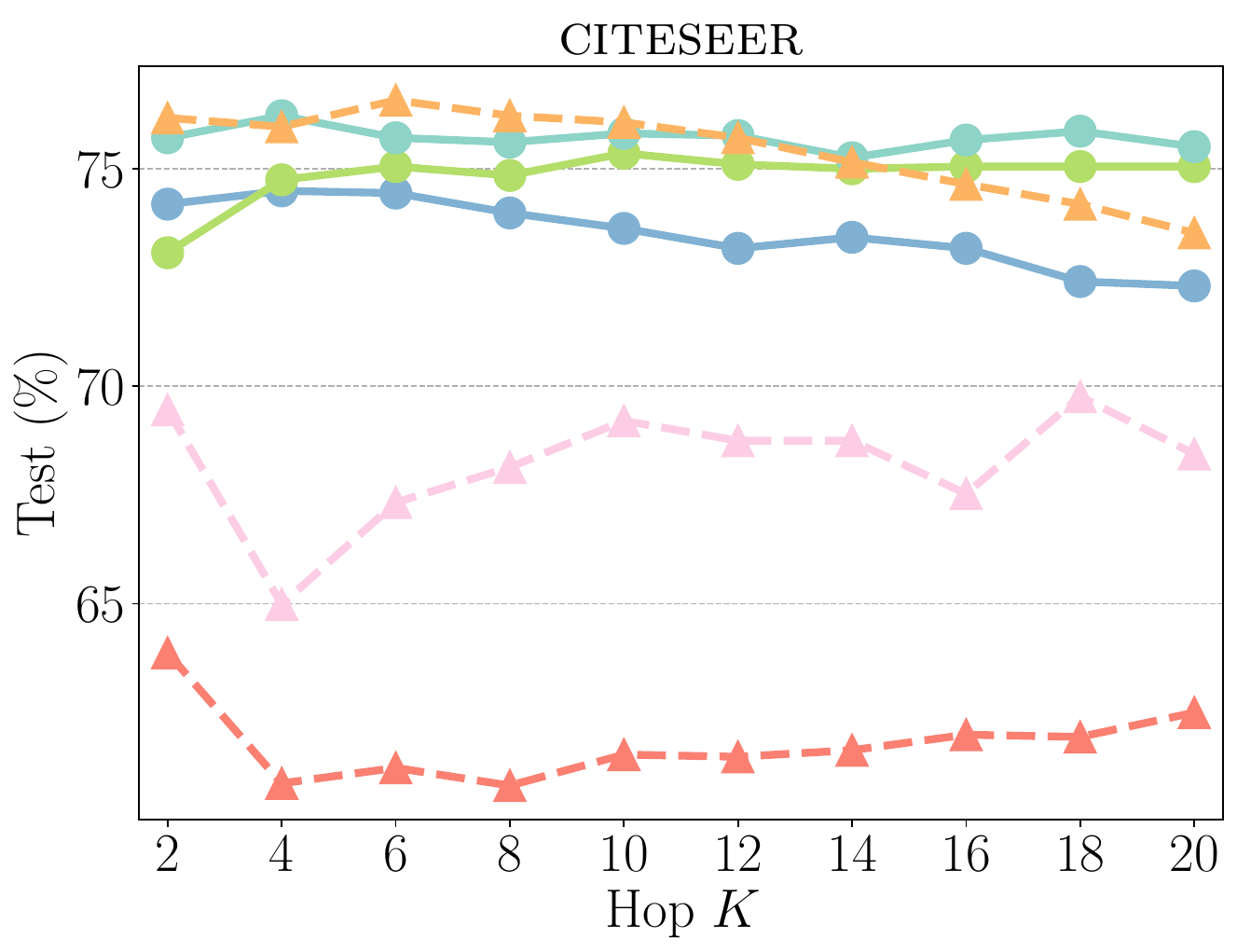}}
    \hfil
    \subcaptionbox{\ds{roman}\label{ffig:hop_fix_roman_empire}}%
    [0.49\columnwidth]{\includegraphics[height=1.2in]{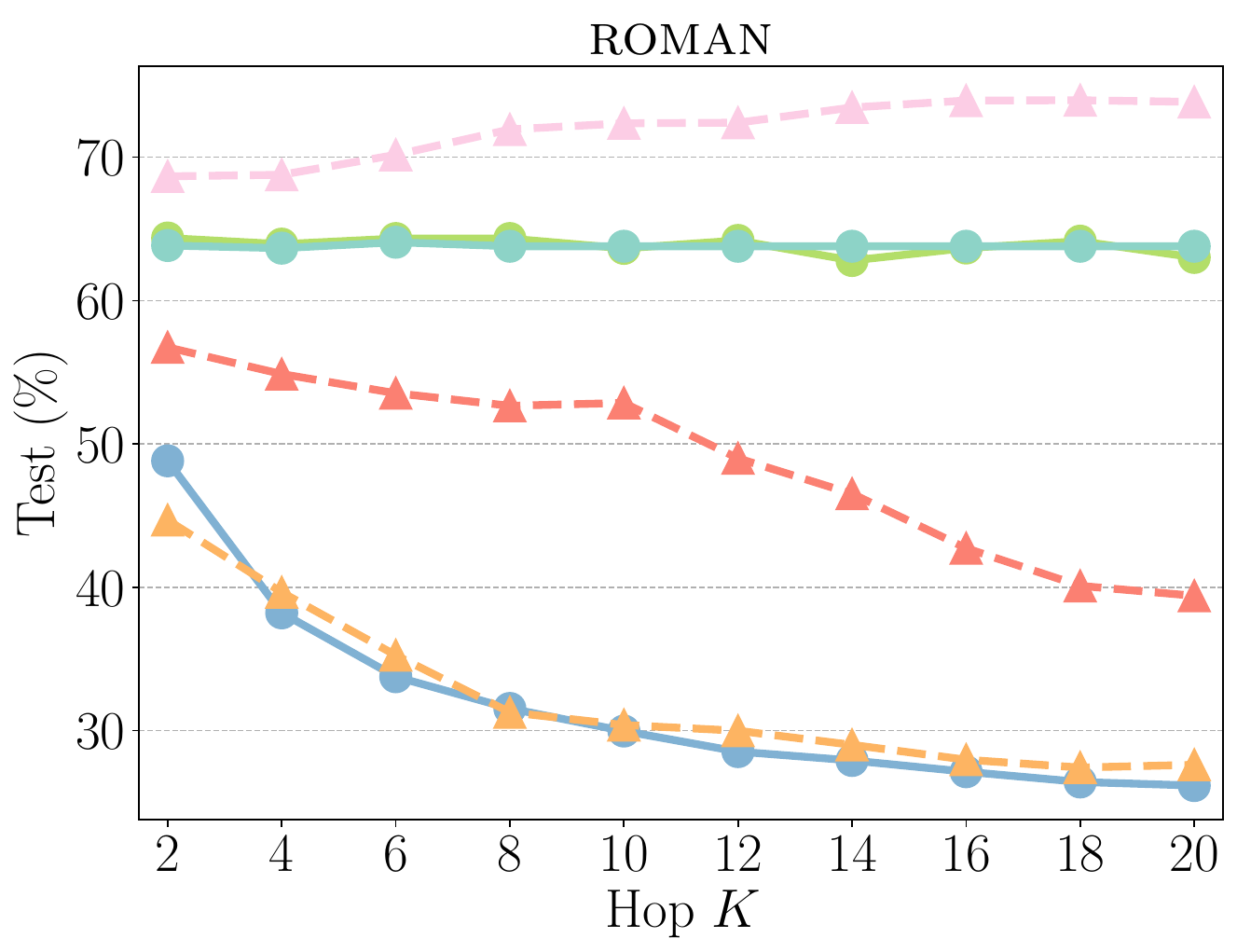}}
    \caption{Effect of propagation hops $K$ on representative full-batch fixed and variable filters. }
  \label{fig:hop}
\end{figure}

\subsubsection{Clustering Visualization.}
The t-SNE method provides an intuitive way to understand the decision boundary learned by spectral GNNs \cite{wang2019e}, which is essentially related to their prediction performance in \cref{res:acc}. \cref{fig:tsne} showcases visualizations of representative filters. Generally, embeddings for most filters on the homophilous \ds{CORA} are well-clustered with sharp boundaries, explaining the similar classification accuracy.
On the heterophilous \ds{CHAMELEON}, a few filters are able to maintain the ability to form clear clusters and consequently satisfy prediction performance (\ft{Monomial}, \ft{Chebyshev}). In comparison, the dispersed and overlapping clusters of \ft{PPR} and \ft{ChebInterp} correspond to the noise introduced by heterophilous graph topology, which undermines their effectiveness. Filters such as \ft{Impulse} and \ft{Jacobi} generate scattered clusters with notable outliers, rendering difficulties in fitting graph signals and producing meaningful representations.
In summary, the evaluation also implies that \textbf{spectral expressiveness outweighs variable designs} in terms of filter efficacy when conducting classification tasks under different graph patterns.

\begin{figure}[!t]
\captionsetup[subfigure]{skip=0pt}
\centering
\begin{subfigure}[b]{0.325\columnwidth}
    \centering
    \includegraphics[height=0.96in]{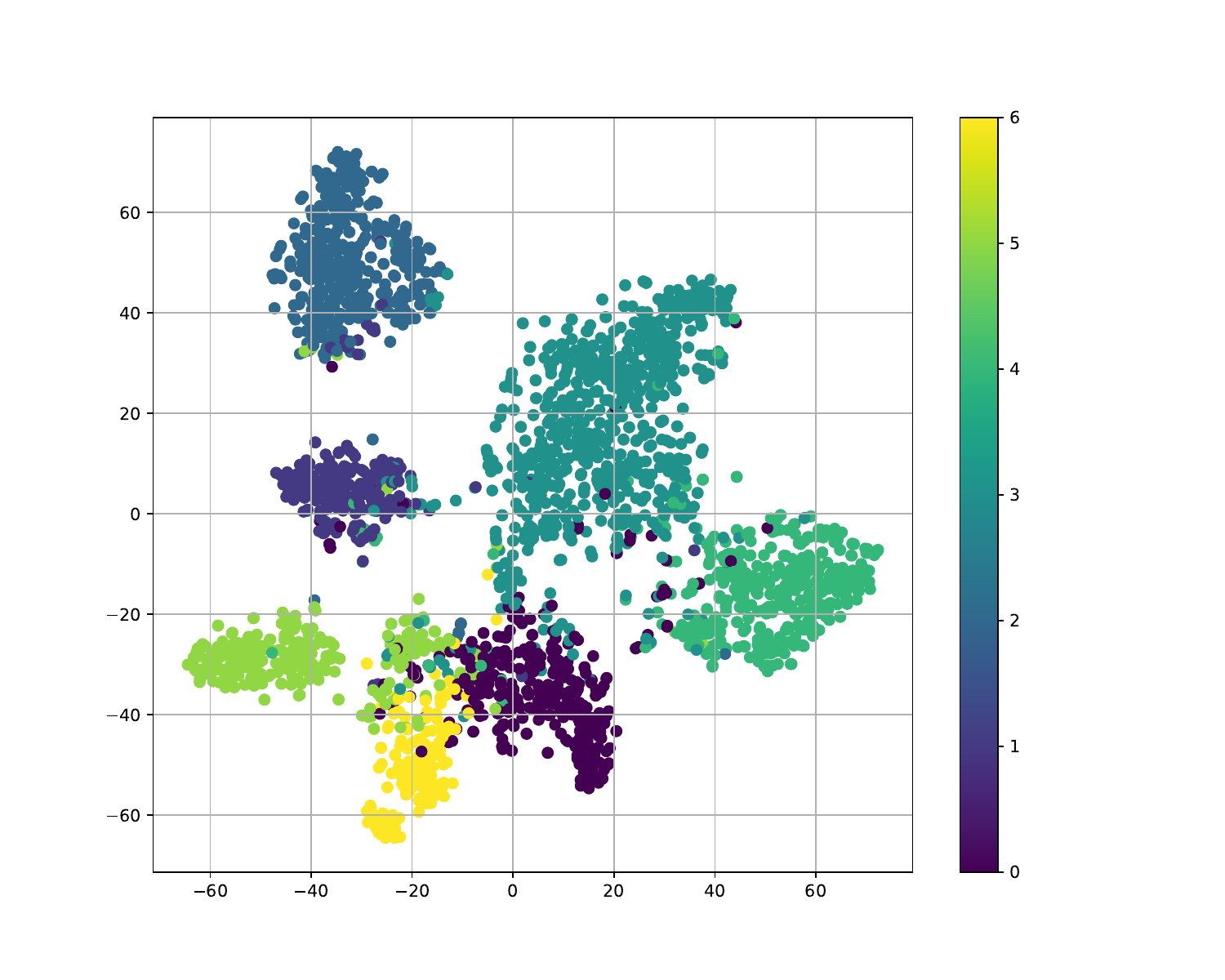}
    \caption{\ft{PPR}, \ds{cora}}
    \label{fig:cora_tsne_appr}
\end{subfigure}
\hfil
\begin{subfigure}[b]{0.325\columnwidth}
    \centering
    \includegraphics[height=0.96in]{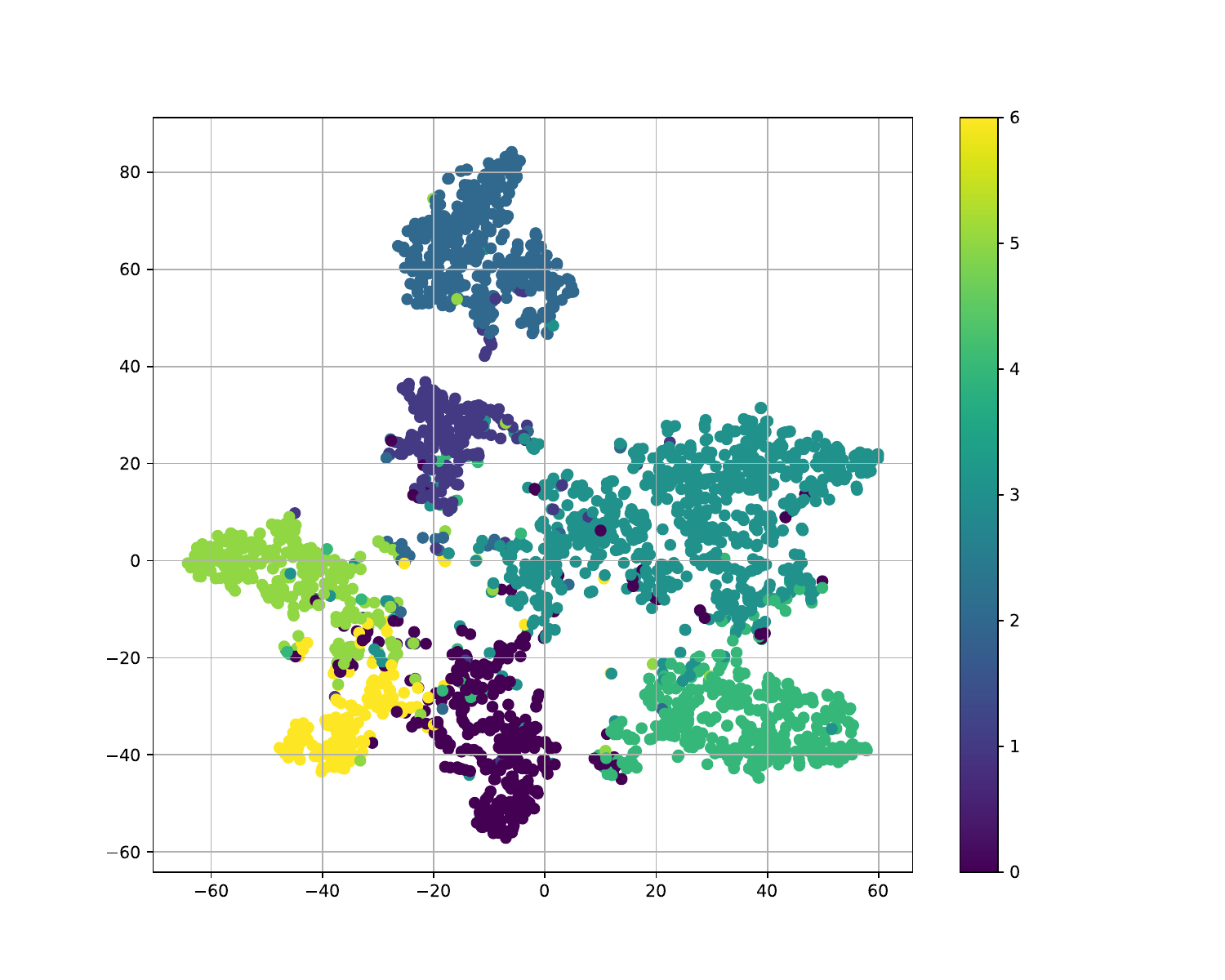}
    \caption{\ft{Cheb}, \ds{cora}}
    \label{fig:cora_tsne_chebconv}
\end{subfigure}
\hfil
\begin{subfigure}[b]{0.325\columnwidth}
    \centering
    \includegraphics[height=0.96in]{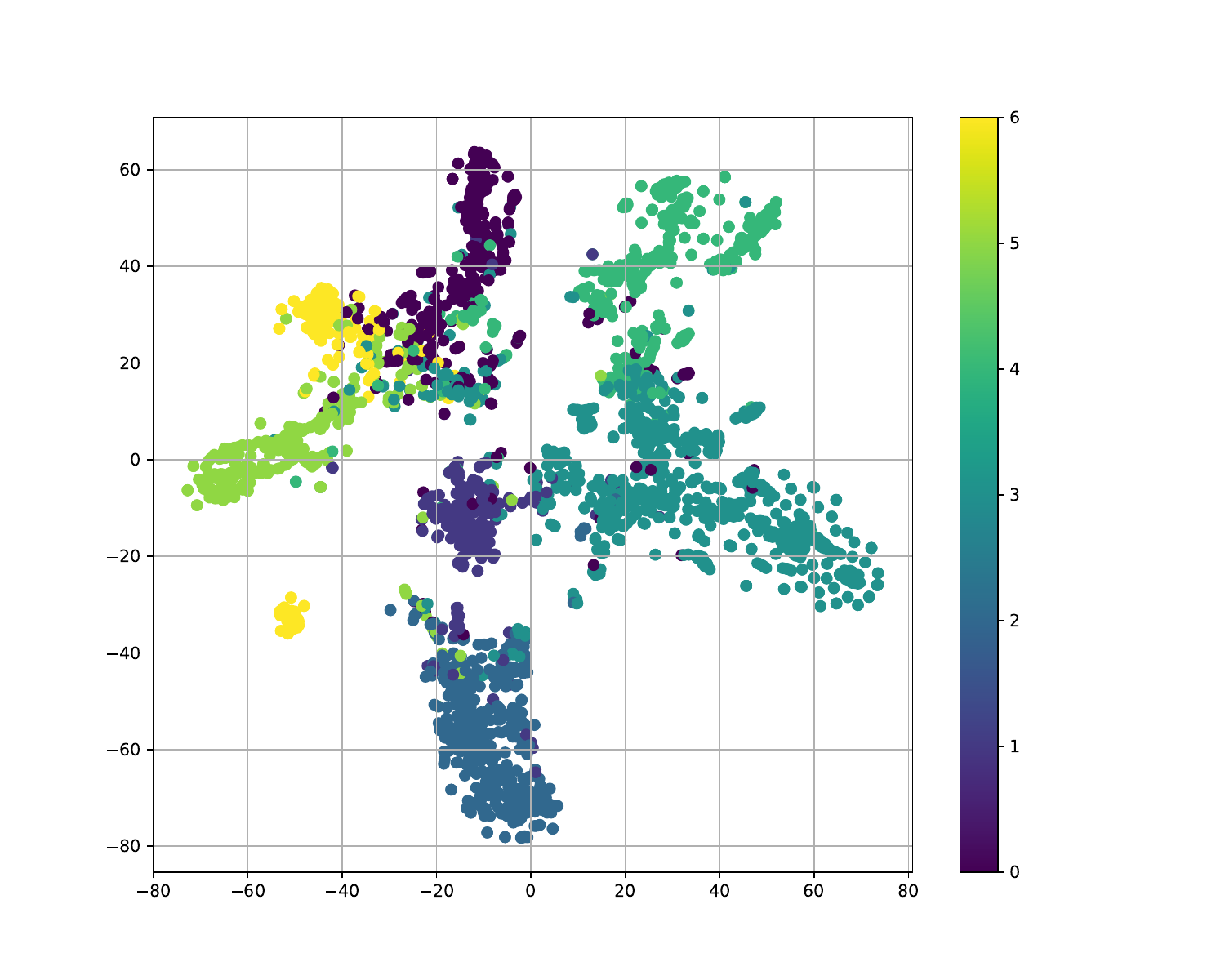}
    \caption{\ft{Jacobi}, \ds{cora}}
    \label{fig:cora_tsne_jacobiconv}
\end{subfigure}
\\[-4pt]
\begin{subfigure}[b]{0.325\columnwidth}
    \centering
    \includegraphics[height=0.96in]{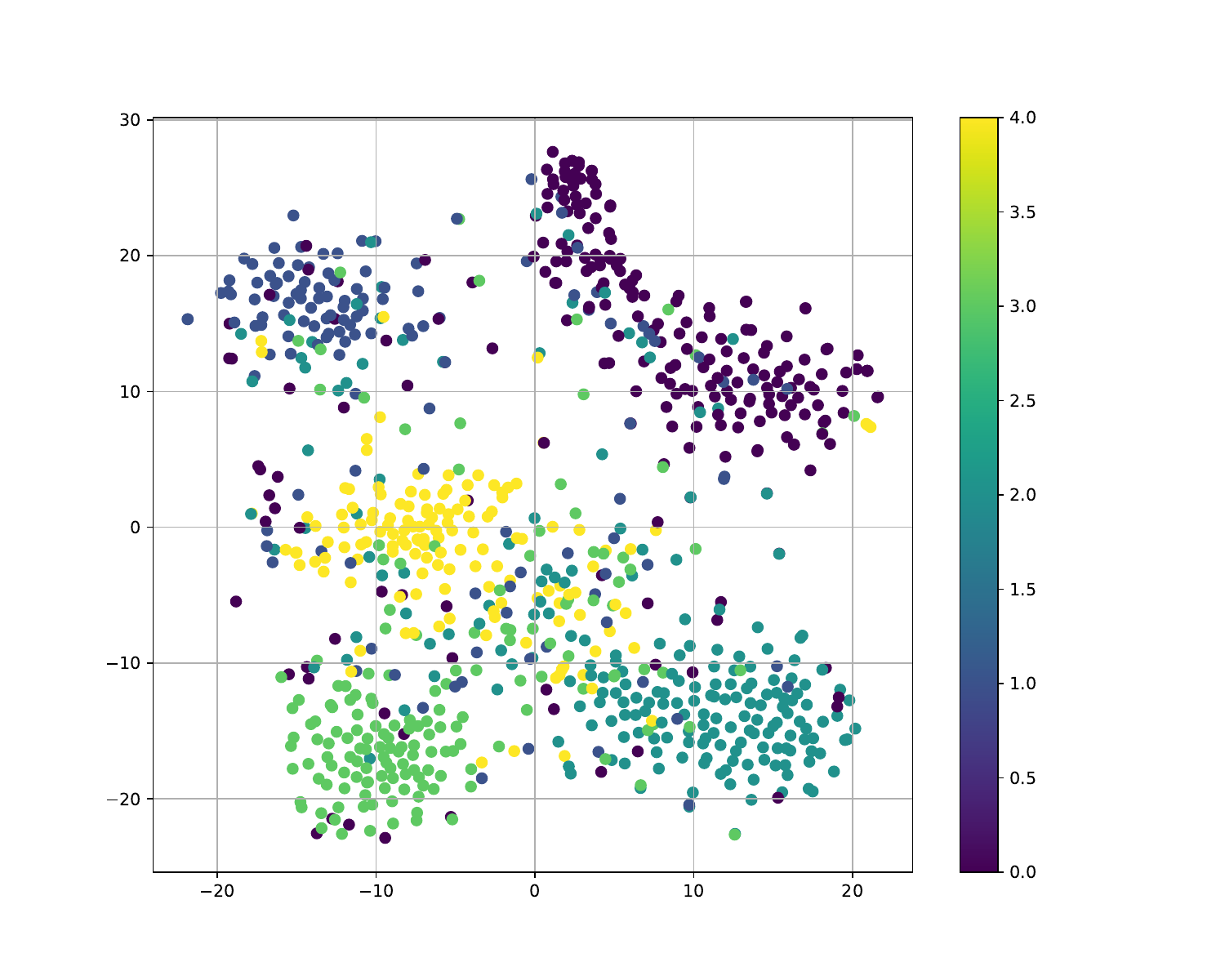}
    \caption{\ft{PPR}, \ds{chameleon}}
    \label{fig:chameleon_tsne_appr}
\end{subfigure}
\hfil
\begin{subfigure}[b]{0.325\columnwidth}
    \centering
    \includegraphics[height=0.96in]{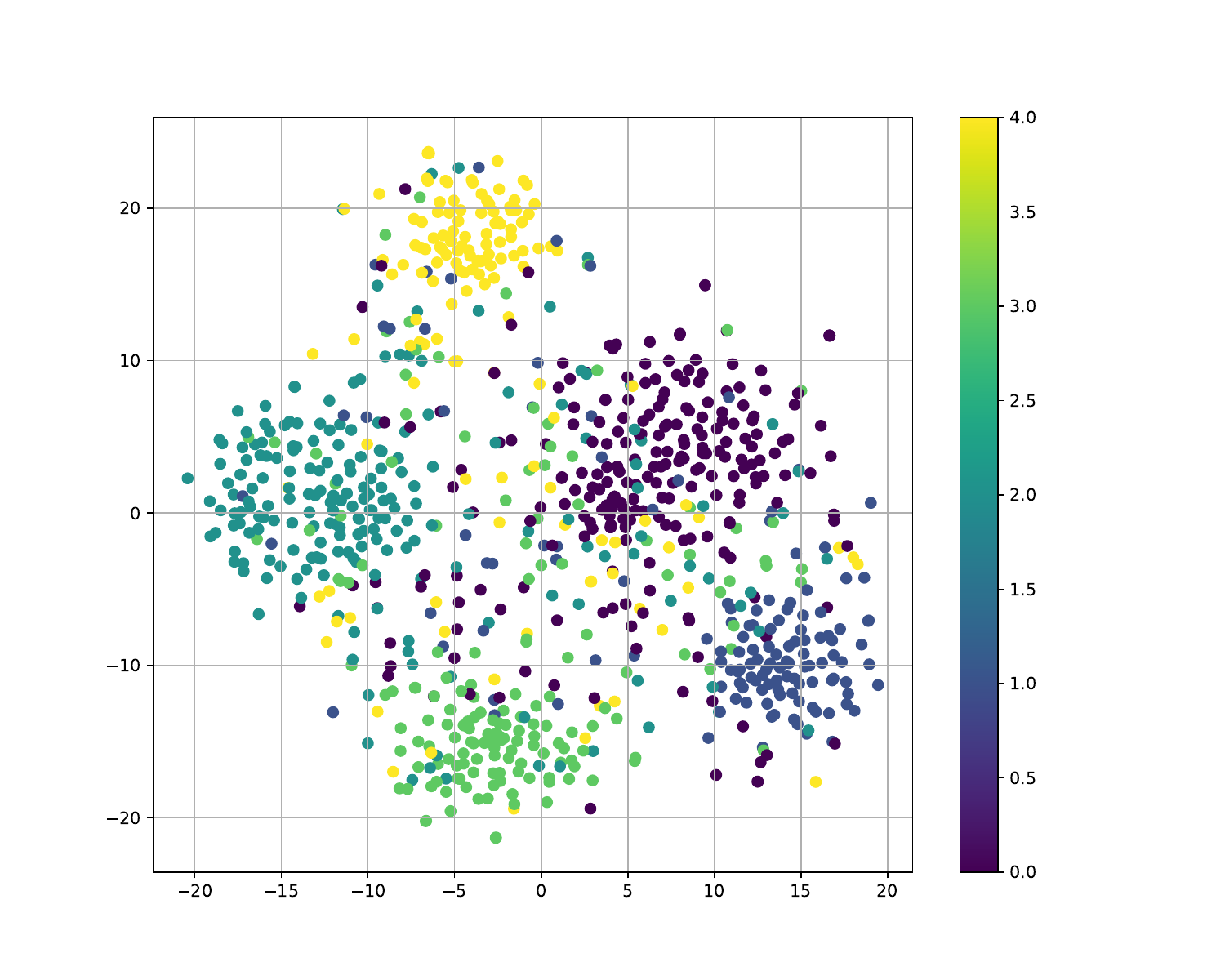}
    \caption{\ft{Cheb}, \ds{chameleon}}
    \label{fig:chameleon_tsne_chebconv}
\end{subfigure}
\hfil
\begin{subfigure}[b]{0.325\columnwidth}
    \centering
    \includegraphics[height=0.96in]{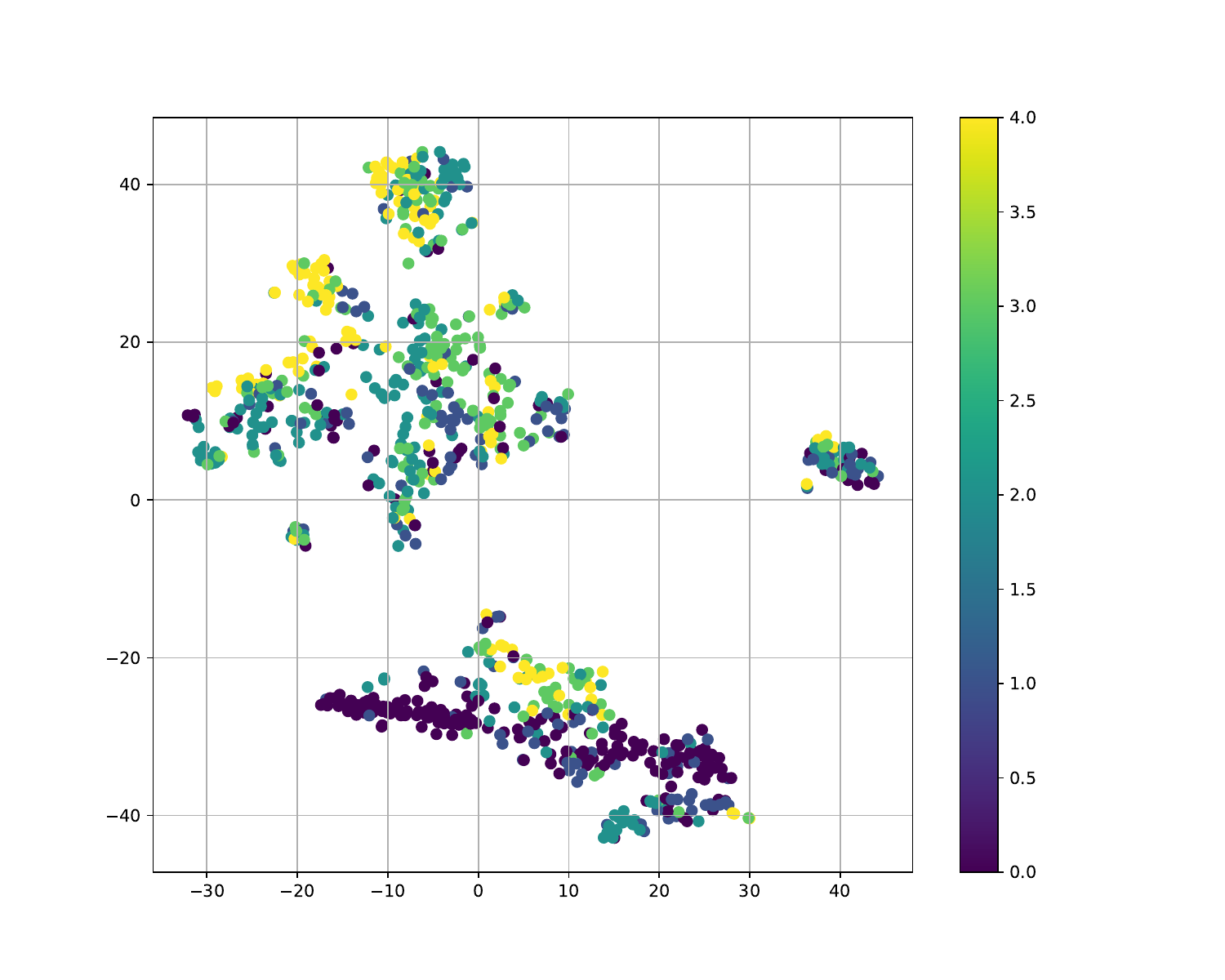}
    \caption{\small{\ft{Jacobi}, \ds{chameleon}}}
    \label{fig:chameleon_tsne_jacobiconv}
\end{subfigure}
\caption{t-SNE clusters of learning results for selected filters and datasets. Nodes are colored by ground truth labels.
}
\label{fig:tsne}
\end{figure}

\begin{figure*}[!t]
\begin{subcaptionblock}{0.48\textwidth}
    \hspace{-1em}
    \captionsetup{skip=-2pt,margin={0em,-2em}}
    \includegraphics[height=1.48in]{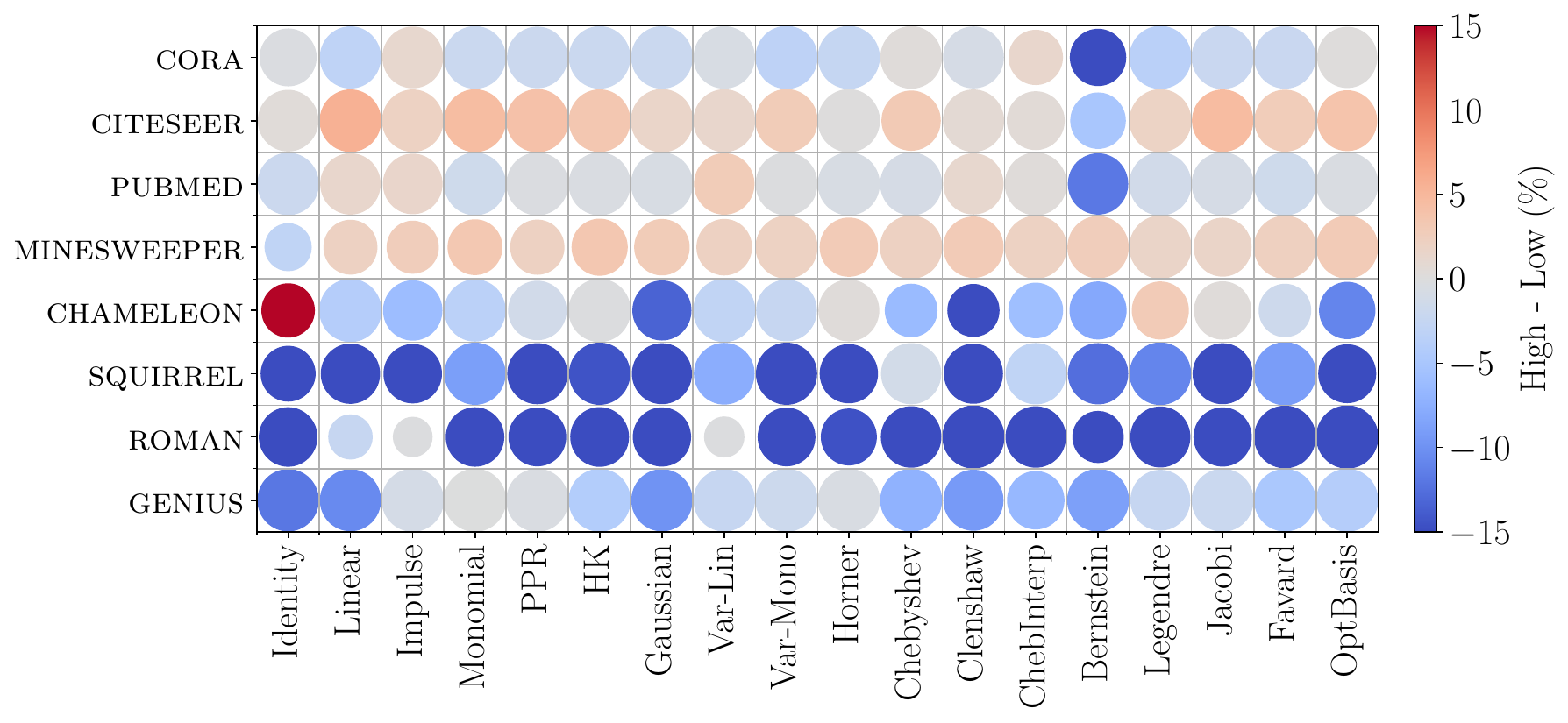}
    \caption{Accuracy gap on homophilous and heterophilous graphs.}\label{ffig:deg_best}
\end{subcaptionblock}%
\begin{subcaptionblock}{0.26\textwidth}
    \hspace{-2.3em}
    \captionsetup{skip=-2pt,margin={0em,-1.5em}}
    \includegraphics[height=1.48in]{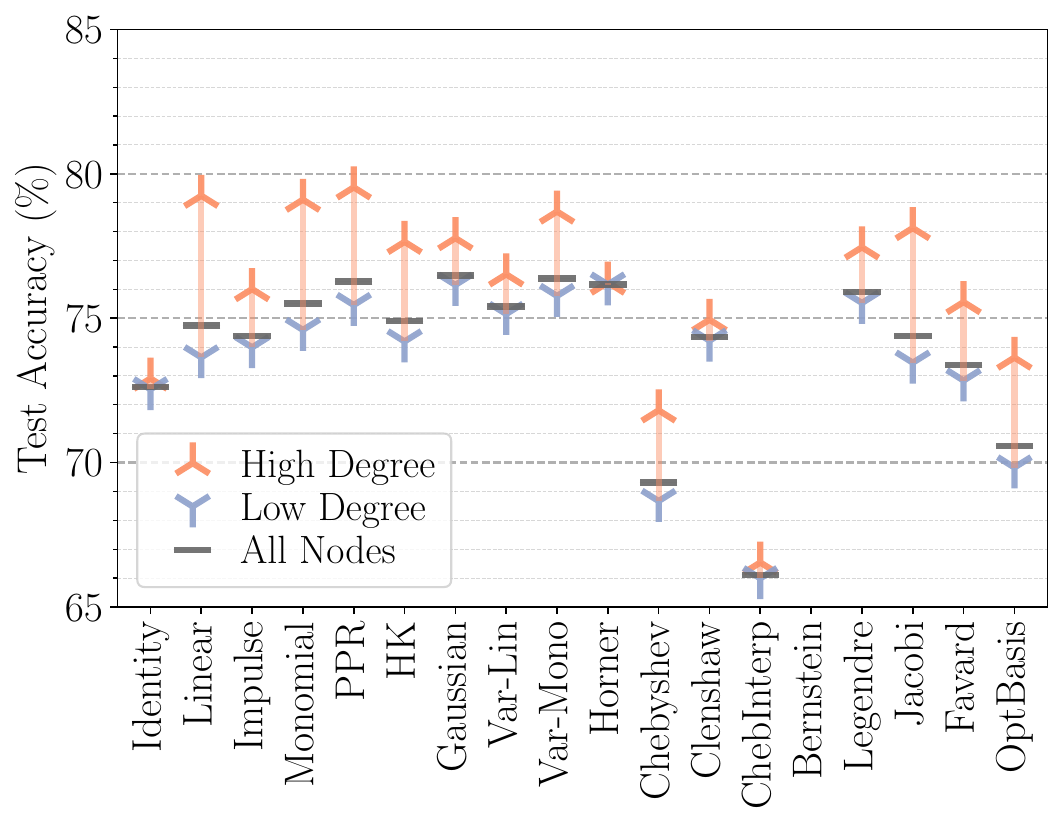}
    \caption{Respective accuracy on \ds{citeseer}.}\label{ffig:deg_citeseer}
\end{subcaptionblock}%
\begin{subcaptionblock}{0.26\textwidth}
    \hspace{-1.1em}
    \captionsetup{skip=-2pt,margin={0em,-1em}}    
    \includegraphics[height=1.48in]{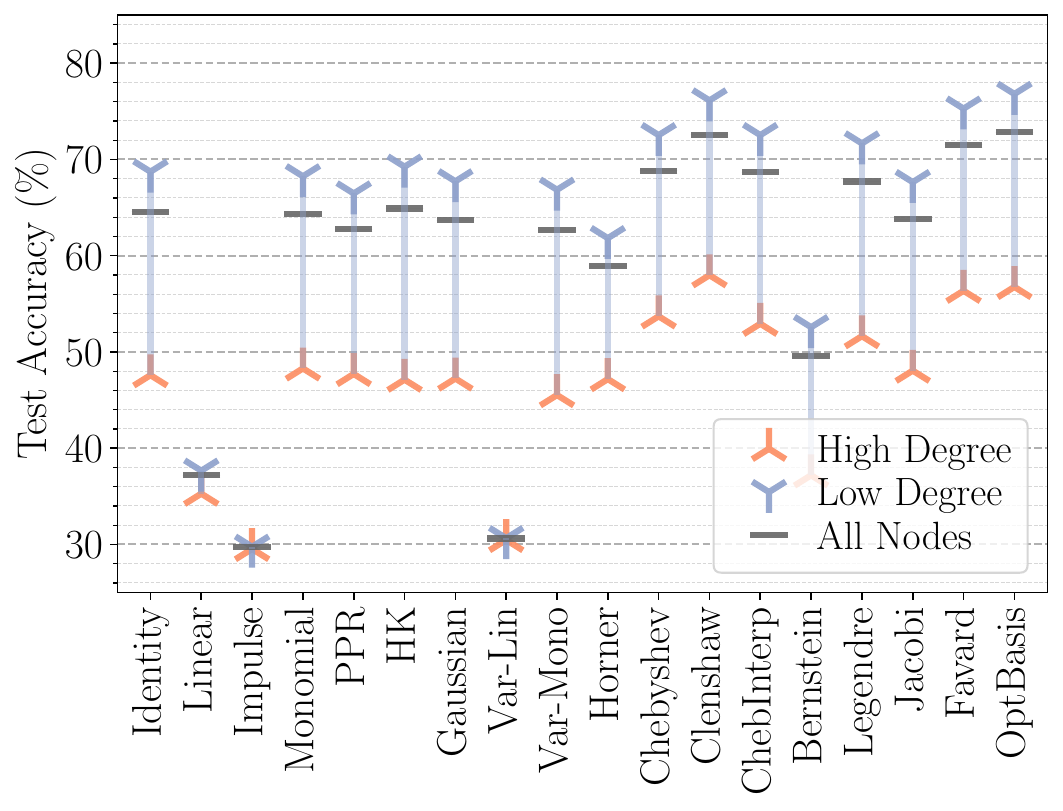}
    \caption{Respective accuracy on \ds{roman}.}\label{ffig:deg_roman}
\end{subcaptionblock}%
\captionsetup{font={stretch=0.98}}
\caption{\textbf{(a)} Accuracy gap between high- and low-degree nodes on homophilous (upper four) and heterophilous (bottom four) graphs. Color of each data point indicates the difference value, while size of the circle represents the relative overall accuracy among all filters in each dataset. \textbf{(b)-(c)} Respective accuracy of high- and low-degree nodes on \ds{citeseer} and \ds{roman}. } 
\label{fig:deg_best}
\vspace{.4ex}
\end{figure*}

\subsection{Degree-specific Effectiveness}
\label{ssec:deg}
A recent trend examines node-wise performance and discovers that the learning effectiveness of vanilla GNNs varies with respect to nodes of differing degree levels \cite{Tang2020,yan2022,Chen2022ba,Liao2023SAILOR,li2024,lai2024}.
Unlike previous investigations assuming homophily, we extend the degree-wise evaluation to heterophilous graphs. \cref{fig:deg_best} displays the difference between prediction accuracy on high- and low-degree nodes across representative datasets. The extended experimental evaluation enables us to relate the degree-specific performance to overall model accuracy in \cref{ssec:main-effect}.

\subsubsection{Relation with Filter Effectiveness.}
\label{ob:deg_effective}~
\begin{rquestionul}\label{rq:deg-effective}
What is the degree-wise effectiveness under homophily and heterophily, and how does it affect the filter efficacy?
\end{rquestionul}
It can be observed in \cref{fig:deg_best} that most filters behave distinctively on homophilous and heterophilous datasets. On \textit{homophilous} graphs (\ds{citeseer} in \cref{ffig:deg_citeseer}), the performance of high-degree nodes is generally on par with or higher than low-degree ones, which echoes earlier studies. From the spectral domain view, high-degree nodes are commonly located in clusters, which is more related to low frequency in the graph spectrum and is preferable for homophilous GNNs throughout learning.
Contrarily, the accuracy of high-degree nodes is usually significantly lower under \textit{heterophily}, as shown in \cref{ffig:deg_roman} for \ds{roman}. In this case, the hypothesis that higher degrees are naturally favored by GNNs no longer holds true. Although these nodes aggregate more information from the neighborhood, it is not necessarily beneficial, and heterophilous connections may carry destructive bias and hinder the prediction. As a more precise amendment to the conclusion in prior works, we state that the \textbf{degree-wise bias is more \textit{sensitive}, but not necessarily beneficial, to high-degree nodes} with regard to varying graph conditions.

It is also notable in \cref{fig:deg_best} that the degree-specific difference is correlated with the test accuracy of models within each dataset, especially under heterophily. For example, in \cref{ffig:deg_roman} \ds{roman}, filters with greater bias achieve relatively better overall accuracy. In comparison, filters such as \ft{Linear} and \ft{Impulse} fail to distinguish between high- and low-degree nodes, resulting in lower test accuracy.
From the observation, we deduce that \textbf{GNNs tend to recognize and adapt to the majority of low-degree nodes} in order to achieve higher performance. This preference is potentially exaggerated by the heterophily-oriented filter design, which pays more attention to the high-frequency components correlating to low-degree nodes while compromising the performance of high-degree nodes. It is thus advisable to explore filter formulations balancing different frequency ranges, which are promising for addressing the current drawback and further advancing overall model accuracy.

\begin{figure}[!b]
\vspace{1ex}
\captionsetup{font={stretch=0.98}}
    \centering
    \includegraphics[width=0.86\columnwidth]{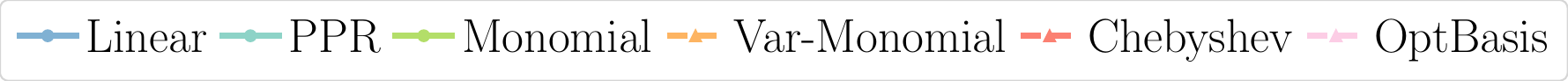}
    \subcaptionbox{\ds{cora}\label{ffig:degng_cora}}%
    [0.49\columnwidth]{\includegraphics[height=1.3in]{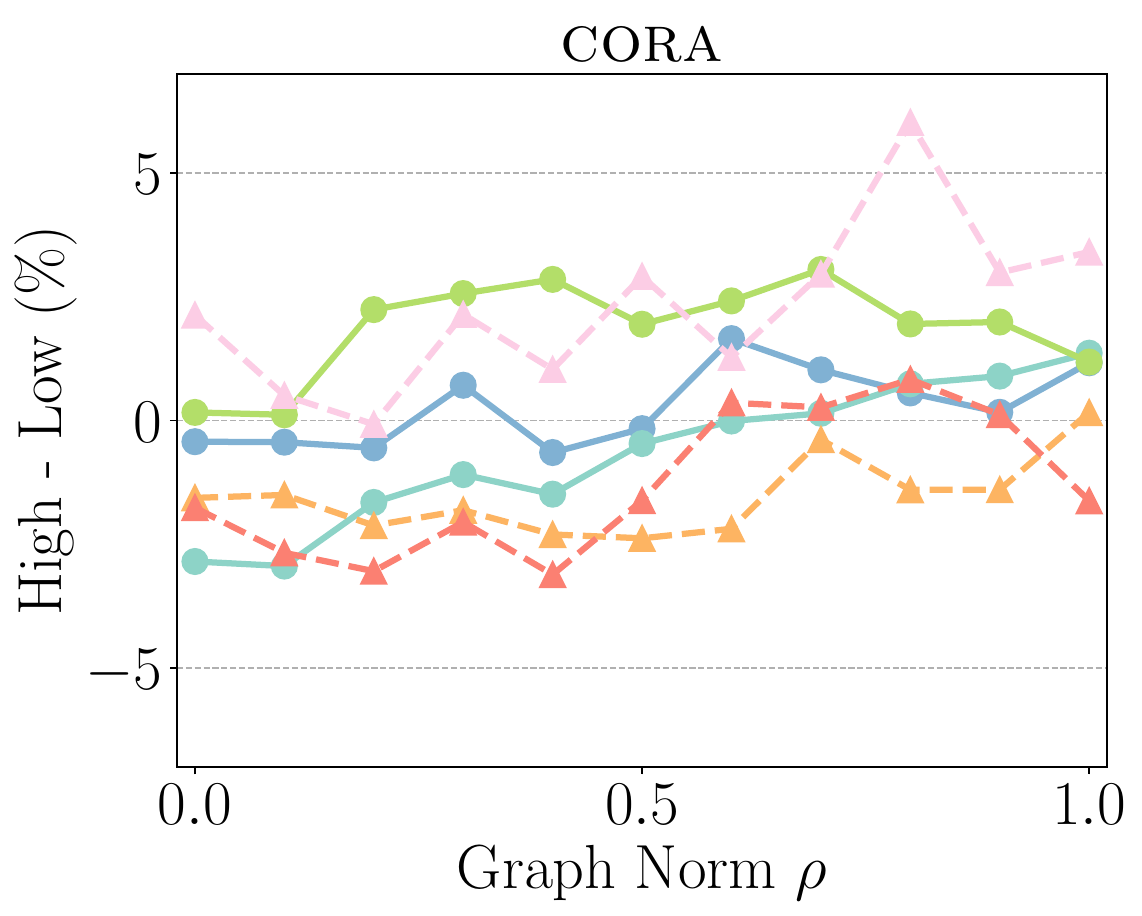}}
    \hfil
    \subcaptionbox{\ds{roman}\label{ffig:degng_roman_empire}}%
    [0.49\columnwidth]{\includegraphics[height=1.3in]{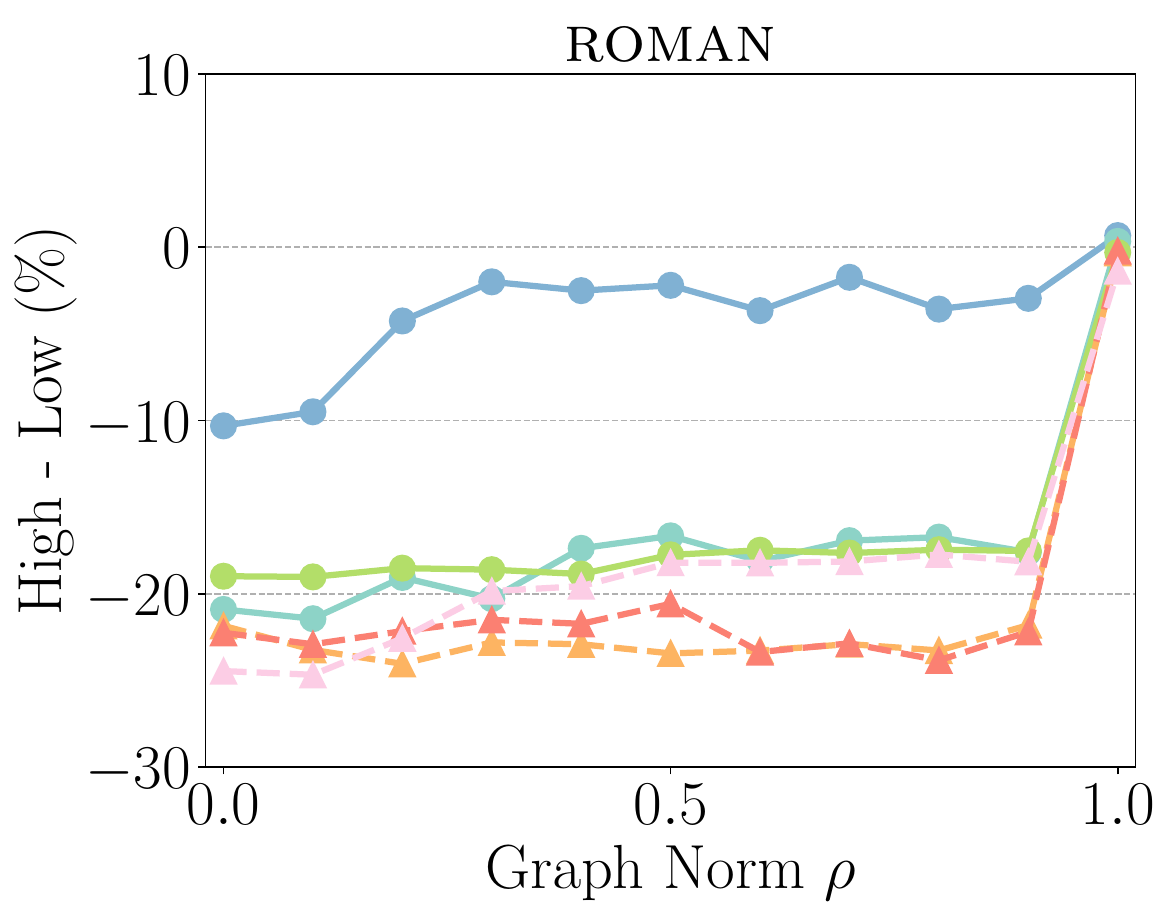}}
    \caption{Effect of graph normalization $\rho$ on the accuracy difference between high- and low-degree nodes. }
  \label{fig:degng}
\end{figure}

\subsubsection{Effect of Graph Normalization.}~
\begin{rquestionul}\label{rq:deg-norm}
How to utilize the graph information to control degree-wise effectiveness, and therefore affect overall accuracy?
\end{rquestionul}
Recall that the normalized adjacency $\Tilde{\bmu{A}} = \Bar{\bmu{D}}^{\rho-1} \Bar{\bmu{A}} \Bar{\bmu{D}}^{-\rho}$ explicitly encodes degree information during propagation, we are thence motivated to use it to impact filter effectiveness. In previous studies, it is revealed that the normalization in form of $\Tilde{\bmu{A}} = \Bar{\bmu{D}}^{-1} \Bar{\bmu{A}}$ affects the node-wise performance on homophilous graphs \cite{zhang2021c,li2024}, while we extend to the continuous range of $\rho \in [0,1]$. Specially, $\rho = 1/2$ is the symmetric normalization with equal contributions from both in- and out-edges.

We present experimental result of the accuracy gap between high- and low-degree nodes when varying the graph normalization hyperparameter $\rho$ in \cref{fig:degng}.
It can be observed that \textbf{a larger $\rho$ increases the difference value}, i.e., improves the relative accuracy of high-degree nodes for both fixed and variable filters on \ds{citeseer} and \ds{roman}. This suggests that inbound information is preferred for model inference on high-degree nodes, and adjusting the aggregation through $\rho$ could be an practical attempt to alleviate degree-wise differences and achieve favorable performance under homophily. On graphs such as \ds{chameleon} and \ds{actor}, where connection utility is hindered by heterophily, this pattern is weaker, and the performance gap becomes more unstable due to the complexity of graph conditions.

\vspace{1ex}
\subsection{Key Conclusions}
\label{ssec:extra-ob}
\begin{enumerate}[label=C\arabic*.]
\setcounter{enumi}{5}
    \item Filter effectiveness can be explained by their spectral properties, especially the response on different frequencies. Sophisticated designs are only effective when their frequency components match the graph signal. (\cref{rq:extra-spectral})
    \item Spectral models exhibit dissimilar effectiveness on nodes of high and low degrees under different graph conditions, challenging the assumptions in previous studies. The efficacy on high-degree nodes degrades under heterophily, potentially indicating greater sensitivity to the graph learning process on these nodes. (\cref{rq:deg-effective})
    \item The degree-wise bias is also relevant to the overall effectiveness of filters and can be controlled by altering the influence of graph topology. Dedicated filtering designs that benefit both low- and high-degree performance remain underexplored. (\cref{rq:deg-norm})
\end{enumerate}

\section{Conclusion and Open Questions}
In this study, we conduct extensive evaluations of spectral GNNs concerning both effectiveness and efficiency. We first provide a thorough analysis of the graph kernels in existing GNNs and categorize them by spectral designs. These filters are then implemented under a unified framework with highly efficient learning schemes. Comprehensive experiments are conducted to assess the model performance, with discussions covering heterophily, graph scales, spectral properties, and degree-wise bias.
Our benchmark observations also identify several open questions worthy of further research toward better spectral GNNs:
\begin{itemize}
    \item \textit{How to properly obtain effective and efficient spectral filters?} As analyzed in \cref{rq:balance}, it is possible to achieve both effectiveness and efficiency for some graphs, while finding the suitable filter is largely empirical. Our observation suggests that despite heterophily, latent patterns of graph data distribution also affect the spectral performance. Identifying and characterizing these factors is beneficial for designing more powerful filters based on graph knowledge.
    \item \textit{How to further improve filter efficiency and scalability?}
    In \cref{rq:efficiency-filter}, the spectral GNN design outlines a solution for performing learning on large graphs, although existing variable and composed filter computations are far from optimized. The GNN scalability issue calls for a persistent pursuit of more efficient designs, which is of great practical interest. Our study is helpful in uncovering impact factors and bottlenecks of large-scale deployment, and underscores the potential of spectral GNNs for achieving efficiency without compromising efficacy.
    \item \textit{How to address the degree-specific bias to enhance overall effectiveness?}
    In \cref{rq:deg-effective}, we discover that the degree-wise difference is related to graph heterophily and affects model precision. As current endeavors to mitigate GNN bias largely assume homophily, it is promising to search for specialized schemes that improve the accuracy of both low- and high-degree nodes, considering various graph conditions, which also benefits overall model performance.
\end{itemize}


\clearpage
\bibliographystyle{ACM-Reference-Format}
\bibliography{ref}
\balance

\clearpage
\appendix
\section{Theoretical Derivation}
\label{seca:model}
\subsection{Spatial and Spectral Convolutions}
\noindentparagraph{Filter Operations.}
Spatial methods are identified by their direct operation $f$ on the graph topology $\Tilde{\bmu{A}}$. A line of studies have revealed the relationship between spatial and spectral operations in the context of GNN learning, showing that spectral formulation can be derived from the spatial one, and vice versa \cite{kipf2016semi,chen2023,bo2023b}. 
Without loss of generality, by expanding $T(\Tilde{\bmu{L}})$ and substituting $\Tilde{\bmu{L}}=\bmu{I}-\Tilde{\bmu{A}}$, the \textit{spectral} filter defined in \cref{eq:sgnn} can be written as a polynomial based on $\Tilde{\bmu{A}}$ as $f(\Tilde{\bmu{A}}) = \sum_{j=0}^{J} \xi_j\, \Tilde{\bmu{A}}^j$, which exactly corresponds to the \textit{spatial} convolution under mean aggregation. In other words, the polynomial spectral filters can be equivalently achieved by a series of recurrent graph propagations. In cases where the spatial and spectral orders align with each other, there is $J=K$. Examples of deriving between $f(\Tilde{\bmu{A}})$ and $g(\Tilde{\bmu{L}})$ formulations are given in \cref{sec:taxonomy}. 

In terms of multi-layer GNNs, if for the $k$-th layer, the single-layer filter in spectral domain is $g^{(k)}$, then the overall spectral filter is as expressive as $g = g^{(k)} \ast g^{(k-1)} \ast\cdots\ast g^{(0)}$ \cite{feng2022b,wang2022b}. 
Conversely, if the explicit polynomial is known, the model can be calculated in the iterative form by recursively multiply the diffusion matrix (or matrices) to acquire the $k$-order basis, and adding to the result with respective weight parameter. Hence, we use iterative or explicit formulas interchangeably to express filters. 


\begin{table}[bp]
\caption{Summary of primary symbols and notations in this work.}
\label{tab:notation}
\centering
\setlength{\tabcolsep}{3pt}
\renewcommand{\arraystretch}{1.}
\begin{adjustbox}{max width=\columnwidth}
\begin{tabular}{@{}c|l@{}} \toprule
    \textbf{Notation} & \textbf{Description} \\ \midrule
    $\mathcal{G}, \mathcal{V}, \mathcal{E}$ & Graph, node set, and edge set \\
    $\mathcal{N}(u)$ & Neighboring node set of node $u$ \\
    $f, \varphi$ & Spatial propagation and transformation \\
    $\Hat{g}, g$ & Full and truncated spectral convolution \\
\midrule
    $n, m, F$ & Node number, edge number, and feature number \\
    $J, K, Q$ & Spatial propagation hops, spectral filter order, number of filters \\
    $\rho$ & General graph normalization coefficient \\
    $\xi_j$ & Spatial propagation parameter of hop $j$ with regard to $\Tilde{\bmu{A}}$ \\
    $\theta_k$ & Spectral filter parameter of order $k$ with regard to polynomial $T^{(k)}(\Tilde{\bmu{L}})$ \\
    $\gamma_q$ & Weight parameter of filter $q$ \\
\midrule
    $\bmu{A},\, \Bar{\bmu{A}},\, \Tilde{\bmu{A}}$ & Raw, self-looped, and normalized adjacency matrix of graph $\mathcal{G}$ \\
    ${\bmu{L}},\, \Tilde{\bmu{L}}$ & Raw and normalized Laplacian matrix of graph $\mathcal{G}$ \\
    $\bmu{D},\, \Bar{\bmu{D}}$ & Raw and self-looped diagonal degree matrix \\
    $\bmu{H}, \bmu{W}$ & Network representation, weight matrix \\
    $\bmu{X},\, \bmu{x}$ & Node attribute matrix and feature-wise vector \\
    $\bmu{\Lambda},\, \lambda_i$ & Diagonal eigenvalue matrix, $i$-th eigenvalue of graph Laplacian \\
\bottomrule
\end{tabular}
\end{adjustbox}
\end{table}

\noindentparagraph{Model Architectures.}
We further elaborate the multi-layer formulation in different model architectures. \textit{Iterative} architecture implies that each hop of propagation is associated with an explicit transformation computation $\varphi$, usually implemented by multiplying a learnable weight matrix $\bmu{W} \in\RR{F\times F}$ and applying non-linear activation. For simplicity, here we assume the dimension of input and output features to be constant across all layers. 
In the common case, the layer representation $\bmu{H}^{(j+1)}$ is dependent on its predecessor $\bmu{H}^{(j)}$. When the $j$-th propagation is $f^{(j)}$, the model is computed in the recursive or iterative form: $\bmu{H}^{(j+1)} = \varphi( f^{(j)}(\Tilde{\bmu{A}}) \cdot \bmu{H}^{(j)})$. Typically, applying such spatial convolution of $J$ iterations leads to a receptive field of $J$ hops on the graph. 

\textit{Decoupled} designs perform all hops of graph-related propagation together without invoking weight tensors, which implies that the parameters $\theta_k$ in \cref{eq:sgnn} are all scalars. Since it also receives $K$-hop information, this does not change the spectral expressiveness of the filtering process. A general spatial formulation for the network pipeline is $\bmu{H} = \varphi_1( f(\Tilde{\bmu{A}}) \cdot \varphi_0(\bmu{X}))$, where $f(\Tilde{\bmu{A}})$ contains all $J$-hop propagations, and $\varphi_0, \varphi_1$ are pre- and post-transformations, usually implemented as MLPs. Its spectral function is explicitly written as $g(\Tilde{\bmu{L}})$ of order $K$. 

\subsection{Relation to Other Taxonomy}
The spectral domain of graphs has long been incorporated into graph learning tasks \cite{ortega2018graph,dong2020graph,gandhi2021}.
In this study, we mainly focus on the form of $K$-order polynomial $g(\Tilde{\bmu{L}}) = \sum_{k=0}^{K} \theta_k\, T^{(k)}(\Tilde{\bmu{L}})$ as a feasible filter approximation for $\hat{g}$, since directly acquiring eigen-decomposition is prohibitive for graph-scale matrices in practice. 
Previous studies show that the expressiveness of polynomial filters is satisfactory for estimating arbitrary smooth signals \cite{shuman2013emerging,chen_rational_2018}. 
\rrc{The formulation is also advantageous in representing filter bank models in the same form.}

Other taxonomies are also proposed for categorizing spectral filters. The most inclusive description of spectral graph filters is derived from the full graph spectrum as presented in \cref{eq:sconv}. However, directly acquiring the graph spectrum through eigen-decomposition is prohibitive due to computational cost, rendering this framework distant from the practical graph filtering process in GNNs. As shown in \cref{sec:related}, \cref{eq:sgnn} serves as a $K$-order approximation to the full spectrum filtering. Relevant works are discussed in \cref{sseca:related}. 

\noindentparagraph{Three types in \cite{chen2023}.}
\cite{chen2023} develops a taxonomy containing linear, polynomial, and rational functions. Intuitively, both its linear and polynomial functions can be expressed by our polynomial formulation \cref{eq:sgnn}, as the linear function corresponds to the special case where $K=1$. The rational function is a more powerful filter formulation being capable of learning non-smooth signals, typically discontinuous ones. Nonetheless, computing the inverse of the graph matrix is impractical, and \textbf{rational filters are also reduced to the polynomial form in practical computation}. This can be achieved by the Neumann series on matrix inverse that $(\bmu{I} - \bmu{A})^{-1} = \sum_{k=0}^{\infty} \bmu{A}^k$, and truncate to $K$-order approximation. In \cref{seca:summary}, filters such as \ft{PPR}, \ft{HK}, and \ft{Horner} showcase such transformation. 

\noindentparagraph{Decomposition-based framework in \cite{bo2023b}.}
\rrc{\cite{bo2023b} mostly identifies spectral GNNs by the eigen-decomposition. In its theoretical framework and code implementation, this is performed by explicitly acquiring the graph spectrum $\Tilde{\bmu{L}} = \bmu{U} \bmu{\Lambda} \bmu{U}\tran$ and applying graph convolution $\hat{g}(\bmu{\Lambda})$ accordingly. 
When applied to graph signals, the full $n$-order eigen-decomposition performs the spectral convolution following \cref{eq:sconv}. In \cref{ssec:spectral_gnn}, we elaborate that \textbf{the polynomial form in \cref{eq:sgnn} serves as an effective truncated $K$-order approximation}. In fact, realistic graph information are largely available in the low-frequency domain with small $k$-s, while high-frequency components are mostly noise. The spectral information is further compressed during spectral processing on the $F$-dimension representation. These considerations render our framework based on the polynomial form a practical and effective approach in characterizing spectral filters.}

\subsection{Other Relevant GNN Models}
\label{sseca:related}

\noindentparagraph{Spectral Decomposition.}
A range of works are only applicable to the full eigen-decomposition. 
{SpectralCNN}~\cite{bruna2014spectral} employs trainable weight matrices as filters on the eigenspace, while {LanczosNet}~\cite{liao2019lanczosnet} integrates simple spatial operations and appends multi-scale capability. {SIGN}~\cite{SIGN} employs a sign invariant and permutation equivariant on the  Laplacian eigenvectors, which are viewed as the positional encodings and added as additional node features in the graphs. 
Alternatively, {Specformer}~\cite{bo2023specformer} exploits a transformer model to establish expressive spectral filtering. 
Nonetheless, we note that the full eigen-decomposition is largely prohibitive, resulting in weaker applicability of these models, especially on large graphs. Hence, we do not include them in our benchmark evaluation, while comparison for these models can be found in \cite{bo2023b} and \cite{chen2023}. 

\noindentparagraph{Alternative Spectral Filters.}
Signal processing techniques beyond the graph Fourier transform have been introduced to build spectral models, such as the wavelet and pyramid transforms \cite{xu2019graph,zheng2021framelets,Geng2023Pyramid}. Other research also sets the stage for adapting spectral properties or filtering pipelines to specific tasks and graph variants \cite{gao2023,stablegcn,SpecGN,DSGC}. Although these methods expand the scope of spectral GNNs, they are less typical considering their reliance on highly customized operations, usually without available implementation for large-scale datasets. 

\noindentparagraph{Task-Specific Models.} 
These methods set the stage for adapting the spectral properties or filtering pipelines to particular tasks, instead of focusing on novel filter designs. \textbf{GHRN}~\cite{gao2023} explores the heterophily in the spatial domain and the frequency in the spectral domain, which efficiently addresses the heterophily issue for graph anomaly detection. \textbf{StableGCNN}~\cite{stablegcn} provides a theoretical spectral perspective of existing GNN models which aims to analyze their stability and establish their generalization guarantees. 
\textbf{SpecGN}~\cite{SpecGN} applies a smoothing function on the graph spectrum to alleviate correlation among signals. 
\textbf{DSGC}~\cite{DSGC} explores the utility of the spectrum of both node and attribute afﬁnity graphs.

\section{Details on Spectral Filters}
\label{seca:summary}

\subsection{Fixed Filter}
\label{sseca:summary1}
\paragraph{\ft{Linear.}}
A layer of \textbf{GCN}~\cite{kipf2016semi} propagation $f(\Tilde{\bmu{A}}) = \bmu{I} + \Tilde{\bmu{A}}$ is equivalent to a single-hop of spectral convolution, \rrb{which represents a linear filter without high-order terms}. Recall that $\Tilde{\bmu{L}} = \bmu{I} - \Tilde{\bmu{A}}$, the filter can be expressed as:
\begin{equation*}
    g(\Tilde{\bmu{L}}) = 2\bmu{I} - \Tilde{\bmu{L}}.
\end{equation*}

\paragraph{\ft{Impulse.}}
\textbf{SGC}~\cite{wu19sim} and \textbf{gfNN}~\cite{nt2019revisiting} adopt a pre-propagation decoupled architecture, while \textbf{GZoom}~\cite{deng2020} applies post-propagation to achieve expansion of the closed-form filter $(\bmu{I}+\Tilde{\bmu{L}})^{-1}$. All these models result in a $J$-hop spatial diffusion operation as $f(\bmu{A}) = \Tilde{\bmu{A}}^J$, \rrb{which corresponds to the impulse signal with only the $J$-th term.}
By respectively examining bases $T^{(k)}(\Tilde{\bmu{L}}) = (\bmu{I} - \Tilde{\bmu{L}})^k$ and $T^{(k)}(\Tilde{\bmu{L}}) = \Tilde{\bmu{L}}^k$, we have two equivalent formulations of the filter:
\begin{alignat*}{2}
    g(\Tilde{\bmu{L}}) &= (\bmu{I} - \Tilde{\bmu{L}})^K,\quad 
    &&T^{(k)}(\Tilde{\bmu{L}}) = (\bmu{I} - \Tilde{\bmu{L}})^k,\;\\
    & &&\theta_0 = \theta_1 = \cdots = \theta_{K-1} = 0,\; \theta_K = 1;\text{ or}\\
    g(\Tilde{\bmu{L}}) &= \sum_{k=0}^{K} \theta_k\, \Tilde{\bmu{L}}^k,\quad
    &&T^{(k)}(\Tilde{\bmu{L}}) = \Tilde{\bmu{L}}^k,\;
    \theta_k = \binom{K}{k} (-1)^k,
\end{alignat*}
where $\binos{K}{k}$ is the binomial coefficient. 

\paragraph{\ft{Monomial.}}
\textbf{S$^2$GC}~\cite{zhu2021a} summarizes $K$-hop propagation results with uniform weights in decouple precomputation, \rrb{which is classifies as the monomial propagation $f(\bmu{A}) = \sum_{j=1}^{J} \xi_j \Tilde{\bmu{A}}^j$ with equal parameters $\xi_j = 1/(J+1)$}. \textbf{GRAND+}~\cite{feng2022} examines an approximate propagation to acquire the filter. Similarly, there are two commonly used spectral interpretations based on two bases:
\begin{alignat*}{2}
\small
    g(\Tilde{\bmu{L}}) &= \sum_{k=0}^{K} \theta_k\, (\bmu{I} - \Tilde{\bmu{L}})^k,\quad 
    &&T^{(k)}(\Tilde{\bmu{L}}) = (\bmu{I} - \Tilde{\bmu{L}})^k,\;\\[-4pt]
    & &&\theta_0 = \cdots = \theta_K = \frac{1}{K+1};\text{ or}\\
    g(\Tilde{\bmu{L}}) &= \sum_{k=0}^{K} \theta_k\, \Tilde{\bmu{L}}^k,\quad
    &&T^{(k)}(\Tilde{\bmu{L}}) = \Tilde{\bmu{L}}^k,\;\\[-4pt]
    & &&\theta_k = \frac{1}{K+1} \sum_{j=k}^{J} \binom{j}{k} (-1)^k,
\end{alignat*}

\paragraph{\ft{Personalized PageRank (PPR).}} 
\textbf{GLP}~\cite{li2019e} derives a closed-form $\hat{f}(\bmu{A}) = (\bmu{I}+\alpha \bmu{L})^{-1}$ from the \ft{auto regressive (AR)} filter \cite{tremblay2018}, while \rrb{\textbf{PPNP}~\cite{Klicpera2019} solves PPR~\cite{Page1999} as $\hat{f}(\Tilde{\bmu{A}}) = \alpha\big(\bmu{I} + (1-\alpha)\Tilde{\bmu{A}}\big)^{-1}$. While targeting at different problems, these two graph processing techniques are equivalent in essence.} $\alpha \in [0,1]$ is the coefficient for balancing the strength of neighbor propagation, that a larger $\alpha$ results in stronger node identity and weaker neighboring impact, and vice versa. 
In both works, the filter is approximated by a recursive calculation $\bmu{H}^{(j+1)} = \varphi( (1-\alpha)\Tilde{\bmu{A}} \bmu{H}^{(j)} + \alpha\bmu{H}^{(0)})$, which is widely accepted in later studies such as \textbf{GCNII}~\cite{ming2020}.  
The explicit spatial and spectral interpretations of the polynomial approximation are respectively:
\begin{equation*}
    f(\Tilde{\bmu{A}}) = \sum_{j=0}^J \alpha(1-\alpha)^j \Tilde{\bmu{A}}^j ;\quad
    g(\Tilde{\bmu{L}}) 
    = \sum_{k=0}^K \theta_k (\bmu{I} - \Tilde{\bmu{L}})^k,\quad
    \theta_k = \alpha(1-\alpha)^k.
\end{equation*}

In addition, approximate computations have been proposed by representative works including GDC~\cite{gasteiger2019}, PPRGo~\cite{bojchevski2020}, GRAND+~\cite{feng2022}. GBP~\cite{Chen2020a} and \textbf{AGP}~\cite{wang2021b} explored the adjacency aggregation under graph normalization. 

\paragraph{\ft{Heat Kernel (HK).}} 
\rrb{\textbf{GDC}~\cite{gasteiger2019} inspects the heat kernel PageRank (HKPR)~\cite{hkpr2007}} replacing the PPR calculation by an exponential parameter. Equivalent \ft{HK} filters are also studied in DGC~\cite{NEURIPS2021_2d95666e} and AGP~\cite{wang2021b}. Let $\alpha>0$ be the temperature coefficient, the filter $\hat{f}(\Tilde{\bmu{A}}) = e^{-\alpha \Tilde{\bmu{L}}}$ is expanded in spatial and spectral forms as:
\begin{equation*}
    f(\Tilde{\bmu{A}}) = \sum_{j=0}^J \frac{e^{-\alpha}\alpha^j}{j!} \Tilde{\bmu{A}}^j ;\quad
    g(\Tilde{\bmu{L}}) 
    = \sum_{k=0}^K \theta_k (\bmu{I} - \Tilde{\bmu{L}})^k,\quad
    \theta_k = \frac{e^{-\alpha}\alpha^k}{k!}.
\end{equation*}

\paragraph{\ft{Gaussian.}} 
\rrb{\textbf{G$^2$CN}~\cite{li2022e} uses the Gaussian filter~\cite{calcaterra2008approximating}} for better flexibility on capturing local information. When concentrating on low frequency, the closed-form propagation is $\hat{f}(\Tilde{\bmu{A}}) = e^{-\alpha(2\bmu{I} - \Tilde{\bmu{L}})}$.
Invoking Taylor expansion for the filter leads to:
\begin{equation*}
    f^{(j)}(\Tilde{\bmu{A}}) = \bmu{I} - \frac{\alpha}{J} (2\bmu{I} - \Tilde{\bmu{L}});\quad
    g(\Tilde{\bmu{L}}) 
    = \sum_{k=0}^K \theta_k (2\bmu{I} - \Tilde{\bmu{L}})^{k},\quad
    \theta_k = \frac{\alpha^k}{k!}.
\end{equation*}

\subsection{Variable Filter}
\label{sseca:summary2}
\paragraph{\ft{Linear.}}
\textbf{GIN}~\cite{xu2019} alters the iterative ridged adjacency propagation with a learnable scaling parameter $\theta > 0$ controlling the strength of self loops, i.e., skip connections. It proves that its propagation $f(\Tilde{\bmu{A}}) = (1+\xi)\bmu{I} + \Tilde{\bmu{A}}$ is more expressive with regard to the Weisfeiler-Lehman (WL) test. 
\textbf{AKGNN}~\cite{AKGNN} elaborates the expression in spectral domain, that the scaling parameter adaptively balances the threshold between high and low frequency.
We thereby obtain the linear spectral function for each layer as:
\begin{equation*}
    g(\Tilde{\bmu{L}}; \theta) = (1 + \theta)\bmu{I} - \Tilde{\bmu{L}}.
\end{equation*}


\paragraph{\ft{Monomial.}}
\rrb{The monomial scheme signifies the same operations on parameters of all terms.}
\textbf{DAGNN}~\cite{liu2020b} studies the scheme of assigning learnable parameters to each hop of the propagation, but implements a costly concatenation-based scheme. \textbf{GPRGNN}~\cite{chien2021} considers the iterative generalized PageRank computation \cite{NEURIPS2019_9ac1382f} under heterophily, which also produces the same variable monomial spectral filter formulated as $f(\bmu{A}) = \sum_{j=0}^{J} \xi_j \Tilde{\bmu{A}}^j$. We present two bases and corresponding relationship between spatial and spectral parameters when $J=K$:
\begin{alignat*}{2}
    g(\Tilde{\bmu{L}}; \theta) &= \sum_{k=0}^{K} \theta_k\, (\bmu{I} - \Tilde{\bmu{L}})^k,\quad 
    &&T^{(k)}(\Tilde{\bmu{L}}) = (\bmu{I} - \Tilde{\bmu{L}})^k,\;\\[-2pt]
    & &&\theta_k = \xi_j;\text{ or}\\
    g(\Tilde{\bmu{L}}; \theta) &= \sum_{k=0}^{K} \theta_k\, \Tilde{\bmu{L}}^k,\quad
    &&T^{(k)}(\Tilde{\bmu{L}}) = \Tilde{\bmu{L}}^k,\;\\[-4pt]
    & &&\theta_k = \sum_{j=k}^{J} \binom{j}{k} (-1)^k \xi_j.
\end{alignat*}
\cite{chien2021} also inspects the effect of initialization of the parameters $\xi_j,\theta_k$ on fitting heterophilous graph signals. 

\paragraph{\ft{Horner.}}
\rrb{Horner's method \cite{horner1815new} is a recursive algorithm to compute the summation of monomial bases with residual connections.} Consider adding a residual term in the layer-wise propagation corresponding to the \ft{Monomial} filter $\bmu{H}^{(j+1)} = \varphi( \Tilde{\bmu{A}} \bmu{H}^{(j)} + \xi_j\bmu{H}^{(0)})$. When the balancing parameter is fixed $\xi_j=\alpha/(1-\alpha)$, it is equivalent to the PPR computation in \cref{sseca:summary1}. When $\xi_j$ is variable, the model is regarded as \textbf{HornerGCN} which is introduced by \cite{guo2023a}. 
Alternatively, \textbf{ARMAGNN}~\cite{arma} utilizes the Auto Regressive Moving Average (ARMA) filter $\hat{f}(\bmu{A}) = \beta(\bmu{I}-\alpha\Tilde{\bmu{A}})^{-1}$ \cite{tremblay2018} as an approach to describe the residual connection learned by respective weights in iterative architecture. Their spectral filter is:
\begin{equation*}
    g(\Tilde{\bmu{L}}; \theta) 
    = \sum_{k=0}^{K} \theta_k\, (\bmu{I} - \Tilde{\bmu{L}})^k,\quad
    \theta_k = \xi_{K-k} .
\end{equation*}
Although it shares an identical spectral interpretation with the \ft{Monomial} filter, the explicit residual connection proves beneficial in guiding the learnable parameters to recognize node identity and alleviate over-smoothing throughout propagation.

\paragraph{\ft{Chebyshev.}}
\rrb{The Chebyshev basis is widely accepted for graph signal processing, which is powerful in producing a minimax polynomial approximation for the analytic functions \cite{shuman2011chebyshev,HAMMOND2011129}.} \textbf{ChebNet}~\cite{defferrard2016convolutional} utilizes it to replace the adjacency propagation in iterative network architecture so that $f^{(j)}(\Tilde{\bmu{A}}) = T^{(j)}(\bmu{I} - \Tilde{\bmu{A}})$. To adapt the basis to decoupled propagation with explicit variable parameters, \cite{he2022a} proposes \textbf{ChebBase}. The spectral expressiveness of these two models are the same, and the filter is:
\begin{align*}
    g(\Tilde{\bmu{L}}; \theta) 
    = \sum_{k=0}^K \theta_k T^{(k)}(\Tilde{\bmu{L}}),\quad
    &T^{(k)}(\Tilde{\bmu{L}}) = 2 \Tilde{\bmu{L}} T^{(k-1)}(\Tilde{\bmu{L}}) - T^{(k-2)}(\Tilde{\bmu{L}}),\,\\
    &T^{(1)}(\Tilde{\bmu{L}}) = \Tilde{\bmu{L}},\, T^{(0)}(\Tilde{\bmu{L}}) = \bmu{I}.
\end{align*}
The Chebyshev polynomial is expressed in the three-term recurrence relation, which is favorable for the GNN iterative propagation. One can also write the Chebyshev basis of the first kind as the closed-form expression $T^{(k)}(\lambda) = \cos(k \arccos{\lambda})$. 

\paragraph{\ft{Chebyshev Interpolation (ChebInterp).}}
\rrb{\textbf{ChebNetII}~\cite{he2022a} utilizes Chebyshev interpolation \cite{gil2007numerical} to modify the \ft{Chebyshev} filter parameter} for better approximation with generally decaying weights. For each Chebyshev basis $T^{(k)}(\Tilde{\bmu{L}})$, it appends the basis with $K$-order interpolation:
\begin{equation*}
    g(\Tilde{\bmu{L}}; \theta) 
    = \frac{2}{K+1} \sum_{k=0}^{K} \sum_{\kappa=0}^{K} \theta_\kappa T^{(k)}(x_\kappa)\, T^{(k)}(\Tilde{\bmu{L}}),\quad
    x_\kappa = \cos\left(\frac{\kappa+1 / 2}{K+1} \pi\right),
\end{equation*}
where $T^{(k)}(x_\kappa), T^{(k)}(\Tilde{\bmu{L}})$ follow the Chebyshev basis, and $x_\kappa$ are the Chebyshev nodes of $T^{(K+1)}$.

\paragraph{\ft{Clenshaw.}}
Similar to Horner's method, \textbf{ClenshawGCN}~\cite{guo2023a} applies Clenshaw algorithm on top of the \ft{Chebyshev} filter to incorporate explicit residual connections. Its spatial convolution is obtained as $\bmu{H}^{(j+1)} = \varphi\big( 2\Tilde{\bmu{A}} \bmu{H}^{(j)} - \bmu{H}^{(j-1)} + \xi_j\bmu{H}^{(0)} \big), \bmu{H}^{(-1)} = \bmu{H}^{(-2)} = \bmu{O}$. The form of spectral filter is related with Chebyshev polynomials of the second kind:
\begin{align*}
    g(\Tilde{\bmu{L}}; \theta) 
    = \sum_{k=0}^{K} \theta_k T^{(k)}(\Tilde{\bmu{L}}),\quad
    &T^{(k)}(\Tilde{\bmu{L}}) = 2 \Tilde{\bmu{L}} T^{(k-1)}(\Tilde{\bmu{L}}) - T^{(k-2)}(\Tilde{\bmu{L}}),\,\\
    &T^{(1)}(\Tilde{\bmu{L}}) = 2\Tilde{\bmu{L}},\, T^{(0)}(\Tilde{\bmu{L}}) = \bmu{I}.
\end{align*}
Alternatively, the closed-form definition of the Chebyshev basis of the second kind is $T^{(k)}(\cos\lambda) = \frac{\sin((k+1) \lambda)}{\sin{\lambda}}$. The relation to spatial parameters is $\theta_k = \xi_{K-k}$. 

\paragraph{\ft{Bernstein.}}
\textbf{BernNet}~\cite{he2021} pursues more interpretable spectral filters by thte \rrb{Bernstein polynomial approximation \cite{FAROUKI2012379}} and invokes constraints form prior knowledge to avoid ill-posed variable parameters. The spatial propagation is special as it applies two graph matrices instead of one. The filter with regard to Bernstein basis $T^{(k)}$ is:
\begin{equation*}
    g(\Tilde{\bmu{L}}; \theta) 
    = \sum_{k=0}^{K} \frac{\theta_k}{2^{K}}\, T^{(k)}(\Tilde{\bmu{L}}),\quad
    T^{(k)}(\Tilde{\bmu{L}}) = \binom{K}{k} (2\bmu{I} - \Tilde{\bmu{L}})^{K-k}\, \Tilde{\bmu{L}}^k,
\end{equation*}
where learnable parameters are initialized as $\theta_k = T^{(k)}(k/K)$. 

\paragraph{\ft{\ft{Legendre}.}} 
\textbf{LegendreNet}~\cite{Chen2023Improved} exploits the Legendre polynomials in an accumulation form of calculation similar to BernNet: 
\begin{equation*}
    g(\Tilde{\bmu{L}}; \theta) 
    = \sum_{k=0}^{K} \theta_k T^{(k)}(\Tilde{\bmu{L}}),\quad
    T^{(k)}(\Tilde{\bmu{L}}) = \frac{(-1)^k}{k!\binos{2k}{k}}\,   (2\bmu{I} - \Tilde{\bmu{L}})^{k}\, \Tilde{\bmu{L}}^k.
\end{equation*}
Based on the relation of Bernstein bases and Legendre polynomials \cite{FAROUKI2000145}, we can also express it in the recurrence form:
\begin{align*}
    g(\Tilde{\bmu{L}}; \theta) 
    = \sum_{k=0}^{K} \theta_k T^{(k)}(\Tilde{\bmu{L}}),\quad
    T^{(0)}(\Tilde{\bmu{L}}) = \bmu{I},\;
    T^{(1)}(\Tilde{\bmu{L}}) = \Tilde{\bmu{L}}&, \\
    T^{(k)}(\Tilde{\bmu{L}}) = \frac{2k-1}{k} \Tilde{\bmu{L}} T^{(k-1)}(\Tilde{\bmu{L}}) - \frac{k-1}{k} T^{(k-2)}(\Tilde{\bmu{L}})&.
\end{align*}

\paragraph{\ft{Jacobi.}} 
\textbf{JacobiConv}~\cite{wang2022b} utilizes the more general Jacobi basis, whereas Chebyshev and Legendre polynomials can be regarded as special cases. Intuitively, it provides more flexible weight functions with two hyperparameters $\alpha,\beta$ to adapt to different signals of the spectral graphs. Each polynomial term of JacobiConv can be formulated by the three-term recurrence as following:
\begin{align*}
   g(\Tilde{\bmu{L}}; \theta) 
    &= \sum_{k=0}^{K} \theta_k T^{(k)}(\Tilde{\bmu{L}}),\quad
   T^{(0)}(\Tilde{\bmu{L}}) = \bmu{I},\\
   T^{(1)}(\Tilde{\bmu{L}}) &= \frac{\alpha-\beta}2\bmu{I}+\frac{\alpha+\beta+2}{2} (\bmu{I} - \Tilde{\bmu L}), \\
   T^{(k)}(\Tilde{\bmu{L}}) &= 
  \delta_k (\bmu{I} - \Tilde{\bmu L}) T^{(k-1)}(\Tilde{\bmu{L}})
  + \delta_k^{\prime}\, T^{(k-1)}(\Tilde{\bmu{L}})
  - \delta_k^{\prime\prime}\, T^{(k-2)}(\Tilde{\bmu{L}}),
\end{align*}
where
{\small
\begin{align*}
    & \delta_k =\frac{(2k+\alpha+\beta)(2k+\alpha+\beta-1)}{2k(k+\alpha+\beta)},\, 
    \delta_k^{\prime} =\frac{(2k+\alpha+\beta-1)(\alpha^2-\beta^2)}{2k(k+\alpha+\beta)(2k+\alpha+\beta-2)},\\[2pt]
    & \delta_k^{\prime\prime} =\frac{(k+\alpha-1)(k+\beta-1)(2k+\alpha+\beta)}{k(k+\alpha+\beta)(2k+\alpha+\beta-2)}\, \text{ for } k \ge 2. 
\end{align*}
}

\paragraph{\ft{Favard.}} 
\rrb{\textbf{FavardGNN}~\cite{guo2023} exploits the Favard’s Theorem~\cite{gautschi2004orthogonal} to learn the polynomial basis from available space and ensure orthonormality}. It is achieved by a three-term recurrence form with multiple series of hop-dependent variable parameters $\theta, \alpha, \beta$:
\begin{align*}
\small
     g(\Tilde{\bmu{L}}; \theta) 
    &= \sum_{k=0}^{K} \theta_k T^{(k)}(\Tilde{\bmu{L}}),\;
    T^{(-1)}(\Tilde{\bmu{L}}) = \bmu{O},\;
    T^{(0)}(\Tilde{\bmu{L}}) = \frac{1}{\sqrt{\alpha_0}} \bmu{I},\\
    T^{(k)}(\Tilde{\bmu{L}}) &= \frac{1}{\sqrt{\alpha_k}} \left(
      (\bmu{I} - \Tilde{\bmu L}) T^{(k-1)}(\Tilde{\bmu{L}}) - \beta_k T^{(k-1)}(\Tilde{\bmu{L}}) - \sqrt{\alpha_{k-1}}\, T^{(k-2)}(\Tilde{\bmu{L}})
    \right). 
\end{align*}

\paragraph{\ft{OptBasis.}} 
\rrb{\textbf{OptBasisGNN}~\cite{guo2023} considers a \ul{Basis} to be \ul{Opt}imal} with regard to convergence rate in the graph signal denoising problem. By replacing the learnable parameters in FavardGNN with parameters derived from the current input signal $\bmu{h}^{(k)}$, it can approach the optimal basis without occurring additional overhead. 
\begin{alignat*}{2}
     g(\Tilde{\bmu{L}}; \theta) 
    &= \sum_{k=0}^{K} \theta_k T^{(k)}(\Tilde{\bmu{L}}),\quad
    T^{(-1)}(\Tilde{\bmu{L}}) &&= \bmu{O},\;
    T^{(0)}(\Tilde{\bmu{L}}) = \frac{1}{\|\bmu{h}^{(0)}\|} \bmu{I},\\
    T^{(k)}(\Tilde{\bmu{L}}) &= \frac{1}{\|\bmu{h}^{(k)}\|} \Big(
      (\bmu{I} - \Tilde{\bmu L}) T^{(k-1)}(\Tilde{\bmu{L}}) &&- \beta_{k-1} T^{(k-1)}(\Tilde{\bmu{L}}) \\ & &&- \|\bmu{h}^{(k-1)}\| T^{(k-2)}(\Tilde{\bmu{L}})
    \Big),\\
    \beta_{k-1} &= \langle (\bmu{I} - \Tilde{\bmu L})\bmu{h}^{(k-1)}, \bmu{h}^{(k-1)} \rangle .&& 
\end{alignat*}

\subsection{Filter Bank}
\label{sseca:summary3}

\paragraph{\ft{AdaGNN.}}
AdaGNN~\cite{dong2021} designs $Q=F$ adaptive filters by assigning feature-specific parameters to the \ft{Linear} filter $g_q(\Tilde{\bmu{L}}) = \bmu{I}-\gamma_q\Tilde{\bmu{L}}, 1\le q \le F$. The representation update is performed in an iterative manner as $\bmu{H}^{(j+1)}= {\bmu{H}}^{(j)}- \Tilde{\bmu{L}} {\bmu{H}}^{(j)}\bmu{\Gamma}^{(j)}$, 
where $\bmu{\Gamma}^{(j)} = \diag(\gamma_{1}^{(j)}, ..., \gamma_{F}^{(j)})$ containing the learnable feature-wise parameter for the $j$-th layer. 
Hence, considering a single layer, the corresponding aggregated filter in spectral domain is:
\begin{equation*}
    \mathscr{g}(\Tilde{\bmu{L}}; \gamma) 
    = \bigparallel_{q=1}^{F} (\bmu{I} - \gamma_q \Tilde{\bmu{L}} ),
\end{equation*}
where each filter $g_q(\Tilde{\bmu{L}})$ is only applied to the $q$-th feature, and $\bigparallel$ is the concatenation operator that combines filtering result tensors among all features. 

\paragraph{\ft{FBGNN.}}
FBGNN~\cite{luan2022a} introduces the concept of filter bank for combining multiple filters \cite{wang2020e} in spectral GNN under the context of graph heterophily. It designs a two-channel scheme using graph adjacency $T_1=\Tilde{\bmu{A}}$ as \ft{Linear} low-pass filter (LP) and Laplacian $T_2=\Tilde{\bmu{L}}$ for \ft{Linear} high-pass filter (HP) to learn the smooth and non-smooth components together. FBGNN adopts an iterative form with scalar parameters $\gamma_1, \gamma_2 \in [0,1]$ for weighted sum. By omitting components including transformation weights and non-linear activation functions, we give the equivalent spectral function for each layer as:
\begin{equation*}
    \mathscr{g}(\Tilde{\bmu{L}}; \gamma) 
    = \gamma_1 (\bmu{I} - \Tilde{\bmu{L}}) + \gamma_2\Tilde{\bmu{L}}.
\end{equation*}

\paragraph{\ft{ACMGNN.}}
ACMGNN~\cite{luan2022} extends FBGNN to three filters with the additional identity (ID) diffusion matrix $T_3=\bmu{I}$, which corresponds to an all-pass filter maintaining node identity throughout propagation. It has two variants of applying different relative order between transformation and propagation, which are denoted as ACMGNN-I and ACMGNN-II. 
Both of their single-layer spectral expressions can be simplified as:
\begin{equation*}
    \mathscr{g}(\Tilde{\bmu{L}}; \gamma) 
    = \gamma_1 (\bmu{I} - \Tilde{\bmu{L}}) + \gamma_2\Tilde{\bmu{L}} + \gamma_3\bmu{I}.
\end{equation*}

\paragraph{\ft{FAGCN.}} 
FAGCN~\cite{bo2021} combines two \ft{Linear} filters with bias for capturing low- and high-frequency signals as $T_1 = (\beta+1)\bmu{I}-\Tilde{\bmu{L}}$ and $T_2 = (\beta-1)\bmu{I} + \Tilde{\bmu{L}}$, where $\beta \in [0,1]$ is the scaling coefficient. Since both filters are linear to $\Tilde{\bmu{L}}$, the fused representation of each layer can be computed using only one propagation by employing attention mechanism. We generally write the channel-wise parameters as $\gamma_1,\gamma_2 \in [0,1]$ and $\gamma_1 + \gamma_2 = 1$. Then the spectral expression for one layer is:
\begin{equation*}
    \mathscr{g}(\Tilde{\bmu{L}}; \gamma) = \gamma_1 ((\beta+1)\bmu{I}-\Tilde{\bmu{L}}) + \gamma_2 ((\beta-1)\bmu{I} + \Tilde{\bmu{L}}).
\end{equation*}

\paragraph{\ft{G$^2$CN.}}
G$^2$CN~\cite{li2022e} derives the \ft{Gaussian} filters for high- and low-frequency concentration centers. Specifically, it adopts 2-hop propagation in each layer. Utilizing the decay coefficient $\alpha \in [0,1]$ and the scaling coefficient $\beta \in [0,1]$, we rewrite the layer-wise propagation as $f^{(j)}(\Tilde{\bmu{A}}) = \bmu{I} - {\alpha} ((1 \pm \beta)\bmu{I} - \Tilde{\bmu{L}})^2 /{J}$. The model integrates the $Q=2$ channels after the decoupled propagation, hence:
\begin{align*}
    \mathscr{g}(\Tilde{\bmu{L}}; \gamma) &= \gamma_1\sum_{k=0}^{\lfloor K/2\rfloor} \theta_{1,k}\, T_1^{(k)} + \gamma_2\sum_{k=0}^{\lfloor K/2\rfloor} \theta_{2,k}\, T_2^{(k)},\\ 
    & T_1^{(k)} = ((1+\beta_1)\bmu{I} - \Tilde{\bmu{L}})^{2k},\quad 
    \theta_{1,k} = \frac{\alpha_1^k}{k!}, \\
    & T_2^{(k)} = ((1-\beta_2)\bmu{I} - \Tilde{\bmu{L}})^{2k},\quad 
    \theta_{2,k} = \frac{\alpha_2^k}{k!}.
\end{align*}

\paragraph{\ft{GNN-LF/HF.}}
GNN-LF/HF~\cite{zhu2021} proposes a pair of generalized GNN propagations based on low- and high-passing filtering (LF/HF) on top of its unified optimization framework depicting a range of GNN designs. Intuitively, its filter formulation is similar to the \ft{PPR} scheme, except a $(\bmu{I} \pm \beta\Tilde{\bmu{L}})$ factor applying to the input signal for distinguishing low- and high-frequency components. 
We transform the original dual filters into one model with $Q=2$ channels by learning the balancing coefficients $\gamma_1,\gamma_2 \in [0,1]$ which adjust the relative strength between node identity and low/high frequency features.  
Consequently, the channels can be implemented in a shared adjacency-based propagation. We formulate the filter bank version of GNN-LF/HF as:
\begin{align*}
    \mathscr{g}(\Tilde{\bmu{L}}; \gamma) &= \gamma_1\sum_{k=0}^{K} \theta_{1,k}\, T_1^{(k)} + \gamma_2\sum_{k=0}^{K} \theta_{2,k}\, T_2^{(k)},\\
    & T_1^{(k)} = (\bmu{I} - \beta_1\Tilde{\bmu{L}})(\bmu{I} - \Tilde{\bmu{L}})^k,\quad
    \theta_{1,k} = \alpha_1(1-\alpha_1)^k, \\
    & T_2^{(k)} = (\bmu{I} + \beta_2\Tilde{\bmu{L}})(\bmu{I} - \Tilde{\bmu{L}})^k,\quad
    \theta_{2,k} = \alpha_2(1-\alpha_2)^k,
\end{align*}
where there are $\alpha_1, \alpha_2 \in [0,1], \beta_1 \in [0, 1/2], \beta_2 \in (0, +\infty)$ according to the optimization objective. 

\paragraph{\ft{FiGURe.}}
FiGURe~\cite{ekbote2023figure} suggests using filter bank to adapt the unsupervised settings where the graph information can assist filter formation. It considers up to $Q=4$ filters, i.e., \ft{Identity} $\bmu{I}$, \ft{Monomial} $\bmu{I}-\Tilde{\bmu{L}}$, \ft{Chebyshev}, and \ft{Bernstein} bases for the filter bank. 
In the first unsupervised stage, an embedding function $\gamma_q: \RR{n\times F} \rightarrow \RR{n\times F}$ is learned for each filter by maximizing the mutual information across all channels. The second stage of supervised graph representation learning fine-tunes another scalar weight $\gamma_q'$ to tailor for the downstream task, along with other trainable model parameters. Formulation of the eventual FiGURe filter can be written as:
\begin{equation*}
    \mathscr{g}(\Tilde{\bmu{L}}; \gamma, \theta) 
    = \sum_{q=1}^{\mathit{Q}} \gamma_q' \gamma_q \cdot g_q(\Tilde{\bmu{L}}; \theta),\quad
    g_q(\Tilde{\bmu{L}}; \theta) = \sum_{k=0}^{K} \theta_{q,k}\, T_q^{(k)}(\Tilde{\bmu{L}}),
\end{equation*}

\begin{figure}[tb]
\centering
    \includegraphics[width=0.95\columnwidth]{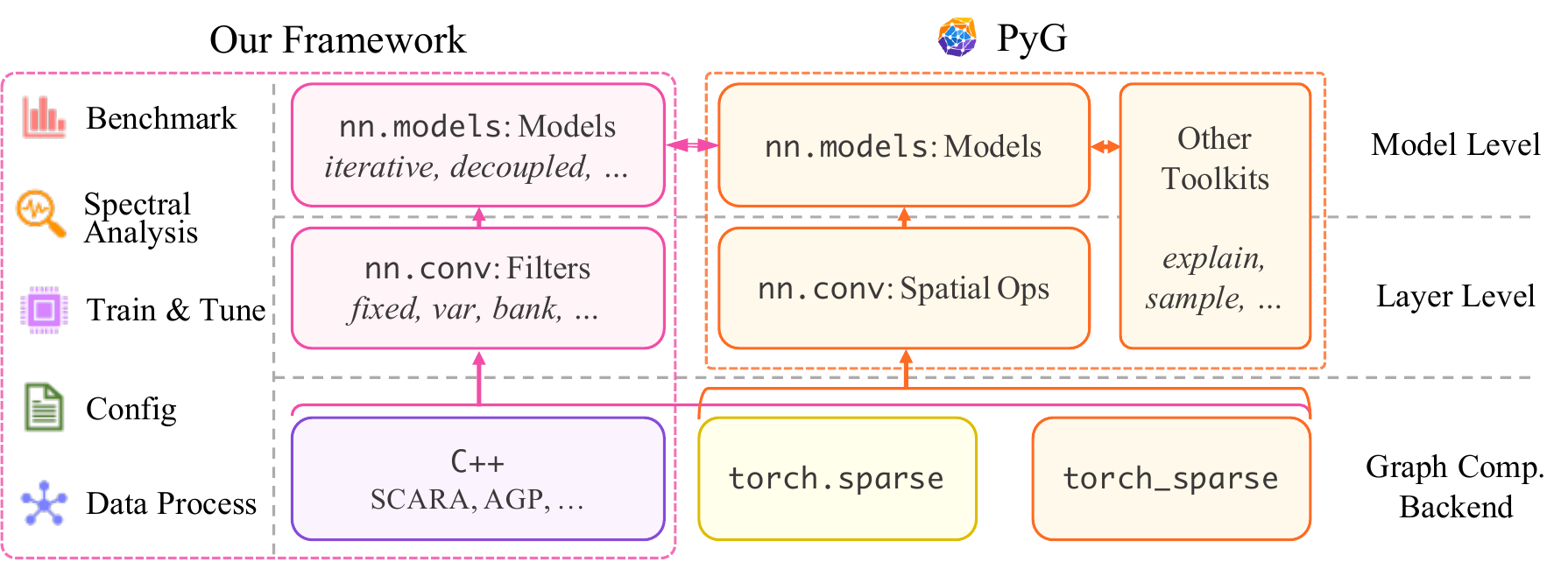}
    \caption{Code structure and relation to PyG. }
    \label{plot:code_struct}
\end{figure}

\balance
\section{Implementation Design}
\label{sseca:implement}

Based on the paradigm of spectral GNNs, we develop a unified framework oriented toward the spectral graph filters in particular, which is applicable to multiple GNN learning schemes. Our implementation embraces the modular design principle, offering similar framework arrangement and plug-and-play filter modules that can be seamlessly integrated into the popular graph learning toolkit PyG \cite{pyg}. It is also easily extendable to new filter designs, as only the spectral formulation in the form of \cref{eq:sgnn} needs to be implemented, while other model architectures and learning configurations can be effectively reused. Additionally, we include our reproducible benchmark pipeline for training and evaluating the models, with a particular focus on spectral analysis as well as scalable mini-batch training. 

Our framework design is based on the popular graph learning library PyTorch Geometric (PyG)~\cite{pyg} with the same arrangement of components. We list the following highlights of our framework compared to PyTorch Geometric and similar works \cite{duan2022comprehensive,ma2024}:
\begin{itemize}
    \item \textbf{Plug-and-play modules:} The model and layer implementation in our framework follows the same design as PyTorch Geometric, implying that they can be seamlessly integrated into PyG-based programs. Our rich collection of spectral models and filters greatly extends the PyG model zoo. 
    \item \textbf{Separated spectral kernels:} We decouple non-spectral designs and feature the pivotal graph filters. Most of the filters are thence applicable to be combined with various network architectures, learning schemes, and other toolkits, including those provided by PyG and PyG-based frameworks. 
    \item \textbf{High scalability:} As spectral GNNs are inherently suitable for large-scale learning, our framework is feasible to common scalable learning schemes and acceleration techniques. Several spectral-oriented approximation algorithms are also supported. 
\end{itemize}

We further elaborate on our code structure in \cref{plot:code_struct} by comparing corresponding components in PyG: 
\begin{itemize}
    \item We implement the spectral filters as \texttt{nn.conv} modules, which are pivotal and basic blocks for building spectral GNNs. This is similar to the implementation of spatial layers in PyG. 
    \item On the model level, common architectures formulated in \cref{ssec:model} are included in \texttt{nn.models}. Since these models are built in PyG style, they can seamlessly access other utility interfaces offered by PyG and PyG-based toolkits. 
    \item Regarding the underlying backend for sparse matrix multiplication in graph propagation, we support PyTorch-based computations inherited from PyG, as well as extendable interfaces for scalable C++-based algorithms specialized for spectral filtering. 
    \item On top of the framework, we also offer utilities for benchmarking, efficiency evaluation, and spectral analysis. 
\end{itemize}

\section{Full Experiment Results}
\label{seca:exp}


\subsection{Full-batch Learning}
\label{sseca:exp_full}
Efficiency evaluation of all full-batch models is shown in \cref{resa:eff_full}. More results on accuracy variance and efficiency difference are available in \cref{figa:box_full,figa:plat_full}, respectively. Note that \ds{obgn-arxiv} and \ds{arxiv-year} share the same underlying graph structure, so we only present the performance on \ds{obgn-arxiv}. 



\subsection{Mini-batch Learning}
\label{sseca:exp_mini}
\cref{resa:acc_mini} displays the efficacy results of applicable spectral filters under mini-batch training across all datasets, with accuracy variance shown in \cref{figa:boxmb_full}. 
Mini-batch efficiency is shown in \cref{resa:eff_mini} with the separated precomputation process. 

\onecolumn
\makeatletter
\setlength{\@fptop}{0pt plus 1fil}
\setlength{\@fpbot}{0pt plus 1fil}
\makeatother

\begin{table}[p]
\caption{Time and memory efficiency of \textit{full-batch} training on medium and large datasets. 
``Train'' and ``Infer'' respectively refer to average training time per epoch and inference time ($\si{ms}$), while ``RAM'', and ``GPU'' respectively refer to peak RAM and GPU memory ($\si{GB}$). 
For each column, highlighted ones are results ranked \rkat{first}, \rkbt{second}, and \rkct{third} within the error interval. 
}
\label{resa:eff_full}
\centering
\small
\newcommand{\dsh}[2]{\multicolumn{4}{c#1}{\ds{#2}}}
\renewcommand{\arraystretch}{0.9}
\setlength{\tabcolsep}{2pt}
\begin{adjustbox}{max width=\linewidth}
\begin{tabular}{c@{}c|cccc|cccc|cccc|cccc}
\toprule
    & \multirow{2}{*}{Filter} 
    & \dsh{|}{flickr} & \dsh{|}{penn94} & \dsh{|}{arxiv} & \dsh{}{twitch} \\ 
  ~ & ~ & \dshh{Train} & \dshh{Infer} & \dshh{RAM} & \dshh{GPU} 
  & \dshh{Train} & \dshh{Infer} & \dshh{RAM} & \dshh{GPU} 
  & \dshh{Train} & \dshh{Infer} & \dshh{RAM} & \dshh{GPU} 
  & \dshh{Train} & \dshh{Infer} & \dshh{RAM} & \dshh{GPU} 
\\\midrule
	& \rko{\ft{Identity}} & \rko{6.3\tpm{1.0}} & \rko{2.1\tpm{0.2}} & \rko{1.7\tpm{0.4}} & \rko{0.4\tpm{0.0}} & \rko{7.9\tpm{0.8}} & \rko{2.0\tpm{0.5}} & \rko{3.4\tpm{0.3}} & \rko{0.9\tpm{0.0}} & \rko{8.9\tpm{0.9}} & \rko{1.7\tpm{0.1}} & \rko{1.5\tpm{0.3}} & \rko{0.5\tpm{0.0}} & \rko{6.9\tpm{0.8}} & \rko{2.3\tpm{0.0}} & \rko{1.8\tpm{0.3}} & \rko{0.4\tpm{0.0}} \\
	& \ft{Linear} & 46.3\tpm{1.0} & 7.8\tpm{0.5} & \rka{1.7\tpm{0.4}} & 0.6\tpm{0.0} & 66.0\tpm{1.1} & \rka{4.7\tpm{0.2}} & \rka{3.5\tpm{0.4}} & \rkb{1.1\tpm{0.0}} & 95.8\tpm{0.8} & 7.0\tpm{0.4} & \rka{1.6\tpm{0.3}} & \rkc{0.9\tpm{0.0}} & 165.4\tpm{0.7} & 7.4\tpm{0.2} & \rka{1.8\tpm{0.4}} & \rkc{0.9\tpm{0.0}} \\
	& \ft{Impulse} & \rka{38.5\tpm{1.0}} & \rka{6.7\tpm{0.1}} & \rka{1.7\tpm{0.4}} & 0.6\tpm{0.0} & \rka{59.0\tpm{1.1}} & \rka{4.8\tpm{0.1}} & \rka{3.4\tpm{0.4}} & \rka{1.1\tpm{0.0}} & \rka{79.6\tpm{0.9}} & 7.1\tpm{0.2} & \rka{1.6\tpm{0.3}} & \rkc{0.9\tpm{0.0}} & \rkb{141.8\tpm{1.1}} & \rkc{6.8\tpm{0.2}} & \rka{1.8\tpm{0.4}} & \rka{0.9\tpm{0.0}} \\
	& \ft{Monomial} & 39.3\tpm{1.1} & \rkb{6.8\tpm{0.2}} & \rka{1.7\tpm{0.4}} & 0.6\tpm{0.0} & \rka{59.1\tpm{0.7}} & \rka{4.8\tpm{0.3}} & \rka{3.4\tpm{0.4}} & \rkb{1.1\tpm{0.0}} & \rka{79.6\tpm{0.5}} & 7.3\tpm{0.1} & \rka{1.6\tpm{0.3}} & \rkc{0.9\tpm{0.0}} & 143.2\tpm{1.0} & 7.5\tpm{0.1} & \rka{1.8\tpm{0.4}} & \rkc{0.9\tpm{0.0}} \\
	& \ft{PPR} & \rka{38.3\tpm{1.1}} & \rkb{6.9\tpm{0.0}} & \rka{1.7\tpm{0.4}} & \rkb{0.6\tpm{0.0}} & 63.4\tpm{1.9} & \rka{4.8\tpm{0.3}} & \rka{3.5\tpm{0.4}} & \rkb{1.1\tpm{0.0}} & \rka{79.1\tpm{0.6}} & \rkb{6.8\tpm{0.3}} & \rka{1.6\tpm{0.3}} & \rkc{0.9\tpm{0.0}} & \rkb{142.4\tpm{1.0}} & \rka{6.7\tpm{0.0}} & \rka{1.8\tpm{0.4}} & \rkc{0.9\tpm{0.0}} \\
	& \ft{HK} & \rka{38.1\tpm{1.0}} & 7.1\tpm{0.6} & \rka{1.7\tpm{0.4}} & 0.6\tpm{0.0} & \rka{59.3\tpm{1.1}} & 9.5\tpm{7.8} & \rka{3.4\tpm{0.4}} & \rkb{1.1\tpm{0.0}} & \rka{79.0\tpm{1.0}} & \rkb{6.8\tpm{0.0}} & \rka{1.6\tpm{0.3}} & \rka{0.7\tpm{0.0}} & \rkb{142.4\tpm{1.0}} & \rkc{7.0\tpm{0.8}} & \rka{1.7\tpm{0.5}} & \rkc{0.9\tpm{0.0}} \\
	& \ft{Gaussian} & \rka{38.7\tpm{1.3}} & 7.1\tpm{0.5} & \rka{1.7\tpm{0.4}} & 0.6\tpm{0.0} & 68.5\tpm{2.7} & \rka{4.8\tpm{0.2}} & \rka{3.5\tpm{0.4}} & \rkb{1.1\tpm{0.0}} & 80.2\tpm{0.7} & 7.1\tpm{0.1} & \rka{1.6\tpm{0.3}} & \rkc{0.9\tpm{0.0}} & \rka{140.7\tpm{0.9}} & 7.1\tpm{0.6} & \rka{1.8\tpm{0.4}} & \rka{0.9\tpm{0.0}} \\
\midrule
	& \ft{Linear} & 52.8\tpm{1.2} & 7.9\tpm{0.6} & \rka{1.7\tpm{0.4}} & 1.4\tpm{0.0} & 71.6\tpm{1.4} & 7.5\tpm{0.4} & \rka{3.4\tpm{0.4}} & 1.5\tpm{0.0} & 106.7\tpm{1.0} & 7.4\tpm{0.4} & \rka{1.6\tpm{0.3}} & 2.4\tpm{0.0} & 176.1\tpm{0.2} & 7.1\tpm{0.1} & \rka{1.8\tpm{0.3}} & 2.5\tpm{0.0} \\
	& \ft{Monomial} & 41.4\tpm{1.1} & 7.1\tpm{0.6} & \rka{1.7\tpm{0.4}} & 0.9\tpm{0.0} & 61.8\tpm{1.0} & 7.3\tpm{0.3} & \rka{3.5\tpm{0.4}} & 1.3\tpm{0.0} & 84.3\tpm{1.0} & 7.2\tpm{0.3} & \rka{1.6\tpm{0.3}} & 1.5\tpm{0.0} & 147.2\tpm{0.6} & 7.2\tpm{0.2} & \rka{1.8\tpm{0.4}} & 1.7\tpm{0.0} \\
	& \ft{Horner} & 54.6\tpm{1.5} & 10.5\tpm{0.7} & \rka{1.7\tpm{0.4}} & 1.1\tpm{0.0} & 68.7\tpm{1.1} & 10.0\tpm{0.4} & \rka{3.4\tpm{0.4}} & 1.3\tpm{0.0} & 109.9\tpm{1.0} & 9.2\tpm{0.2} & \rka{1.6\tpm{0.3}} & 1.9\tpm{0.0} & 170.6\tpm{0.6} & 9.6\tpm{0.4} & \rka{1.8\tpm{0.4}} & 1.7\tpm{0.0} \\
	& \ft{Chebyshev} & 46.0\tpm{1.2} & 9.2\tpm{0.5} & \rka{1.7\tpm{0.4}} & 1.0\tpm{0.0} & 63.2\tpm{0.9} & 8.8\tpm{0.1} & \rka{3.4\tpm{0.3}} & 1.3\tpm{0.0} & 94.0\tpm{1.0} & 10.6\tpm{4.9} & \rka{1.6\tpm{0.3}} & 1.6\tpm{0.0} & 153.5\tpm{0.3} & 8.5\tpm{0.0} & \rka{1.8\tpm{0.4}} & 1.8\tpm{0.0} \\
	& \ft{Clenshaw} & 61.9\tpm{1.1} & 10.9\tpm{0.1} & \rka{1.7\tpm{0.4}} & 1.2\tpm{0.0} & 72.2\tpm{0.9} & 11.7\tpm{0.3} & \rka{3.4\tpm{0.4}} & 1.3\tpm{0.0} & 123.7\tpm{1.0} & 10.8\tpm{0.6} & \rka{1.6\tpm{0.3}} & 1.8\tpm{0.0} & 185.4\tpm{0.8} & 10.4\tpm{0.1} & \rka{1.8\tpm{0.3}} & 1.9\tpm{0.0} \\
	& \ft{ChebInterp} & 67.3\tpm{1.9} & 14.1\tpm{0.1} & \rka{1.7\tpm{0.4}} & 1.0\tpm{0.0} & 82.9\tpm{1.1} & 11.2\tpm{0.4} & \rka{3.5\tpm{0.4}} & 1.3\tpm{0.0} & 126.0\tpm{0.5} & 71.6\tpm{99.9} & \rka{1.6\tpm{0.3}} & 1.6\tpm{0.0} & 175.8\tpm{2.5} & 15.6\tpm{4.3} & \rka{1.8\tpm{0.4}} & 1.8\tpm{0.0} \\
	& \ft{Bernstein} & 80.3\tpm{0.7} & 9.5\tpm{0.6} & \rka{1.7\tpm{0.4}} & 1.0\tpm{0.0} & 122.1\tpm{1.1} & 10.2\tpm{0.4} & \rka{3.5\tpm{0.4}} & 1.3\tpm{0.0} & 204.2\tpm{3.0} & 9.7\tpm{0.4} & \rka{1.6\tpm{0.3}} & 1.8\tpm{0.0} & 308.3\tpm{0.6} & 9.8\tpm{0.1} & \rka{1.8\tpm{0.4}} & 1.8\tpm{0.0} \\
	& \ft{Legendre} & 49.2\tpm{1.3} & 8.9\tpm{0.1} & \rka{1.7\tpm{0.4}} & 1.0\tpm{0.0} & 66.0\tpm{1.4} & 9.8\tpm{0.4} & \rka{3.3\tpm{0.5}} & 1.3\tpm{0.0} & 117.9\tpm{0.6} & 14.0\tpm{7.5} & \rka{1.6\tpm{0.3}} & 1.6\tpm{0.0} & 162.7\tpm{1.3} & 9.4\tpm{0.2} & \rka{1.8\tpm{0.4}} & 1.8\tpm{0.0} \\
	& \ft{Jacobi} & 55.7\tpm{1.4} & 9.2\tpm{0.3} & \rka{1.9\tpm{0.0}} & 1.0\tpm{0.0} & 68.5\tpm{1.6} & 10.1\tpm{0.0} & \rka{3.3\tpm{0.4}} & 1.3\tpm{0.0} & 113.1\tpm{0.9} & 9.5\tpm{0.2} & \rka{1.6\tpm{0.3}} & 1.7\tpm{0.0} & 173.2\tpm{1.0} & 9.5\tpm{0.6} & \rka{1.8\tpm{0.3}} & 1.8\tpm{0.0} \\
	& \ft{Favard} & 71.3\tpm{1.3} & 10.4\tpm{0.5} & \rka{1.7\tpm{0.4}} & 1.5\tpm{0.0} & 75.9\tpm{1.7} & 11.3\tpm{0.1} & \rka{3.4\tpm{0.3}} & 1.5\tpm{0.0} & 135.6\tpm{1.1} & 9.5\tpm{0.1} & \rka{1.6\tpm{0.3}} & 2.6\tpm{0.0} & 197.1\tpm{2.4} & 9.8\tpm{0.5} & \rka{1.8\tpm{0.4}} & 2.7\tpm{0.0} \\
	& \ft{OptBasis} & 154.9\tpm{1.1} & 12.8\tpm{0.6} & \rka{1.7\tpm{0.4}} & 2.3\tpm{0.0} & 113.6\tpm{1.1} & 13.3\tpm{0.8} & \rka{3.5\tpm{0.4}} & 1.9\tpm{0.0} & 349.1\tpm{0.8} & 13.2\tpm{0.1} & \rka{1.6\tpm{0.3}} & 4.2\tpm{0.0} & 413.3\tpm{0.8} & 12.7\tpm{0.2} & \rka{1.9\tpm{0.4}} & 4.1\tpm{0.0} \\
\midrule
	& \ft{AdaGNN} & 43.2\tpm{1.6} & \rkb{6.8\tpm{0.3}} & \rka{1.7\tpm{0.4}} & 1.0\tpm{0.0} & 79.9\tpm{1.1} & 6.6\tpm{0.1} & \rka{3.5\tpm{0.4}} & 1.3\tpm{0.0} & 87.8\tpm{1.4} & \rka{6.1\tpm{0.2}} & \rka{1.6\tpm{0.3}} & 1.6\tpm{0.0} & 148.6\tpm{1.5} & \rka{6.2\tpm{0.6}} & \rka{1.9\tpm{0.3}} & 1.7\tpm{0.0} \\
	& \ft{FBGNNI} & 147.3\tpm{1.7} & 13.0\tpm{0.4} & \rka{1.7\tpm{0.4}} & 2.9\tpm{0.0} & 185.6\tpm{1.6} & 12.4\tpm{2.1} & \rka{3.5\tpm{0.4}} & 2.2\tpm{0.0} & 290.4\tpm{0.9} & 10.8\tpm{0.4} & \rka{1.6\tpm{0.3}} & 5.2\tpm{0.0} & 434.6\tpm{2.8} & 10.0\tpm{0.1} & \rka{1.9\tpm{0.4}} & 5.3\tpm{0.0} \\
	& \ft{FBGNNII} & 149.1\tpm{0.7} & 12.4\tpm{0.3} & \rka{1.7\tpm{0.4}} & 3.8\tpm{0.0} & 176.9\tpm{24.2} & 10.2\tpm{1.1} & \rka{3.4\tpm{0.4}} & 2.6\tpm{0.0} & 289.9\tpm{0.6} & 13.1\tpm{1.1} & \rka{1.6\tpm{0.3}} & 7.0\tpm{0.0} & 435.6\tpm{2.1} & 10.8\tpm{1.3} & \rka{1.9\tpm{0.4}} & 7.0\tpm{0.0} \\
	& \ft{ACMGNNI} & 188.1\tpm{1.5} & 14.2\tpm{1.0} & \rka{1.7\tpm{0.4}} & 3.9\tpm{0.0} & 181.9\tpm{3.0} & 13.2\tpm{1.9} & \rka{3.5\tpm{0.4}} & 2.6\tpm{0.0} & 364.3\tpm{1.1} & 15.7\tpm{2.8} & \rka{1.6\tpm{0.3}} & 7.1\tpm{0.0} & 508.0\tpm{1.1} & 13.0\tpm{1.0} & \rka{1.8\tpm{0.4}} & 7.1\tpm{0.0} \\
	& \ft{ACMGNNII} & 189.7\tpm{1.6} & 14.8\tpm{0.4} & \rka{1.7\tpm{0.4}} & 4.8\tpm{0.0} & 230.4\tpm{1.3} & 13.1\tpm{0.3} & \rka{3.5\tpm{0.4}} & 3.1\tpm{0.0} & 366.1\tpm{0.7} & 14.9\tpm{1.5} & \rka{1.6\tpm{0.3}} & 8.9\tpm{0.0} & 510.7\tpm{1.2} & 13.7\tpm{0.6} & \rka{1.8\tpm{0.4}} & 8.9\tpm{0.0} \\
	& \ft{FAGNN} & 89.5\tpm{0.8} & 8.5\tpm{0.2} & \rka{1.7\tpm{0.4}} & \rkc{0.6\tpm{0.0}} & 128.4\tpm{1.1} & 5.8\tpm{0.2} & \rka{3.5\tpm{0.3}} & 1.1\tpm{0.0} & 186.6\tpm{0.7} & 9.0\tpm{0.1} & \rka{1.6\tpm{0.3}} & 1.0\tpm{0.0} & 334.9\tpm{1.4} & 8.8\tpm{1.1} & \rka{1.8\tpm{0.4}} & 1.0\tpm{0.0} \\
	& \ft{G$^2$CN} & 44.8\tpm{1.4} & \rkb{7.0\tpm{0.3}} & \rka{1.7\tpm{0.4}} & \rka{0.5\tpm{0.0}} & 70.7\tpm{0.8} & 5.6\tpm{0.1} & \rka{3.5\tpm{0.4}} & 1.1\tpm{0.0} & 94.3\tpm{1.1} & 8.0\tpm{0.2} & \rka{1.6\tpm{0.3}} & \rkb{0.9\tpm{0.0}} & 170.1\tpm{1.1} & 7.6\tpm{0.3} & \rka{1.8\tpm{0.4}} & 1.0\tpm{0.0} \\
	& \ft{GNN-LF/HF} & 80.7\tpm{1.1} & 8.9\tpm{0.4} & \rka{1.7\tpm{0.4}} & 0.7\tpm{0.0} & 123.9\tpm{0.8} & 6.0\tpm{0.5} & \rka{3.5\tpm{0.4}} & 1.1\tpm{0.0} & 170.0\tpm{0.9} & 8.6\tpm{1.0} & \rka{1.6\tpm{0.3}} & 1.0\tpm{0.0} & 318.8\tpm{0.7} & 7.8\tpm{0.0} & \rka{1.8\tpm{0.4}} & 1.1\tpm{0.0} \\
	& \ft{FiGURe} & 161.4\tpm{1.6} & 14.3\tpm{0.1} & \rka{1.7\tpm{0.4}} & 2.1\tpm{0.0} & 260.9\tpm{22.7} & 12.8\tpm{0.1} & \rka{3.4\tpm{0.4}} & 1.8\tpm{0.0} & 346.3\tpm{0.7} & 13.3\tpm{0.9} & \rka{1.6\tpm{0.3}} & 3.6\tpm{0.0} & 618.7\tpm{1.0} & 16.2\tpm{2.7} & \rka{1.9\tpm{0.4}} & 3.8\tpm{0.0} \\
\toprule
    & Filter
    & \dsh{|}{genius} & \dsh{|}{mag} & \dsh{|}{pokec} & \dsh{}{snap} \\
\midrule
        & \rko{\ft{Identity}} & \rko{11.0\tpm{0.9}} & \rko{3.1\tpm{0.1}} & \rko{1.4\tpm{0.1}} & \rko{0.9\tpm{0.0}} & \rko{60.3\tpm{0.9}} & \rko{4.5\tpm{0.7}} & \rko{2.9\tpm{0.3}} & \rko{7.3\tpm{0.0}} & \rko{46.3\tpm{0.9}} & \rko{2.9\tpm{0.2}} & \rko{5.3\tpm{0.3}} & \rko{4.1\tpm{0.0}} & \rko{108.1\tpm{1.9}} & \rko{3.1\tpm{1.5}} & \rko{10.3\tpm{0.4}} & \rko{8.8\tpm{0.0}} \\
	& \ft{Linear} & 114.2\tpm{0.9} & \rkc{7.1\tpm{0.2}} & \rka{1.4\tpm{0.1}} & 2.0\tpm{0.0} & 508.1\tpm{2.8} & 10.9\tpm{0.7} & \rka{2.9\tpm{0.3}} & \rka{7.2\tpm{0.0}} & 1071.8\tpm{0.7} & \rkc{8.7\tpm{0.0}} & \rka{5.3\tpm{0.4}} & \rkc{8.2\tpm{0.0}} & 1386.8\tpm{1.4} & \rkb{71.9\tpm{2.4}} & \rka{10.3\tpm{0.4}} & \rkc{15.6\tpm{0.0}} \\
	& \ft{Impulse} & \rka{84.0\tpm{0.9}} & 7.4\tpm{0.1} & \rka{1.4\tpm{0.1}} & \rka{1.7\tpm{0.0}} & \rka{431.8\tpm{5.0}} & 10.7\tpm{0.5} & \rka{2.9\tpm{0.3}} & \rka{7.1\tpm{0.1}} & \rkb{896.7\tpm{1.3}} & \rkb{8.3\tpm{0.1}} & \rka{5.3\tpm{0.4}} & \rkc{8.2\tpm{0.0}} & \rka{1133.8\tpm{10.4}} & \rkb{68.4\tpm{10.9}} & \rka{10.3\tpm{0.4}} & \rkc{15.6\tpm{0.0}} \\
	& \ft{Monomial} & \rka{84.3\tpm{1.0}} & \rkc{6.9\tpm{0.5}} & \rka{1.4\tpm{0.1}} & 2.0\tpm{0.0} & \rka{434.7\tpm{0.8}} & \rkb{10.4\tpm{0.1}} & \rka{3.0\tpm{0.3}} & \rka{7.2\tpm{0.0}} & \rkb{897.0\tpm{0.7}} & \rkc{8.4\tpm{0.4}} & \rka{5.3\tpm{0.4}} & \rkc{8.2\tpm{0.0}} & \rka{1134.7\tpm{4.9}} & \rka{10.9\tpm{4.1}} & \rka{10.3\tpm{0.4}} & \rka{14.0\tpm{0.0}} \\
	& \ft{PPR} & \rka{84.6\tpm{1.0}} & 7.3\tpm{0.1} & \rka{1.4\tpm{0.1}} & 2.0\tpm{0.0} & \rka{434.7\tpm{1.8}} & 10.9\tpm{0.9} & \rka{2.9\tpm{0.3}} & \rka{7.1\tpm{0.1}} & \rkb{897.3\tpm{1.1}} & 8.9\tpm{0.2} & \rka{5.3\tpm{0.4}} & \rkc{8.2\tpm{0.0}} & \rka{1142.7\tpm{1.6}} & \rkb{65.3\tpm{1.7}} & \rka{10.3\tpm{0.4}} & \rkc{15.6\tpm{0.0}} \\
	& \ft{HK} & \rka{84.6\tpm{1.1}} & \rkc{7.2\tpm{0.1}} & \rka{1.4\tpm{0.1}} & 2.0\tpm{0.0} & \rka{432.5\tpm{2.3}} & 10.6\tpm{0.3} & \rka{2.9\tpm{0.3}} & \rka{7.2\tpm{0.0}} & \rka{892.5\tpm{0.9}} & \rkc{8.5\tpm{0.1}} & \rka{5.3\tpm{0.4}} & \rka{6.3\tpm{0.0}} & \rka{1141.8\tpm{7.2}} & \rkb{64.9\tpm{3.5}} & \rka{10.3\tpm{0.4}} & \rkc{15.6\tpm{0.0}} \\
	& \ft{Gaussian} & \rka{83.8\tpm{1.0}} & \rkc{6.9\tpm{0.5}} & \rka{1.4\tpm{0.1}} & \rka{1.7\tpm{0.0}} & \rka{435.3\tpm{1.9}} & 10.7\tpm{0.5} & \rka{2.9\tpm{0.3}} & \rka{7.0\tpm{0.2}} & \rkb{897.0\tpm{1.1}} & \rkc{8.7\tpm{0.3}} & \rka{5.3\tpm{0.4}} & \rkb{7.2\tpm{0.0}} & \rka{1140.5\tpm{9.8}} & \rkb{50.0\tpm{33.7}} & \rka{10.3\tpm{0.4}} & \rkc{15.6\tpm{0.0}} \\
\midrule
        & \ft{Linear} & 138.3\tpm{0.8} & \rkc{7.2\tpm{0.5}} & \rka{1.4\tpm{0.2}} & 5.5\tpm{0.0} & 551.3\tpm{0.5} & 10.7\tpm{0.4} & \rka{2.9\tpm{0.3}} & 14.5\tpm{0.0} & \multicolumn{4}{c|}{(OOM)} & \multicolumn{4}{c}{(OOM)} \\
	& \ft{Monomial} & 96.3\tpm{1.1} & \rka{6.6\tpm{0.3}} & \rka{1.4\tpm{0.2}} & 3.6\tpm{0.0} & 447.4\tpm{1.5} & 10.7\tpm{0.3} & \rka{3.0\tpm{0.3}} & 10.3\tpm{0.0} & 944.0\tpm{0.9} & \rkc{8.8\tpm{0.5}} & \rka{5.3\tpm{0.4}} & 14.4\tpm{0.0} & \multicolumn{4}{c}{(OOM)} \\
	& \ft{Horner} & 157.3\tpm{1.8} & 9.6\tpm{0.4} & \rka{1.4\tpm{0.1}} & 4.4\tpm{0.0} & 563.7\tpm{2.9} & 13.5\tpm{0.8} & \rka{3.0\tpm{0.3}} & 10.9\tpm{0.0} & 1178.4\tpm{0.7} & 20.1\tpm{17.6} & \rka{5.3\tpm{0.4}} & 16.6\tpm{0.0} & \multicolumn{4}{c}{(OOM)} \\
	& \ft{Chebyshev} & 120.2\tpm{12.6} & 8.3\tpm{0.6} & \rka{1.4\tpm{0.1}} & 3.6\tpm{0.0} & 489.2\tpm{6.0} & 13.2\tpm{2.7} & \rka{2.9\tpm{0.3}} & 10.7\tpm{0.0} & 1013.2\tpm{1.0} & 9.3\tpm{0.4} & \rka{5.3\tpm{0.4}} & 15.2\tpm{0.0} & \multicolumn{4}{c}{(OOM)} \\
	& \ft{Clenshaw} & 200.6\tpm{4.9} & 10.8\tpm{0.1} & \rka{1.5\tpm{0.0}} & 4.6\tpm{0.0} & 631.4\tpm{6.9} & 14.2\tpm{0.9} & \rka{2.9\tpm{0.3}} & 10.9\tpm{0.0} & 1318.3\tpm{0.6} & 89.7\tpm{10.7} & \rka{5.3\tpm{0.4}} & 18.3\tpm{0.0} & \multicolumn{4}{c}{(OOM)} \\
	& \ft{ChebInterp} & 152.6\tpm{3.2} & 14.1\tpm{0.6} & \rka{1.4\tpm{0.2}} & 3.6\tpm{0.0} & 522.2\tpm{6.4} & 17.8\tpm{1.7} & \rka{2.9\tpm{0.3}} & 10.8\tpm{0.0} & 1081.4\tpm{0.7} & 13.2\tpm{0.1} & \rka{5.2\tpm{0.4}} & 14.2\tpm{0.0} & \multicolumn{4}{c}{(OOM)} \\
	& \ft{Bernstein} & 244.1\tpm{0.5} & 11.0\tpm{0.4} & \rka{1.4\tpm{0.1}} & 4.0\tpm{0.0} & 972.1\tpm{0.7} & 13.5\tpm{0.8} & \rka{2.9\tpm{0.3}} & 10.9\tpm{0.0} & 2238.9\tpm{0.8} & 44.7\tpm{0.7} & \rka{5.3\tpm{0.4}} & 17.5\tpm{0.0} & \multicolumn{4}{c}{(OOM)} \\
	& \ft{Legendre} & 168.2\tpm{1.3} & 9.5\tpm{0.3} & \rka{1.4\tpm{0.1}} & 3.8\tpm{0.0} & 519.7\tpm{1.1} & 12.8\tpm{0.4} & \rka{2.9\tpm{0.3}} & 10.5\tpm{0.0} & 1090.7\tpm{1.1} & 11.2\tpm{0.6} & \rka{5.3\tpm{0.4}} & 15.2\tpm{0.0} & \multicolumn{4}{c}{(OOM)} \\
	& \ft{Jacobi} & 160.9\tpm{2.3} & 10.7\tpm{0.0} & \rka{1.4\tpm{0.1}} & 3.8\tpm{0.0} & 566.0\tpm{1.6} & 12.8\tpm{0.5} & \rka{2.9\tpm{0.3}} & 10.6\tpm{0.0} & 1191.6\tpm{1.1} & 53.9\tpm{3.5} & \rka{5.3\tpm{0.4}} & 15.9\tpm{0.0} & \multicolumn{4}{c}{(OOM)} \\
	& \ft{Favard} & 230.7\tpm{3.7} & 24.3\tpm{13.4} & \rka{1.4\tpm{0.1}} & 6.1\tpm{0.0} & 673.1\tpm{0.5} & 12.8\tpm{0.6} & \rka{2.9\tpm{0.3}} & 14.9\tpm{0.0} & \multicolumn{4}{c|}{(OOM)} & \multicolumn{4}{c}{(OOM)} \\
	& \ft{OptBasis} & 941.5\tpm{2.3} & 12.8\tpm{0.9} & \rka{1.5\tpm{0.1}} & 10.0\tpm{0.0} & \multicolumn{4}{c|}{(OOM)} & \multicolumn{4}{c|}{(OOM)} & \multicolumn{4}{c}{(OOM)} \\
\midrule
        & \ft{AdaGNN} & 100.9\tpm{1.0} & \rka{6.2\tpm{0.6}} & \rka{1.4\tpm{0.1}} & 3.6\tpm{0.0} & 461.3\tpm{2.1} & \rka{9.6\tpm{0.6}} & \rka{2.9\tpm{0.3}} & 10.4\tpm{0.0} & 959.4\tpm{0.9} & \rka{7.2\tpm{0.5}} & \rka{5.3\tpm{0.4}} & 13.9\tpm{0.0} & \multicolumn{4}{c}{(OOM)} \\
	& \ft{FBGNNI} & 471.8\tpm{1.3} & 11.2\tpm{0.2} & \rka{1.4\tpm{0.1}} & 12.5\tpm{0.0} & \multicolumn{4}{c|}{(OOM)} & \multicolumn{4}{c|}{(OOM)} & \multicolumn{4}{c}{(OOM)} \\
	& \ft{FBGNNII} & 471.0\tpm{0.6} & 11.1\tpm{1.4} & \rka{1.4\tpm{0.1}} & 17.0\tpm{0.0} & \multicolumn{4}{c|}{(OOM)} & \multicolumn{4}{c|}{(OOM)} & \multicolumn{4}{c}{(OOM)} \\
	& \ft{ACMGNNI} & 649.1\tpm{0.7} & 14.4\tpm{0.1} & \rka{1.4\tpm{0.1}} & 17.2\tpm{0.0} & \multicolumn{4}{c|}{(OOM)} & \multicolumn{4}{c|}{(OOM)} & \multicolumn{4}{c}{(OOM)} \\
	& \ft{ACMGNNII} & 649.8\tpm{1.0} & 12.6\tpm{0.7} & \rka{1.4\tpm{0.1}} & 21.7\tpm{0.0} & \multicolumn{4}{c|}{(OOM)} & \multicolumn{4}{c|}{(OOM)} & \multicolumn{4}{c}{(OOM)} \\
	& \ft{FAGNN} & 221.2\tpm{0.8} & 9.3\tpm{0.1} & \rka{1.4\tpm{0.1}} & \rkc{1.9\tpm{0.0}} & 933.1\tpm{0.9} & 11.7\tpm{0.3} & \rka{2.9\tpm{0.3}} & \rka{7.2\tpm{0.0}} & 2163.3\tpm{0.2} & 9.4\tpm{0.1} & \rka{5.3\tpm{0.4}} & 9.0\tpm{0.0} & 2744.4\tpm{6.4} & \rkb{80.1\tpm{14.0}} & \rka{10.3\tpm{0.4}} & \rkb{15.3\tpm{0.0}} \\
	& \ft{G$^2$CN} & 98.5\tpm{0.8} & 7.8\tpm{0.1} & \rka{1.4\tpm{0.1}} & 2.2\tpm{0.0} & 502.0\tpm{1.0} & \rkb{10.4\tpm{0.2}} & \rka{2.9\tpm{0.3}} & \rka{7.2\tpm{0.0}} & 1072.9\tpm{0.8} & 9.1\tpm{0.1} & \rka{5.3\tpm{0.3}} & 9.0\tpm{0.0} & 1351.4\tpm{9.5} & 135.3\tpm{4.7} & \rka{10.3\tpm{0.4}} & 16.9\tpm{0.0} \\
	& \ft{GNN-LF/HF} & 181.4\tpm{1.4} & 8.9\tpm{1.1} & \rka{1.4\tpm{0.2}} & 2.2\tpm{0.0} & 860.1\tpm{2.8} & 11.4\tpm{0.3} & \rka{2.9\tpm{0.3}} & 7.4\tpm{0.1} & 2007.4\tpm{1.4} & 9.8\tpm{1.8} & \rka{5.3\tpm{0.4}} & 9.7\tpm{0.0} & 2480.8\tpm{14.4} & 141.1\tpm{7.9} & \rka{10.3\tpm{0.4}} & 16.0\tpm{0.0} \\
	& \ft{FiGURe} & 394.0\tpm{1.1} & 14.2\tpm{0.1} & \rka{1.4\tpm{0.1}} & 8.4\tpm{0.0} & 1763.1\tpm{2.5} & 105.9\tpm{17.1} & \rka{2.9\tpm{0.3}} & 19.1\tpm{0.0} & \multicolumn{4}{c|}{(OOM)} & \multicolumn{4}{c}{(OOM)} \\
\bottomrule
\end{tabular}
\end{adjustbox}
\end{table}

\clearpage

\begin{table}[p]
\captionsetup{font={small}}
\caption{Effectiveness results (\%) and standard deviations of spectral filters with \textit{mini-batch} training on all datasets. \rrc{For each dataset, results are highlighted based on the relative effectiveness among filters, where \rkoo{green} results are better.}
}
\label{resa:acc_mini}
    \centering
    \setlength{\tabcolsep}{1.2pt}
\begin{adjustbox}{max width=\linewidth}
\begin{tabular}{@{}c@{}c|ccccc cccc|c ccccc ccccc ccc@{}}
\toprule
    & Filter & \ds{cora} & \ds{citeseer} & \ds{pubmed} & \ds{mine} & \ds{questions} & \ds{tolokers} & \ds{arxiv} & \ds{mag} & \ds{products} & \ds{chameleon} & \ds{squirrel} & \ds{actor} & \ds{roman} & \ds{ratings} & \ds{flickr} & \ds{year} & \ds{penn94} & \ds{genius} & \ds{twitch} & \ds{pokec} & \ds{snap} & \ds{wiki}
\\ \midrule
	& \ft{Identity} & \tz{fde4d5}{66.54\tpm{2.35}} & \tz{fde4d5}{67.10\tpm{1.79}} & \tz{fcfeee}{87.48\tpm{0.56}} & \tz{fde4d5}{50.18\tpm{2.13}} & \tz{fdfeef}{66.39\tpm{1.49}} & \tz{fffdec}{74.39\tpm{0.31}} & \tz{fff8e5}{47.05\tpm{0.12}} & \tz{fff7e3}{55.28\tpm{0.31}} & \tz{fde4d5}{10.61\tpm{nan}} & \tz{fdfef0}{26.62\tpm{0.04}} & \tz{fee7d7}{29.10\tpm{5.13}} & \tz{fde4d5}{24.98\tpm{2.41}} & \tz{d7ebdb}{34.86\tpm{1.46}} & \tz{e8f4df}{63.64\tpm{0.47}} & \tz{d6ebda}{44.49\tpm{1.27}} & \tz{fee6d6}{35.69\tpm{0.25}} & \tz{ffffef}{72.27\tpm{0.49}} & \tz{fff3df}{85.31\tpm{1.29}} & \tz{cce5d7}{99.98\tpm{0.01}} & \tz{fffae7}{61.71\tpm{0.08}} & \tz{feedda}{30.39\tpm{0.03}} & \tz{f6fbe7}{35.21\tpm{0.13}} \\
	& \ft{Linear} & \tz{f0f8e3}{85.29\tpm{1.27}} & \tz{f9fcea}{74.66\tpm{1.11}} & \tz{fffbe9}{84.96\tpm{0.50}} & \tz{fffae8}{65.64\tpm{2.16}} & \tz{fff9e5}{63.13\tpm{2.83}} & \tz{e8f4df}{78.84\tpm{1.44}} & \tz{e3f2de}{52.58\tpm{0.63}} & \tz{eef7e1}{68.24\tpm{0.21}} & \tz{fee9d8}{13.06\tpm{0.92}} & \tz{fefef0}{26.59\tpm{0.01}} & \tz{fafdeb}{38.09\tpm{3.13}} & \tz{daeddc}{38.33\tpm{2.46}} & \tz{fff5e1}{25.99\tpm{1.07}} & \tz{feefdc}{31.28\tpm{1.14}} & \tz{fff7e2}{34.69\tpm{2.49}} & \tz{deefdd}{49.07\tpm{0.26}} & \tz{ffffef}{72.38\tpm{0.34}} & \tz{f7fbe8}{88.37\tpm{0.15}} & \tz{fee7d7}{72.77\tpm{2.62}} & \tz{feeedb}{54.98\tpm{0.38}} & \tz{fdfeef}{45.33\tpm{0.19}} & \tz{ecf6df}{37.69\tpm{0.42}} \\
	& \ft{Impulse} & \tz{e8f4df}{86.04\tpm{1.76}} & \tz{ffffef}{74.14\tpm{1.04}} & \tz{fffae7}{84.03\tpm{0.51}} & \tz{fff8e4}{62.99\tpm{0.79}} & \tz{f8fce9}{67.53\tpm{1.31}} & \tz{fde4d5}{66.54\tpm{5.00}} & \tz{fcfdee}{50.55\tpm{0.62}} & \tz{d8ecdb}{70.75\tpm{0.15}} & \tz{fffae8}{22.42\tpm{1.69}} & \tz{fffcea}{26.16\tpm{0.75}} & \tz{e8f4df}{39.83\tpm{2.21}} & \tz{cee7d8}{38.94\tpm{1.78}} & \tz{feefdc}{24.32\tpm{1.60}} & \tz{feecd9}{28.63\tpm{0.87}} & \tz{fffae8}{35.81\tpm{2.37}} & \tz{fffeee}{45.23\tpm{0.15}} & \tz{fff5e1}{60.52\tpm{0.49}} & \tz{f7fbe7}{88.38\tpm{0.34}} & \tz{fff0dd}{77.50\tpm{0.67}} & \tz{fff3e0}{57.84\tpm{0.11}} & \tz{e0f0dd}{53.56\tpm{0.18}} & \tz{fefef1}{32.88\tpm{1.47}} \\
	& \ft{Monomial} & \tz{e4f2de}{86.30\tpm{1.54}} & \tz{f9fcea}{74.66\tpm{1.88}} & \tz{daeddc}{89.46\tpm{0.47}} & \tz{fff1de}{58.46\tpm{5.64}} & \tz{cce5d7}{73.70\tpm{2.08}} & \tz{cce5d7}{81.61\tpm{1.30}} & \tz{fafceb}{50.80\tpm{0.23}} & \tz{d9ecdb}{70.66\tpm{0.11}} & \tz{e0f0dd}{32.79\tpm{0.84}} & \tz{fefef1}{26.59\tpm{0.00}} & \tz{fdfef0}{37.70\tpm{2.16}} & \tz{fffeed}{34.91\tpm{0.89}} & \tz{eff7e2}{32.64\tpm{1.41}} & \tz{e7f3de}{64.14\tpm{0.51}} & \tz{dbeddc}{43.83\tpm{0.89}} & \tz{dff0dd}{48.93\tpm{0.31}} & \tz{f9fcea}{75.03\tpm{0.32}} & \tz{fbfded}{88.15\tpm{0.25}} & \tz{ecf6e0}{94.36\tpm{6.64}} & \tz{fffceb}{63.82\tpm{0.10}} & \tz{fffded}{42.83\tpm{1.12}} & \tz{fee6d6}{23.07\tpm{2.46}} \\
	& \ft{PPR} & \tz{d7ebdb}{87.19\tpm{1.72}} & \tz{e9f5df}{75.53\tpm{1.50}} & \tz{eaf5df}{88.73\tpm{0.46}} & \tz{e8f4df}{81.53\tpm{2.86}} & \tz{fffeed}{65.36\tpm{1.91}} & \tz{d4e9d9}{80.90\tpm{1.19}} & \tz{fefef1}{50.31\tpm{0.11}} & \tz{dff0dd}{70.02\tpm{0.40}} & \tz{fff1de}{17.02\tpm{3.99}} & \tz{f7fbe7}{26.91\tpm{0.44}} & \tz{dbeddc}{40.73\tpm{2.78}} & \tz{fffeed}{34.89\tpm{1.29}} & \tz{f0f8e3}{32.49\tpm{1.22}} & \tz{d1e8d9}{69.81\tpm{0.59}} & \tz{f6fbe6}{39.72\tpm{1.00}} & \tz{e4f2de}{48.56\tpm{0.31}} & \tz{f9fcea}{75.03\tpm{0.36}} & \tz{e5f3de}{89.06\tpm{0.49}} & \tz{dceedc}{97.34\tpm{0.32}} & \tz{fffae8}{62.18\tpm{0.04}} & \tz{f6fbe7}{47.83\tpm{3.67}} & \tz{fde4d5}{22.40\tpm{2.00}} \\
	& \ft{HK} & \tz{e0f0dd}{86.60\tpm{1.39}} & \tz{f7fbe8}{74.76\tpm{0.96}} & \tz{dceedc}{89.39\tpm{0.41}} & \tz{fffdec}{68.48\tpm{1.02}} & \tz{dbeddc}{71.99\tpm{1.90}} & \tz{f7fbe7}{76.79\tpm{0.89}} & \tz{fbfdec}{50.69\tpm{0.27}} & \tz{fffae8}{58.91\tpm{0.43}} & \tz{fffeee}{25.72\tpm{0.67}} & \tz{fefef0}{26.61\tpm{0.76}} & \tz{cce5d7}{41.63\tpm{3.14}} & \tz{f7fbe8}{36.37\tpm{1.94}} & \tz{edf6e1}{32.88\tpm{1.39}} & \tz{d9ecdb}{67.82\tpm{0.49}} & \tz{fffceb}{36.66\tpm{1.16}} & \tz{f0f8e2}{47.50\tpm{0.30}} & \tz{ffffef}{72.77\tpm{0.76}} & \tz{e4f2de}{89.10\tpm{0.14}} & \tz{daeddb}{97.79\tpm{1.10}} & \tz{fffae8}{62.04\tpm{0.08}} & \tz{ebf5df}{50.99\tpm{0.16}} & \tz{cce5d7}{43.06\tpm{0.30}} \\
	& \ft{Gaussian} & \tz{fee6d6}{67.86\tpm{1.81}} & \tz{fcfeee}{74.44\tpm{1.23}} & \tz{eaf5df}{88.70\tpm{0.55}} & \tz{e5f3de}{82.16\tpm{1.65}} & \tz{fde4d5}{57.80\tpm{1.95}} & \tz{fee9d8}{67.78\tpm{4.30}} & \tz{f3f9e4}{51.45\tpm{0.24}} & \tz{ecf6df}{68.61\tpm{0.29}} & \tz{fffeee}{25.70\tpm{0.70}} & \tz{fefef1}{26.59\tpm{0.00}} & \tz{f0f8e3}{39.08\tpm{2.93}} & \tz{ffffef}{35.32\tpm{1.60}} & \tz{edf6e1}{32.84\tpm{1.28}} & \tz{e2f1dd}{65.38\tpm{1.55}} & \tz{fff7e3}{34.76\tpm{2.36}} & \tz{eaf5df}{48.03\tpm{0.21}} & \tz{fcfeee}{73.93\tpm{0.45}} & \tz{f6fbe6}{88.44\tpm{0.40}} & \tz{d0e8d8}{99.32\tpm{0.10}} & \tz{fffae8}{62.20\tpm{0.05}} & \tz{e4f2de}{52.58\tpm{0.04}} & \tz{cee6d7}{42.81\tpm{0.45}} \\
\midrule
	& \ft{Linear} & \tz{fff7e2}{76.36\tpm{0.79}} & \tz{fff9e5}{71.79\tpm{1.40}} & \tz{fbfdec}{87.60\tpm{0.47}} & \tz{fde4d5}{50.57\tpm{1.89}} & \tz{fffeed}{65.34\tpm{4.59}} & \tz{fff7e2}{71.50\tpm{0.35}} & \tz{fff7e3}{46.63\tpm{0.18}} & \tz{fff7e3}{54.85\tpm{0.32}} & \tz{ffffef}{26.44\tpm{0.84}} & \tz{ffffef}{26.49\tpm{0.23}} & \tz{ffffef}{37.36\tpm{3.17}} & \tz{fffeee}{34.93\tpm{1.94}} & \tz{cce5d7}{35.72\tpm{0.88}} & \tz{e9f5df}{63.34\tpm{0.79}} & \tz{fffbe9}{36.27\tpm{0.83}} & \tz{fde4d5}{35.03\tpm{0.50}} & \tz{fafdeb}{74.63\tpm{0.67}} & \tz{fff8e4}{86.10\tpm{0.15}} & \tz{fdfeef}{89.90\tpm{5.42}} & \tz{fffae8}{62.21\tpm{0.16}} & \tz{feedda}{30.47\tpm{0.34}} & \tz{e5f2de}{38.98\tpm{0.51}} \\
	& \ft{Monomial} & \tz{d4e9d9}{87.41\tpm{1.24}} & \tz{d8ecdb}{76.26\tpm{1.35}} & \tz{d2e8d9}{89.83\tpm{0.53}} & \tz{fffceb}{67.66\tpm{3.43}} & \tz{fffae7}{63.67\tpm{4.81}} & \tz{fefff1}{75.35\tpm{2.85}} & \tz{d2e8d9}{53.62\tpm{0.41}} & \tz{f7fbe7}{67.05\tpm{0.47}} & \tz{deefdd}{33.11\tpm{0.37}} & \tz{f9fcea}{26.82\tpm{0.42}} & \tz{fff8e4}{33.99\tpm{3.79}} & \tz{fffeee}{35.07\tpm{2.04}} & \tz{d8ecdb}{34.81\tpm{1.34}} & \tz{cee6d7}{70.63\tpm{1.80}} & \tz{fde4d5}{30.84\tpm{4.82}} & \tz{ecf6df}{47.92\tpm{0.41}} & \tz{d5eada}{82.63\tpm{0.54}} & \tz{d6ebda}{89.51\tpm{0.47}} & \tz{fff7e3}{81.87\tpm{6.42}} & \tz{d0e8d8}{78.11\tpm{0.70}} & \tz{edf6e1}{50.35\tpm{0.07}} & \tz{fffdec}{31.34\tpm{0.48}} \\
	& \ft{Horner} & \tz{e0f0dd}{86.64\tpm{1.63}} & \tz{edf6e1}{75.33\tpm{0.80}} & \tz{fffae7}{84.17\tpm{0.46}} & \tz{fff7e3}{62.14\tpm{1.32}} & \tz{fff9e6}{63.35\tpm{2.36}} & \tz{fff6e2}{71.22\tpm{1.21}} & \tz{e5f2de}{52.47\tpm{0.30}} & \tz{fffff2}{65.53\tpm{0.63}} & \tz{f0f8e3}{30.04\tpm{0.47}} & \tz{fefef1}{26.59\tpm{0.00}} & \tz{fffdec}{36.40\tpm{2.74}} & \tz{f4fae5}{36.62\tpm{1.69}} & \tz{fff3df}{25.49\tpm{1.47}} & \tz{feecd9}{28.40\tpm{1.69}} & \tz{fde4d5}{30.91\tpm{4.71}} & \tz{fffeed}{44.84\tpm{0.19}} & \tz{fde4d5}{47.09\tpm{0.65}} & \tz{f7fbe7}{88.38\tpm{0.14}} & \tz{fff0dd}{77.82\tpm{0.67}} & \tz{fffbe9}{62.87\tpm{0.25}} & \tz{fff9e6}{38.73\tpm{1.71}} & \tz{fffdec}{31.42\tpm{0.37}} \\
	& \ft{Chebyshev} & \tz{cce5d7}{87.89\tpm{0.87}} & \tz{e5f2de}{75.71\tpm{1.63}} & \tz{cce5d7}{90.06\tpm{0.33}} & \tz{d1e8d9}{88.36\tpm{4.03}} & \tz{f2f9e4}{68.64\tpm{4.79}} & \tz{daeddb}{80.33\tpm{1.35}} & \tz{fffae8}{47.99\tpm{1.14}} & \tz{d5eada}{71.05\tpm{0.21}} & \tz{fff6e2}{19.54\tpm{5.50}} & \tz{fefef1}{26.58\tpm{0.01}} & \tz{fff8e5}{34.27\tpm{3.39}} & \tz{fffceb}{34.23\tpm{1.21}} & \tz{fffeed}{29.89\tpm{1.82}} & \tz{fff2de}{35.06\tpm{2.65}} & \tz{fffeed}{37.15\tpm{2.22}} & \tz{ffffef}{45.54\tpm{0.49}} & \tz{d0e8d8}{83.40\tpm{0.40}} & \tz{fffae8}{86.70\tpm{0.19}} & \tz{cce5d7}{99.92\tpm{0.05}} & \tz{cce5d7}{78.95\tpm{0.13}} & \tz{fffded}{42.83\tpm{0.96}} & \tz{feead9}{24.26\tpm{2.39}} \\
	& \ft{Clenshaw} & \tz{fffcea}{80.83\tpm{2.22}} & \tz{fffbe9}{72.60\tpm{0.96}} & \tz{fde4d5}{75.75\tpm{1.07}} & \tz{feebd9}{54.17\tpm{2.07}} & \tz{fffbe9}{63.99\tpm{1.80}} & \tz{ffffef}{75.03\tpm{1.64}} & \tz{fde4d5}{41.35\tpm{0.99}} & \tz{fff7e2}{54.34\tpm{0.15}} & \tz{f2f9e4}{29.74\tpm{0.94}} & \tz{fefef1}{26.59\tpm{0.00}} & \tz{fffae7}{35.06\tpm{2.57}} & \tz{fffeee}{35.18\tpm{0.97}} & \tz{feeedb}{23.93\tpm{0.55}} & \tz{fde4d5}{21.42\tpm{0.60}} & \tz{cce5d7}{45.63\tpm{1.05}} & \tz{fffded}{44.64\tpm{2.77}} & \tz{fde5d6}{48.39\tpm{2.59}} & \tz{fffbe9}{86.98\tpm{0.07}} & \tz{feead9}{74.34\tpm{4.28}} & \tz{fee9d8}{52.79\tpm{2.66}} & \tz{f8fce9}{47.18\tpm{0.20}} & \tz{fffdec}{31.25\tpm{0.10}} \\
	& \ft{ChebInterp} & \tz{f7fbe7}{84.70\tpm{3.59}} & \tz{f8fce9}{74.73\tpm{1.89}} & \tz{dfefdd}{89.28\tpm{0.43}} & \tz{cde6d7}{89.48\tpm{0.95}} & \tz{fde5d6}{58.05\tpm{1.94}} & \tz{feedda}{68.83\tpm{0.97}} & \tz{cfe7d8}{53.79\tpm{0.64}} & \tz{fefef0}{65.73\tpm{0.89}} & \tz{f9fcea}{28.26\tpm{0.84}} & \tz{cce5d7}{28.11\tpm{1.17}} & \tz{fde4d5}{28.37\tpm{3.09}} & \tz{fffded}{34.62\tpm{1.98}} & \tz{ffffef}{30.36\tpm{1.22}} & \tz{cce5d7}{70.95\tpm{0.69}} & \tz{fffae7}{35.58\tpm{2.22}} & \tz{fffdec}{44.28\tpm{0.29}} & \tz{fff8e4}{63.59\tpm{0.70}} & \tz{fffeee}{87.65\tpm{0.38}} & \tz{cee6d7}{99.71\tpm{0.45}} & \tz{cee6d7}{78.60\tpm{0.68}} & \tz{fcfdee}{45.77\tpm{0.56}} & \tz{fee9d8}{23.85\tpm{1.84}} \\
	& \ft{Bernstein} & \tz{fffded}{82.05\tpm{1.46}} & \tz{feeedb}{69.18\tpm{1.83}} & \tz{fffceb}{85.46\tpm{0.45}} & \tz{f0f8e3}{77.98\tpm{1.77}} & \tz{fff7e3}{62.56\tpm{3.79}} & \tz{fffceb}{74.01\tpm{1.19}} & \tz{fff0dd}{44.58\tpm{2.06}} & \tz{fde4d5}{39.92\tpm{0.71}} & \tz{feefdc}{16.07\tpm{0.32}} & \tz{fefef1}{26.59\tpm{0.17}} & \tz{eaf5df}{39.61\tpm{3.21}} & \tz{fefef1}{35.68\tpm{2.35}} & \tz{fefef0}{30.68\tpm{1.05}} & \tz{ffffef}{53.87\tpm{1.09}} & \tz{fde5d6}{31.24\tpm{2.14}} & \tz{feecda}{37.34\tpm{1.11}} & \tz{e5f2de}{79.72\tpm{0.34}} & \tz{fcfeee}{88.12\tpm{0.47}} & \tz{dceedc}{97.43\tpm{0.55}} & \tz{e7f3de}{73.61\tpm{0.02}} & \tz{f3f9e5}{48.67\tpm{0.28}} & \tz{feecd9}{24.54\tpm{1.95}} \\
	& \ft{Legendre} & \tz{d4e9d9}{87.43\tpm{1.00}} & \tz{daeddb}{76.17\tpm{1.02}} & \tz{d1e8d9}{89.84\tpm{0.29}} & \tz{fff8e5}{63.32\tpm{1.81}} & \tz{fffbe9}{64.14\tpm{2.24}} & \tz{fcfeee}{75.78\tpm{1.02}} & \tz{fcfeee}{50.53\tpm{0.20}} & \tz{cee7d8}{71.65\tpm{0.13}} & \tz{d2e9d9}{34.74\tpm{0.45}} & \tz{f2f9e4}{27.10\tpm{0.36}} & \tz{f4fae5}{38.70\tpm{2.49}} & \tz{fafceb}{36.15\tpm{2.02}} & \tz{ecf6df}{33.05\tpm{1.45}} & \tz{f0f8e2}{61.08\tpm{2.63}} & \tz{f5fae6}{39.94\tpm{1.27}} & \tz{edf6e1}{47.73\tpm{0.30}} & \tz{f2f9e4}{76.88\tpm{0.41}} & \tz{e3f1de}{89.15\tpm{0.42}} & \tz{dceedc}{97.32\tpm{4.52}} & \tz{ebf5df}{72.60\tpm{1.49}} & \tz{fffded}{42.94\tpm{0.54}} & \tz{fffceb}{31.16\tpm{0.40}} \\
	& \ft{Jacobi} & \tz{d8ecdb}{87.17\tpm{1.50}} & \tz{edf6e0}{75.34\tpm{1.26}} & \tz{d4e9d9}{89.75\tpm{0.33}} & \tz{cce5d7}{89.84\tpm{0.91}} & \tz{dff0dd}{71.42\tpm{3.01}} & \tz{f7fbe7}{76.80\tpm{0.72}} & \tz{d8ecdb}{53.29\tpm{0.17}} & \tz{d1e8d9}{71.41\tpm{0.09}} & \tz{dff0dd}{32.96\tpm{0.23}} & \tz{fde4d5}{24.37\tpm{3.14}} & \tz{f6fbe7}{38.54\tpm{3.73}} & \tz{fcfeee}{35.86\tpm{1.80}} & \tz{e9f4df}{33.32\tpm{2.14}} & \tz{cce5d7}{71.05\tpm{0.67}} & \tz{fffeed}{37.06\tpm{1.47}} & \tz{cce5d7}{50.29\tpm{0.23}} & \tz{cce5d7}{84.06\tpm{0.43}} & \tz{cce5d7}{89.79\tpm{0.10}} & \tz{d4eada}{98.61\tpm{2.03}} & \tz{e9f4df}{73.05\tpm{2.02}} & \tz{e1f0dd}{53.37\tpm{2.23}} & \tz{fdfeef}{33.25\tpm{3.10}} \\
	& \ft{OptBasis} & \tz{fffded}{82.20\tpm{2.46}} & \tz{fffceb}{73.11\tpm{1.23}} & \tz{f2f9e4}{88.23\tpm{1.10}} & \tz{d0e8d8}{88.56\tpm{1.13}} & \tz{fde4d5}{57.70\tpm{5.70}} & \tz{e0f0dd}{79.62\tpm{2.38}} & \tz{f9fceb}{50.81\tpm{0.68}} & \tz{dbeddc}{70.46\tpm{0.24}} & \tz{eaf5df}{31.29\tpm{1.68}} & \tz{fffff2}{26.56\tpm{0.06}} & \tz{fafdeb}{38.09\tpm{2.54}} & \tz{fffeee}{35.00\tpm{1.62}} & \tz{fffceb}{29.34\tpm{1.61}} & \tz{fafdec}{56.85\tpm{2.90}} & \tz{e8f4df}{42.15\tpm{1.08}} & \tz{fff6e2}{40.87\tpm{0.89}} & \tz{daeddb}{81.81\tpm{0.99}} & \tz{fefef1}{88.00\tpm{0.57}} & \tz{fffcea}{86.13\tpm{0.50}} & \tz{cee6d7}{78.67\tpm{0.11}} & \tz{cce5d7}{57.49\tpm{2.42}} & \tz{fffcea}{30.87\tpm{3.11}} \\
\midrule
	& \ft{FAGNN} & \tz{dbeddc}{86.95\tpm{1.13}} & \tz{f2f9e4}{75.07\tpm{1.05}} & \tz{fffcea}{85.25\tpm{0.28}} & \tz{fafceb}{73.82\tpm{0.93}} & \tz{fff5e1}{62.01\tpm{2.39}} & \tz{eff7e2}{77.91\tpm{0.59}} & \tz{cce5d7}{53.94\tpm{0.21}} & \tz{deefdd}{70.11\tpm{0.33}} & \tz{cce5d7}{35.55\tpm{0.37}} & \tz{fefef1}{26.59\tpm{0.01}} & \tz{eaf5df}{39.66\tpm{2.47}} & \tz{e2f1dd}{37.91\tpm{2.05}} & \tz{fffae8}{28.23\tpm{1.78}} & \tz{fffae8}{46.59\tpm{1.97}} & \tz{ecf6e0}{41.44\tpm{0.75}} & \tz{fdfeef}{45.85\tpm{0.96}} & \tz{fffcea}{68.49\tpm{0.88}} & \tz{fcfdee}{88.14\tpm{0.17}} & \tz{fff4e0}{79.92\tpm{1.70}} & \tz{fffbe9}{62.60\tpm{0.12}} & \tz{fffae7}{39.29\tpm{0.66}} & \tz{fffceb}{31.23\tpm{0.55}} \\
	& \ft{G$^2$CN} & \tz{e2f1dd}{86.51\tpm{1.79}} & \tz{f2f9e4}{75.05\tpm{1.11}} & \tz{fffae8}{84.42\tpm{0.51}} & \tz{fbfdec}{73.11\tpm{1.09}} & \tz{d5eada}{72.65\tpm{2.74}} & \tz{fff0dd}{69.70\tpm{6.19}} & \tz{f7fbe8}{51.04\tpm{0.24}} & \tz{f4fae5}{67.34\tpm{0.41}} & \tz{f2f9e4}{29.68\tpm{0.27}} & \tz{fdfeef}{26.65\tpm{0.13}} & \tz{daeddb}{40.79\tpm{3.47}} & \tz{cce5d7}{39.08\tpm{2.93}} & \tz{fde4d5}{21.47\tpm{0.88}} & \tz{e2f1dd}{65.53\tpm{0.74}} & \tz{fff3df}{33.84\tpm{1.18}} & \tz{d2e9d9}{49.85\tpm{0.36}} & \tz{e1f0dd}{80.55\tpm{0.45}} & \tz{d6ebda}{89.50\tpm{0.46}} & \tz{fff2de}{78.72\tpm{1.21}} & \tz{fde4d5}{50.69\tpm{0.05}} & \tz{fde4d5}{25.89\tpm{0.16}} & \tz{e7f3de}{38.66\tpm{0.10}} \\
	& \ft{GNN-LF/HF} & \tz{d4eada}{87.36\tpm{1.02}} & \tz{dff0dd}{75.95\tpm{1.35}} & \tz{d3e9d9}{89.76\tpm{0.39}} & \tz{f3f9e5}{76.77\tpm{7.55}} & \tz{cce5d7}{73.70\tpm{1.52}} & \tz{e3f2de}{79.35\tpm{0.95}} & \tz{fffae7}{47.65\tpm{0.45}} & \tz{daeddc}{70.51\tpm{0.06}} & \tz{f3f9e5}{29.41\tpm{0.22}} & \tz{fff9e5}{25.80\tpm{0.91}} & \tz{dceedc}{40.66\tpm{2.14}} & \tz{fffeed}{34.75\tpm{1.98}} & \tz{e3f1de}{33.91\tpm{1.61}} & \tz{fffae7}{45.11\tpm{5.00}} & \tz{fdfef0}{38.04\tpm{1.84}} & \tz{d8ecdb}{49.47\tpm{0.38}} & \tz{f7fbe7}{75.72\tpm{0.44}} & \tz{edf6e0}{88.79\tpm{0.15}} & \tz{cce5d7}{99.91\tpm{0.07}} & \tz{fffbe9}{62.95\tpm{0.65}} & \tz{fefef1}{44.82\tpm{1.81}} & \tz{f3f9e5}{35.97\tpm{1.65}} \\
	& \ft{FiGURe} & \tz{d7ebdb}{87.21\tpm{1.13}} & \tz{cce5d7}{76.67\tpm{1.33}} & \tz{d4e9d9}{89.73\tpm{0.49}} & \tz{fff1de}{57.93\tpm{3.89}} & \tz{e3f2de}{70.88\tpm{1.87}} & \tz{f3f9e4}{77.36\tpm{3.03}} & \tz{e4f2de}{52.51\tpm{0.36}} & \tz{cce5d7}{71.87\tpm{0.18}} & \tz{d4eada}{34.43\tpm{1.13}} & \tz{fffceb}{26.24\tpm{1.03}} & \tz{eef7e1}{39.33\tpm{4.26}} & \tz{ddeedc}{38.21\tpm{2.33}} & \tz{e8f4df}{33.45\tpm{2.07}} & \tz{e9f4df}{63.48\tpm{6.27}} & \tz{f9fcea}{39.09\tpm{1.17}} & \tz{cce5d7}{50.26\tpm{0.33}} & \tz{cee7d8}{83.70\tpm{0.38}} & \tz{fde4d5}{83.21\tpm{0.18}} & \tz{fde4d5}{71.26\tpm{3.64}} & \tz{e7f3de}{73.58\tpm{2.53}} & \tz{fffceb}{41.76\tpm{0.66}} & \tz{e7f3de}{38.57\tpm{0.76}} \\
\bottomrule
\end{tabular}
\end{adjustbox}
\vspace{6mm}
\end{table}

\begin{table}[p]
\captionsetup{font={small}}
\caption{Time and memory efficiency of \textit{mini-batch} training on medium- and large-scale datasets. 
``Pre.'' refers to precomputation time ($\si{s}$), ``Train'' and ``Infer'' respectively refer to average training time per epoch and inference time ($\si{ms}$), while ``RAM'', and ``GPU'' respectively refer to peak RAM and GPU memory ($\si{GB}$). 
For each column, highlighted ones are results ranked \rkat{first}, \rkbt{second}, and \rkct{third} within the error interval. . 
}
\label{resa:eff_mini}
\centering
\newcommand{\dsh}[2]{\multicolumn{5}{c#1}{\ds{#2}}}
\setlength{\tabcolsep}{1.25pt}
\renewcommand{\arraystretch}{0.95}
\begin{adjustbox}{max width=\linewidth}
\begin{tabular}{@{}c@{}c|ccccc|ccccc|ccccc|ccccc|ccccc@{}}
\toprule
    & \multirow{2}{*}{Filter} 
    & \dsh{|}{flickr} & \dsh{|}{penn94} & \dsh{|}{arxiv} & \dsh{|}{twitch} & \dsh{}{genius} \\ 
  ~ & ~ & \dshh{Pre.} & \dshh{Train} & \dshh{Infer} & \dshh{RAM} & \dshh{GPU} 
  & \dshh{Pre.} & \dshh{Train} & \dshh{Infer} & \dshh{RAM} & \dshh{GPU} 
  & \dshh{Pre.} & \dshh{Train} & \dshh{Infer} & \dshh{RAM} & \dshh{GPU} 
  & \dshh{Pre.} & \dshh{Train} & \dshh{Infer} & \dshh{RAM} & \dshh{GPU} 
  & \dshh{Pre.} & \dshh{Train} & \dshh{Infer} & \dshh{RAM} & \dshh{GPU} 
\\\midrule
	& \rko{\ft{Identity}} & \rko{0.0\tpm{0.0}} & \rko{21.3\tpm{2.7}} & \rko{6.0\tpm{0.9}} & \rko{1.8\tpm{0.2}} & \rko{0.0\tpm{0.0}} & \rko{0.0\tpm{0.0}} & \rko{14.8\tpm{0.8}} & \rko{4.1\tpm{0.6}} & \rko{3.6\tpm{0.2}} & \rko{0.2\tpm{0.0}} & \rko{0.0\tpm{0.0}} & \rko{33.1\tpm{1.0}} & \rko{9.8\tpm{2.3}} & \rko{1.7\tpm{0.2}} & \rko{0.1\tpm{0.0}} & \rko{0.0\tpm{0.0}} & \rko{32.6\tpm{1.8}} & \rko{8.8\tpm{0.9}} & \rko{2.5\tpm{0.2}} & \rko{0.0\tpm{0.0}} & \rko{0.0\tpm{0.0}} & \rko{96.8\tpm{13.8}} & \rko{40.1\tpm{13.6}} & \rko{1.4\tpm{0.1}} & \rko{0.0\tpm{0.0}} \\
	& \ft{Linear} & 1.8\tpm{0.0} & \rka{22.1\tpm{2.3}} & \rka{6.8\tpm{0.2}} & \rka{2.4\tpm{0.3}} & \rka{0.0\tpm{0.0}} & 16.0\tpm{0.7} & 17.1\tpm{0.9} & \rka{5.3\tpm{0.7}} & 6.6\tpm{0.2} & \rka{0.2\tpm{0.0}} & 0.9\tpm{0.0} & 41.7\tpm{2.4} & \rkc{13.1\tpm{2.5}} & \rka{1.9\tpm{0.3}} & \rka{0.1\tpm{0.0}} & \rkc{0.3\tpm{0.4}} & 40.1\tpm{5.6} & 14.3\tpm{4.8} & \rka{2.5\tpm{0.2}} & \rkc{0.0\tpm{0.0}} & 0.2\tpm{0.3} & \rka{87.2\tpm{10.2}} & \rka{33.1\tpm{9.1}} & \rka{1.5\tpm{0.1}} & \rka{0.0\tpm{0.0}} \\
	& \ft{Impulse} & \rkc{1.3\tpm{0.1}} & \rkc{24.1\tpm{1.3}} & 7.9\tpm{1.7} & \rka{2.3\tpm{0.3}} & \rka{0.0\tpm{0.0}} & \rka{12.0\tpm{0.4}} & \rka{16.0\tpm{0.5}} & \rka{5.0\tpm{0.9}} & \rka{5.9\tpm{0.2}} & \rka{0.2\tpm{0.0}} & \rka{0.6\tpm{0.1}} & \rkb{40.2\tpm{0.7}} & \rkc{13.4\tpm{0.8}} & \rka{1.8\tpm{0.3}} & \rka{0.1\tpm{0.0}} & \rkc{0.4\tpm{0.4}} & \rkc{37.4\tpm{2.1}} & \rka{12.5\tpm{1.9}} & \rka{2.5\tpm{0.2}} & 0.0\tpm{0.0} & 1.0\tpm{0.6} & 109.9\tpm{31.4} & 46.5\tpm{14.1} & \rka{1.5\tpm{0.1}} & \rka{0.0\tpm{0.0}} \\
	& \ft{Monomial} & \rka{1.2\tpm{0.0}} & \rkc{22.6\tpm{3.8}} & \rka{6.9\tpm{0.6}} & \rka{2.3\tpm{0.3}} & \rka{0.0\tpm{0.0}} & 12.9\tpm{0.5} & \rka{16.5\tpm{0.9}} & \rka{5.3\tpm{0.5}} & \rka{5.9\tpm{0.2}} & \rka{0.2\tpm{0.0}} & \rka{0.6\tpm{0.1}} & \rkb{39.1\tpm{1.5}} & \rkc{14.8\tpm{4.1}} & \rka{1.8\tpm{0.3}} & \rka{0.1\tpm{0.0}} & \rkc{0.3\tpm{0.3}} & \rka{35.5\tpm{0.7}} & \rka{11.9\tpm{0.8}} & \rka{2.5\tpm{0.2}} & \rka{0.0\tpm{0.0}} & 0.4\tpm{0.5} & \rkc{102.5\tpm{16.1}} & \rka{40.0\tpm{15.0}} & \rka{1.5\tpm{0.1}} & \rka{0.0\tpm{0.0}} \\
	& \ft{PPR} & \rkb{1.2\tpm{0.0}} & \rka{20.2\tpm{2.1}} & \rkc{7.0\tpm{0.6}} & \rka{2.3\tpm{0.3}} & \rka{0.0\tpm{0.0}} & \rka{12.4\tpm{0.4}} & \rka{16.6\tpm{0.8}} & \rka{4.7\tpm{0.8}} & \rka{5.8\tpm{0.2}} & \rka{0.2\tpm{0.0}} & \rka{0.6\tpm{0.0}} & \rkb{40.0\tpm{0.8}} & \rka{12.8\tpm{1.6}} & \rka{1.8\tpm{0.3}} & \rka{0.1\tpm{0.0}} & \rka{0.2\tpm{0.0}} & 51.9\tpm{6.8} & 16.9\tpm{4.4} & \rka{2.5\tpm{0.2}} & 0.0\tpm{0.0} & 0.3\tpm{0.4} & \rka{92.5\tpm{10.0}} & \rka{34.3\tpm{7.1}} & \rka{1.5\tpm{0.1}} & \rka{0.0\tpm{0.0}} \\
	& \ft{HK} & \rkc{1.2\tpm{0.0}} & \rkc{24.3\tpm{4.4}} & 7.7\tpm{1.8} & \rka{2.3\tpm{0.3}} & \rka{0.0\tpm{0.0}} & \rka{12.3\tpm{0.6}} & \rka{16.3\tpm{1.1}} & \rka{5.3\tpm{0.4}} & \rka{5.9\tpm{0.2}} & \rka{0.2\tpm{0.0}} & \rka{0.6\tpm{0.0}} & \rkb{40.2\tpm{0.6}} & \rka{12.7\tpm{0.4}} & \rka{1.8\tpm{0.3}} & \rka{0.1\tpm{0.0}} & \rkc{0.4\tpm{0.3}} & \rka{35.0\tpm{2.0}} & \rka{12.0\tpm{1.2}} & \rka{2.5\tpm{0.2}} & \rka{0.0\tpm{0.0}} & 0.6\tpm{0.7} & 113.4\tpm{15.6} & 50.6\tpm{16.8} & \rka{1.5\tpm{0.1}} & \rka{0.0\tpm{0.0}} \\
	& \ft{Gaussian} & \rkc{1.3\tpm{0.1}} & \rkc{24.8\tpm{1.3}} & 8.5\tpm{2.0} & \rka{2.4\tpm{0.3}} & \rka{0.0\tpm{0.0}} & \rka{12.3\tpm{0.6}} & \rka{16.5\tpm{0.5}} & \rka{5.4\tpm{0.8}} & \rka{5.9\tpm{0.2}} & \rka{0.2\tpm{0.0}} & \rka{0.6\tpm{0.0}} & \rka{37.7\tpm{1.0}} & \rkc{15.4\tpm{4.2}} & \rka{1.8\tpm{0.3}} & \rka{0.1\tpm{0.0}} & \rka{0.2\tpm{0.2}} & 42.8\tpm{6.5} & \rka{12.7\tpm{2.3}} & \rka{2.5\tpm{0.2}} & 0.0\tpm{0.0} & 0.2\tpm{0.2} & 111.3\tpm{18.0} & 55.9\tpm{28.9} & \rka{1.5\tpm{0.1}} & \rka{0.0\tpm{0.0}} \\
\midrule
	& \ft{Var-Linear} & 1.4\tpm{0.1} & 55.5\tpm{0.7} & 27.0\tpm{5.8} & 5.4\tpm{0.3} & 0.2\tpm{0.0} & 13.2\tpm{1.1} & 26.1\tpm{2.3} & 15.0\tpm{8.1} & 19.4\tpm{0.2} & 1.8\tpm{0.0} & \rka{0.6\tpm{0.0}} & 67.8\tpm{1.0} & 43.4\tpm{3.8} & 3.5\tpm{0.3} & 0.1\tpm{0.0} & 1.4\tpm{0.0} & 152.7\tpm{29.8} & 67.4\tpm{8.7} & \rka{2.4\tpm{0.3}} & 0.0\tpm{0.0} & 0.3\tpm{0.3} & 183.7\tpm{23.5} & 147.2\tpm{31.6} & 1.9\tpm{0.2} & 0.0\tpm{0.0} \\
	& \ft{Var-Monomial} & \rkc{1.2\tpm{0.2}} & 60.7\tpm{2.3} & 23.0\tpm{5.2} & 5.4\tpm{0.3} & 0.2\tpm{0.0} & \rka{12.4\tpm{1.3}} & 32.5\tpm{0.6} & 11.0\tpm{1.3} & 19.4\tpm{0.2} & 1.8\tpm{0.0} & \rka{0.6\tpm{0.1}} & 76.7\tpm{1.0} & 41.2\tpm{8.9} & 3.4\tpm{0.3} & 0.1\tpm{0.0} & 1.2\tpm{0.3} & 143.6\tpm{3.5} & 47.7\tpm{4.4} & \rka{2.4\tpm{0.3}} & 0.0\tpm{0.0} & 0.3\tpm{0.3} & 190.3\tpm{7.2} & 100.0\tpm{34.0} & 1.9\tpm{0.2} & 0.0\tpm{0.0} \\
	& \ft{Horner} & \rkc{1.2\tpm{0.2}} & 51.0\tpm{0.6} & 30.4\tpm{0.1} & 5.4\tpm{0.3} & 0.2\tpm{0.0} & 13.7\tpm{0.7} & 31.6\tpm{0.8} & 18.2\tpm{3.7} & 19.8\tpm{0.3} & 1.8\tpm{0.0} & \rka{0.7\tpm{0.1}} & 79.6\tpm{3.1} & 61.0\tpm{11.4} & 3.5\tpm{0.4} & 0.1\tpm{0.0} & 1.9\tpm{0.2} & 164.6\tpm{18.2} & 84.0\tpm{23.7} & \rka{2.4\tpm{0.3}} & 0.0\tpm{0.0} & \rkb{0.1\tpm{0.0}} & 207.2\tpm{28.8} & 169.3\tpm{37.6} & 1.9\tpm{0.2} & 0.0\tpm{0.0} \\
	& \ft{Chebyshev} & 1.6\tpm{0.1} & 55.4\tpm{4.0} & 21.6\tpm{3.1} & 5.4\tpm{0.3} & 0.2\tpm{0.0} & 15.1\tpm{0.8} & 32.4\tpm{0.6} & 11.5\tpm{0.5} & 19.3\tpm{0.4} & 1.8\tpm{0.0} & 0.8\tpm{0.1} & 74.6\tpm{2.0} & 35.1\tpm{8.3} & 3.5\tpm{0.4} & 0.1\tpm{0.0} & 1.6\tpm{0.1} & 139.9\tpm{9.8} & 52.2\tpm{13.2} & \rka{2.4\tpm{0.3}} & 0.0\tpm{0.0} & 0.6\tpm{0.3} & 157.7\tpm{14.0} & 81.2\tpm{10.9} & 1.9\tpm{0.2} & 0.0\tpm{0.0} \\
	& \ft{Clenshaw} & 1.9\tpm{0.2} & 60.1\tpm{1.8} & 32.7\tpm{3.6} & 5.4\tpm{0.3} & 0.2\tpm{0.0} & 15.5\tpm{0.6} & 29.0\tpm{1.0} & 16.1\tpm{2.9} & 19.8\tpm{0.4} & 1.8\tpm{0.0} & 0.9\tpm{0.0} & 76.1\tpm{1.3} & 53.5\tpm{5.3} & 3.4\tpm{0.3} & 0.1\tpm{0.0} & 1.9\tpm{0.2} & 155.4\tpm{10.4} & 71.1\tpm{1.9} & \rka{2.4\tpm{0.4}} & 0.0\tpm{0.0} & 0.6\tpm{0.3} & 183.0\tpm{17.9} & 168.8\tpm{35.1} & 1.9\tpm{0.2} & 0.0\tpm{0.0} \\
	& \ft{ChebInterp} & 1.7\tpm{0.1} & 606.5\tpm{13.2} & 180.7\tpm{5.8} & 5.4\tpm{0.3} & 0.2\tpm{0.0} & 13.9\tpm{0.8} & 296.5\tpm{1.6} & 67.0\tpm{0.1} & 19.3\tpm{0.3} & 1.8\tpm{0.0} & 0.8\tpm{0.1} & 917.0\tpm{68.0} & 295.4\tpm{32.9} & 3.4\tpm{0.3} & 0.1\tpm{0.0} & 1.6\tpm{0.1} & 758.9\tpm{4.8} & 393.8\tpm{15.9} & \rka{2.4\tpm{0.3}} & 0.0\tpm{0.0} & \rkb{0.1\tpm{0.0}} & 1632.7\tpm{99.1} & 867.5\tpm{62.5} & 1.9\tpm{0.2} & 0.0\tpm{0.0} \\
	& \ft{Bernstein} & 5.0\tpm{0.4} & 60.3\tpm{5.1} & 25.2\tpm{0.4} & 5.5\tpm{0.3} & 0.2\tpm{0.0} & 67.7\tpm{1.2} & 39.5\tpm{1.8} & 17.5\tpm{1.9} & 19.3\tpm{0.4} & 1.9\tpm{0.0} & 2.5\tpm{0.3} & 90.8\tpm{3.6} & 46.5\tpm{6.7} & 3.5\tpm{0.3} & 0.1\tpm{0.0} & 7.2\tpm{0.7} & 171.6\tpm{26.1} & 79.8\tpm{40.6} & \rka{2.5\tpm{0.4}} & 0.0\tpm{0.0} & 1.4\tpm{1.9} & 180.9\tpm{16.4} & 117.8\tpm{41.7} & 1.9\tpm{0.2} & 0.0\tpm{0.0} \\
	& \ft{Legendre} & 1.8\tpm{0.1} & 51.9\tpm{1.0} & 21.1\tpm{0.7} & 5.4\tpm{0.3} & 0.2\tpm{0.0} & 18.6\tpm{2.3} & 36.5\tpm{3.4} & 13.1\tpm{4.9} & 19.3\tpm{0.4} & 1.8\tpm{0.0} & 0.9\tpm{0.1} & 76.2\tpm{2.6} & 33.2\tpm{2.9} & 3.5\tpm{0.4} & 0.1\tpm{0.0} & 2.6\tpm{0.6} & 170.0\tpm{33.3} & 47.4\tpm{3.2} & \rka{2.4\tpm{0.3}} & 0.0\tpm{0.0} & 0.4\tpm{0.3} & 164.0\tpm{11.1} & 80.6\tpm{10.3} & 2.0\tpm{0.3} & 0.0\tpm{0.0} \\
	& \ft{Jacobi} & 2.3\tpm{0.1} & 54.5\tpm{4.0} & 20.6\tpm{0.6} & 5.4\tpm{0.3} & 0.2\tpm{0.0} & 20.8\tpm{2.8} & 33.3\tpm{1.6} & 15.1\tpm{4.9} & 19.3\tpm{0.4} & 1.8\tpm{0.0} & 1.2\tpm{0.1} & 84.1\tpm{7.9} & 36.9\tpm{9.2} & 3.5\tpm{0.4} & 0.1\tpm{0.0} & 1.8\tpm{1.0} & 136.9\tpm{8.0} & 52.4\tpm{2.5} & \rka{2.4\tpm{0.3}} & 0.0\tpm{0.0} & \rka{0.1\tpm{0.0}} & 126.8\tpm{3.0} & 66.6\tpm{7.3} & 2.1\tpm{0.3} & 0.0\tpm{0.0} \\
	& \ft{OptBasis} & 11.4\tpm{0.1} & 50.4\tpm{1.5} & 20.4\tpm{0.1} & 5.4\tpm{0.3} & 0.2\tpm{0.0} & 108.2\tpm{6.1} & 30.8\tpm{0.7} & 10.1\tpm{1.0} & 19.8\tpm{0.3} & 1.8\tpm{0.0} & 6.8\tpm{0.7} & 79.0\tpm{2.3} & 38.1\tpm{6.6} & 3.4\tpm{0.3} & 0.1\tpm{0.0} & 10.5\tpm{8.5} & 164.4\tpm{23.0} & 45.4\tpm{8.1} & \rka{2.4\tpm{0.4}} & 0.0\tpm{0.0} & 0.9\tpm{0.0} & 130.1\tpm{6.1} & 66.7\tpm{5.8} & 1.9\tpm{0.3} & 0.0\tpm{0.0} \\
\midrule
	& \ft{FAGNN} & 3.8\tpm{0.2} & 31.6\tpm{0.4} & 8.5\tpm{0.0} & 2.7\tpm{0.3} & 0.1\tpm{0.0} & 29.5\tpm{2.6} & 19.7\tpm{1.2} & \rka{5.5\tpm{0.6}} & 7.3\tpm{0.2} & 0.3\tpm{0.0} & 1.3\tpm{0.3} & 47.1\tpm{1.4} & \rkc{14.5\tpm{1.2}} & \rka{2.1\tpm{0.3}} & \rka{0.1\tpm{0.0}} & \rkc{0.3\tpm{0.1}} & 44.1\tpm{3.9} & 13.2\tpm{1.2} & \rka{2.5\tpm{0.2}} & 0.0\tpm{0.0} & 0.3\tpm{0.3} & 125.9\tpm{11.8} & \rka{40.0\tpm{4.5}} & 1.6\tpm{0.2} & \rka{0.0\tpm{0.0}} \\
	& \ft{G$^2$CN} & 1.4\tpm{0.1} & 30.1\tpm{1.6} & 8.5\tpm{0.1} & 2.7\tpm{0.3} & 0.1\tpm{0.0} & 49.6\tpm{1.7} & 18.0\tpm{1.3} & \rka{5.4\tpm{0.7}} & 7.4\tpm{0.2} & 0.3\tpm{0.0} & \rka{0.6\tpm{0.1}} & 50.4\tpm{1.5} & 16.2\tpm{2.6} & \rka{2.0\tpm{0.3}} & \rka{0.1\tpm{0.0}} & 0.6\tpm{0.2} & 42.7\tpm{3.2} & 13.4\tpm{2.0} & \rka{2.5\tpm{0.2}} & 0.0\tpm{0.0} & 0.6\tpm{0.6} & 135.6\tpm{30.2} & 64.6\tpm{18.8} & \rka{1.5\tpm{0.1}} & \rka{0.0\tpm{0.0}} \\
	& \ft{GNN-LF/HF} & 2.8\tpm{0.2} & 30.2\tpm{5.0} & 8.1\tpm{0.1} & \rka{2.5\tpm{0.3}} & 0.1\tpm{0.0} & 24.7\tpm{1.3} & 16.8\tpm{0.4} & \rka{4.8\tpm{0.0}} & 6.9\tpm{0.3} & 0.3\tpm{0.0} & 1.3\tpm{0.0} & 47.3\tpm{3.7} & \rkc{15.1\tpm{1.1}} & \rka{2.1\tpm{0.3}} & \rka{0.1\tpm{0.0}} & \rkc{0.3\tpm{0.0}} & 43.9\tpm{2.6} & 13.7\tpm{1.2} & 3.4\tpm{0.7} & 0.0\tpm{0.0} & \rkb{0.1\tpm{0.1}} & \rkc{100.0\tpm{8.1}} & \rka{35.6\tpm{1.2}} & 1.6\tpm{0.2} & 0.0\tpm{0.0} \\
	& \ft{FiGURe} & 8.6\tpm{0.4} & 143.1\tpm{1.6} & 49.3\tpm{7.3} & 15.1\tpm{0.3} & 0.6\tpm{0.0} & 85.4\tpm{1.9} & 61.8\tpm{1.5} & 20.4\tpm{1.0} & 63.3\tpm{0.2} & 5.3\tpm{0.0} & 5.5\tpm{2.1} & 164.8\tpm{11.4} & 86.6\tpm{16.0} & 8.2\tpm{0.3} & 0.2\tpm{0.0} & 1.2\tpm{0.3} & 173.4\tpm{25.1} & 91.5\tpm{12.4} & 2.9\tpm{0.5} & 0.0\tpm{0.0} & 0.9\tpm{0.5} & 401.1\tpm{35.7} & 248.0\tpm{49.9} & 3.0\tpm{0.3} & 0.0\tpm{0.0} \\
\toprule
    & Filter
    & \dsh{|}{mag} & \dsh{|}{pokec} & \dsh{|}{snap} & \dsh{|}{products} & \dsh{}{wiki} \\
\midrule
	& \rko{\ft{Identity}} & \rko{0.0\tpm{0.0}} & \rko{9.5\tpm{0.9}} & \rko{3.2\tpm{0.5}} & \rko{2.9\tpm{0.2}} & \rko{5.0\tpm{0.0}} & \rko{0.0\tpm{0.0}} & \rko{14.2\tpm{0.4}} & \rko{5.8\tpm{0.6}} & \rko{6.7\tpm{0.3}} & \rko{0.5\tpm{0.0}} & \rko{0.0\tpm{0.0}} & \rko{26.8\tpm{2.3}} & \rko{9.0\tpm{0.6}} & \rko{10.1\tpm{0.3}} & \rko{0.6\tpm{0.0}} & \rko{0.0\tpm{0.0}} & \rko{2.7\tpm{0.6}} & \rko{11.5\tpm{3.8}} & \rko{21.6\tpm{0.3}} & \rko{2.4\tpm{0.0}} & \rko{0.0\tpm{0.0}} & \rko{22.9\tpm{1.4}} & \rko{6.9\tpm{1.7}} & \rko{57.4\tpm{0.3}} & \rko{1.0\tpm{0.0}} \\
	& \ft{Linear} & 4.7\tpm{0.6} & \rka{11.5\tpm{1.1}} & \rkc{5.5\tpm{1.0}} & 3.8\tpm{0.4} & 5.0\tpm{0.0} & 8.5\tpm{0.8} & \rka{13.0\tpm{0.9}} & \rka{6.2\tpm{0.9}} & \rka{6.6\tpm{0.3}} & \rkb{0.5\tpm{0.0}} & 27.3\tpm{1.0} & \rka{21.0\tpm{0.6}} & \rka{8.3\tpm{0.4}} & 20.8\tpm{0.3} & \rkc{0.6\tpm{0.0}} & 19.5\tpm{1.2} & \rka{3.1\tpm{0.8}} & \rka{10.3\tpm{0.4}} & \rka{21.6\tpm{0.4}} & 2.3\tpm{0.0} & 206.5\tpm{11.4} & 22.8\tpm{2.6} & 6.8\tpm{0.9} & \rka{57.5\tpm{0.4}} & \rkb{1.0\tpm{0.0}} \\
	& \ft{Impulse} & \rka{3.9\tpm{0.2}} & \rka{11.4\tpm{0.5}} & \rkc{5.3\tpm{0.5}} & \rka{3.4\tpm{0.3}} & 4.9\tpm{0.0} & 5.4\tpm{0.2} & 14.6\tpm{0.3} & \rka{6.4\tpm{0.3}} & \rka{6.6\tpm{0.3}} & \rkb{0.5\tpm{0.0}} & \rkc{21.0\tpm{1.4}} & \rkc{21.5\tpm{1.0}} & \rka{8.4\tpm{0.2}} & \rka{18.1\tpm{0.4}} & \rkc{0.6\tpm{0.0}} & \rka{15.0\tpm{0.5}} & \rka{3.1\tpm{0.6}} & \rka{10.3\tpm{1.2}} & \rka{21.6\tpm{0.4}} & 2.3\tpm{0.0} & \rka{164.1\tpm{23.0}} & \rkc{21.8\tpm{0.6}} & 7.4\tpm{1.1} & \rka{57.5\tpm{0.4}} & \rkb{1.0\tpm{0.0}} \\
	& \ft{Monomial} & \rka{4.1\tpm{0.0}} & 12.1\tpm{0.2} & \rka{4.7\tpm{0.3}} & \rka{3.4\tpm{0.3}} & 5.0\tpm{0.0} & \rka{5.2\tpm{0.0}} & 14.5\tpm{0.6} & \rka{6.5\tpm{0.6}} & \rka{6.6\tpm{0.3}} & \rkb{0.5\tpm{0.0}} & 21.9\tpm{4.7} & \rkc{22.2\tpm{0.6}} & 8.7\tpm{0.6} & \rka{18.0\tpm{0.4}} & \rka{0.6\tpm{0.0}} & 15.5\tpm{1.2} & \rka{3.1\tpm{0.5}} & \rka{10.4\tpm{0.3}} & \rka{21.6\tpm{0.4}} & 2.3\tpm{0.0} & \rka{172.6\tpm{18.0}} & 23.0\tpm{1.6} & 6.4\tpm{0.9} & \rka{57.4\tpm{0.3}} & \rka{0.9\tpm{0.0}} \\
	& \ft{PPR} & \rka{4.0\tpm{0.1}} & \rka{11.4\tpm{0.6}} & \rkc{5.4\tpm{1.1}} & \rka{3.4\tpm{0.3}} & 5.0\tpm{0.0} & 5.7\tpm{0.3} & \rka{13.7\tpm{0.6}} & \rka{6.3\tpm{0.7}} & \rka{6.6\tpm{0.3}} & \rka{0.5\tpm{0.0}} & \rkc{20.3\tpm{0.9}} & \rka{20.8\tpm{0.6}} & \rka{8.4\tpm{0.5}} & \rka{18.1\tpm{0.4}} & \rka{0.6\tpm{0.0}} & 15.8\tpm{1.5} & \rka{3.1\tpm{0.8}} & \rka{10.6\tpm{0.3}} & \rka{21.6\tpm{0.4}} & \rkb{1.7\tpm{0.0}} & \rka{166.6\tpm{16.9}} & \rkc{22.2\tpm{0.5}} & \rka{6.0\tpm{0.5}} & \rka{57.4\tpm{0.4}} & \rkb{1.0\tpm{0.0}} \\
	& \ft{HK} & \rka{3.9\tpm{0.2}} & 12.1\tpm{1.1} & \rka{4.8\tpm{0.1}} & \rka{3.4\tpm{0.3}} & 5.0\tpm{0.0} & \rka{5.2\tpm{0.1}} & 14.2\tpm{1.2} & \rka{6.3\tpm{0.3}} & \rka{6.6\tpm{0.4}} & \rkb{0.5\tpm{0.0}} & \rka{18.8\tpm{0.7}} & \rkc{22.2\tpm{0.6}} & \rka{8.4\tpm{0.3}} & \rka{18.1\tpm{0.4}} & \rkc{0.6\tpm{0.0}} & \rka{14.8\tpm{0.4}} & \rka{3.1\tpm{0.6}} & 11.1\tpm{0.6} & \rka{21.6\tpm{0.4}} & \rkb{1.9\tpm{0.0}} & \rka{165.6\tpm{9.6}} & \rka{20.9\tpm{0.6}} & \rka{5.8\tpm{0.2}} & \rka{57.4\tpm{0.3}} & \rkb{1.0\tpm{0.0}} \\
	& \ft{Gaussian} & \rka{3.9\tpm{0.3}} & 12.1\tpm{1.4} & \rkc{5.7\tpm{1.8}} & \rka{3.4\tpm{0.3}} & 5.0\tpm{0.0} & \rka{5.2\tpm{0.2}} & \rka{13.6\tpm{1.4}} & 7.7\tpm{0.6} & \rka{6.6\tpm{0.3}} & \rkb{0.5\tpm{0.0}} & \rkb{19.7\tpm{0.1}} & 22.6\tpm{0.1} & 10.0\tpm{0.7} & \rka{18.1\tpm{0.4}} & \rkc{0.6\tpm{0.0}} & \rka{14.7\tpm{0.3}} & \rka{3.0\tpm{0.7}} & \rka{9.8\tpm{0.8}} & \rka{21.6\tpm{0.4}} & 2.1\tpm{0.0} & \rka{156.1\tpm{25.3}} & \rka{20.4\tpm{0.7}} & \rkc{6.1\tpm{0.3}} & \rka{57.5\tpm{0.4}} & \rkb{1.0\tpm{0.0}} \\
\midrule
	& \ft{Linear} & \rka{4.0\tpm{0.2}} & 20.4\tpm{2.6} & 9.4\tpm{0.4} & 10.1\tpm{0.3} & 5.0\tpm{0.0} & 5.7\tpm{0.2} & 25.0\tpm{1.2} & 15.6\tpm{0.8} & 11.8\tpm{0.4} & 1.2\tpm{0.0} & 26.5\tpm{3.4} & 33.9\tpm{0.7} & 22.4\tpm{2.6} & 70.3\tpm{0.3} & 4.9\tpm{0.0} & 16.6\tpm{0.7} & 4.6\tpm{0.6} & 16.2\tpm{0.4} & 25.9\tpm{0.3} & 2.4\tpm{0.2} & \rka{160.2\tpm{28.8}} & 32.8\tpm{2.7} & 15.9\tpm{2.1} & 109.0\tpm{0.3} & 10.8\tpm{0.0} \\
	& \ft{Monomial} & \rka{3.8\tpm{0.4}} & 22.8\tpm{1.2} & 7.6\tpm{0.7} & 10.1\tpm{0.3} & 5.2\tpm{0.0} & 6.0\tpm{0.5} & 30.6\tpm{5.1} & 14.6\tpm{0.9} & 11.8\tpm{0.4} & 1.2\tpm{0.0} & 25.0\tpm{2.0} & 36.3\tpm{0.7} & 17.0\tpm{1.8} & 70.3\tpm{0.3} & 4.9\tpm{0.0} & 15.3\tpm{0.9} & 5.3\tpm{0.7} & 15.4\tpm{1.4} & 25.9\tpm{0.3} & 2.1\tpm{0.0} & \rka{168.1\tpm{7.6}} & 34.4\tpm{2.5} & 15.6\tpm{6.7} & 109.0\tpm{0.3} & 10.8\tpm{0.0} \\
	& \ft{Horner} & \rka{3.9\tpm{0.2}} & 20.8\tpm{0.8} & 12.4\tpm{3.0} & 10.1\tpm{0.3} & 5.1\tpm{0.0} & 6.2\tpm{0.8} & 26.8\tpm{1.9} & 17.8\tpm{0.6} & 11.8\tpm{0.4} & 1.2\tpm{0.0} & 23.5\tpm{2.7} & 35.5\tpm{2.5} & 24.7\tpm{2.9} & 70.3\tpm{0.3} & 4.9\tpm{0.0} & 20.9\tpm{6.2} & 6.8\tpm{1.2} & 25.5\tpm{2.6} & 25.9\tpm{0.3} & 2.5\tpm{0.0} & 183.1\tpm{39.9} & 34.4\tpm{5.0} & 22.4\tpm{4.7} & 109.0\tpm{0.3} & 10.8\tpm{0.0} \\
	& \ft{Chebyshev} & 5.7\tpm{1.1} & 27.0\tpm{1.5} & 9.5\tpm{2.7} & 10.2\tpm{0.2} & 5.1\tpm{0.0} & 7.1\tpm{0.1} & 27.3\tpm{0.9} & 12.0\tpm{0.1} & 12.3\tpm{0.4} & 1.2\tpm{0.0} & 26.3\tpm{1.6} & 32.2\tpm{0.8} & 15.4\tpm{2.4} & 70.5\tpm{0.3} & 4.9\tpm{0.0} & 24.1\tpm{3.9} & 6.4\tpm{0.9} & 17.6\tpm{3.3} & 27.3\tpm{0.3} & 2.5\tpm{0.0} & 186.1\tpm{21.6} & 33.8\tpm{1.0} & 15.6\tpm{2.5} & 114.4\tpm{0.3} & 10.8\tpm{0.0} \\
	& \ft{Clenshaw} & 6.0\tpm{1.9} & 20.1\tpm{2.7} & 13.2\tpm{5.0} & 10.1\tpm{0.2} & 5.1\tpm{0.0} & 7.7\tpm{0.3} & 28.5\tpm{3.0} & 19.1\tpm{1.5} & 11.8\tpm{0.4} & 1.2\tpm{0.0} & 35.1\tpm{3.1} & 43.8\tpm{3.3} & 31.6\tpm{7.3} & 70.3\tpm{0.3} & 4.9\tpm{0.0} & 28.4\tpm{3.9} & 6.3\tpm{0.8} & 23.9\tpm{0.5} & 25.9\tpm{0.3} & 2.7\tpm{0.0} & 204.9\tpm{51.8} & 36.8\tpm{2.1} & 21.6\tpm{4.2} & 109.0\tpm{0.3} & 10.8\tpm{0.0} \\
	& \ft{ChebInterp} & 5.7\tpm{1.3} & 319.8\tpm{2.6} & 47.0\tpm{5.4} & 10.2\tpm{0.3} & \rkc{3.7\tpm{1.0}} & 7.3\tpm{0.5} & 251.2\tpm{1.0} & 77.0\tpm{3.0} & 12.3\tpm{0.4} & 1.2\tpm{0.0} & 35.2\tpm{1.9} & 922.4\tpm{86.8} & 108.6\tpm{7.3} & 70.5\tpm{0.3} & 4.9\tpm{0.0} & 23.5\tpm{5.1} & 70.4\tpm{1.6} & 102.1\tpm{8.4} & 27.3\tpm{0.3} & 2.6\tpm{0.0} & 202.1\tpm{74.7} & 1243.4\tpm{23.5} & 69.5\tpm{17.9} & 114.4\tpm{0.3} & 10.8\tpm{0.0} \\
	& \ft{Bernstein} & 23.0\tpm{2.2} & 31.1\tpm{5.1} & 9.9\tpm{1.3} & 10.4\tpm{0.4} & 4.9\tpm{0.0} & 26.4\tpm{0.8} & 30.9\tpm{0.4} & 15.2\tpm{1.2} & 13.4\tpm{0.6} & 1.6\tpm{0.0} & 100.0\tpm{5.5} & 44.7\tpm{1.6} & 44.3\tpm{37.5} & 71.0\tpm{0.4} & 5.2\tpm{0.0} & 99.7\tpm{1.0} & 6.5\tpm{0.6} & 20.6\tpm{1.1} & 30.7\tpm{1.2} & 2.5\tpm{0.0} & 1069.5\tpm{263.2} & 37.8\tpm{3.2} & 14.2\tpm{1.0} & 124.4\tpm{2.1} & 11.7\tpm{0.0} \\
	& \ft{Legendre} & 6.0\tpm{0.5} & 23.0\tpm{0.7} & 10.8\tpm{4.0} & 9.9\tpm{0.4} & 5.1\tpm{0.0} & 7.5\tpm{0.5} & 31.2\tpm{3.5} & 12.9\tpm{1.1} & 11.8\tpm{0.4} & 1.2\tpm{0.0} & 32.1\tpm{2.4} & 38.2\tpm{1.2} & 15.9\tpm{2.0} & 70.3\tpm{0.3} & 4.9\tpm{0.0} & 22.5\tpm{2.8} & 5.9\tpm{0.7} & 15.0\tpm{0.6} & 25.9\tpm{0.3} & 2.3\tpm{0.0} & \rka{176.6\tpm{36.5}} & 34.6\tpm{3.2} & 18.5\tpm{15.6} & 108.9\tpm{0.3} & 10.8\tpm{0.0} \\
	& \ft{Jacobi} & 7.1\tpm{0.1} & 21.2\tpm{1.2} & 7.0\tpm{1.1} & 9.9\tpm{0.4} & 5.1\tpm{0.0} & 9.0\tpm{0.0} & 30.7\tpm{3.7} & 14.4\tpm{0.4} & 11.8\tpm{0.4} & 1.2\tpm{0.0} & 42.8\tpm{2.8} & 37.4\tpm{0.4} & 19.8\tpm{6.1} & 70.3\tpm{0.4} & 4.9\tpm{0.0} & 23.3\tpm{0.5} & 6.1\tpm{0.7} & 15.7\tpm{0.3} & 25.9\tpm{0.3} & 2.5\tpm{0.0} & 184.4\tpm{29.8} & 35.0\tpm{4.0} & 12.1\tpm{1.7} & 108.9\tpm{0.3} & 10.8\tpm{0.0} \\
	& \ft{OptBasis} & 30.4\tpm{0.9} & 24.4\tpm{3.6} & 10.8\tpm{6.9} & 10.0\tpm{0.4} & \rkb{3.2\tpm{0.0}} & 35.1\tpm{1.7} & 28.2\tpm{0.7} & 13.9\tpm{1.5} & 11.9\tpm{0.4} & 1.2\tpm{0.0} & 269.1\tpm{74.1} & 42.6\tpm{2.7} & 18.7\tpm{2.9} & 70.5\tpm{0.3} & 4.9\tpm{0.0} & 87.2\tpm{8.7} & 6.3\tpm{0.4} & 17.9\tpm{4.7} & 25.9\tpm{0.3} & 2.9\tpm{0.0} & 448.3\tpm{81.9} & 32.9\tpm{2.7} & 10.7\tpm{1.3} & 109.0\tpm{0.3} & 10.8\tpm{0.0} \\
\midrule
	& \ft{FAGNN} & 10.1\tpm{1.2} & 345.0\tpm{19.1} & 121.3\tpm{27.7} & 4.4\tpm{0.4} & 5.0\tpm{0.0} & 15.3\tpm{0.8} & 18.2\tpm{1.8} & 7.7\tpm{1.0} & \rka{6.6\tpm{0.4}} & 0.6\tpm{0.0} & 68.9\tpm{3.8} & 32.0\tpm{1.8} & 13.8\tpm{4.7} & 23.6\tpm{0.3} & 1.0\tpm{0.0} & 46.9\tpm{1.2} & 4.2\tpm{0.8} & 11.8\tpm{2.1} & \rka{21.6\tpm{0.3}} & \rkb{1.5\tpm{0.5}} & 438.8\tpm{66.9} & 26.4\tpm{4.1} & 7.2\tpm{0.3} & \rka{57.2\tpm{0.3}} & 2.0\tpm{0.0} \\
	& \ft{G$^2$CN} & 17.7\tpm{0.6} & 15.1\tpm{0.6} & \rkc{5.5\tpm{0.3}} & 4.3\tpm{0.5} & \rka{2.2\tpm{0.0}} & 28.2\tpm{3.4} & 17.3\tpm{2.0} & \rka{6.6\tpm{0.1}} & \rka{6.6\tpm{0.4}} & 0.6\tpm{0.0} & 91.0\tpm{6.0} & 28.1\tpm{1.9} & 10.1\tpm{0.9} & 23.6\tpm{0.3} & 1.0\tpm{0.0} & 81.9\tpm{3.8} & 4.2\tpm{1.3} & 11.0\tpm{2.0} & \rka{21.5\tpm{0.3}} & \rka{1.0\tpm{0.0}} & 827.7\tpm{141.0} & 29.2\tpm{1.4} & 8.0\tpm{1.2} & \rka{57.2\tpm{0.3}} & 2.0\tpm{0.0} \\
	& \ft{GNN-LF/HF} & 9.0\tpm{0.3} & 15.3\tpm{0.7} & 5.9\tpm{0.5} & 4.4\tpm{0.5} & 4.8\tpm{0.0} & 12.5\tpm{0.4} & 16.7\tpm{0.5} & \rka{6.8\tpm{0.1}} & 7.4\tpm{1.1} & 0.6\tpm{0.0} & 52.7\tpm{2.8} & 29.5\tpm{1.3} & 11.0\tpm{0.5} & 21.8\tpm{0.5} & 1.0\tpm{0.0} & 38.0\tpm{2.9} & 4.0\tpm{0.8} & 11.3\tpm{1.8} & 23.8\tpm{2.1} & \rkb{1.7\tpm{0.1}} & 385.5\tpm{46.7} & 26.5\tpm{0.7} & 7.3\tpm{0.5} & 59.7\tpm{4.5} & 2.0\tpm{0.0} \\
	& \ft{FiGURe} & 26.5\tpm{3.0} & 1220.1\tpm{118.7} & 420.6\tpm{45.3} & 31.5\tpm{0.4} & 5.1\tpm{0.0} & 46.6\tpm{8.5} & 52.7\tpm{0.7} & 26.6\tpm{1.5} & 37.1\tpm{0.7} & 3.5\tpm{0.0} & 175.1\tpm{7.0} & 79.4\tpm{7.6} & 40.4\tpm{4.7} & 230.8\tpm{0.5} & 14.5\tpm{0.0} & 145.2\tpm{9.1} & 11.4\tpm{0.9} & 35.6\tpm{1.7} & 84.6\tpm{0.7} & 5.7\tpm{0.0} & 1405.6\tpm{231.2} & 107.8\tpm{9.3} & 45.1\tpm{7.2} & 359.8\tpm{2.4} & 20.4\tpm{0.9} \\
\bottomrule
\end{tabular}
\end{adjustbox}
\end{table}

\clearpage

\begin{figure}[p]
    \centering
    \subcaptionbox{\ds{citeseer}\label{ffiga:box_citeseer_full}}%
    [0.24\linewidth]{\includegraphics[height=1.2in]{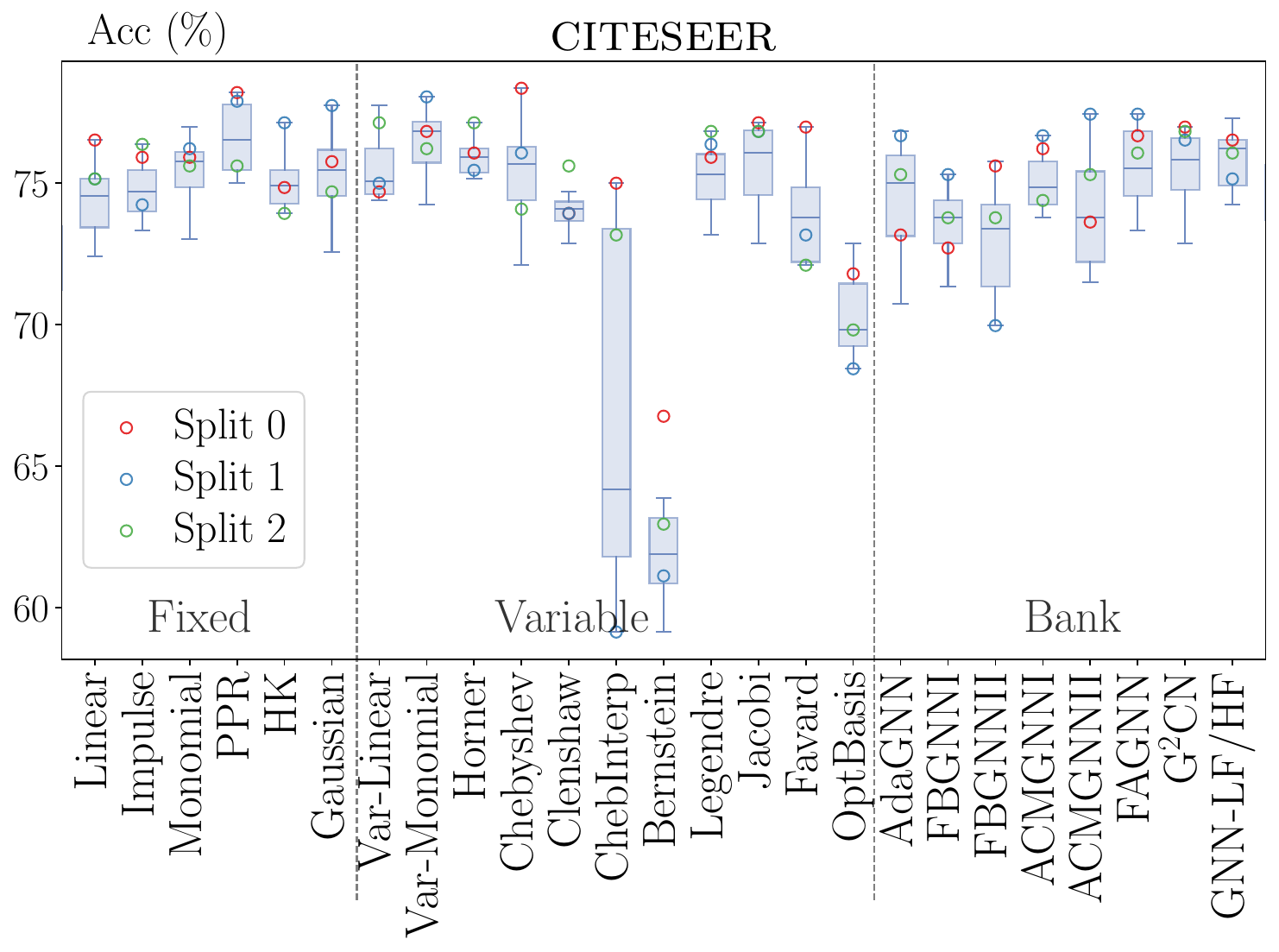}}
    \hfil
    \subcaptionbox{\ds{pubmed}\label{ffiga:box_pubmed_full}}%
    [0.24\linewidth]{\includegraphics[height=1.2in]{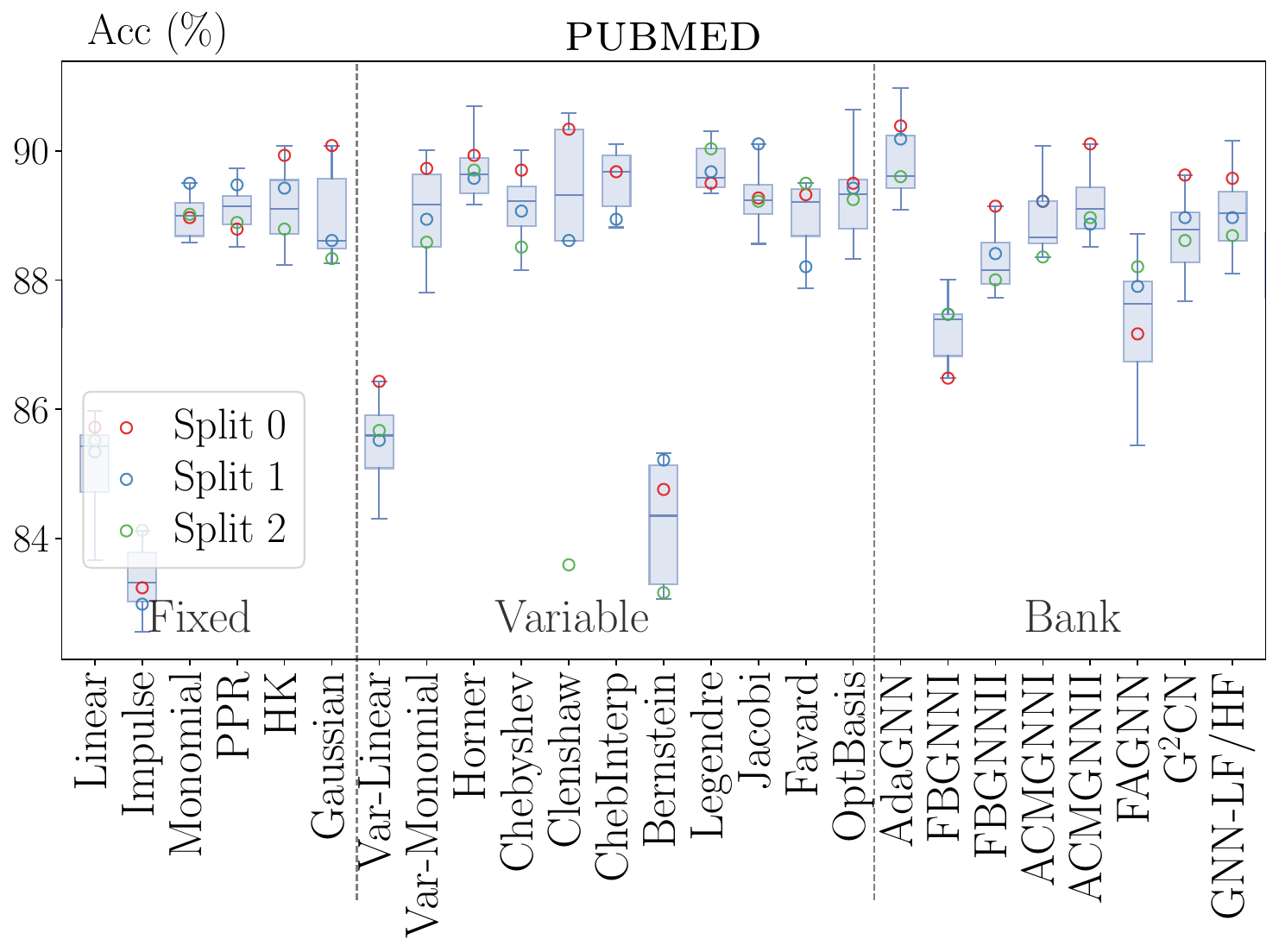}}
    \hfil
    \subcaptionbox{\ds{minesweeper}\label{ffiga:box_minesweeper_full}}%
    [0.24\linewidth]{\includegraphics[height=1.2in]{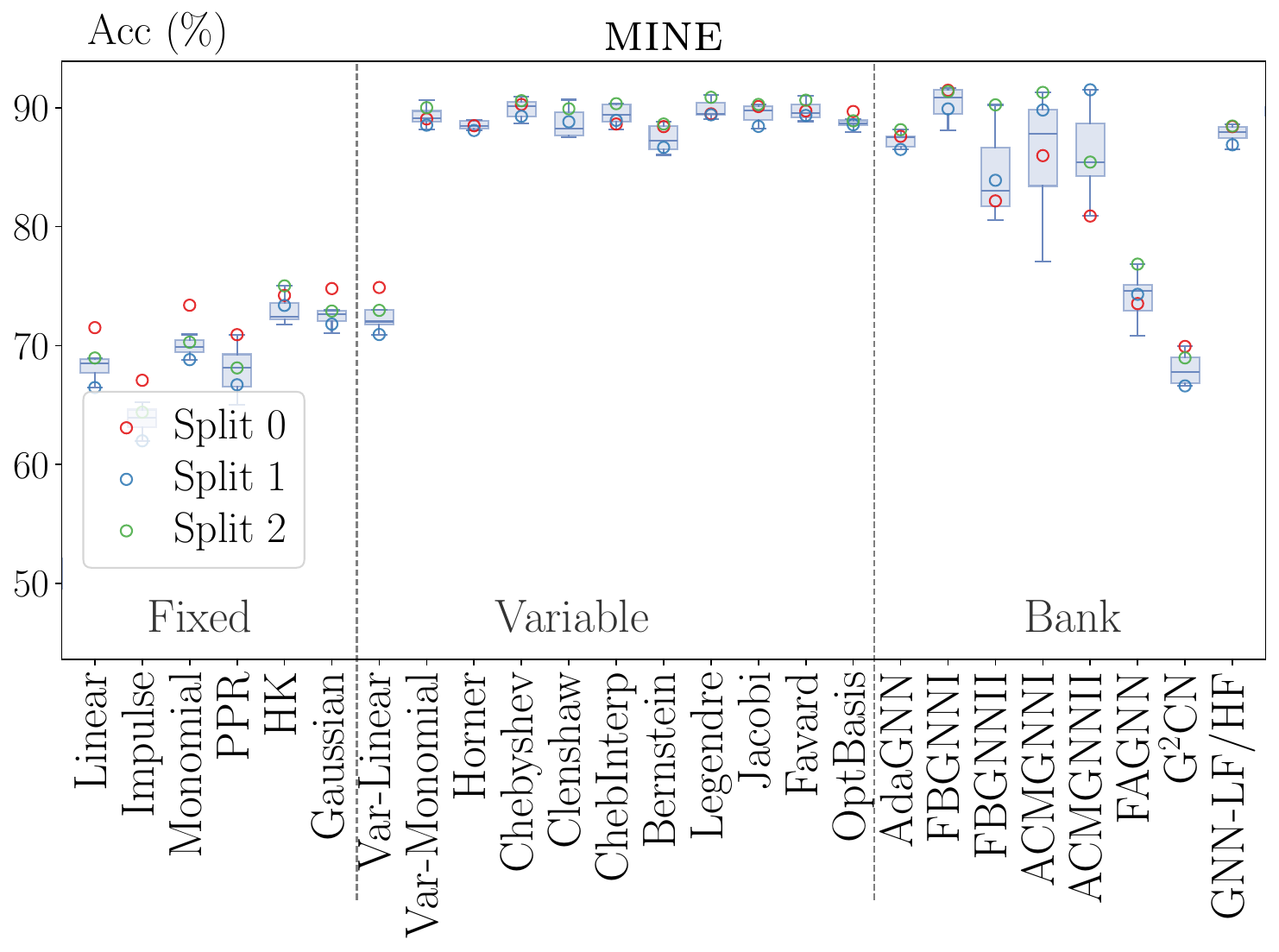}}
    \hfil
    \subcaptionbox{\ds{flickr}\label{ffiga:box_flickr_full}}%
    [0.24\linewidth]{\includegraphics[height=1.2in]{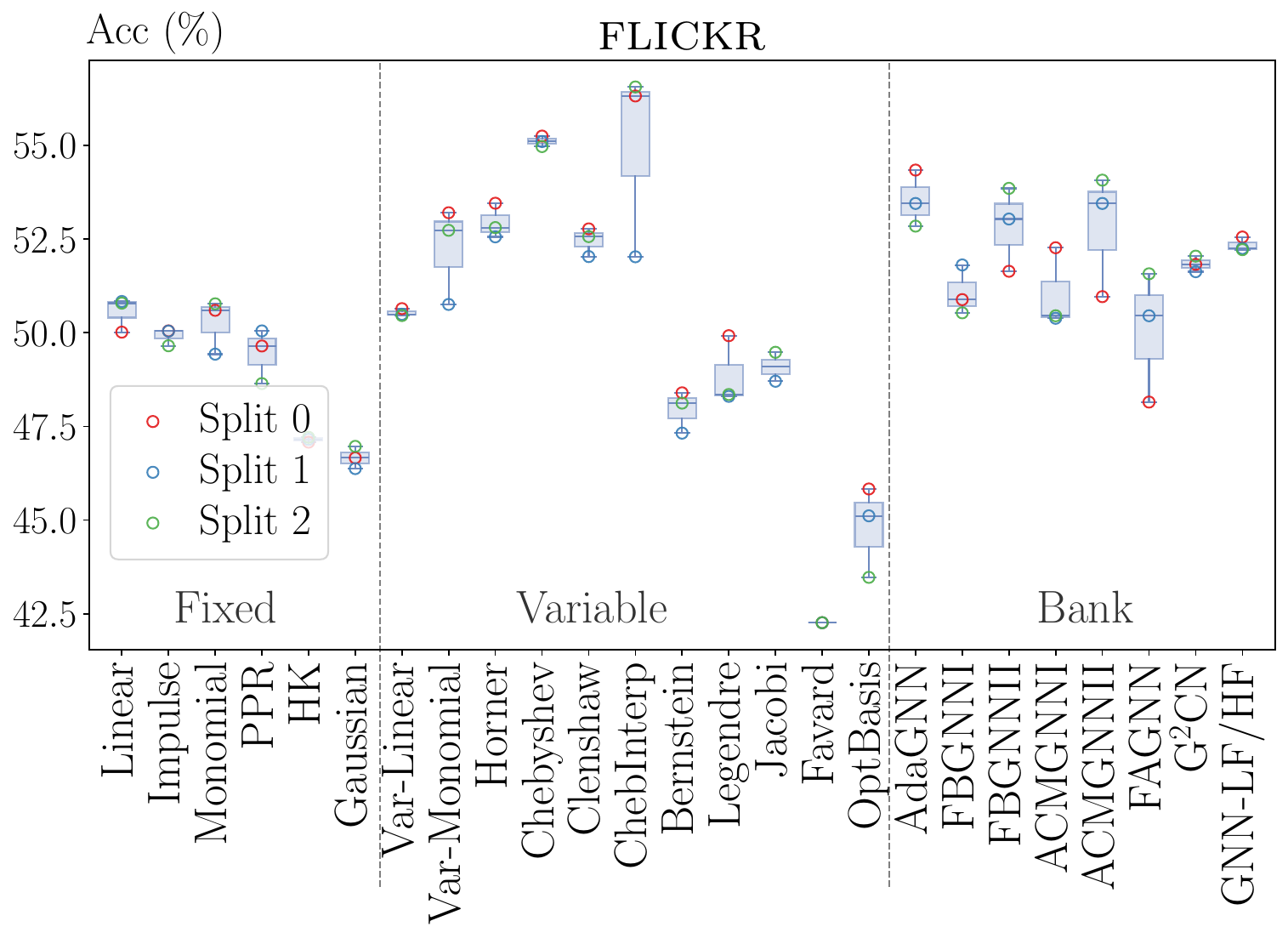}}
    \\ \vspace{8pt}
    
   \subcaptionbox{\ds{chameleon}\label{ffiga:box_chameleon_filtered_full}}%
    [0.24\linewidth]{\includegraphics[height=1.2in]{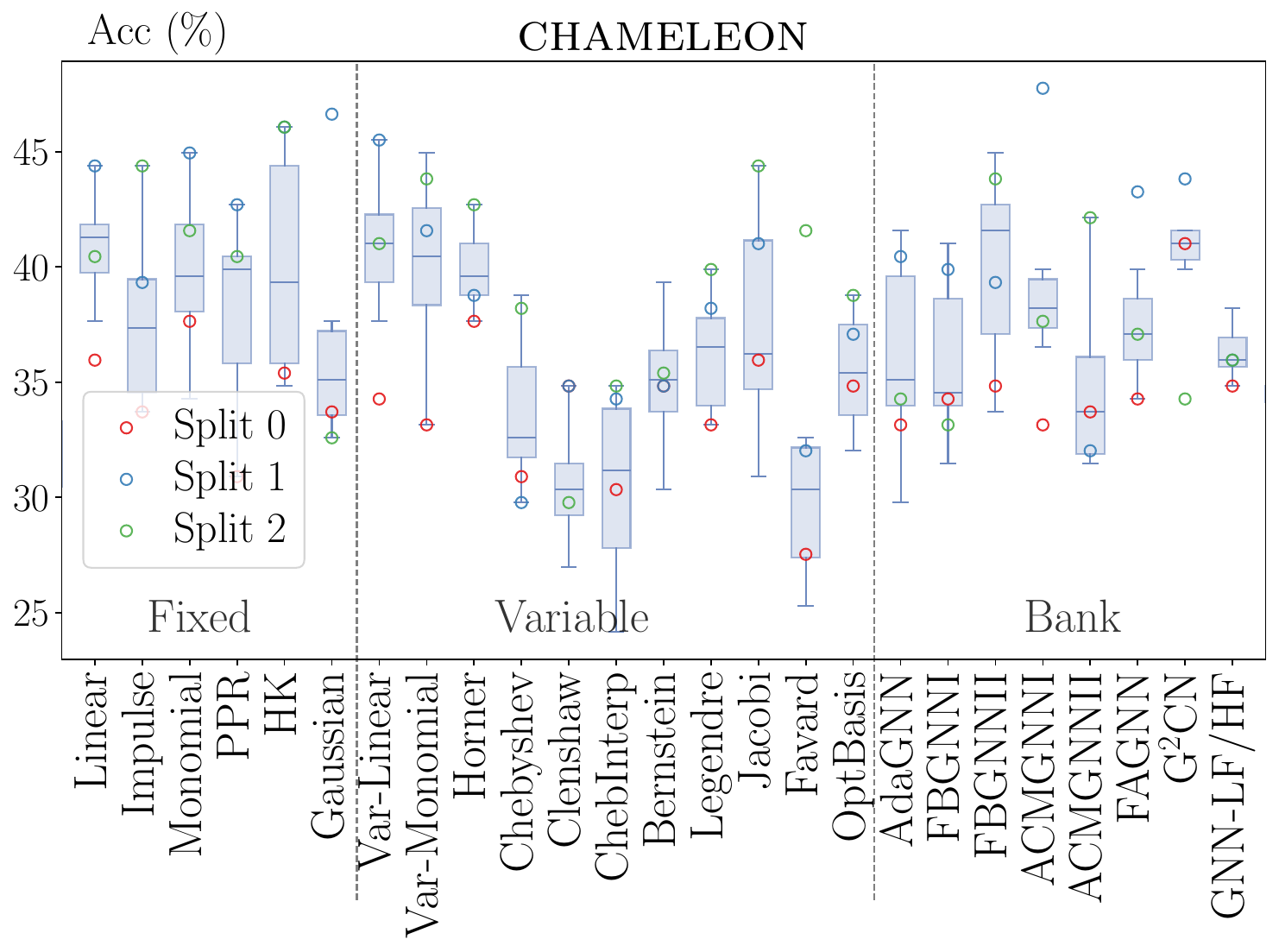}}
    \hfil
    \subcaptionbox{\ds{squirrel}\label{ffiga:box_squirrel_filtered_full}}%
    [0.24\linewidth]{\includegraphics[height=1.2in]{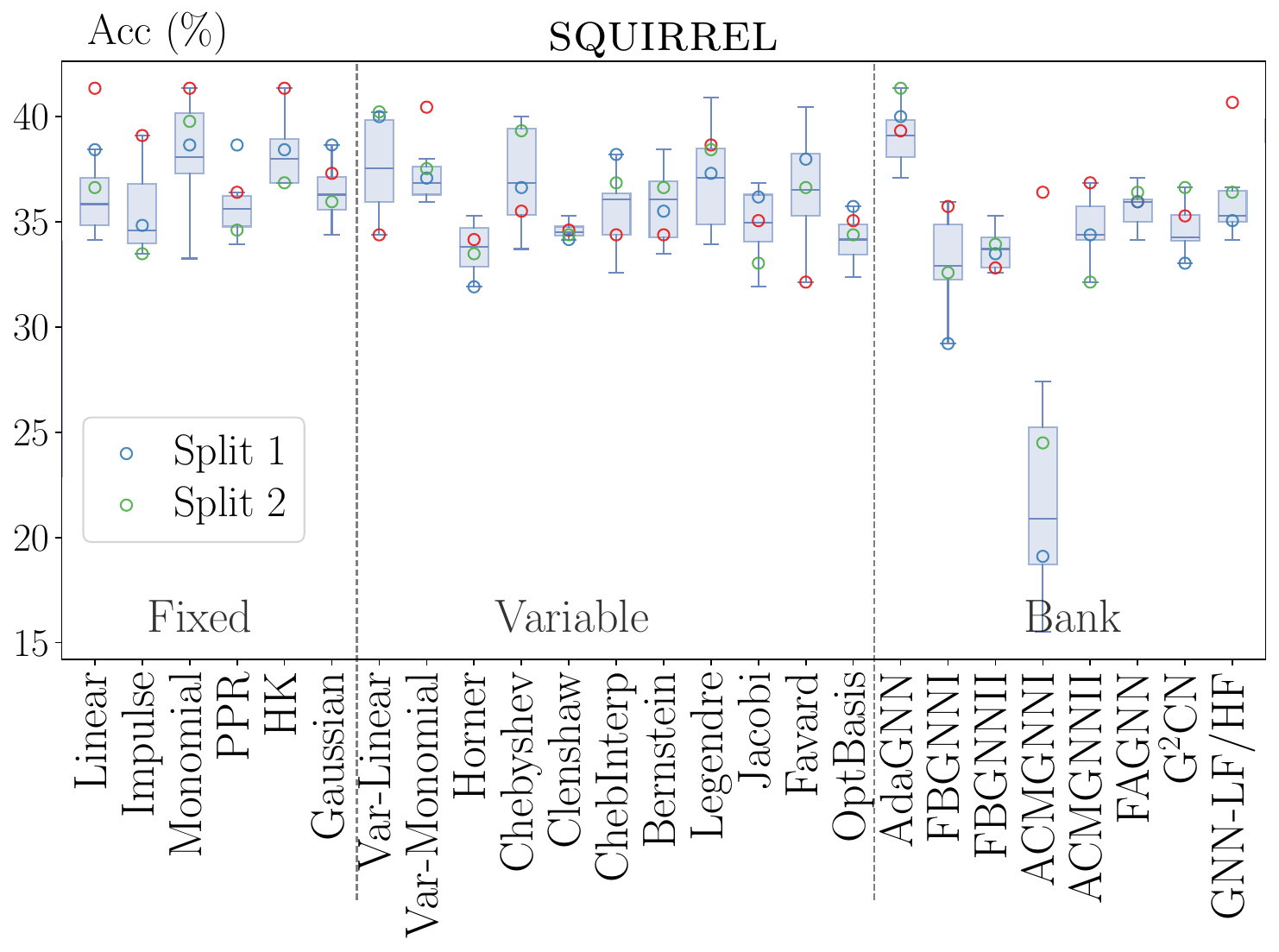}}
    \hfil
    \subcaptionbox{\ds{roman}\label{ffiga:box_roman_empire_full}}%
    [0.24\linewidth]{\includegraphics[height=1.2in]{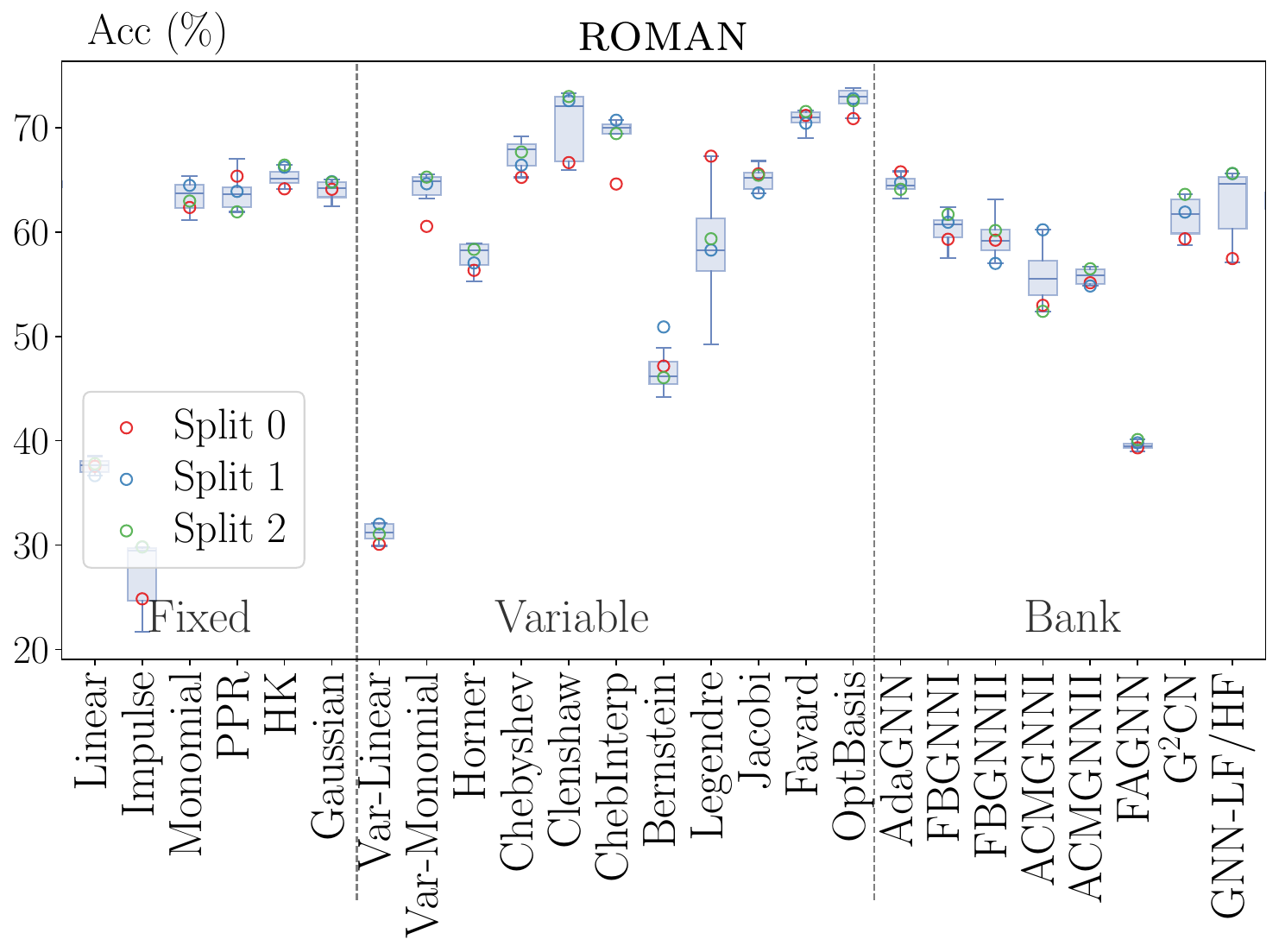}}
    \hfil
    \subcaptionbox{\ds{year}\label{ffiga:box_arxiv-year_full}}%
    [0.24\linewidth]{\includegraphics[height=1.2in]{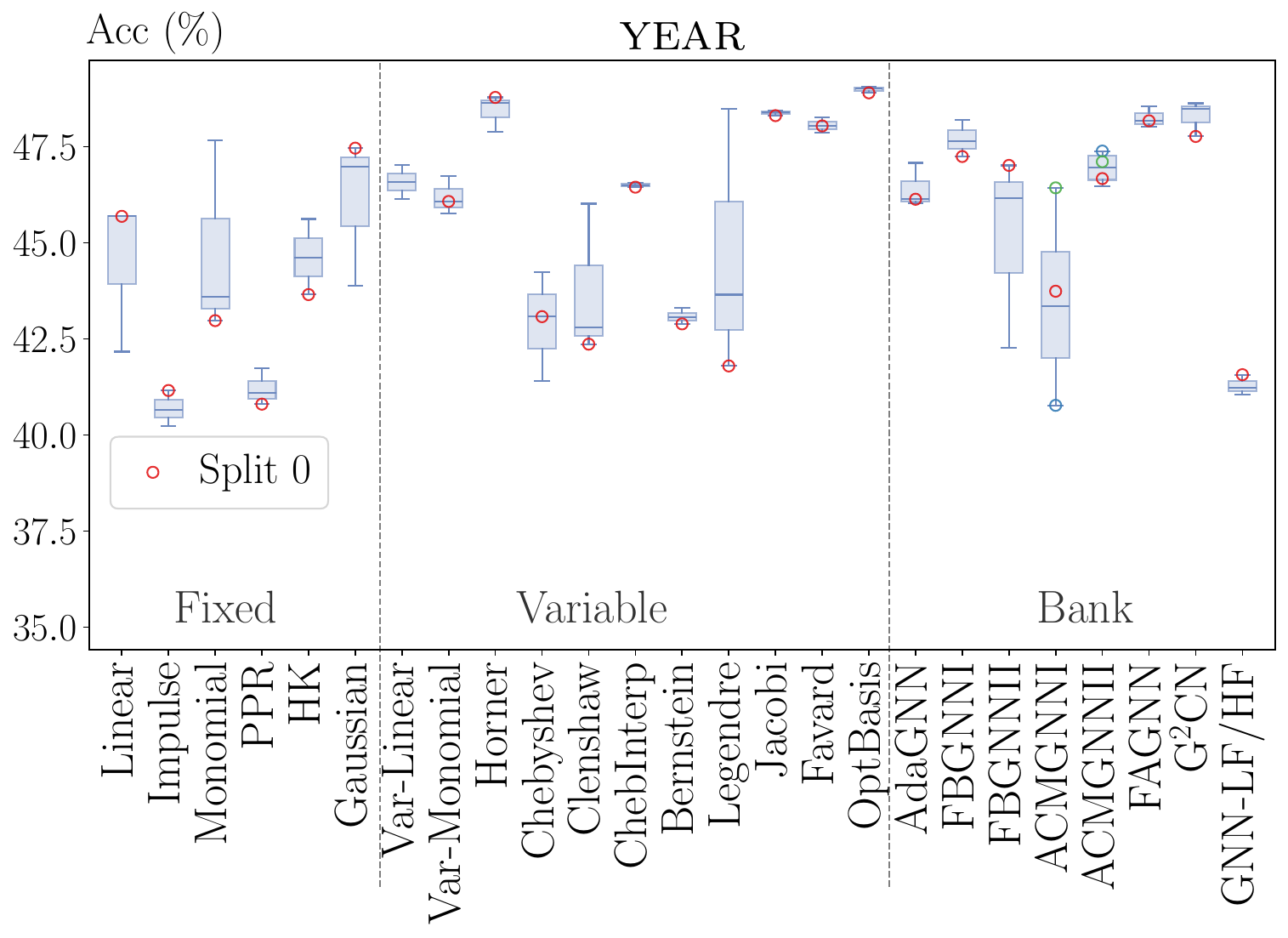}}
    \\ \vspace{8pt}
    \caption{Effectiveness variance of \textit{full-batch} training on 4 homophilous and 4 heterophilous datasets.}
  \label{figa:box_full}
\vspace{4mm}
\end{figure}
\begin{figure}[p]
    \centering
    \subcaptionbox{\ds{citeseer}\label{ffiga:boxmb_citeseer_full}}%
    [0.24\linewidth]{\includegraphics[height=1.31in]{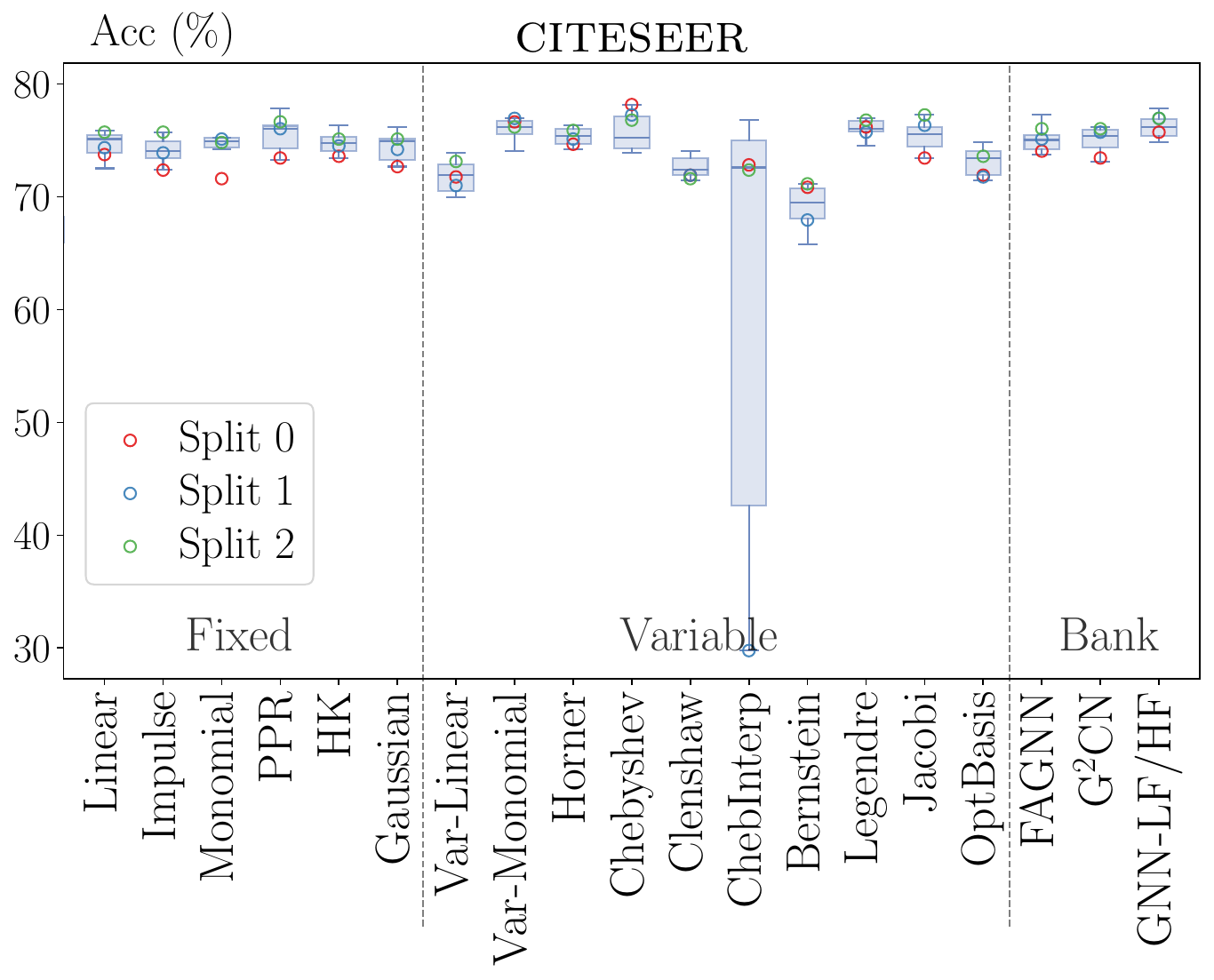}}
    \hfil
    \subcaptionbox{\ds{pubmed}\label{ffiga:boxmb_pubmed_full}}%
    [0.24\linewidth]{\includegraphics[height=1.31in]{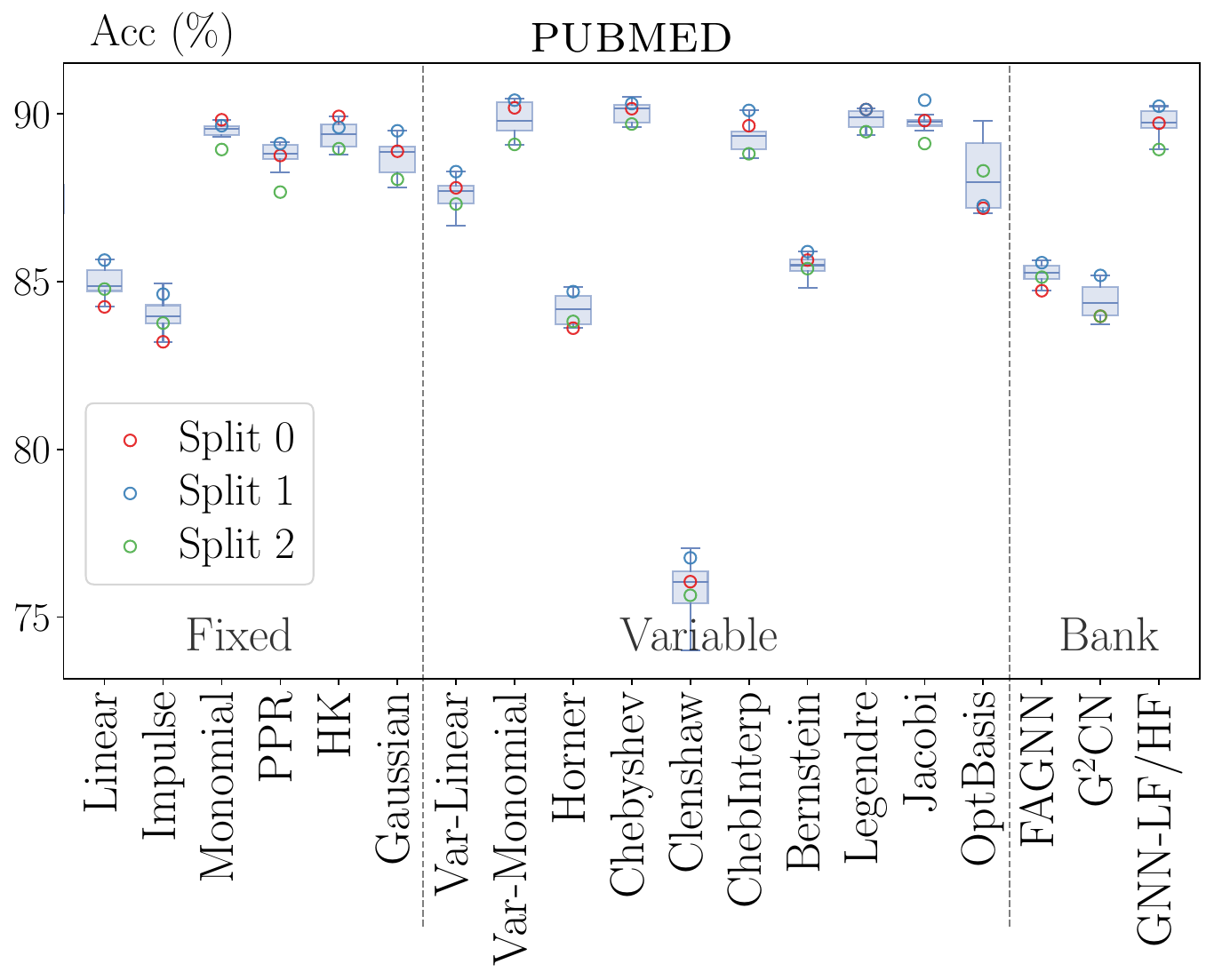}}
    \hfil
    \subcaptionbox{\ds{minesweeper}\label{ffiga:boxmb_minesweeper_full}}%
    [0.24\linewidth]{\includegraphics[height=1.31in]{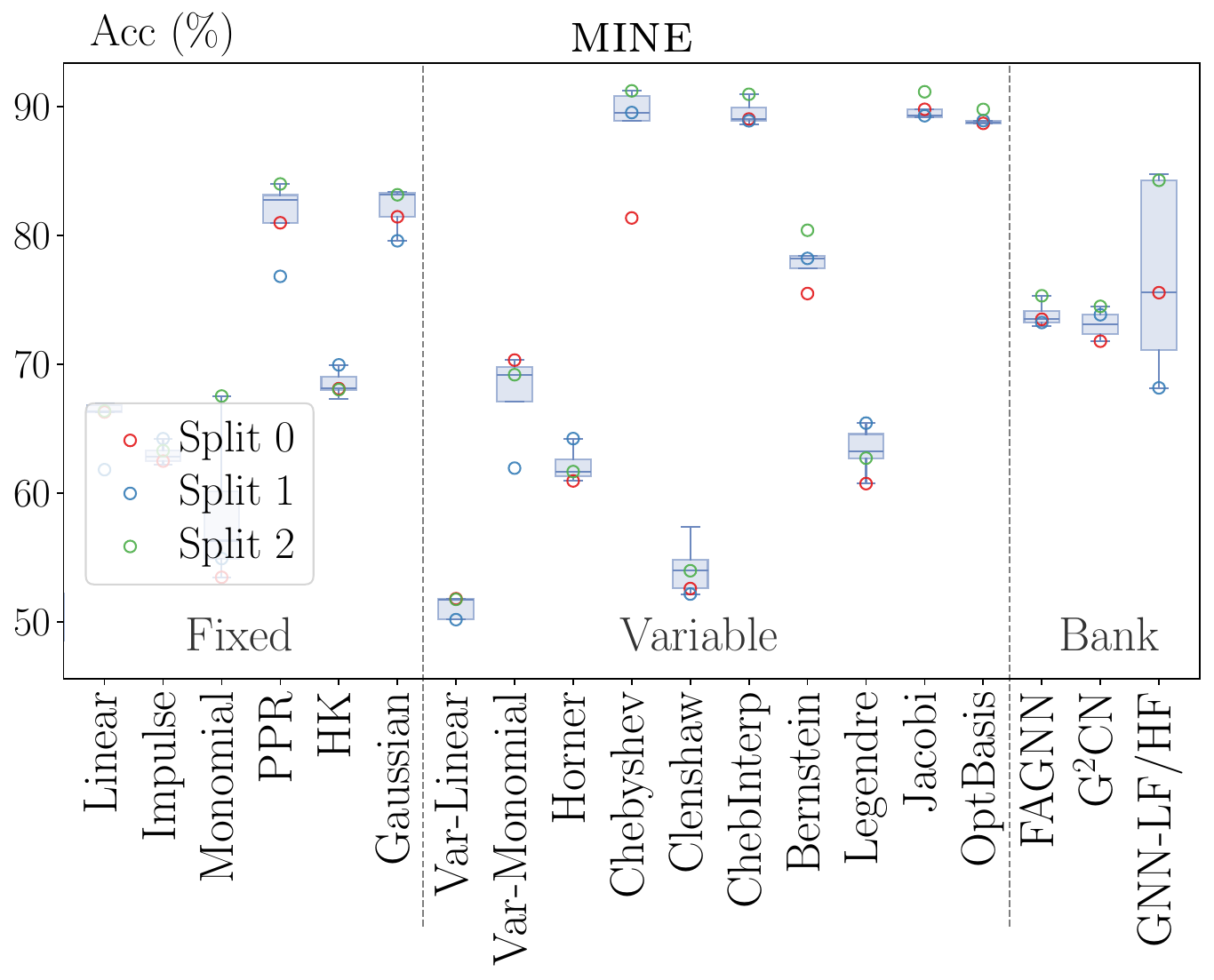}}
    \hfil
    \subcaptionbox{\ds{flickr}\label{ffiga:boxmb_flickr_full}}%
    [0.24\linewidth]{\includegraphics[height=1.31in]{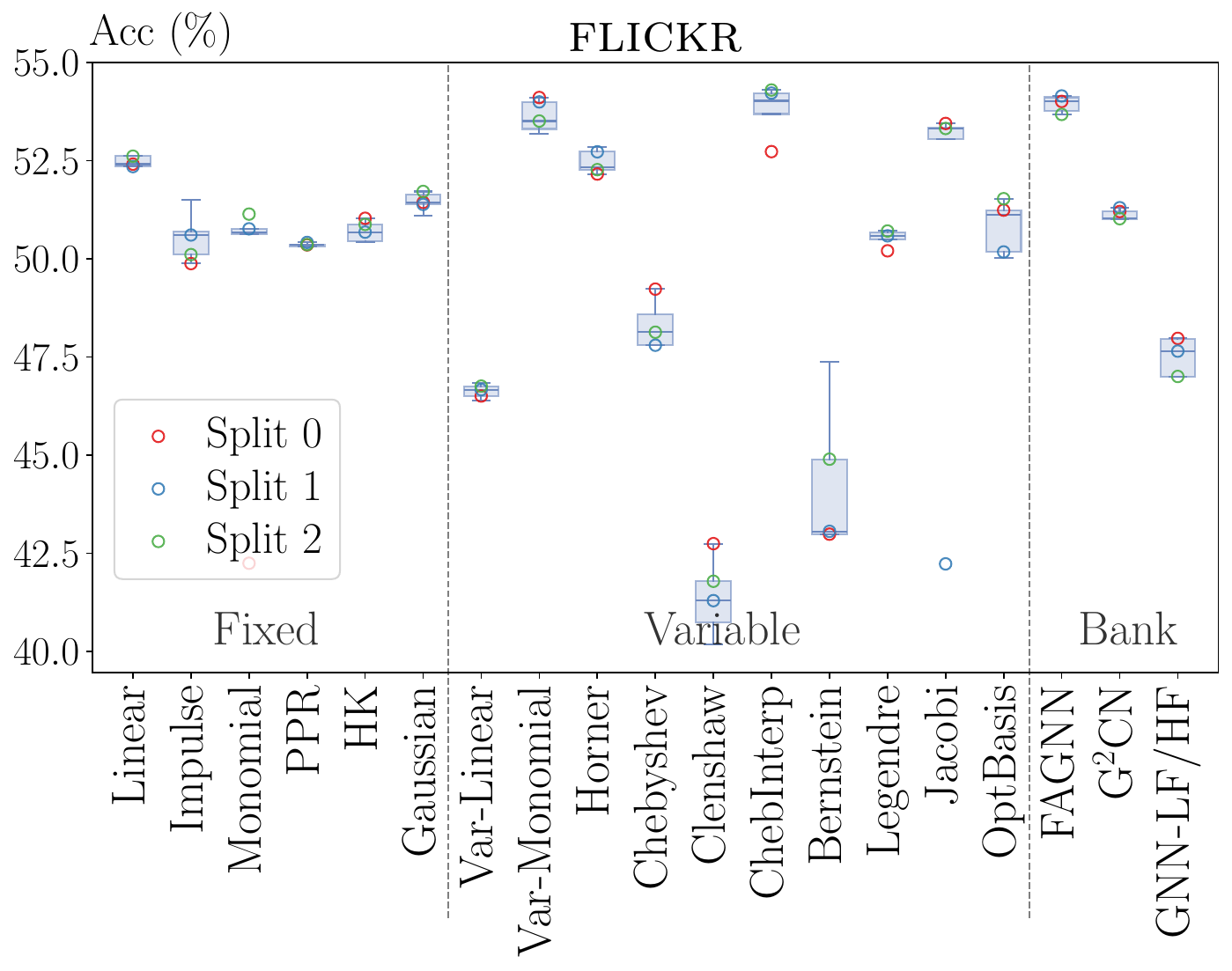}}
    \\ \vspace{8pt}
    \subcaptionbox{\ds{chameleon}\label{ffiga:boxmb_chameleon_filtered_full}}%
    [0.24\linewidth]{\includegraphics[height=1.31in]{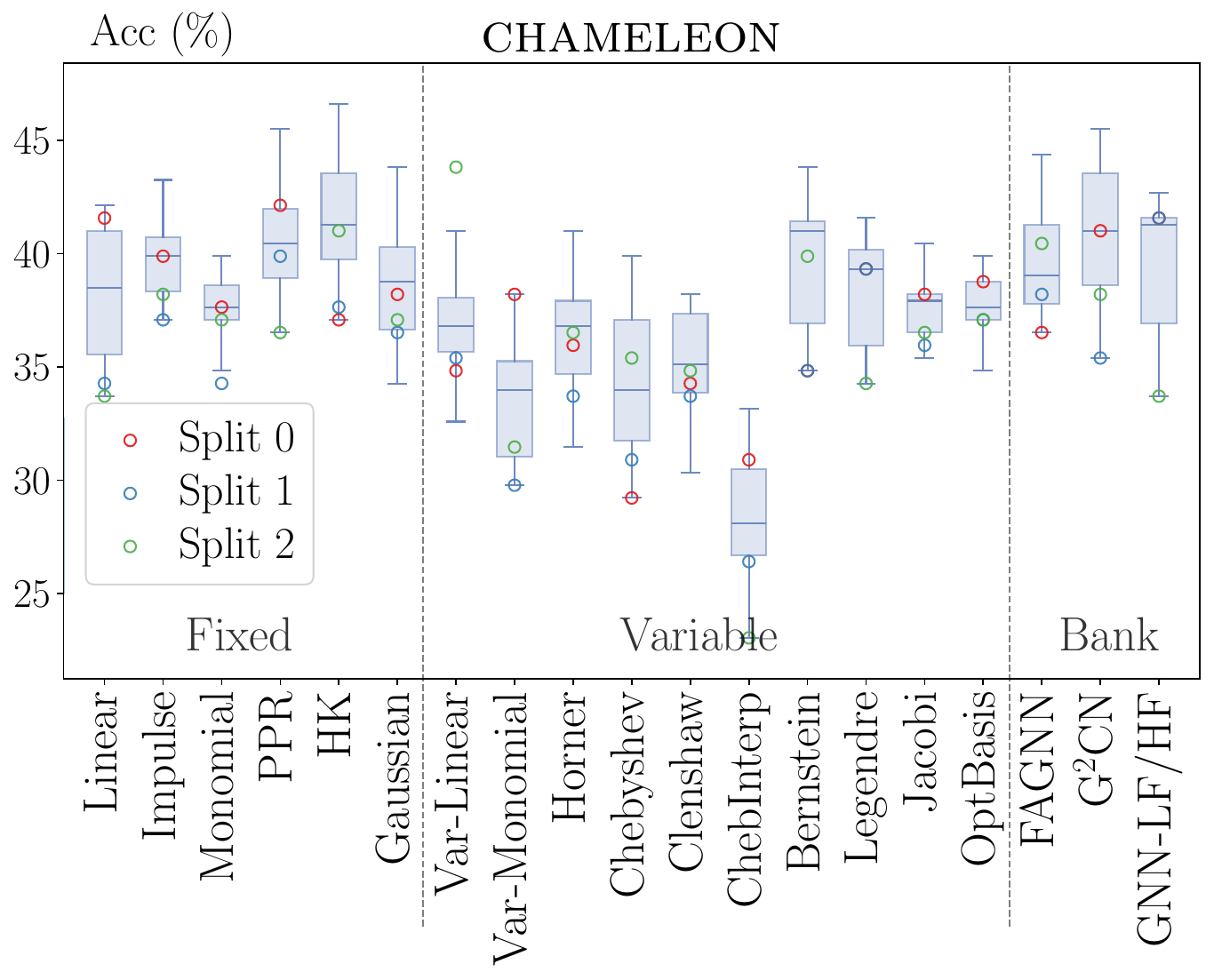}}
    \hfil
    \subcaptionbox{\ds{squirrel}\label{ffiga:boxmb_squirrel_filtered_full}}%
    [0.24\linewidth]{\includegraphics[height=1.31in]{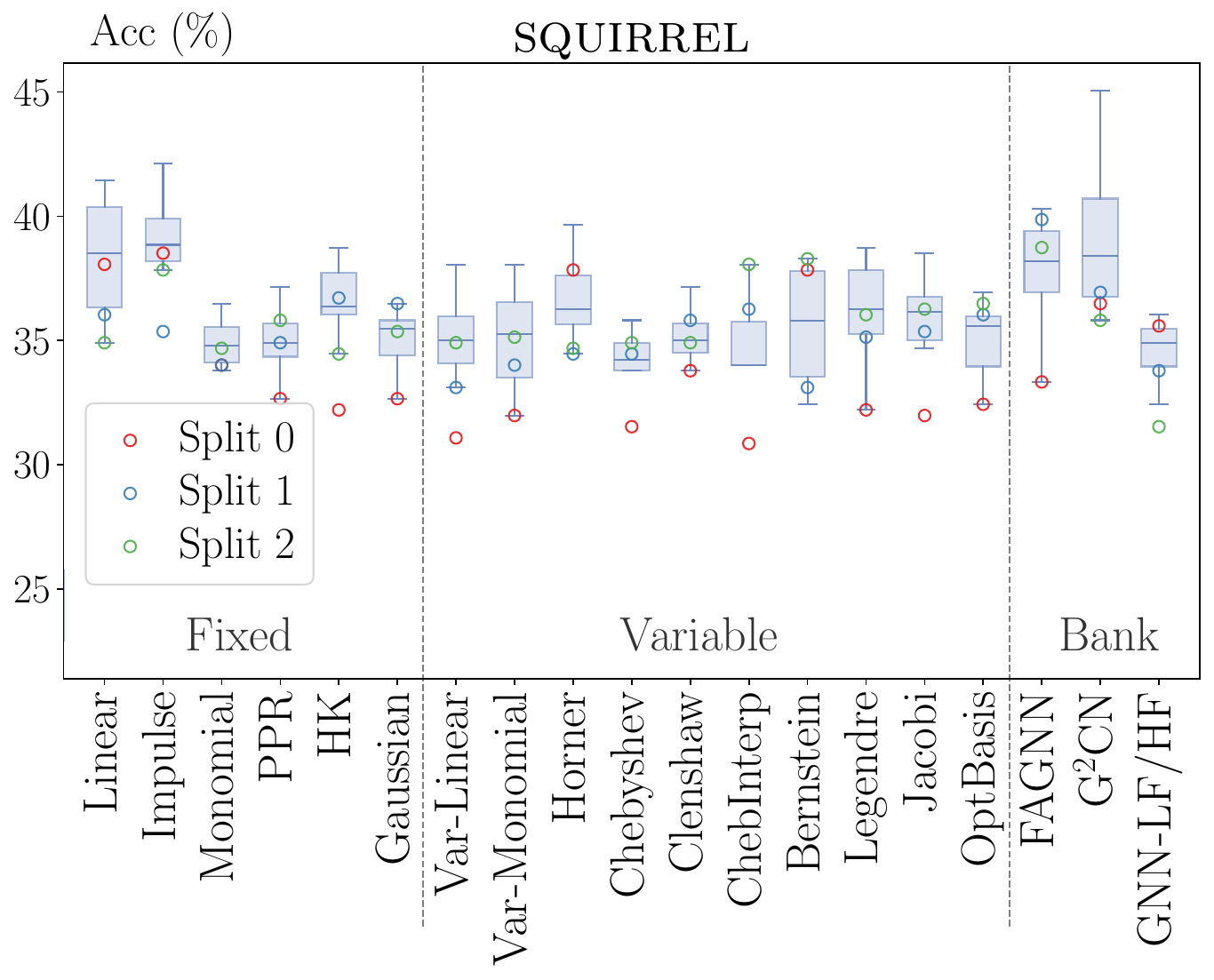}}
    \hfil
    \subcaptionbox{\ds{roman}\label{ffiga:boxmb_roman_empire_full}}%
    [0.24\linewidth]{\includegraphics[height=1.31in]{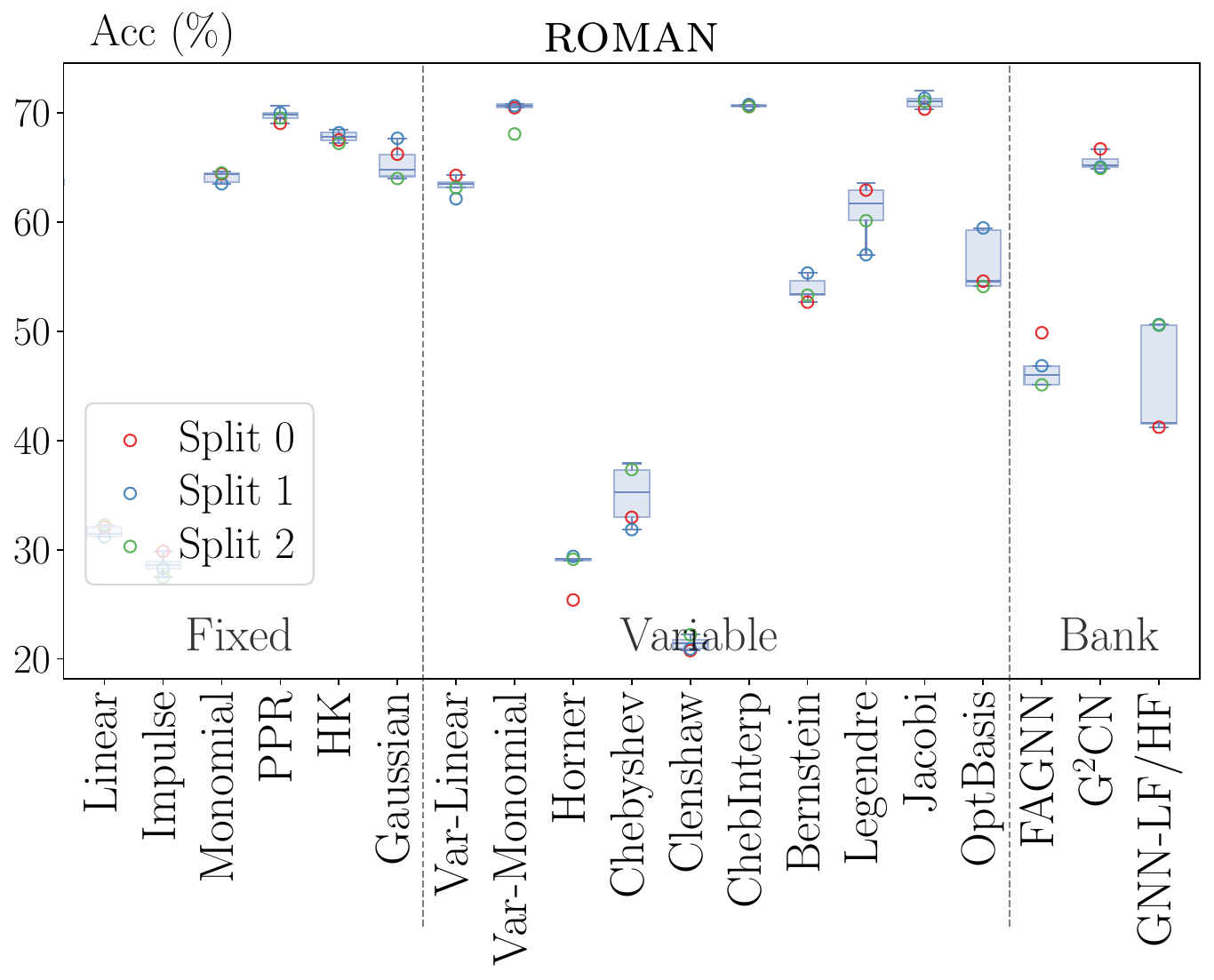}}
    \hfil
    \subcaptionbox{\ds{year}\label{ffiga:boxmb_arxiv-year_full}}%
    [0.24\linewidth]{\includegraphics[height=1.31in]{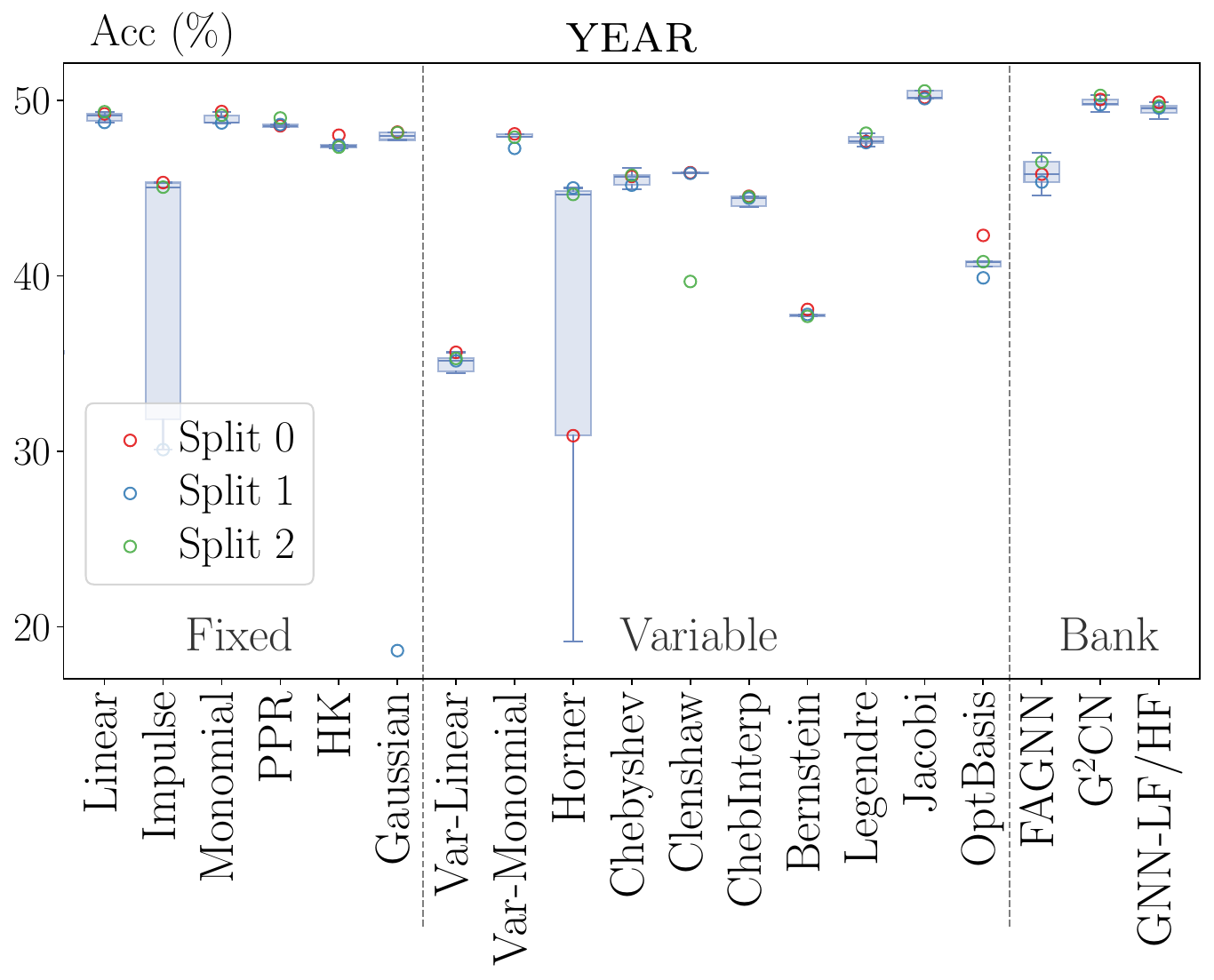}}
    \\ \vspace{8pt}
    \subcaptionbox{\ds{cora}\label{ffiga:boxmb_cora_full}}%
    [0.24\linewidth]{\includegraphics[height=1.31in]{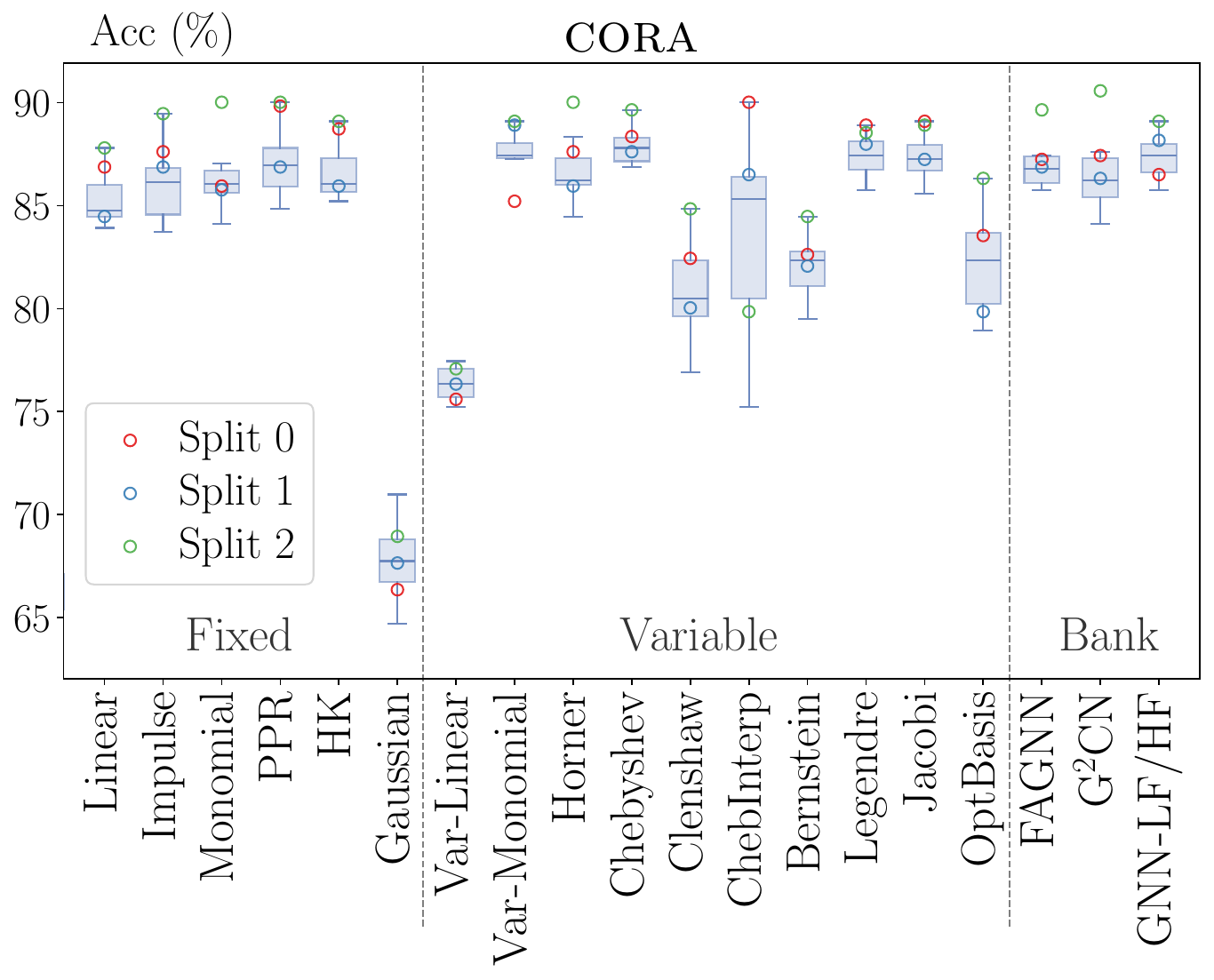}}
    \hfil
    \subcaptionbox{\ds{arxiv}\label{ffiga:boxmb_arxiv_full}}%
    [0.24\linewidth]{\includegraphics[height=1.31in]{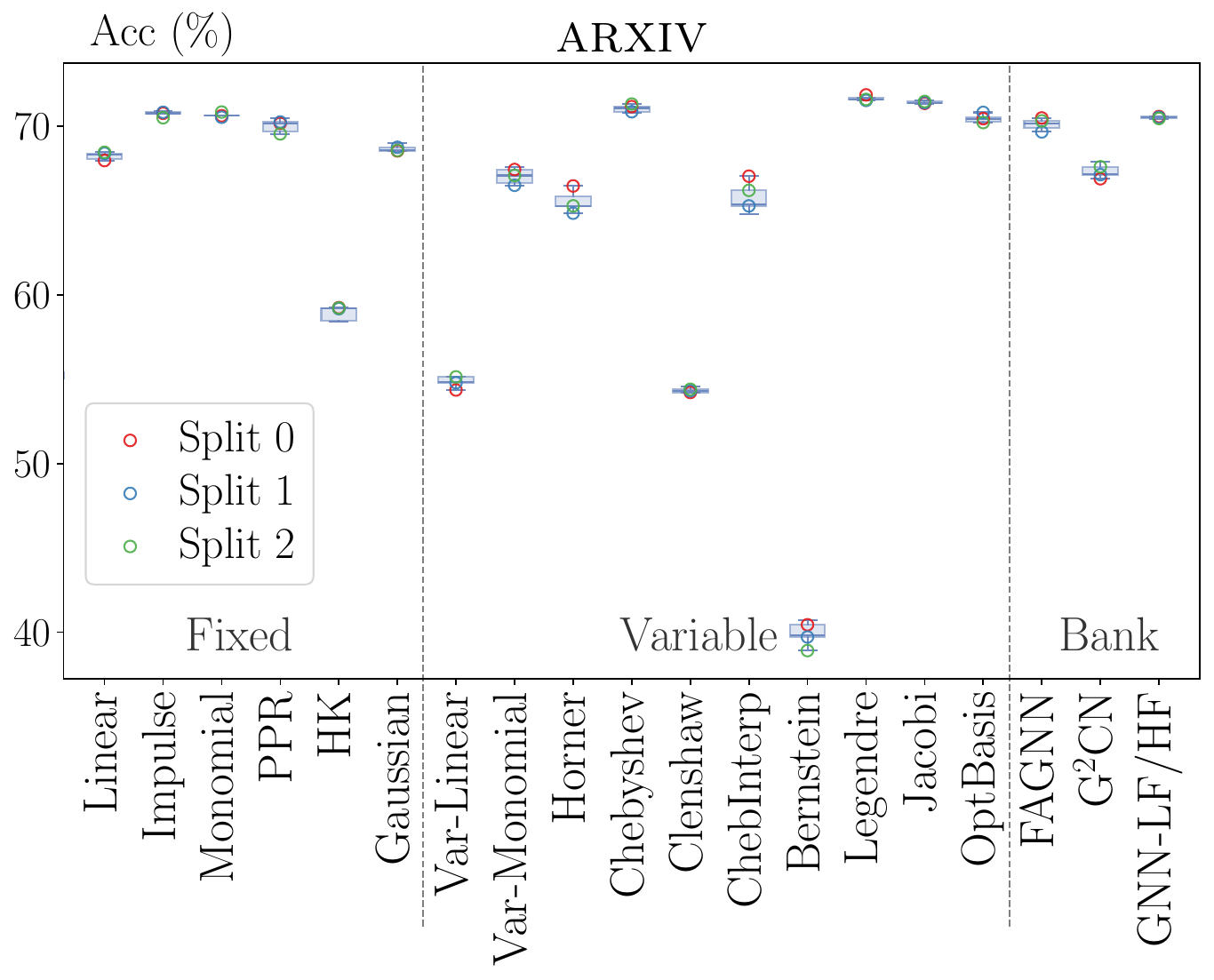}}
    \hfil
    \subcaptionbox{\ds{penn94}\label{ffiga:boxmb_penn94_full}}%
    [0.24\linewidth]{\includegraphics[height=1.31in]{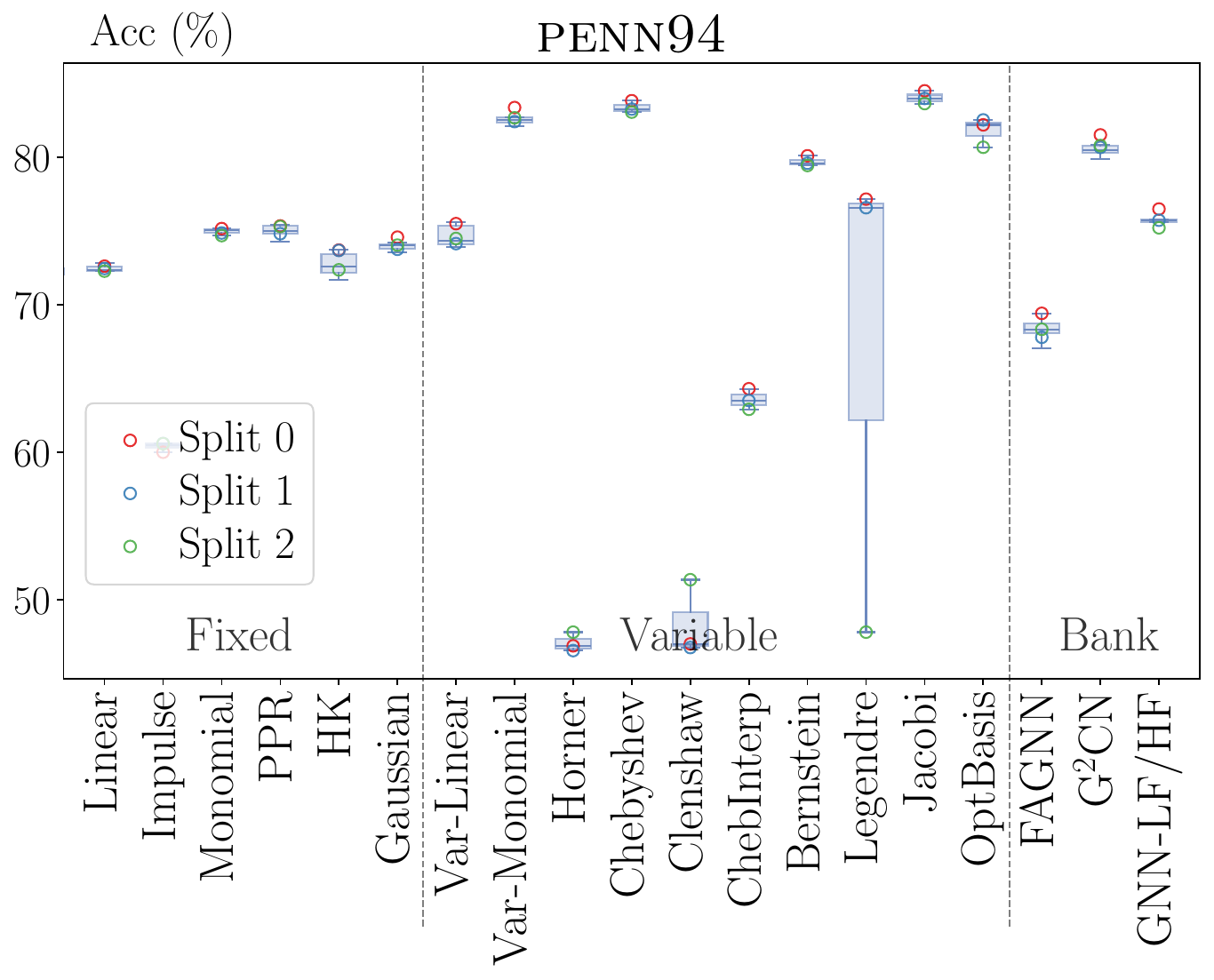}}
    \hfil
    \subcaptionbox{\ds{genius}\label{ffiga:boxmb_genius_full}}%
    [0.24\linewidth]{\includegraphics[height=1.31in]{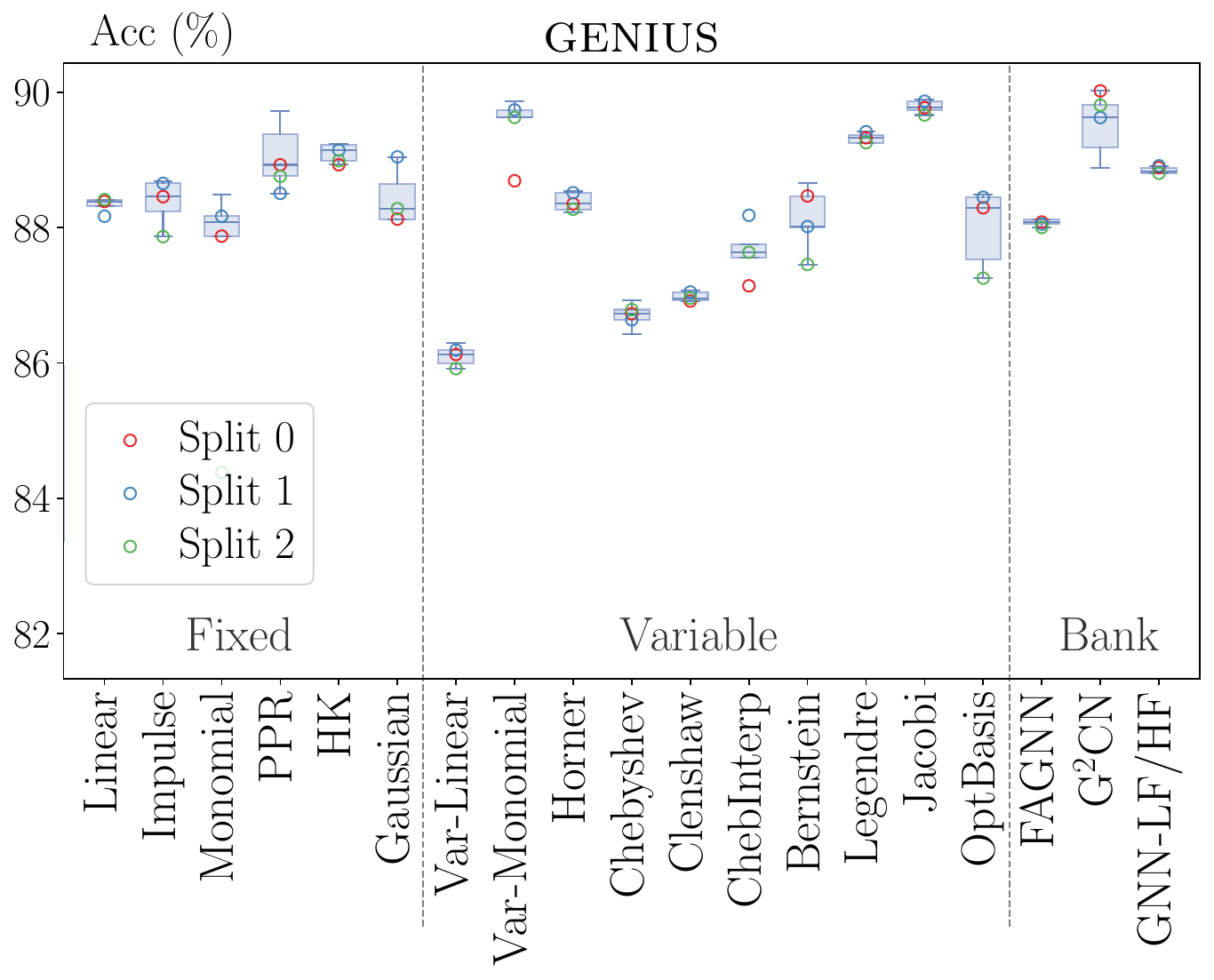}}
    \\ \vspace{8pt}
    \caption{Effectiveness variance of \textit{mini-batch} training on 6 homophilous and 6 heterophilous datasets.}
  \label{figa:boxmb_full}
\end{figure}

\clearpage

\begin{figure*}[tp]

\section{Specific Evaluation}
\label{seca:extra}

\subsection{Effect of Propagation Hop}
    \centering
    \subcaptionbox{Fixed filters on \ds{cora}\label{ffiga:hop_fix_cora}}%
    [0.24\linewidth]{\includegraphics[height=1.05in]{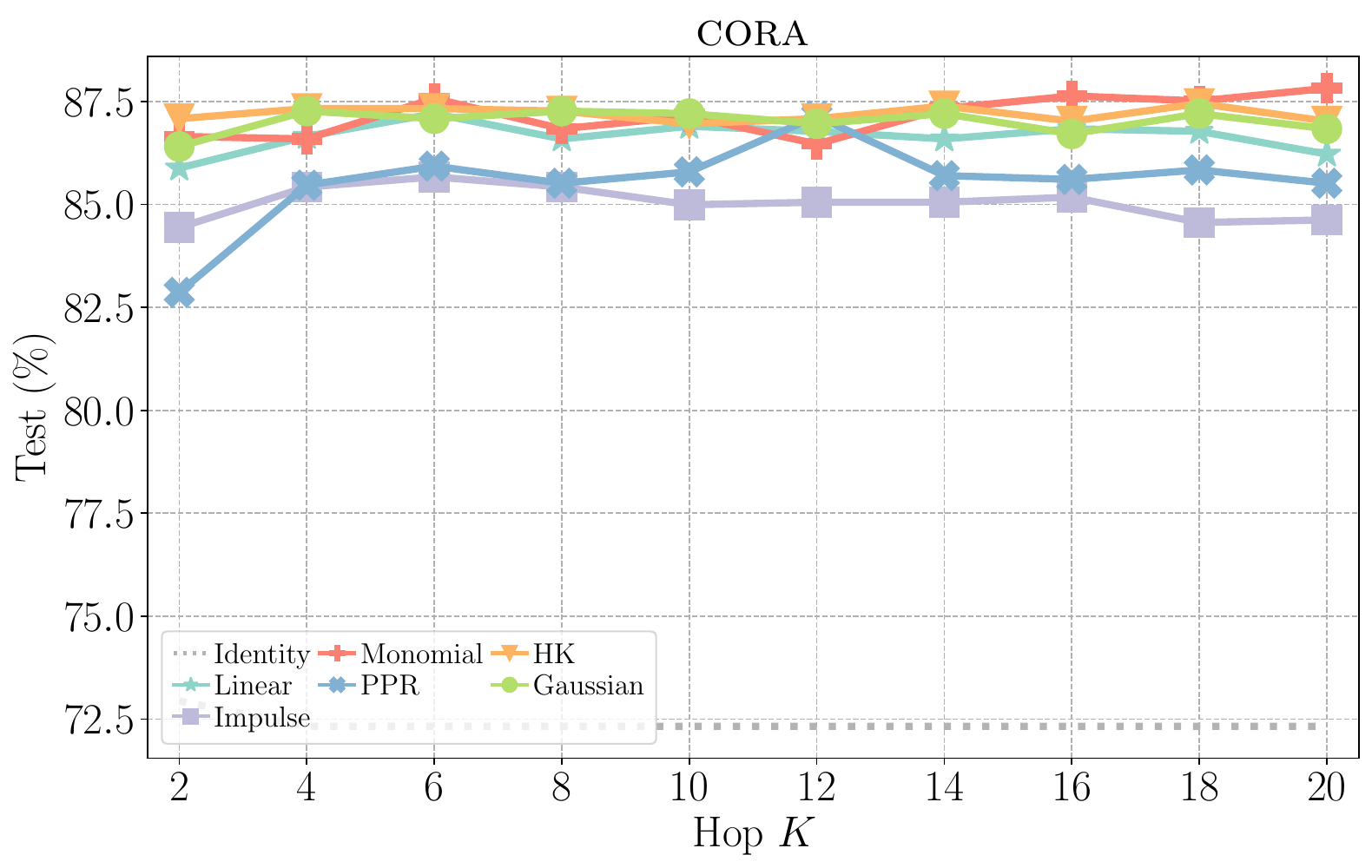}}
    \hfil
    \subcaptionbox{Variable filters on \ds{cora}\label{ffiga:hop_var_cora}}%
    [0.24\linewidth]{\includegraphics[height=1.05in]{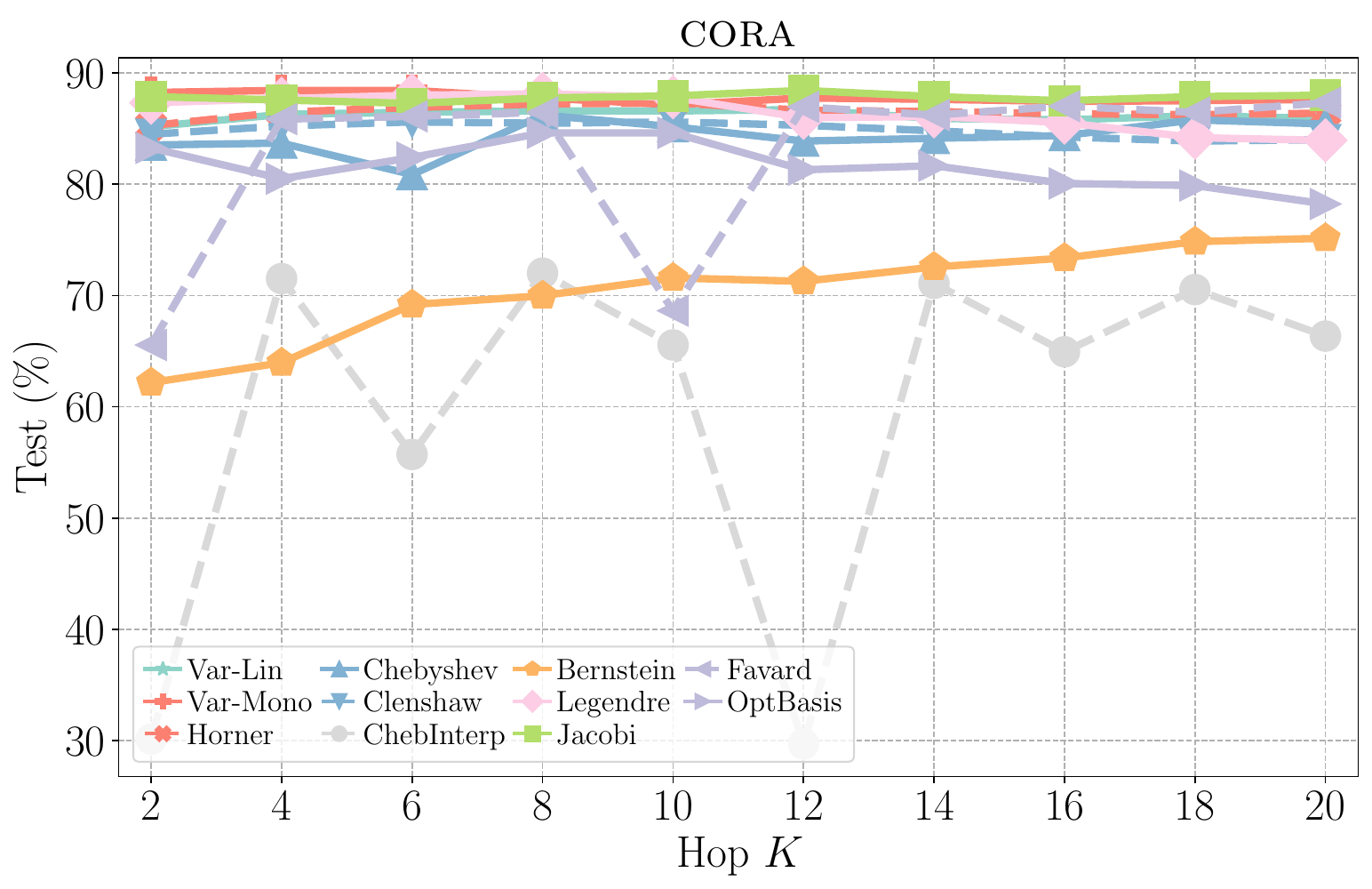}}
    \hfil
    \subcaptionbox{Fixed filters on \ds{citeseer}\label{ffiga:hop_fix_citeseer}}%
    [0.24\linewidth]{\includegraphics[height=1.05in]{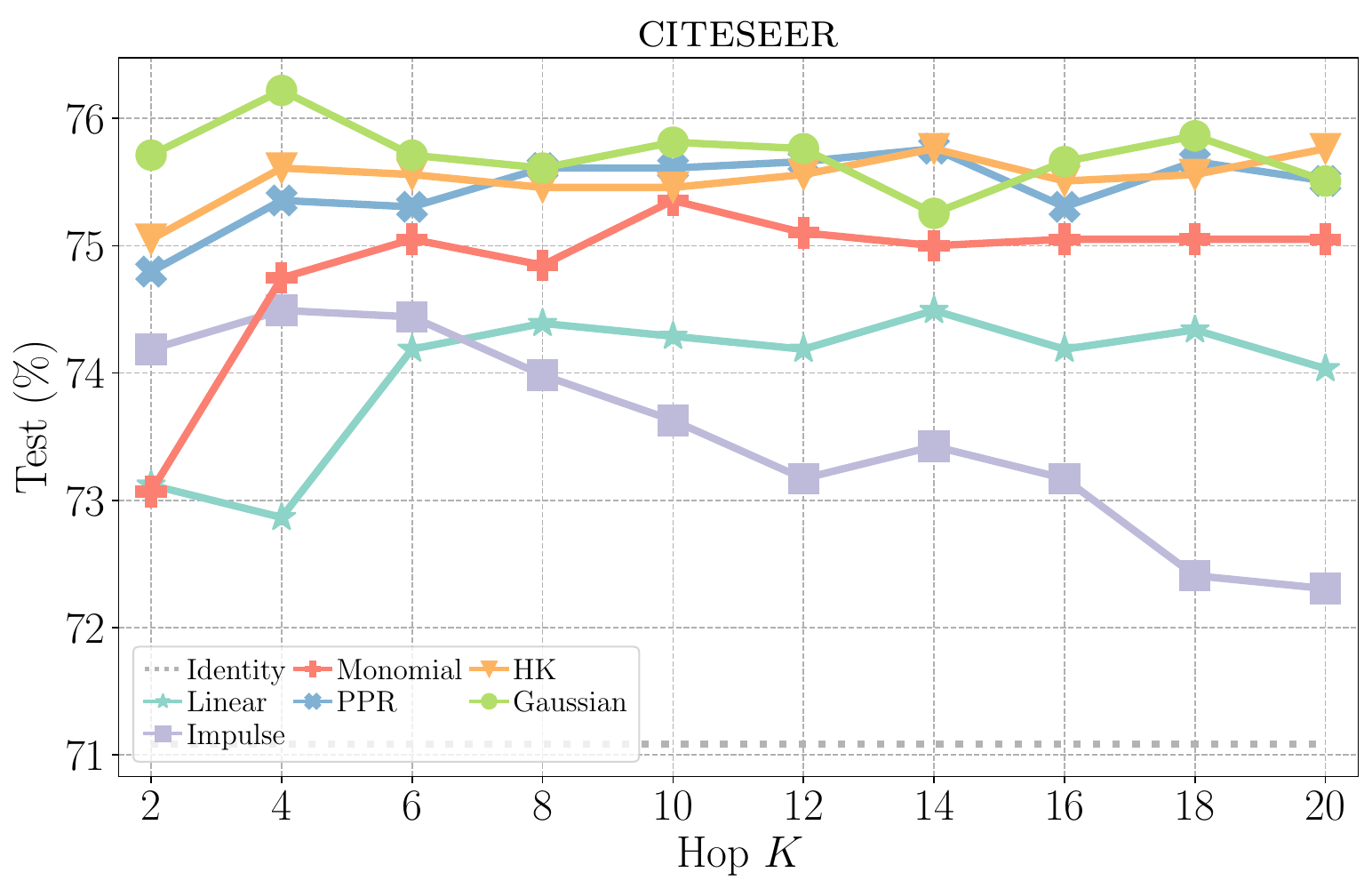}}
    \hfil
    \subcaptionbox{Variable filters on \ds{citeseer}\label{ffiga:hop_var_citeseer}}%
    [0.24\linewidth]{\includegraphics[height=1.05in]{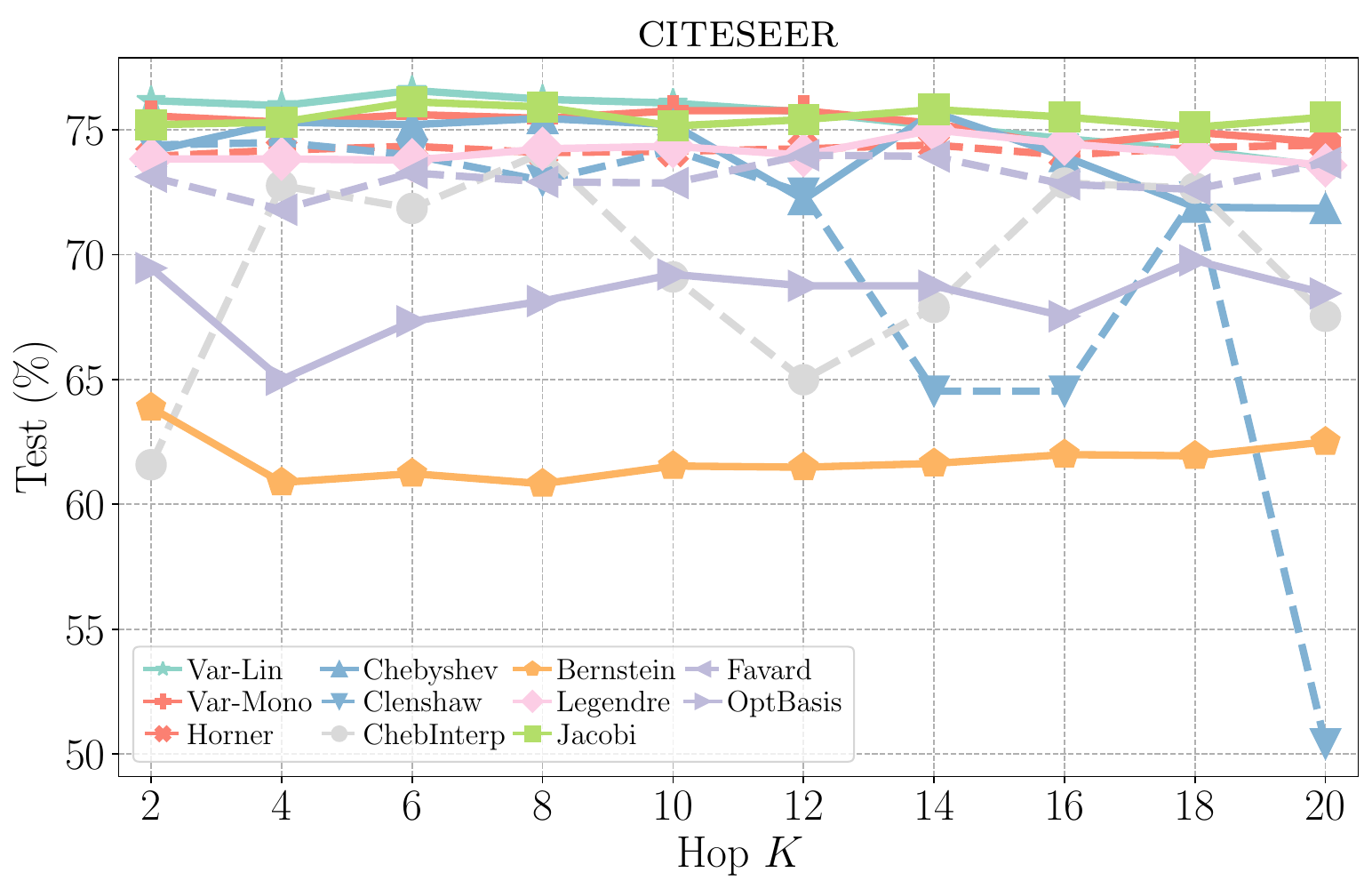}}
    \\ 
    \subcaptionbox{Fixed filters on \ds{pubmed}\label{ffiga:hop_fix_pubmed}}%
    [0.24\linewidth]{\includegraphics[height=1.05in]{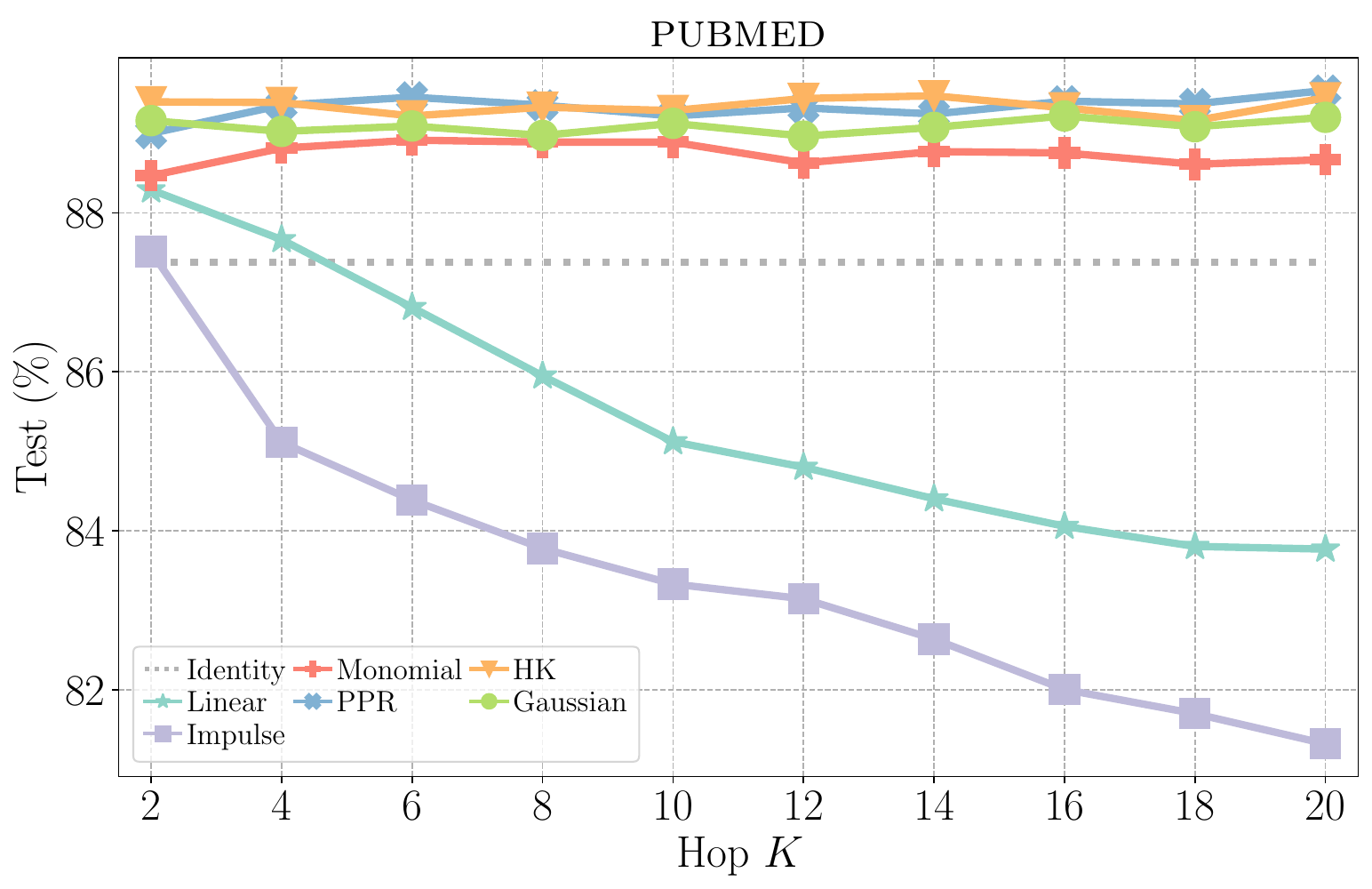}}
    \hfil
    \subcaptionbox{Variable filters on \ds{pubmed}\label{ffiga:hop_var_pubmed}}%
    [0.24\linewidth]{\includegraphics[height=1.05in]{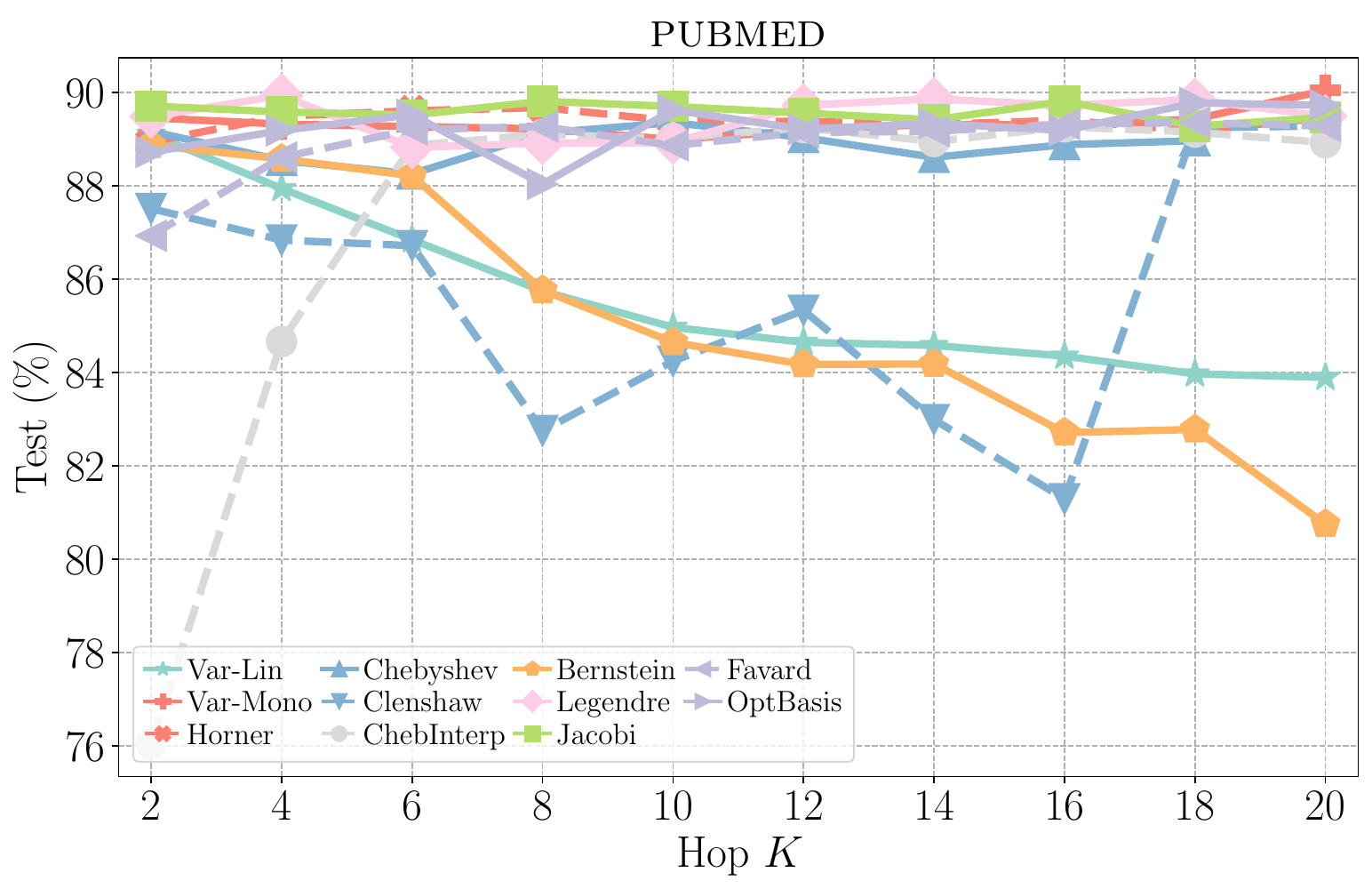}}
    \hfil
    \subcaptionbox{Fixed filters on \ds{minesweeper}\label{ffiga:hop_fix_minesweeper}}%
    [0.24\linewidth]{\includegraphics[height=1.05in]{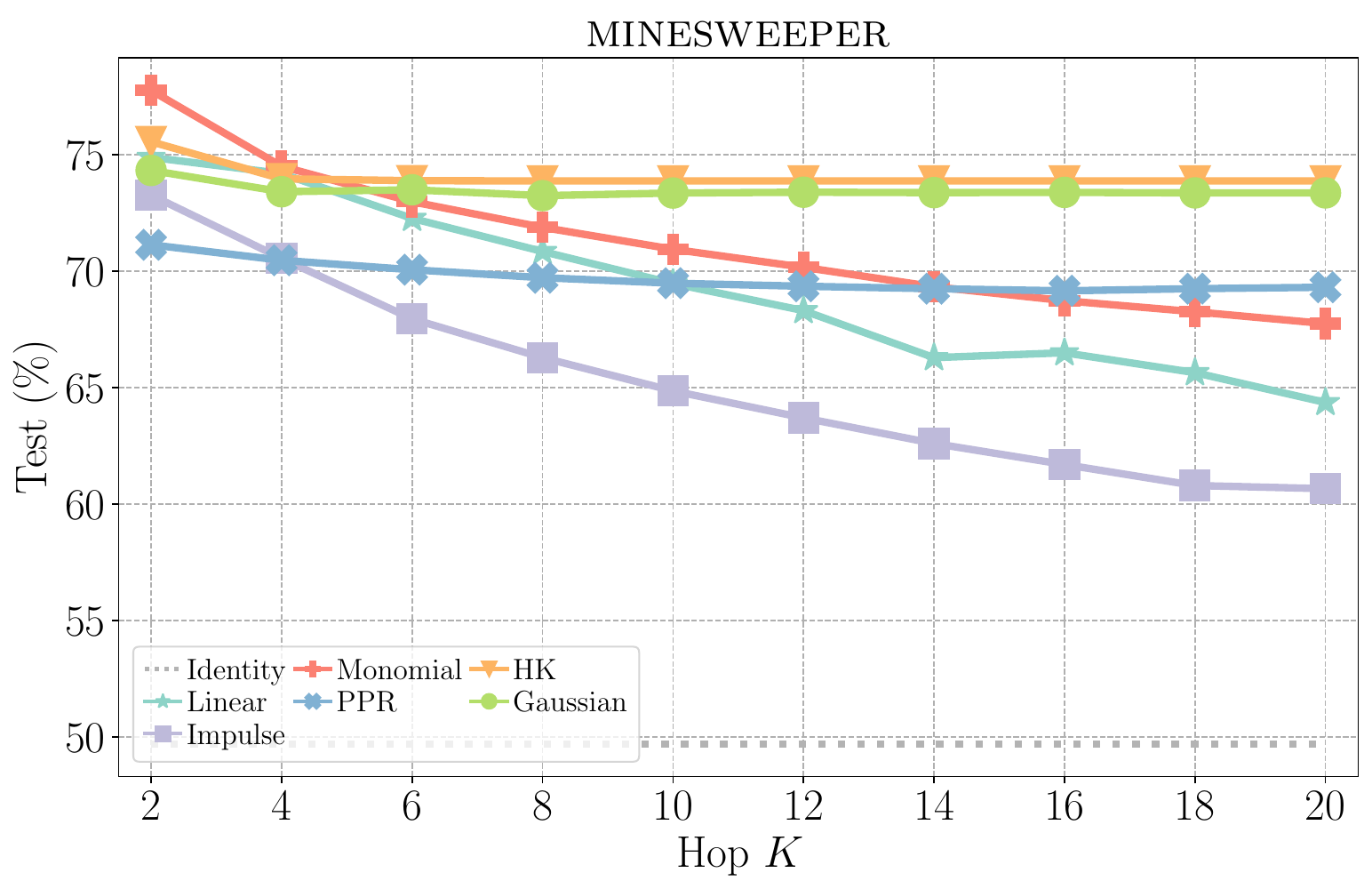}}
    \hfil
    \subcaptionbox{Variable filters on \ds{minesweeper}\label{ffiga:hop_var_minesweeper}}%
    [0.24\linewidth]{\includegraphics[height=1.05in]{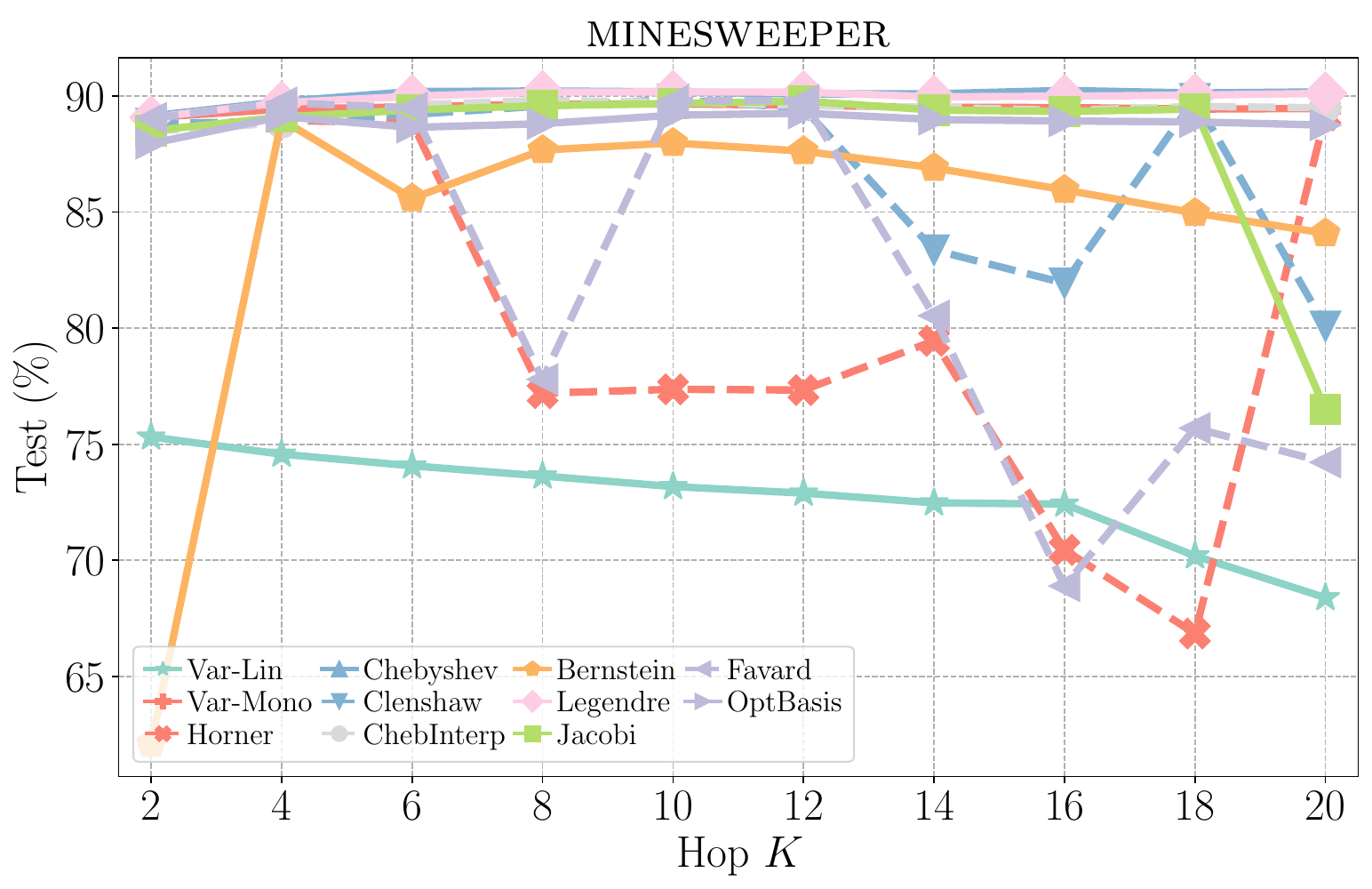}}
    \\
    \subcaptionbox{Fixed filters on \ds{questions}\label{ffiga:hop_fix_questions}}%
    [0.24\linewidth]{\includegraphics[height=1.05in]{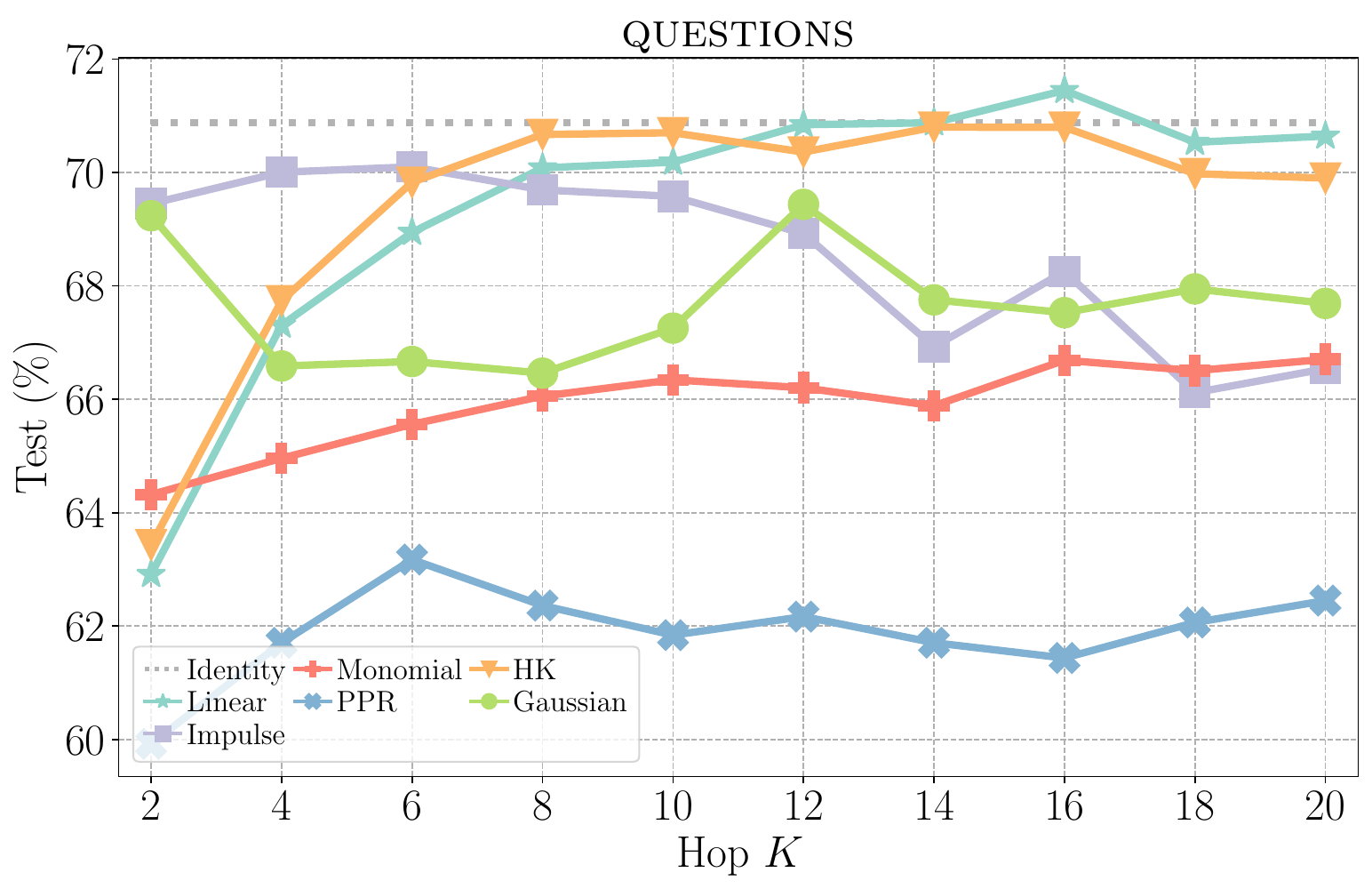}}
    \hfil
    \subcaptionbox{Variable filters on \ds{questions}\label{ffiga:hop_var_questions}}%
    [0.24\linewidth]{\includegraphics[height=1.05in]{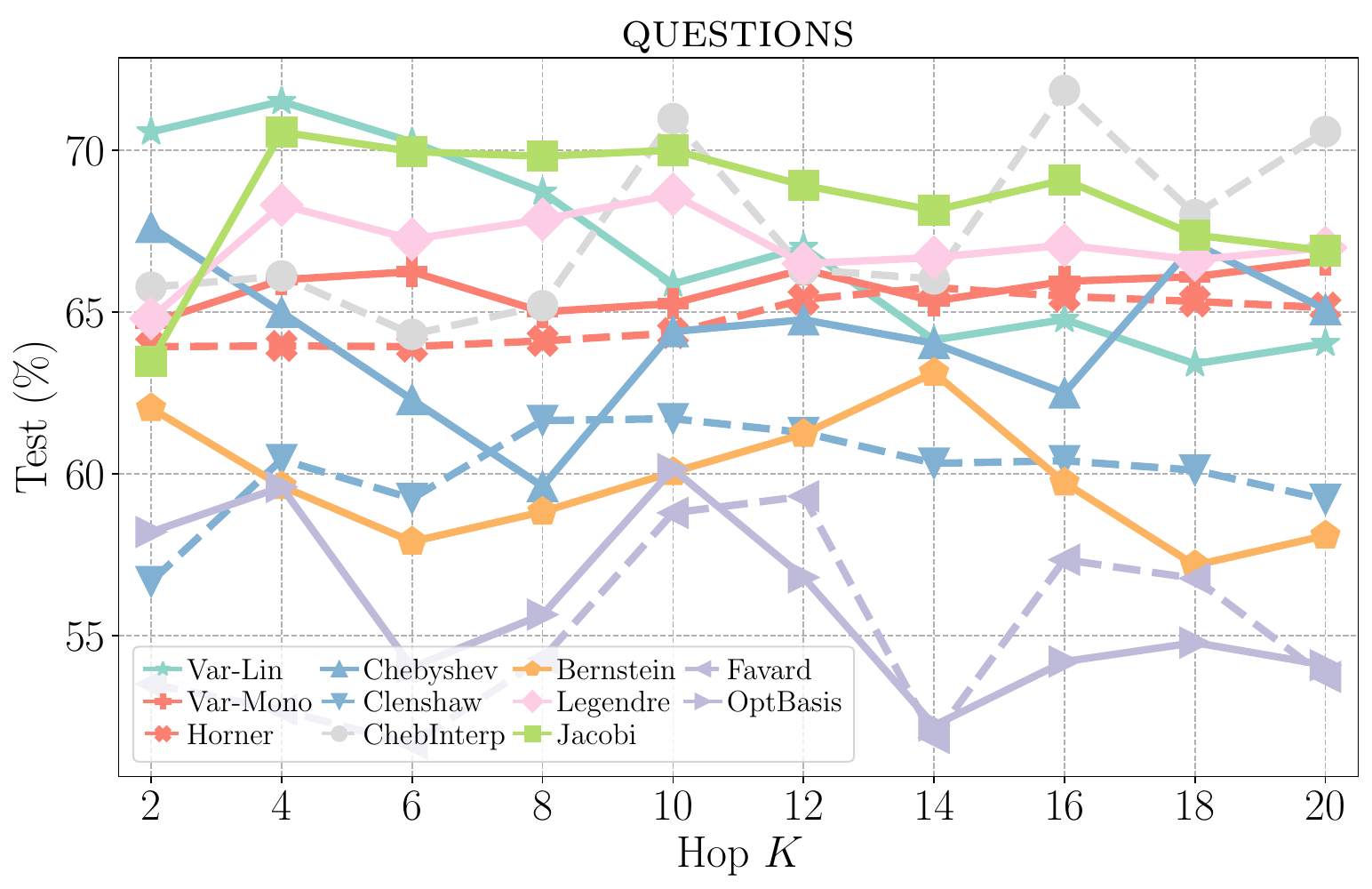}}
    \hfil
    \subcaptionbox{Fixed filters on \ds{tolokers}\label{ffiga:hop_fix_tolokers}}%
    [0.24\linewidth]{\includegraphics[height=1.05in]{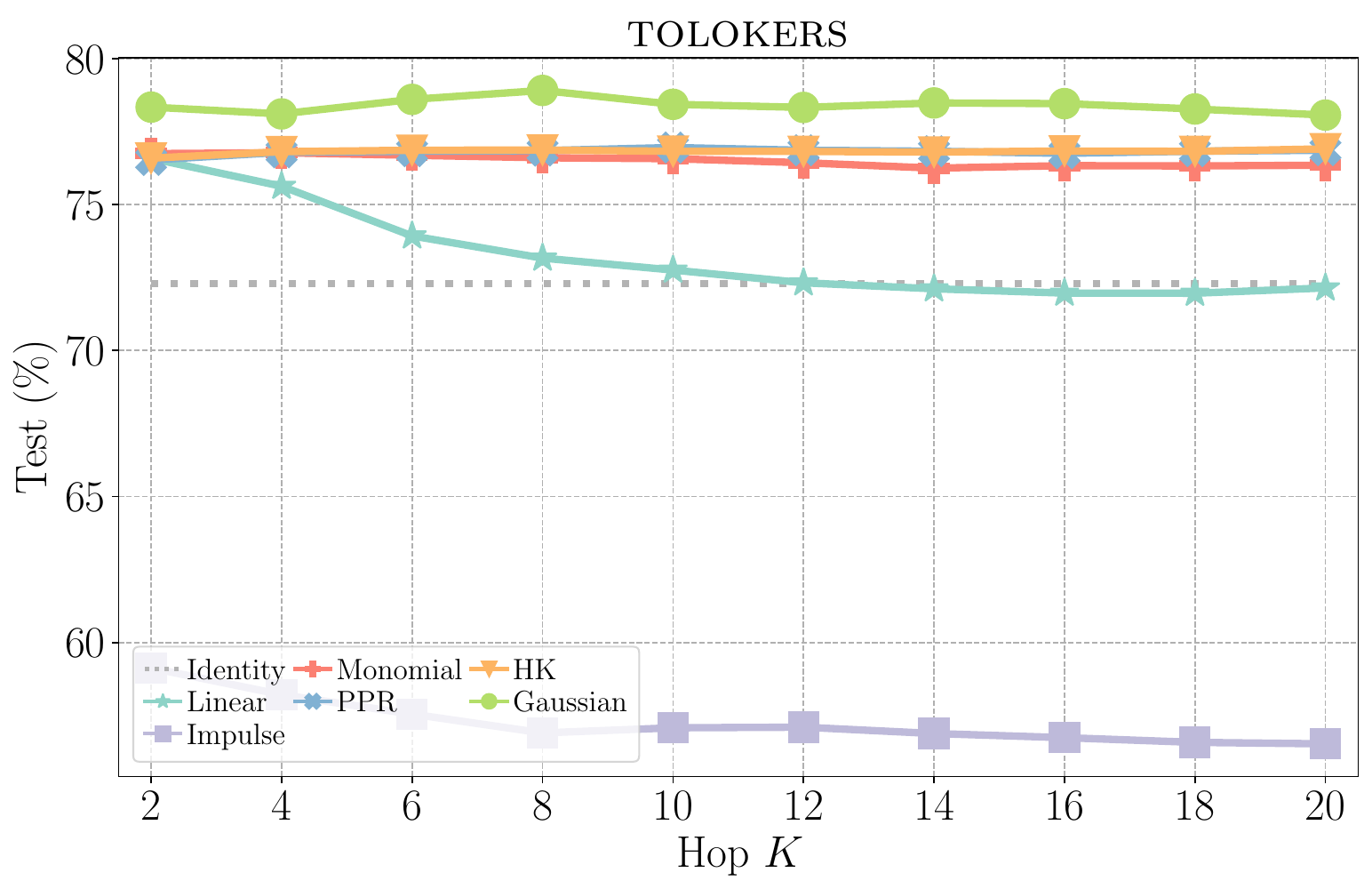}}
    \hfil
    \subcaptionbox{Variable filters on \ds{tolokers}\label{ffiga:hop_var_tolokers}}%
    [0.24\linewidth]{\includegraphics[height=1.05in]{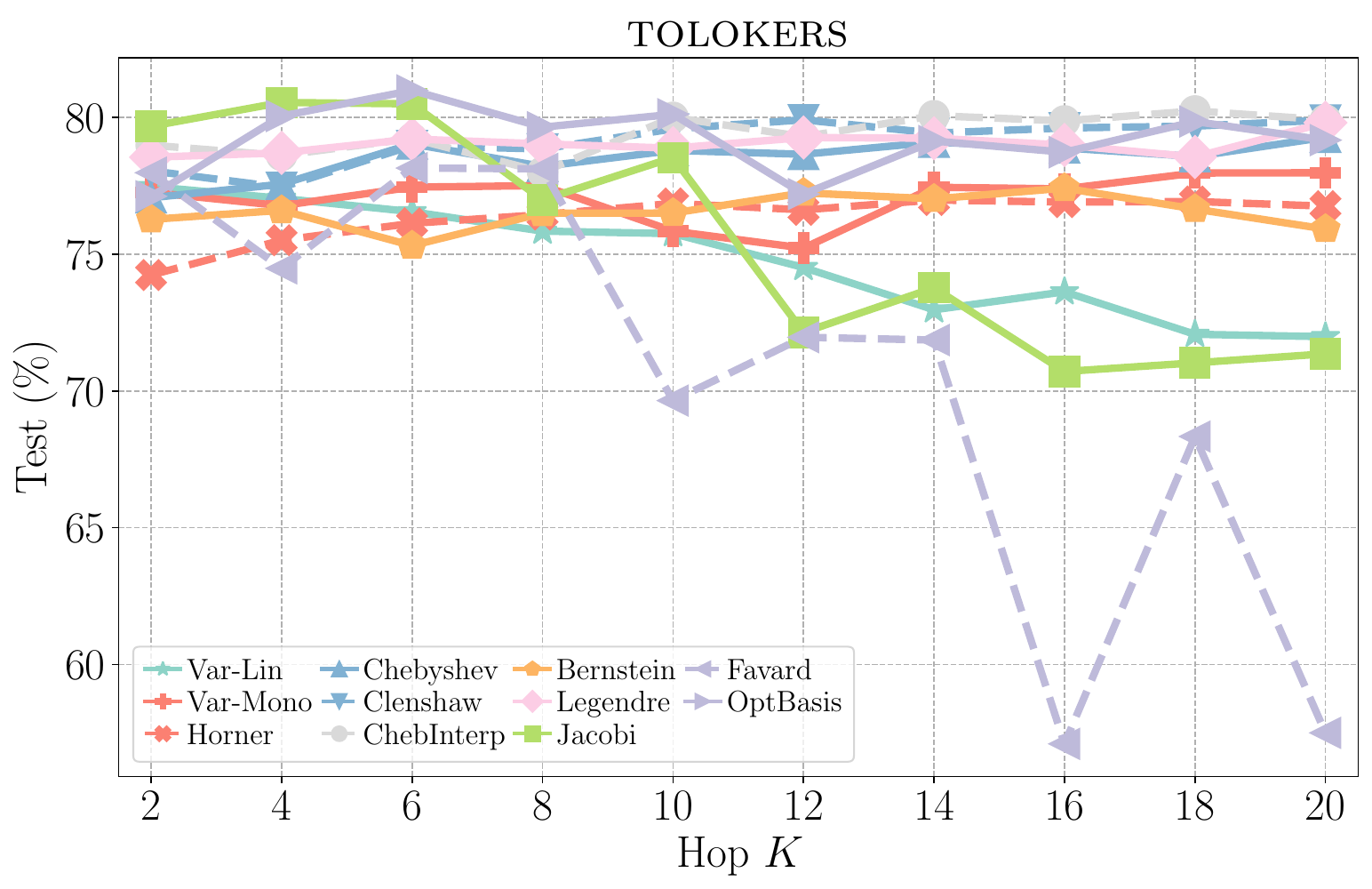}}
    \caption{Effect of \textit{propagation hops} $K$ of full-batch fixed and variable filters on homophilous datasets. }
  \label{figa:hop_homo}
\vspace{4mm}
\end{figure*}

\begin{figure*}[bp]
    \centering
    \subcaptionbox{Fixed filters on \ds{chameleon}\label{ffiga:hop_fix_chameleon_filtered}}%
    [0.24\linewidth]{\includegraphics[height=1.05in]{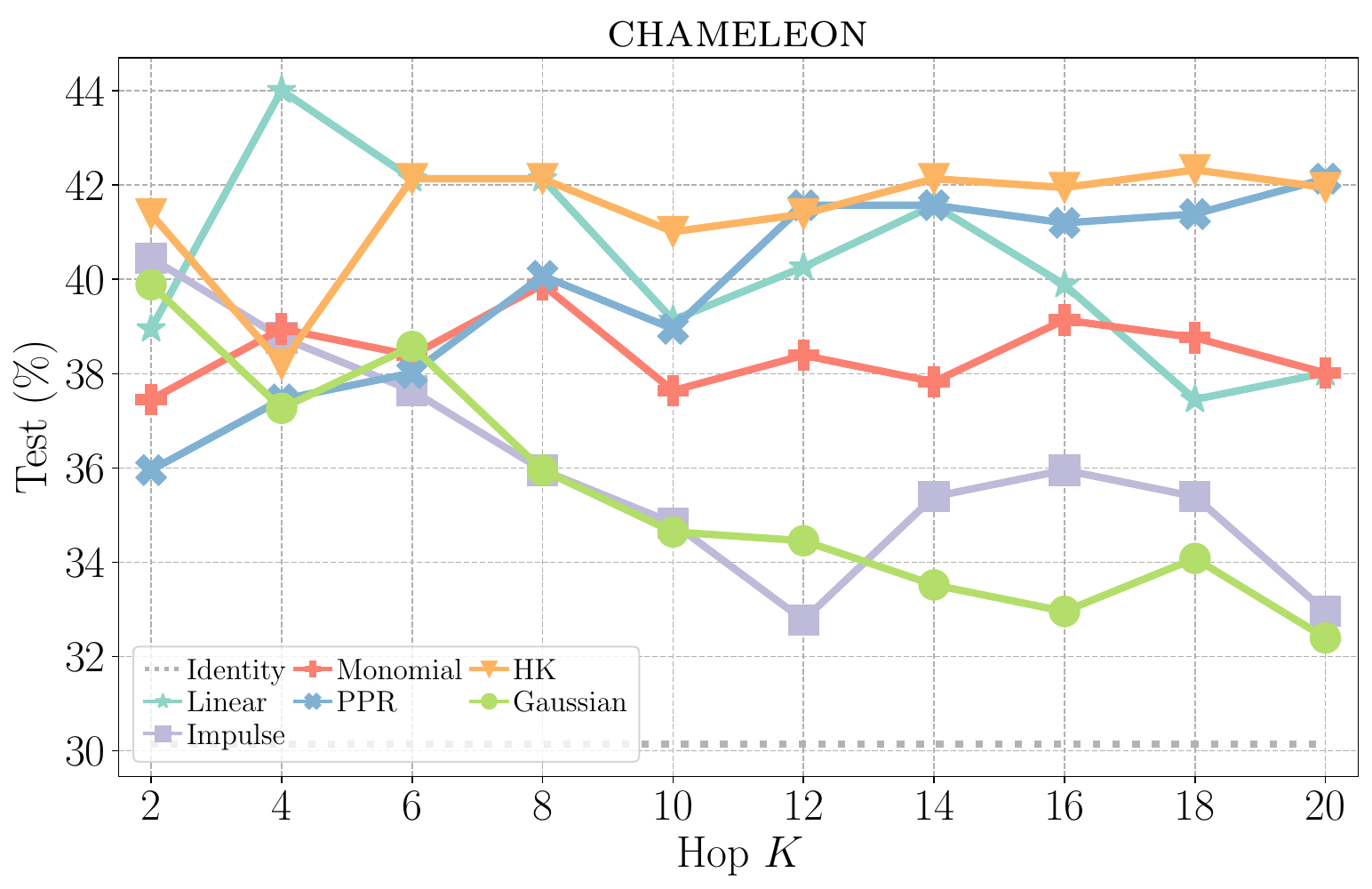}}
    \hfil
    \subcaptionbox{Variable filters on \ds{chameleon}\label{ffiga:hop_var_chameleon_filtered}}%
    [0.24\linewidth]{\includegraphics[height=1.05in]{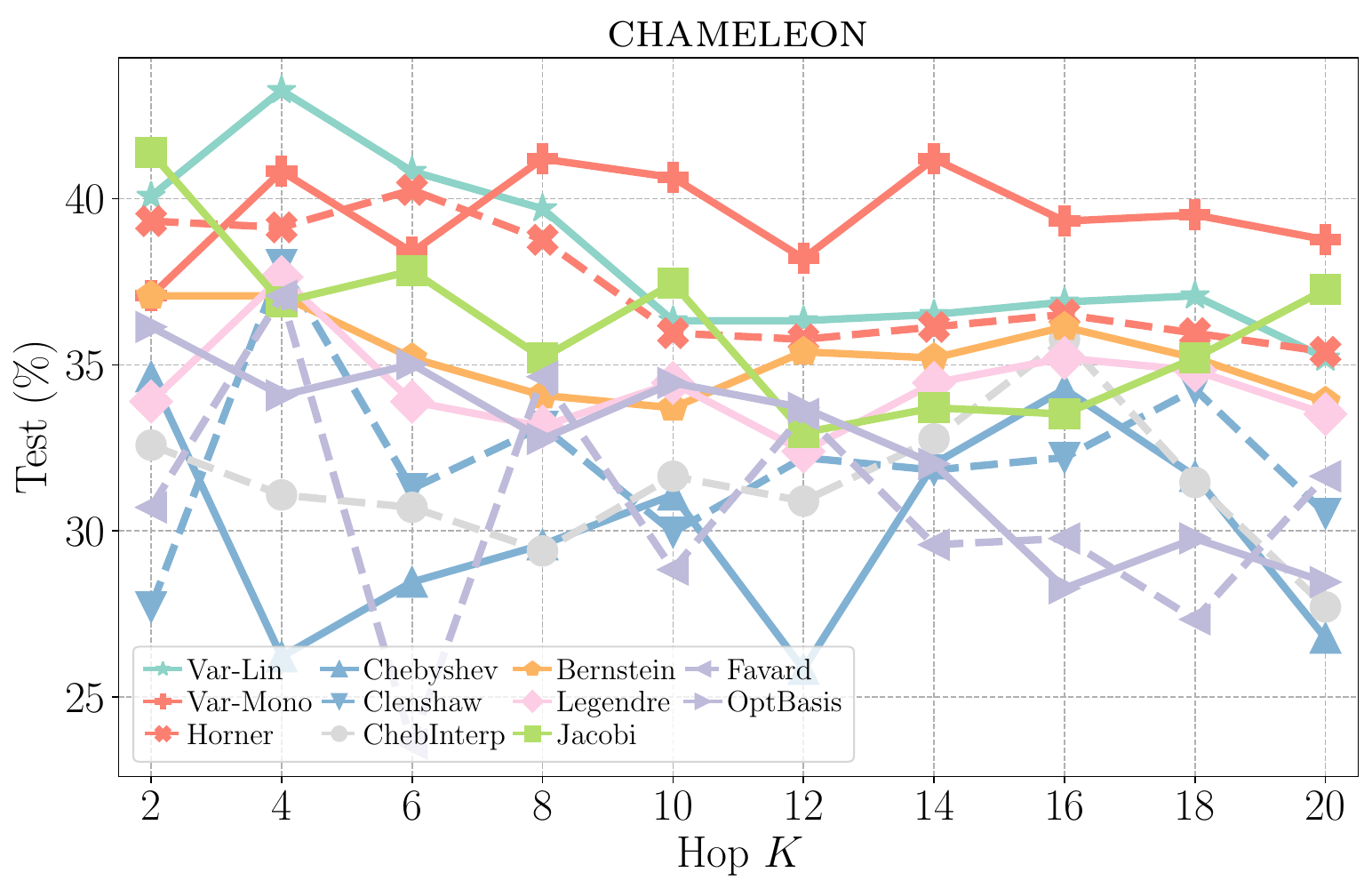}}
    \hfil
    \subcaptionbox{Fixed filters on \ds{squirrel}\label{ffiga:hop_fix_squirrel_filtered}}%
    [0.24\linewidth]{\includegraphics[height=1.05in]{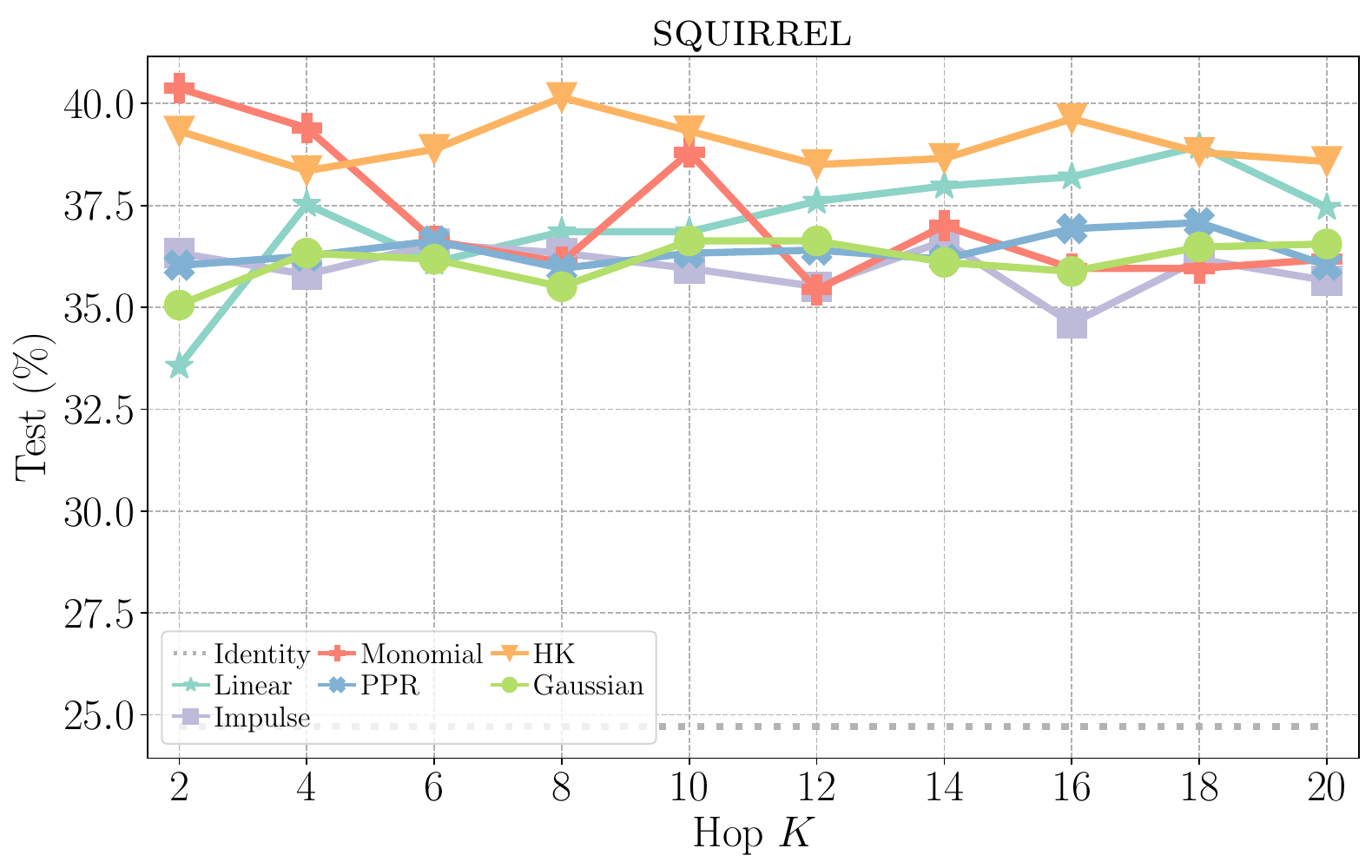}}
    \hfil
    \subcaptionbox{Variable filters on \ds{squirrel}\label{ffiga:hop_var_squirrel_filtered}}%
    [0.24\linewidth]{\includegraphics[height=1.05in]{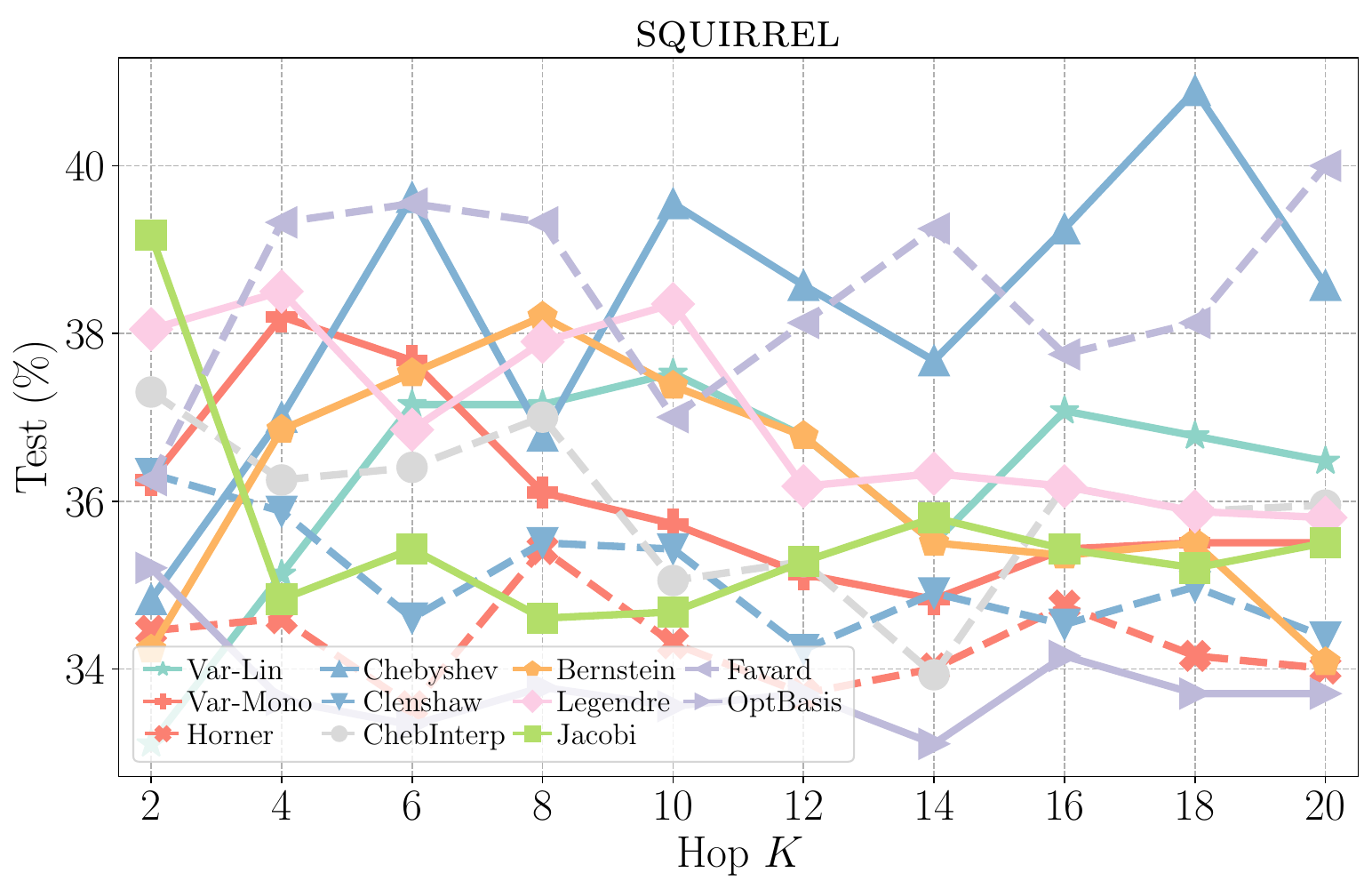}}
    \\ 
    \subcaptionbox{Fixed filters on \ds{actor}\label{ffiga:hop_fix_actor}}%
    [0.24\linewidth]{\includegraphics[height=1.05in]{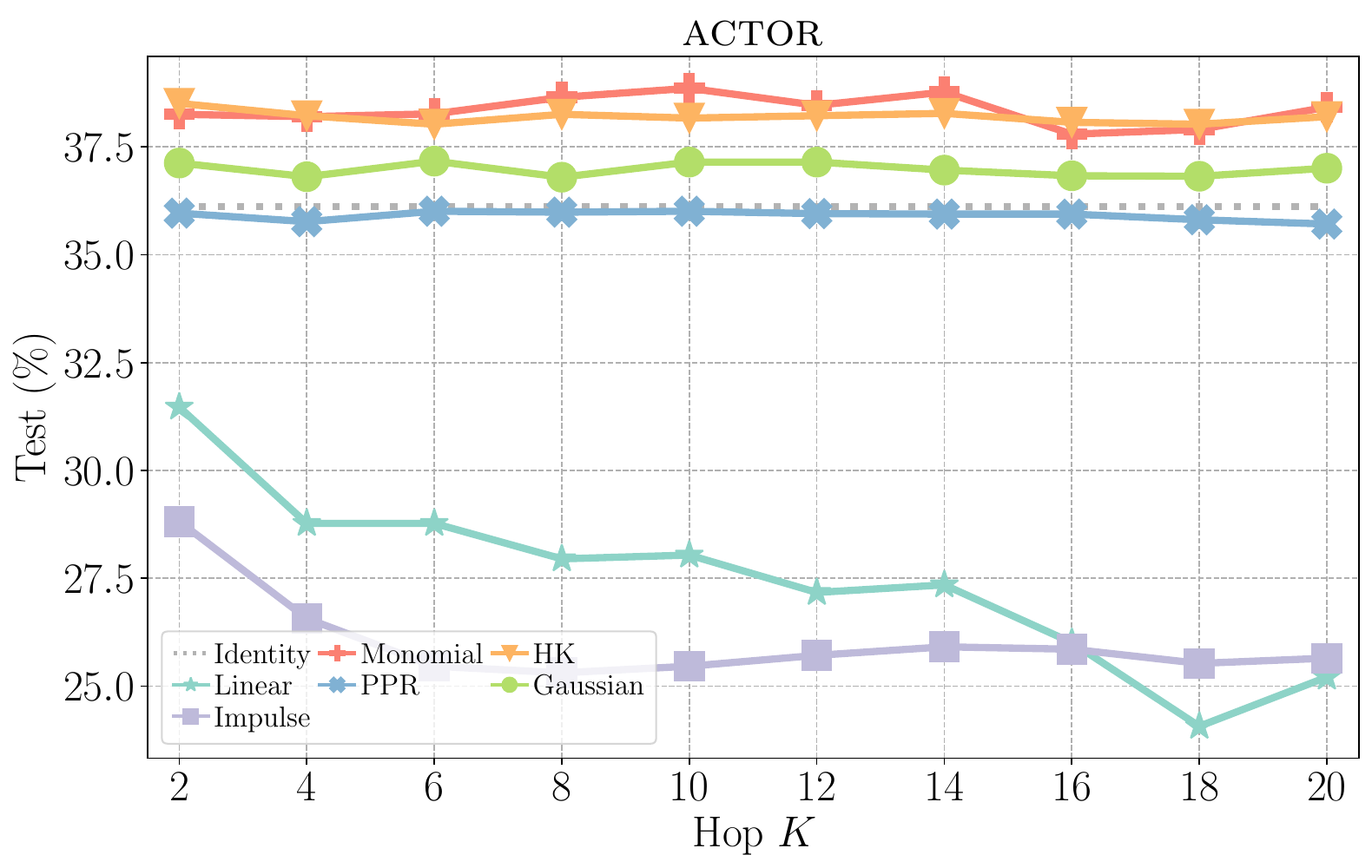}}
    \hfil
    \subcaptionbox{Variable filters on \ds{actor}\label{ffiga:hop_var_actor}}%
    [0.24\linewidth]{\includegraphics[height=1.05in]{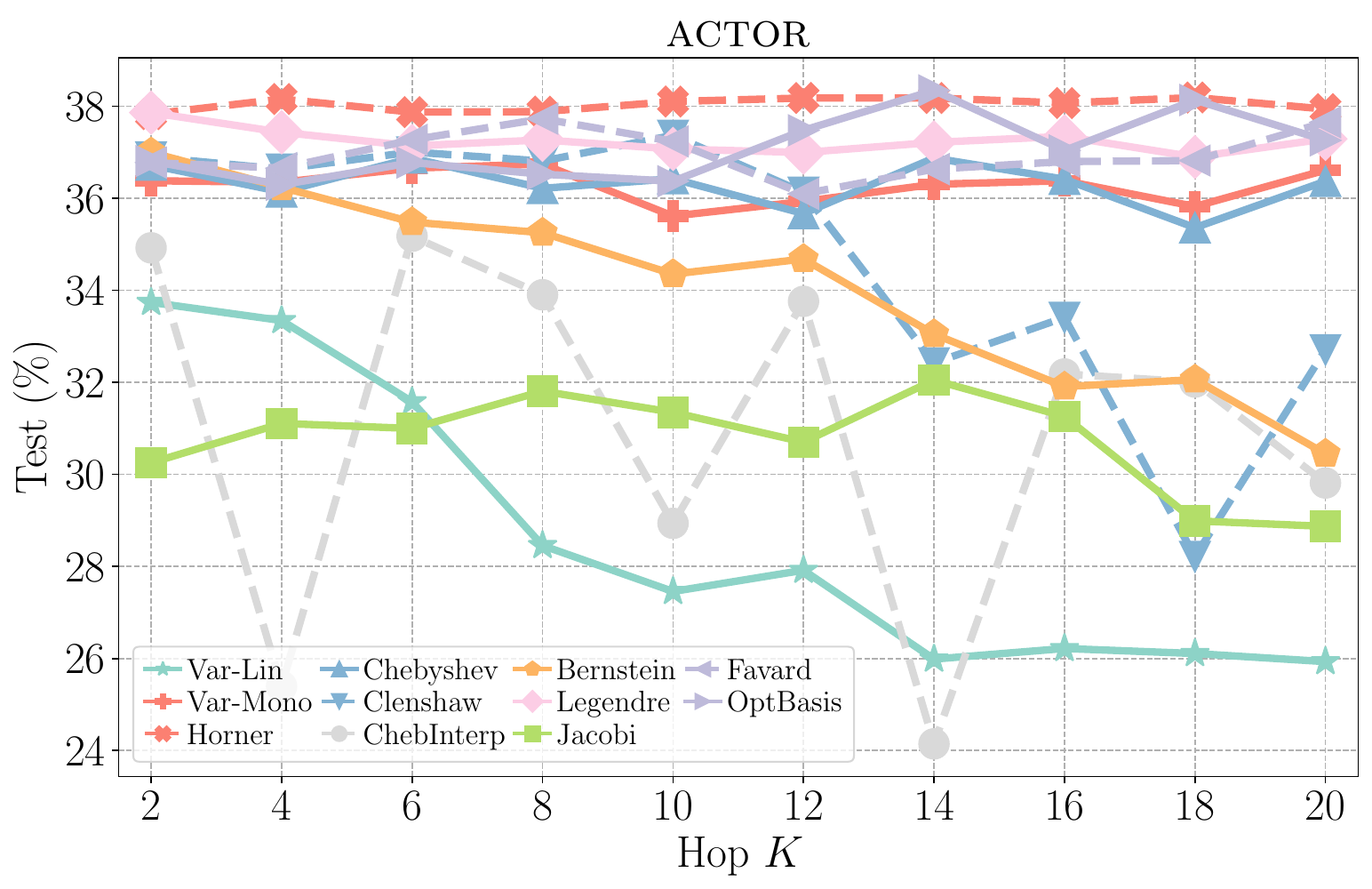}}
    \hfil 
    \subcaptionbox{Fixed filters on \ds{roman}\label{ffiga:hop_fix_roman_empire}}%
    [0.24\linewidth]{\includegraphics[height=1.05in]{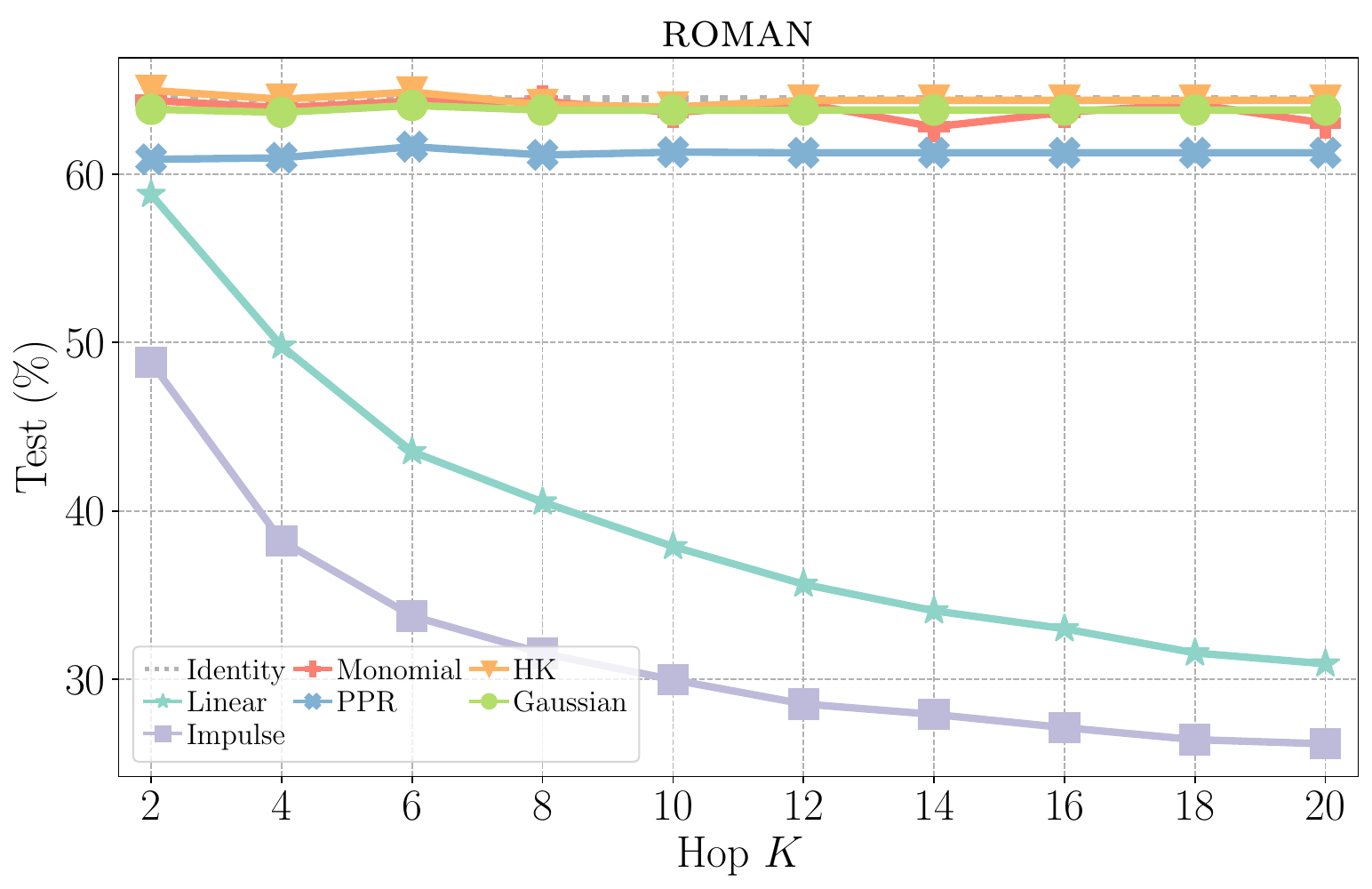}}
    \hfil
    \subcaptionbox{Variable filters on \ds{roman}\label{ffiga:hop_var_roman_empire}}%
    [0.24\linewidth]{\includegraphics[height=1.05in]{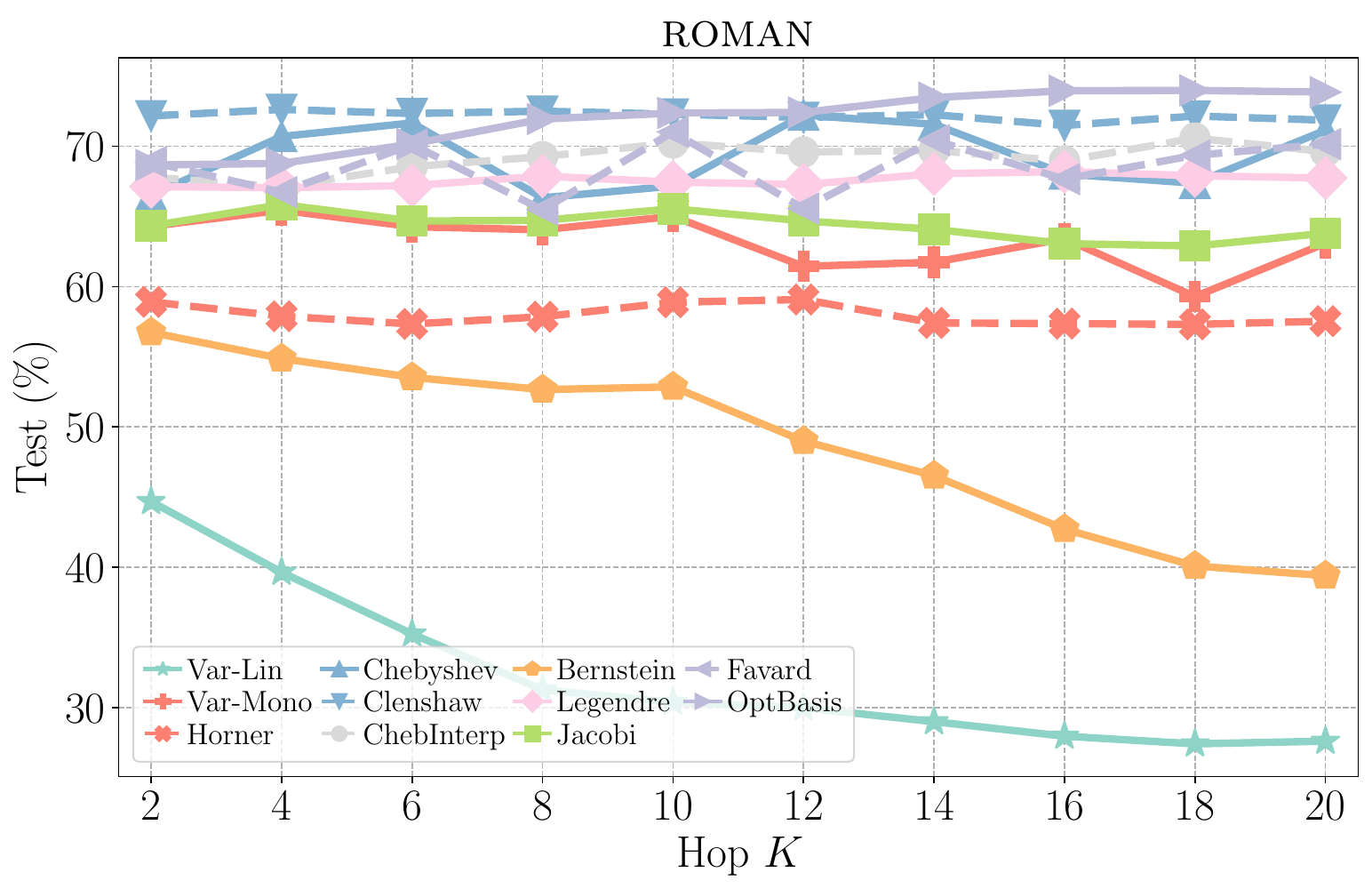}}
    \\
    \subcaptionbox{Fixed filters on \ds{ratings}\label{ffiga:hop_fix_amazon_ratings}}%
    [0.24\linewidth]{\includegraphics[height=1.05in]{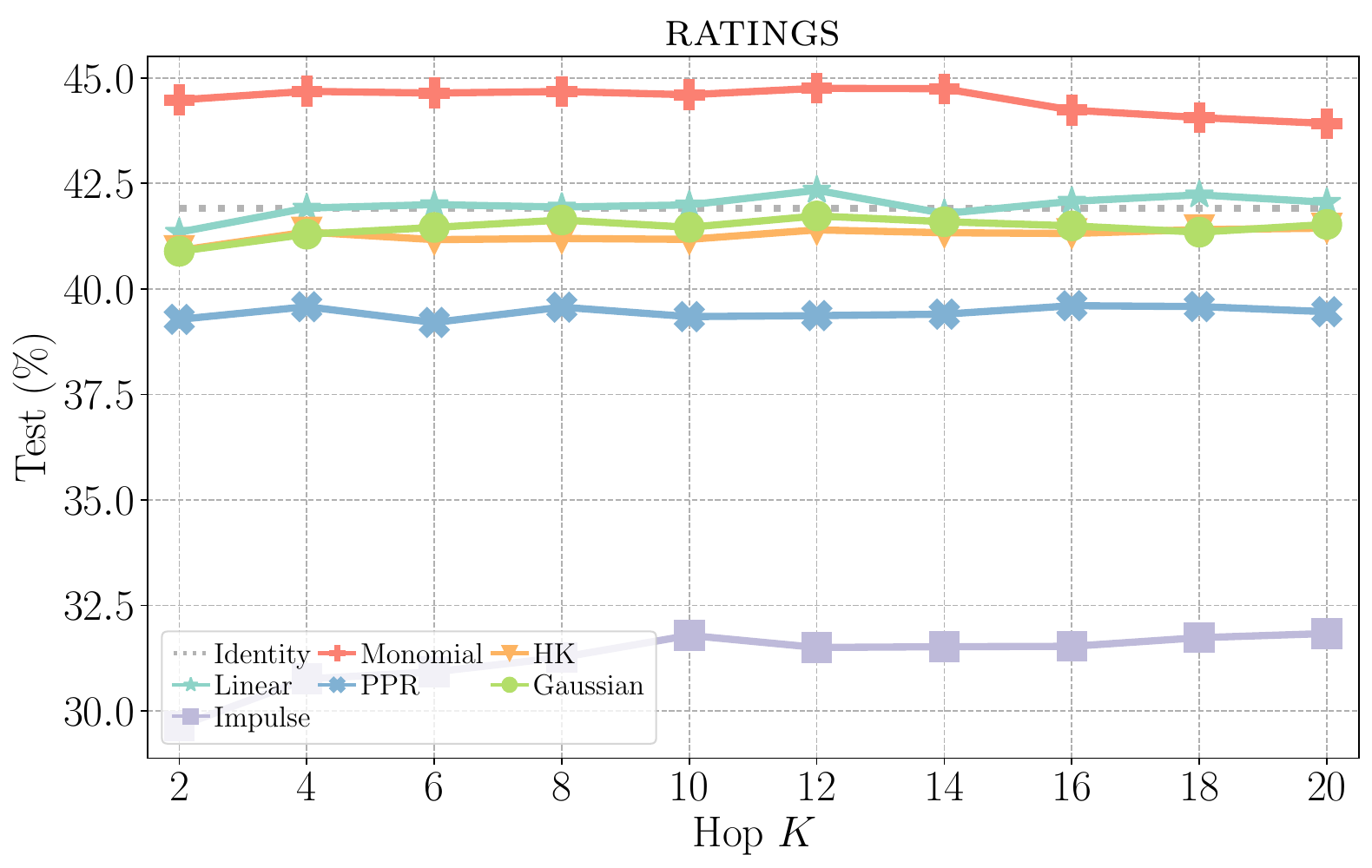}}
    \hfil
    \subcaptionbox{Variable filters on \ds{ratings}\label{ffiga:hop_var_amazon_ratings}}%
    [0.24\linewidth]{\includegraphics[height=1.05in]{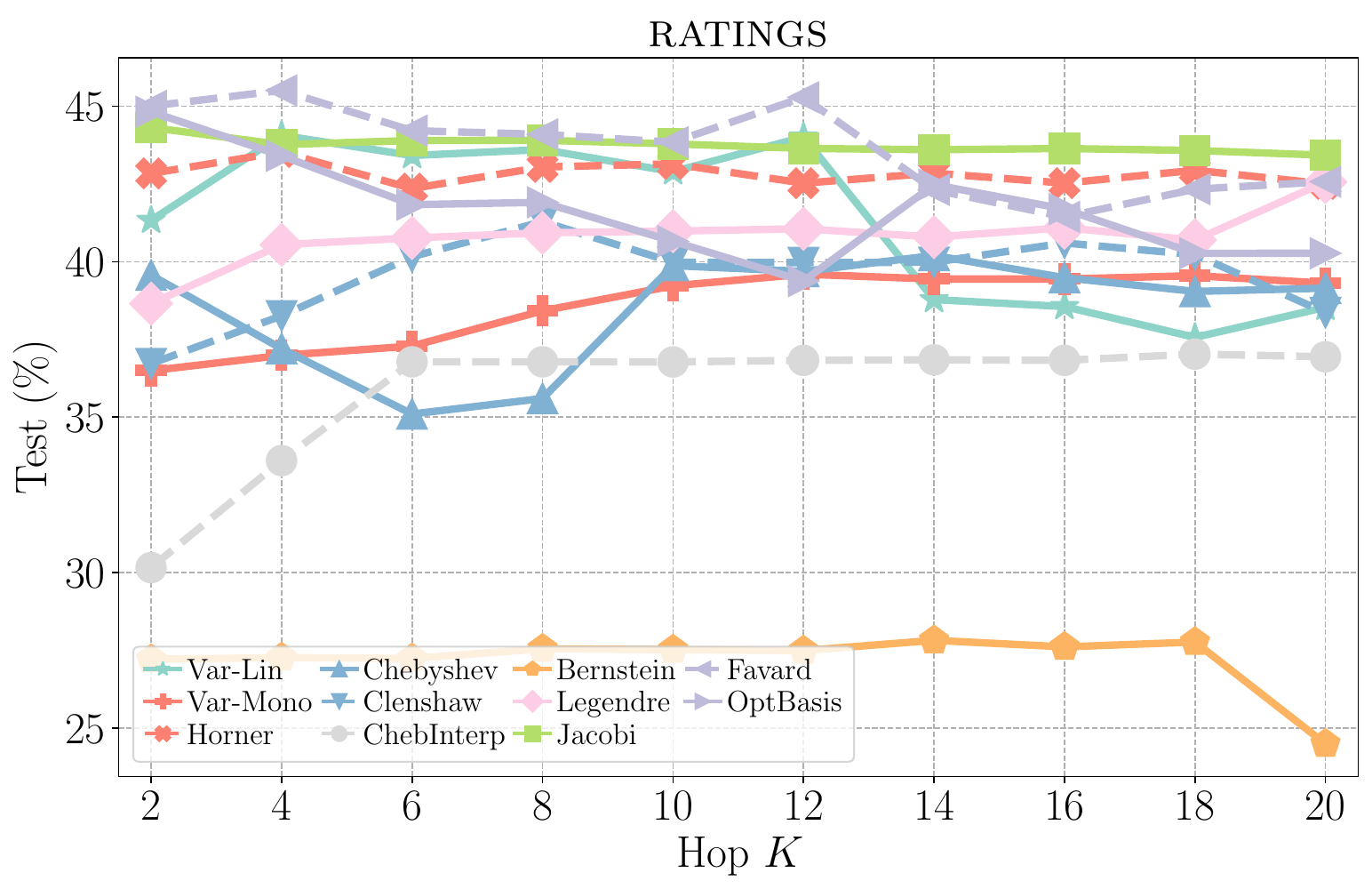}}
    \hfil
    \subcaptionbox{Fixed filters on \ds{flickr}\label{ffiga:hop_fix_flickr}}%
    [0.24\linewidth]{\includegraphics[height=1.05in]{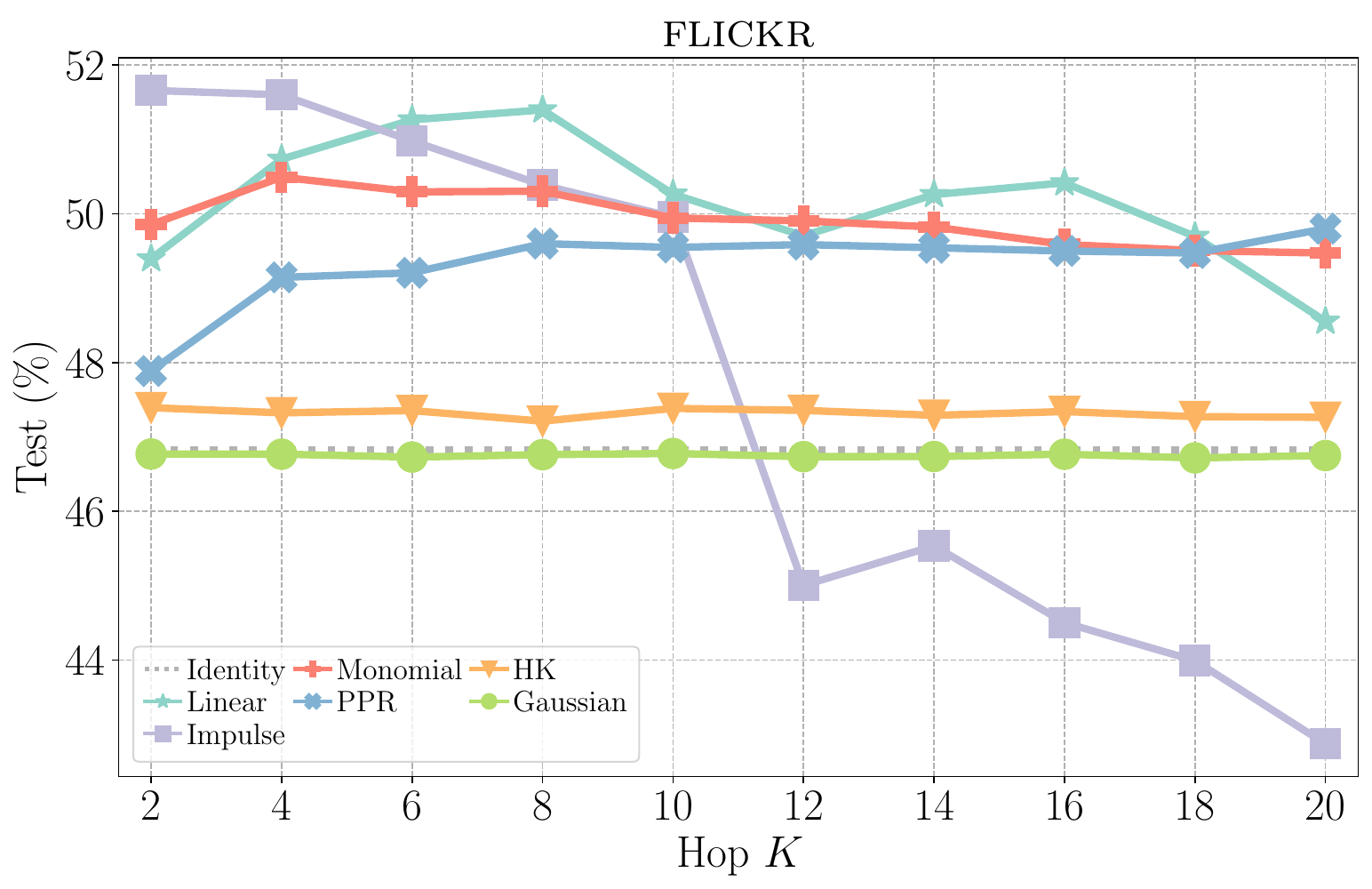}}
    \hfil
    \subcaptionbox{Variable filters on \ds{flickr}\label{ffiga:hop_var_flickr}}%
    [0.24\linewidth]{\includegraphics[height=1.05in]{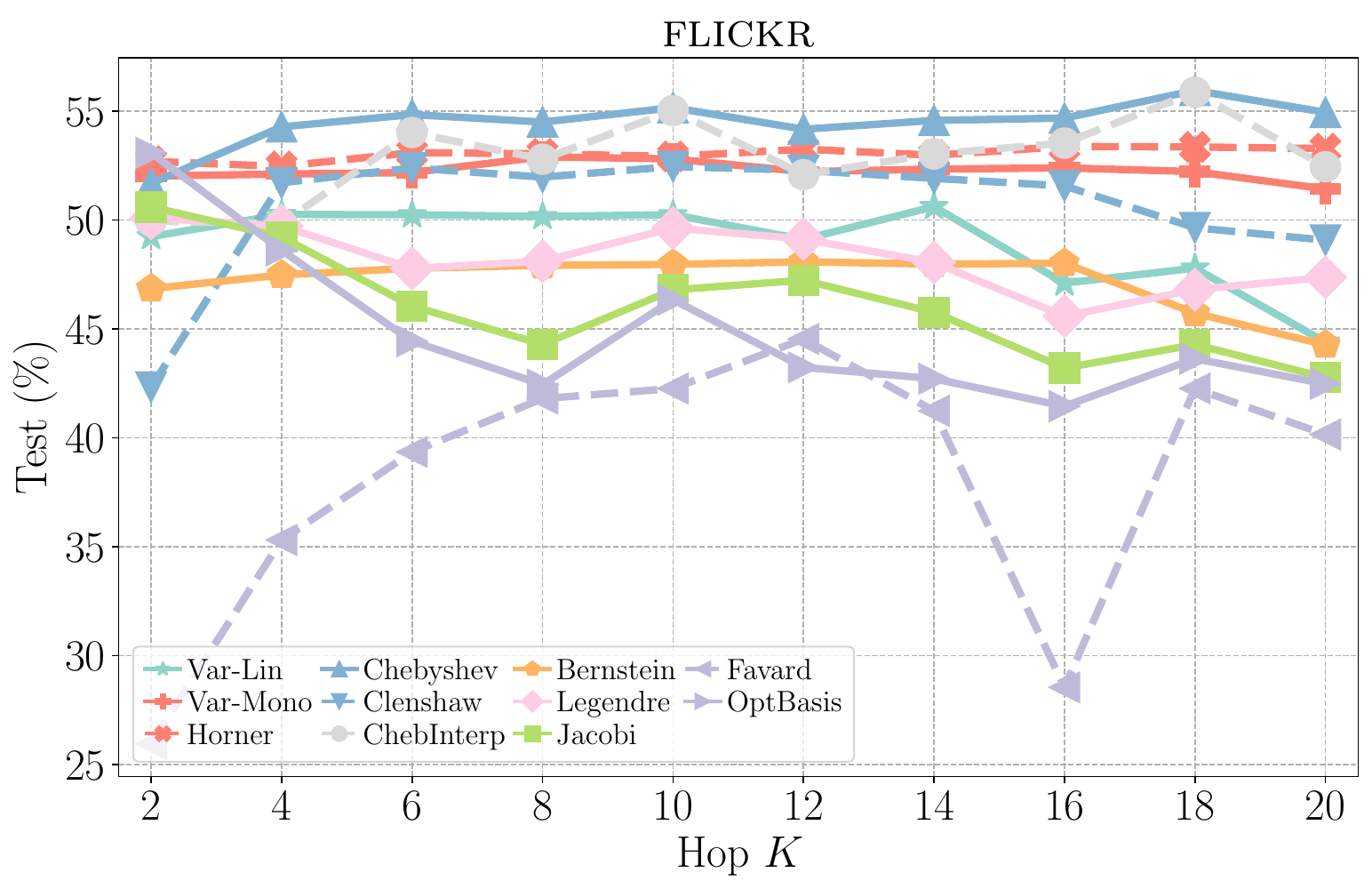}}    
    \caption{Effect of \textit{propagation hops} $K$ of full-batch fixed and variable filters on heterophilous datasets. }
  \label{figa:hop_hetero}
\end{figure*}

\clearpage



\begin{figure*}[!b]
\subsection{Clustering Visualization}
\label{sseca:extra_tsne}
\centering
\begin{subfigure}[b]{0.19\textwidth}
    \centering
    \includegraphics[width=1.1\textwidth]{figs/tsne/cora/DecoupledFixed_AdjConv-appr.pdf}
    \caption{PPR}
    \label{figa:cora_tsne_appr}
\end{subfigure}
\hfill
\begin{subfigure}[b]{0.19\textwidth}
    \centering
    \includegraphics[width=1.1\textwidth]{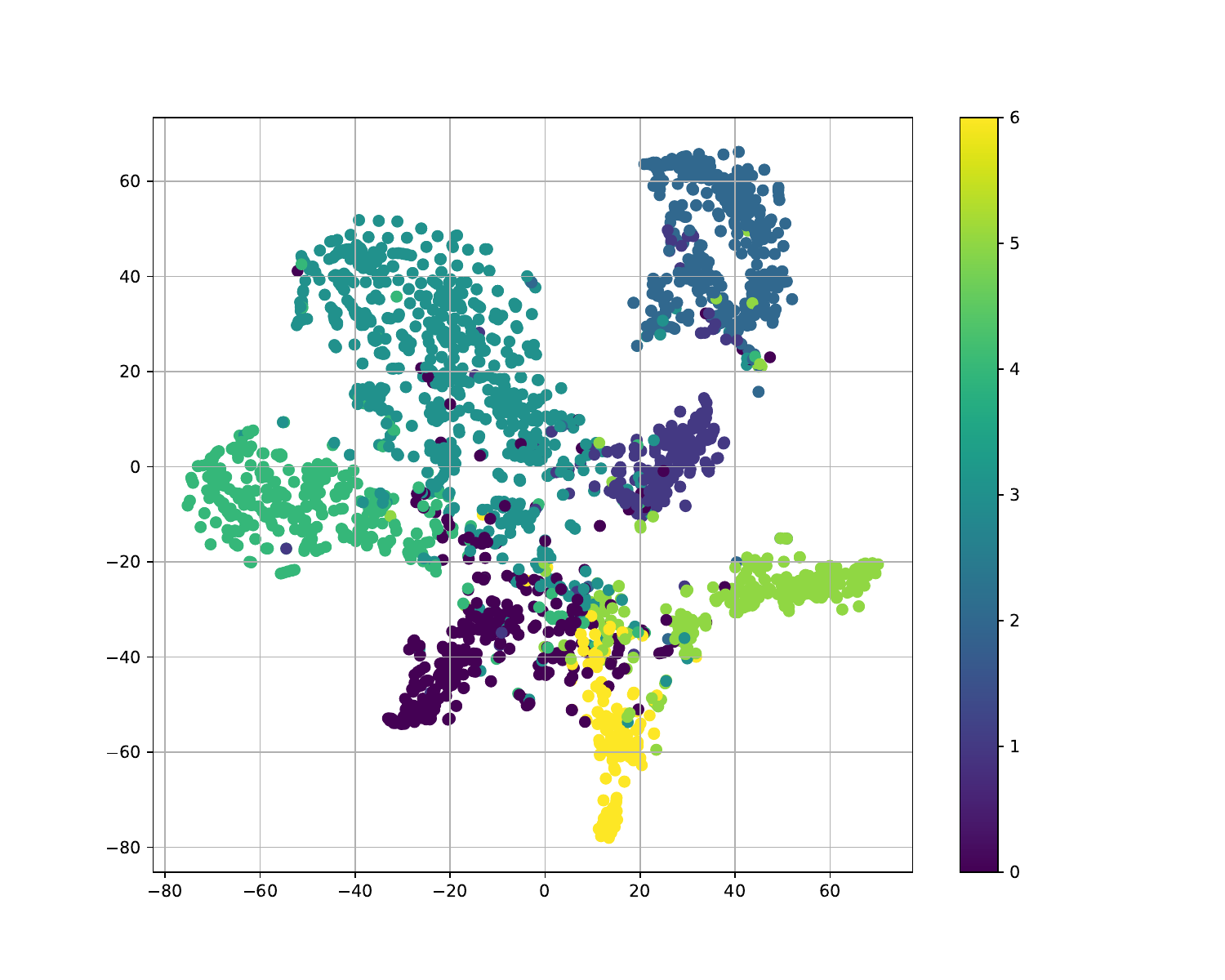}
    \caption{Gaussian}
    \label{figa:cora_tsne_gaussian}
\end{subfigure}
\hfill
\begin{subfigure}[b]{0.19\textwidth}
    \centering
    \includegraphics[width=1.1\textwidth]{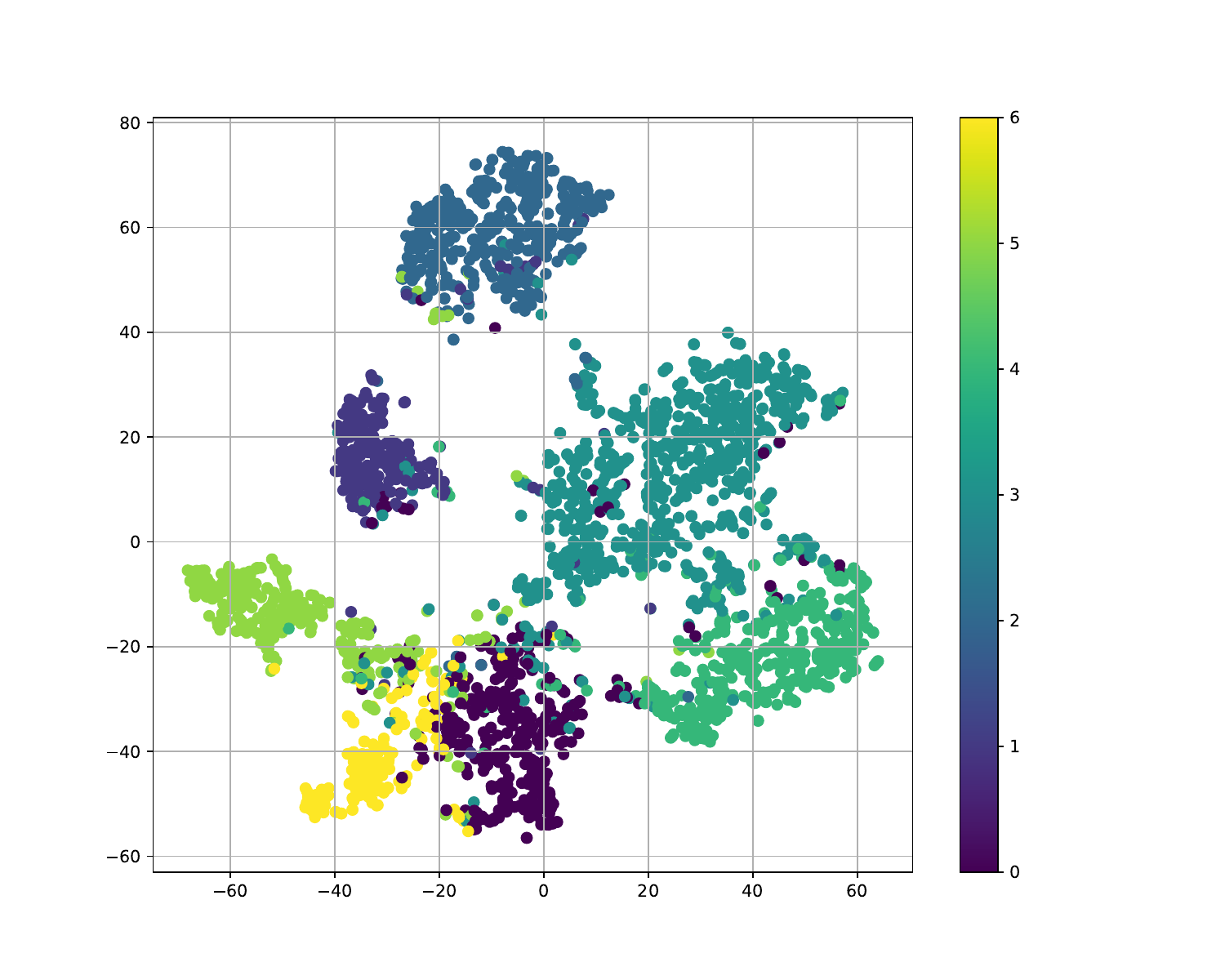}
    \caption{HK}
    \label{figa:cora_tsne_hk}
\end{subfigure}
\hfill
\begin{subfigure}[b]{0.19\textwidth}
    \centering
    \includegraphics[width=1.1\textwidth]{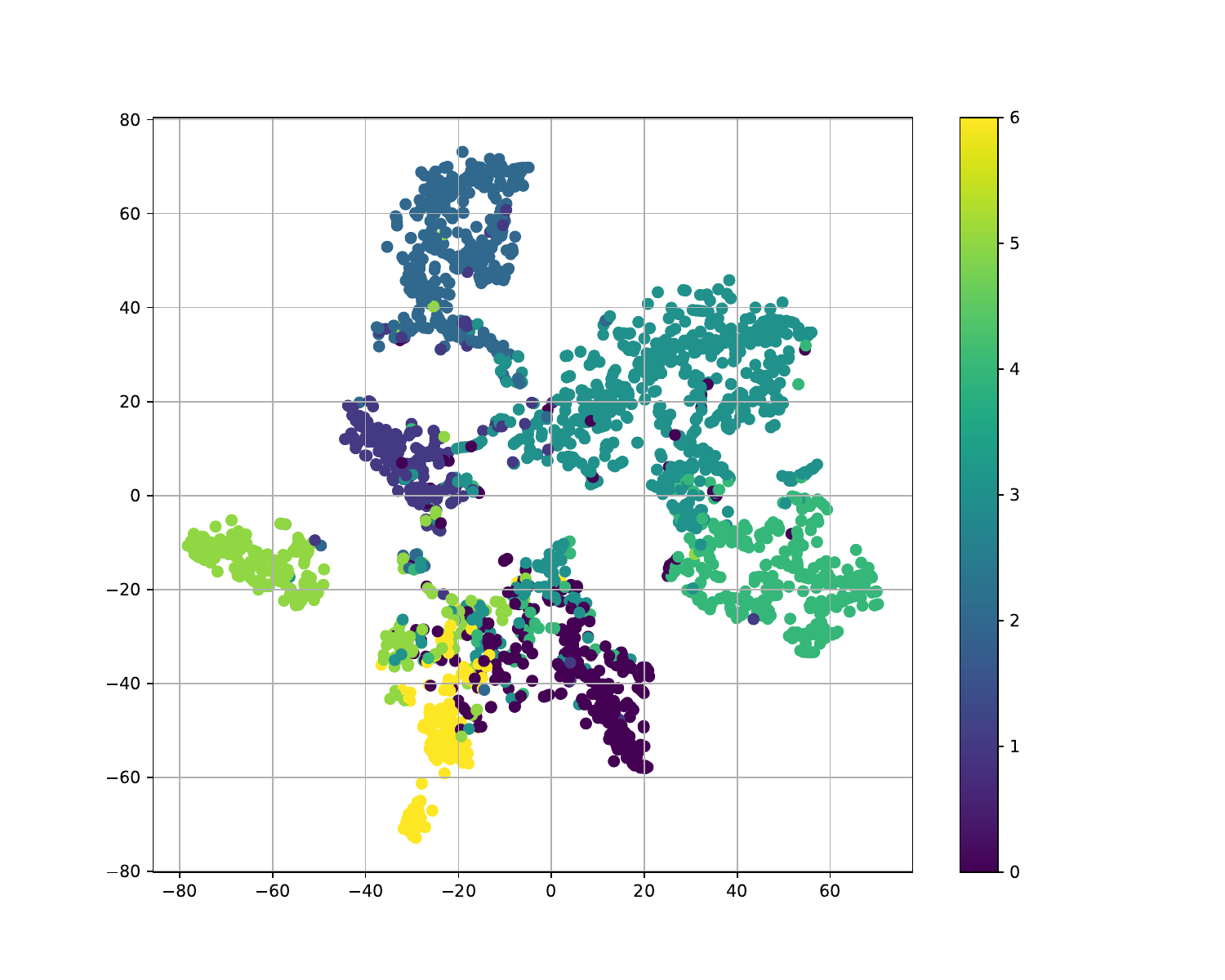}
    \caption{Impulse}
    \label{figa:cora_tsne_impulse}
\end{subfigure}
\hfill
\begin{subfigure}[b]{0.19\textwidth}
    \centering
    \includegraphics[width=1.1\textwidth]{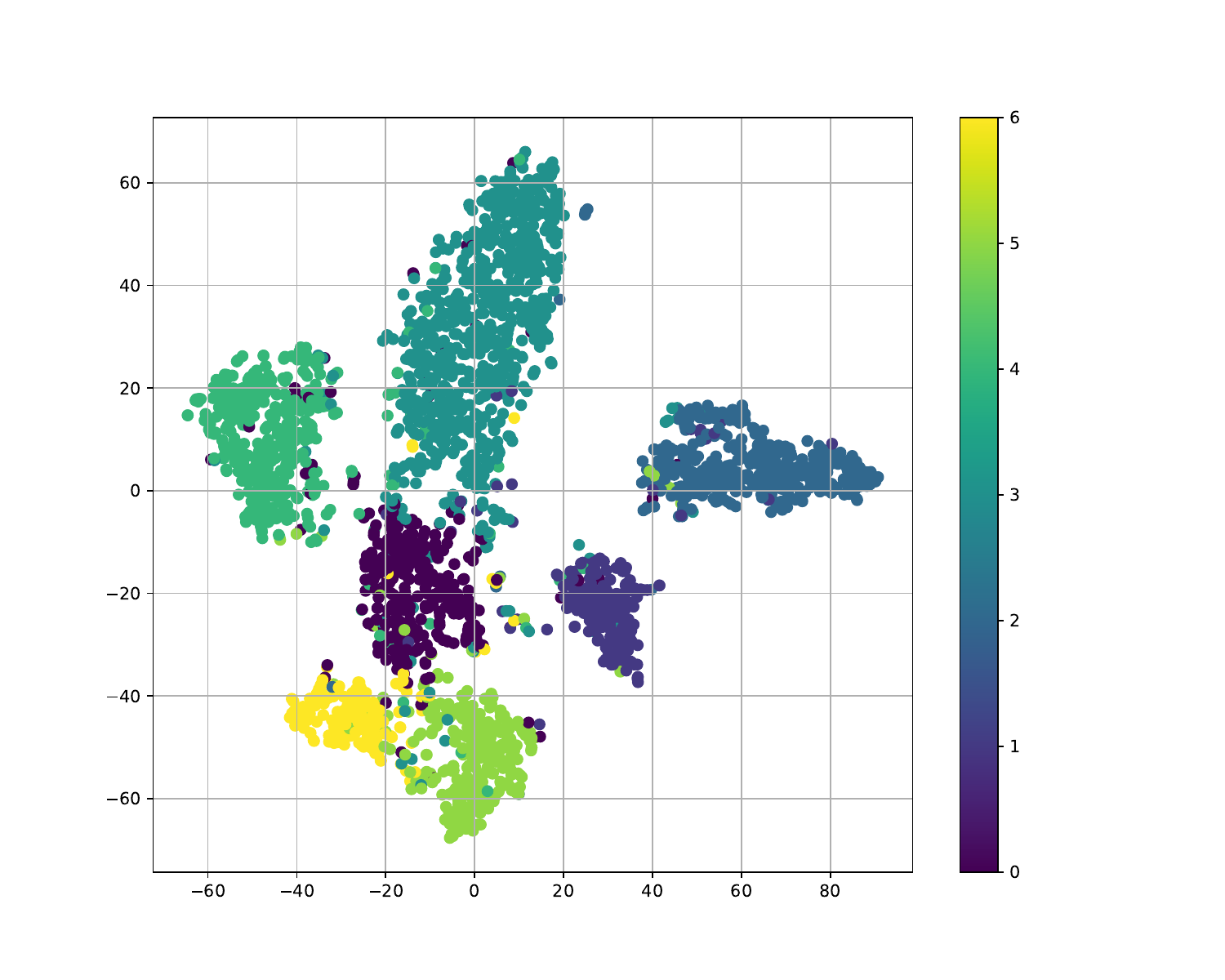}
    \caption{Mono}
    \label{figa:cora_tsne_mono}
\end{subfigure}
\begin{subfigure}[b]{0.19\textwidth}
    \centering
    \includegraphics[width=1.1\textwidth]{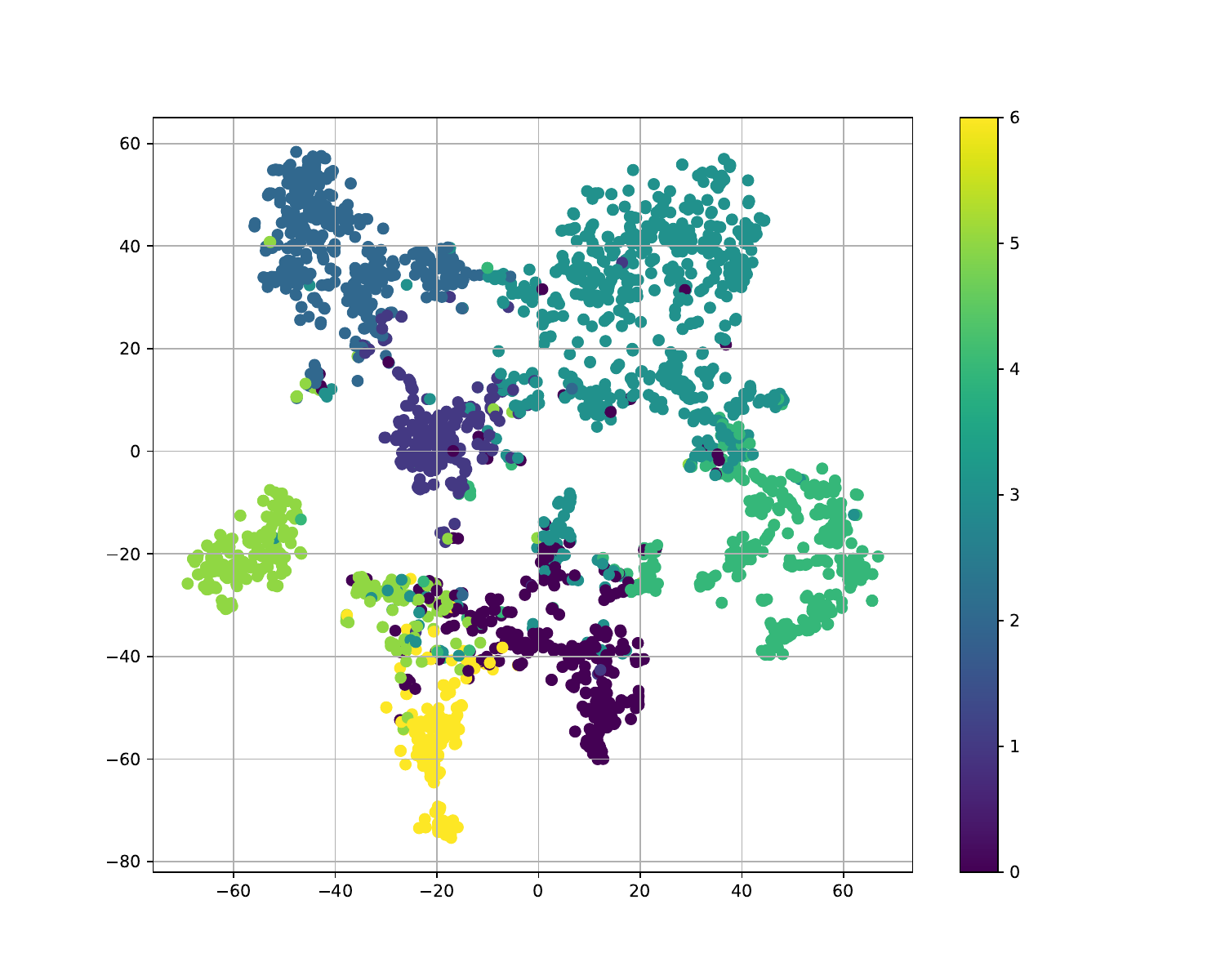}
    \caption{Var-Mono}
    \label{figa:cora_tsne_adjconv}
\end{subfigure}
\hfill
\begin{subfigure}[b]{0.19\textwidth}
    \centering
    \includegraphics[width=1.1\textwidth]{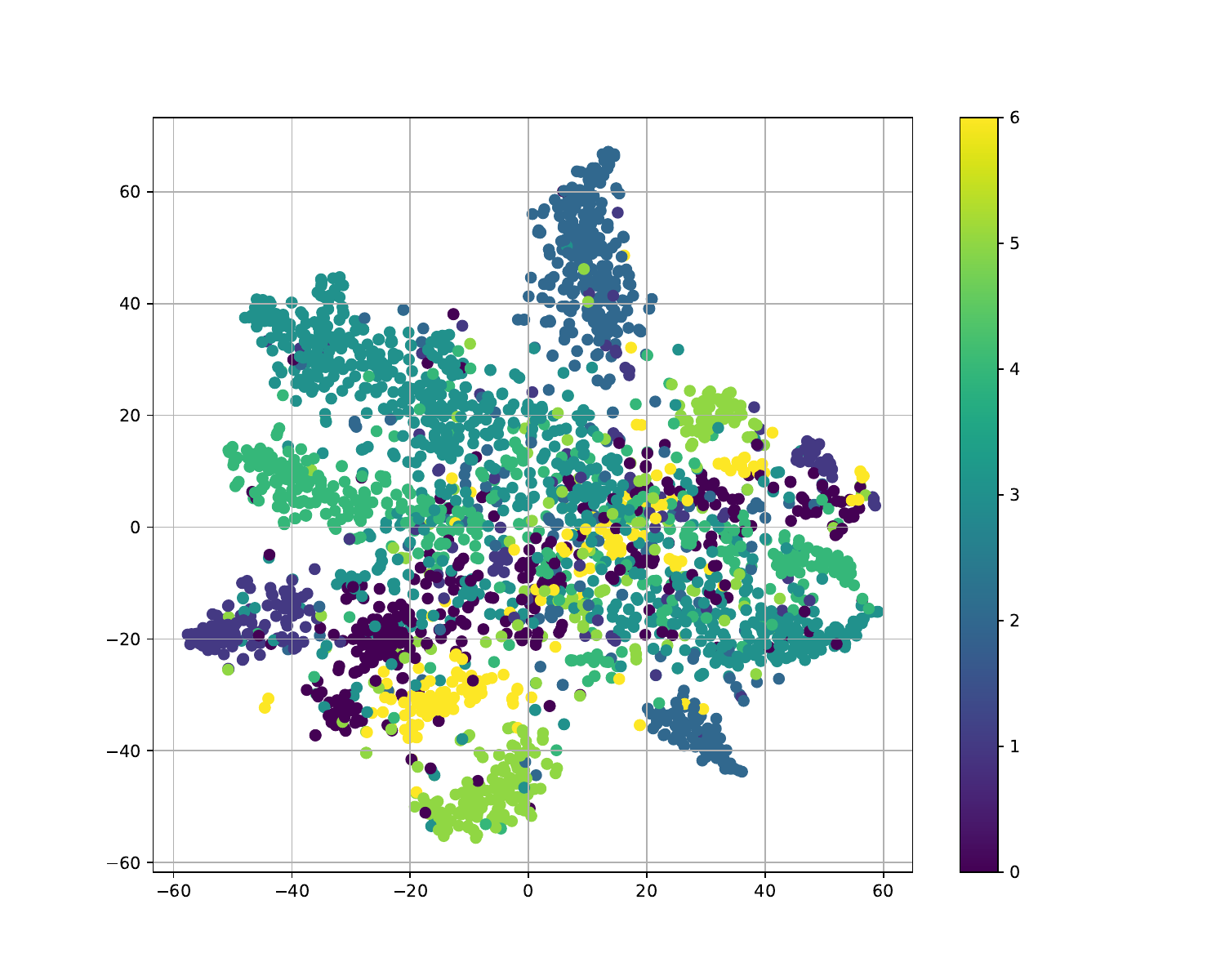}
    \caption{Bern}
    \label{figa:cora_tsne_bernconv}
\end{subfigure}
\hfill
\begin{subfigure}[b]{0.19\textwidth}
    \centering
    \includegraphics[width=1.1\textwidth]{figs/tsne/cora/DecoupledVar_ChebConv.pdf}
    \caption{Cheb}
    \label{figa:cora_tsne_chebconv}
\end{subfigure}
\hfill
\begin{subfigure}[b]{0.19\textwidth}
    \centering
    \includegraphics[width=1.1\textwidth]{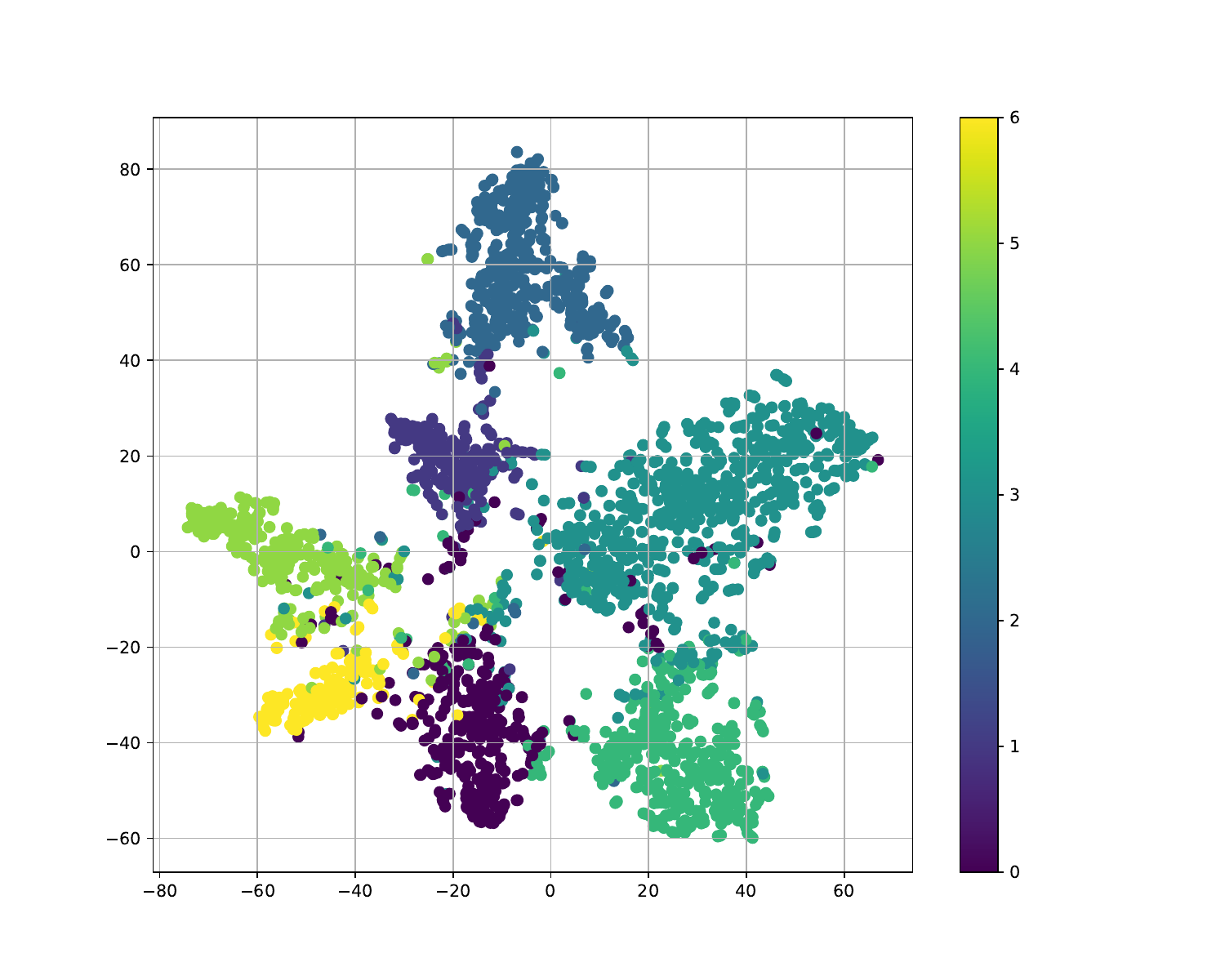}
    \caption{ChebInterp}
    \label{figa:cora_tsne_chebiiconv}
\end{subfigure}
\hfill
\begin{subfigure}[b]{0.19\textwidth}
    \centering
    \includegraphics[width=1.1\textwidth]{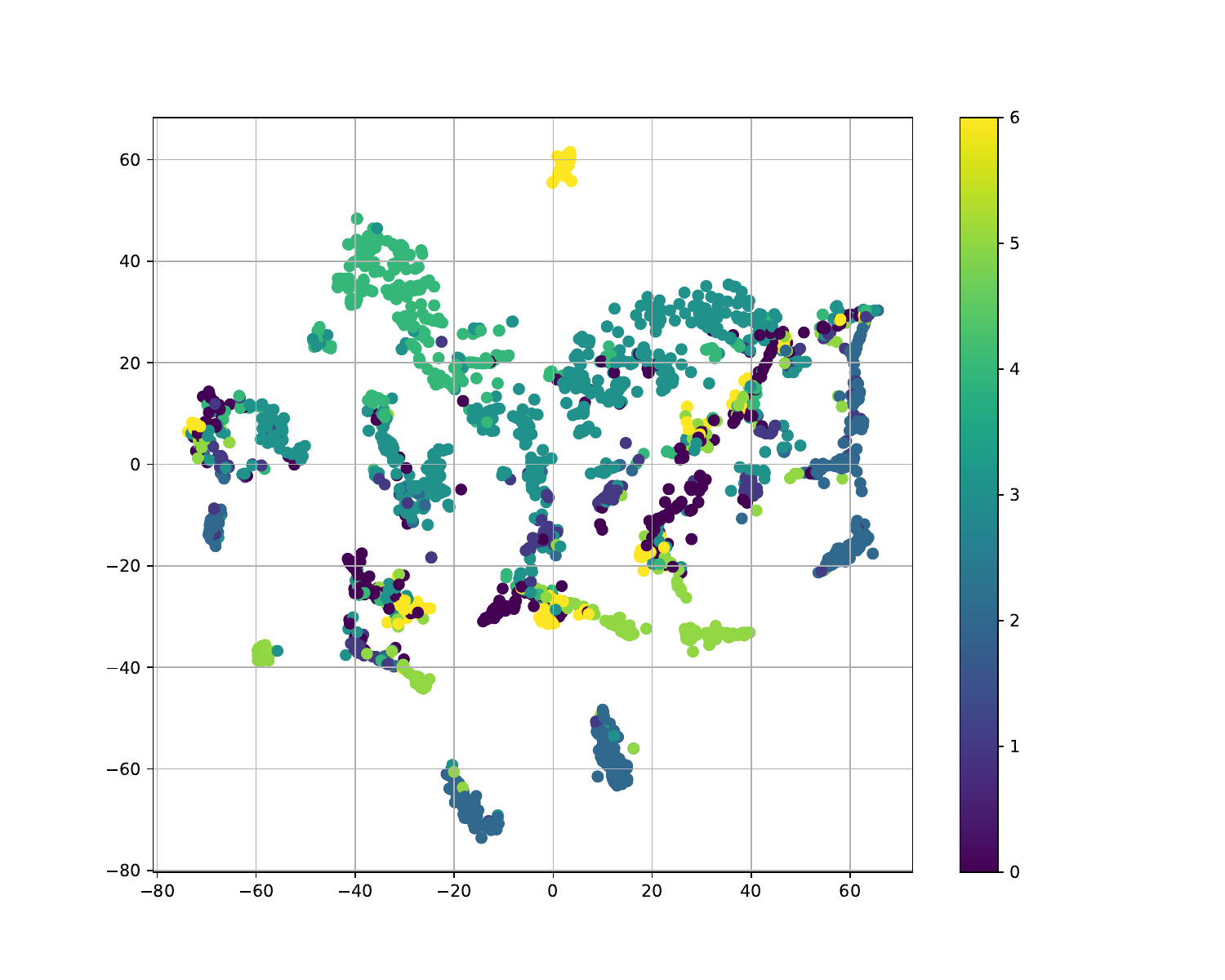}
    \caption{Clenshaw}
    \label{figa:cora_tsne_clenshawconv}
\end{subfigure}
\begin{subfigure}[b]{0.19\textwidth}
    \centering
    \includegraphics[width=1.1\textwidth]{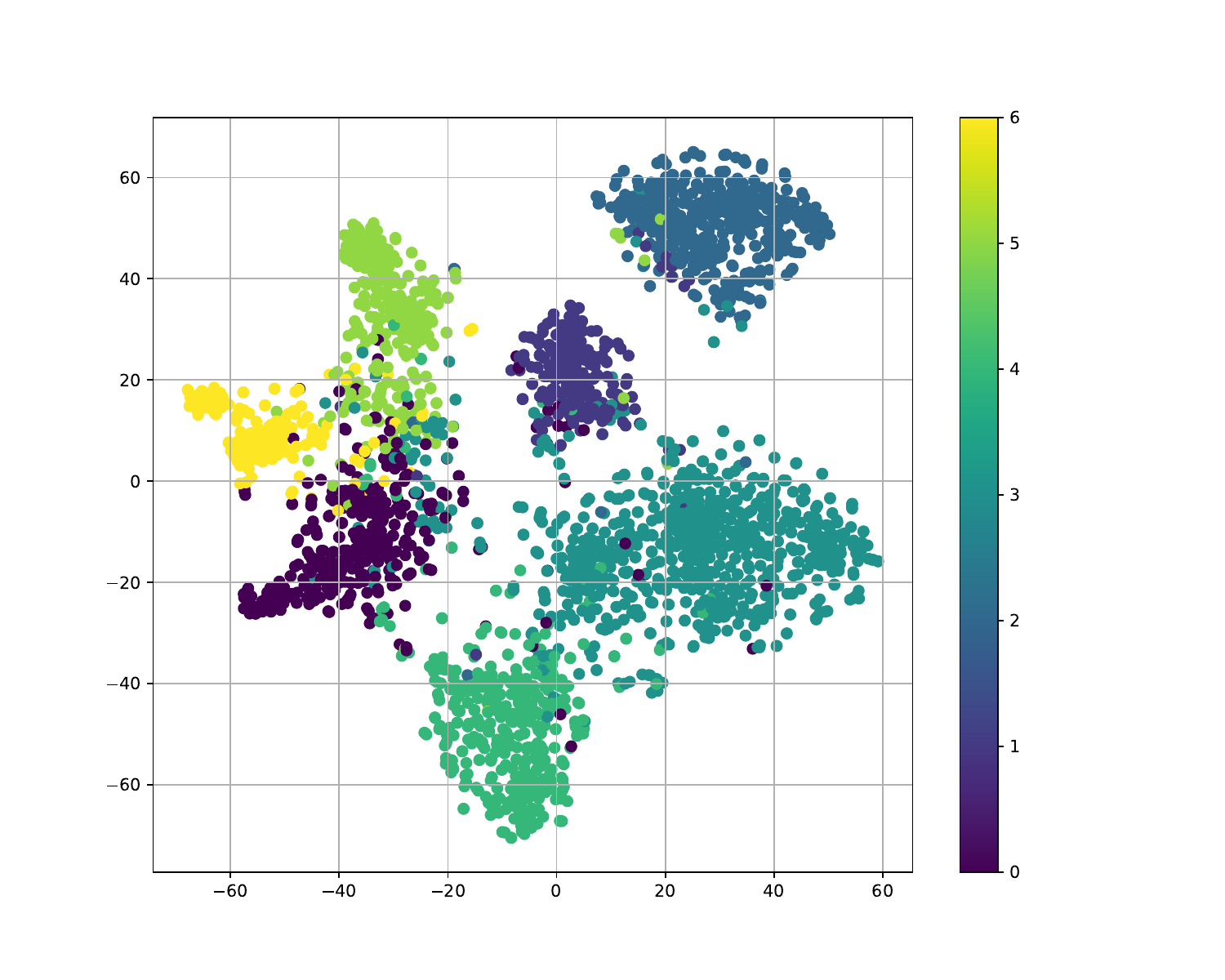}
    \caption{Favard}
    \label{figa:cora_tsne_favardconv}
\end{subfigure}
\hfill
\begin{subfigure}[b]{0.19\textwidth}
    \centering
    \includegraphics[width=1.1\textwidth]{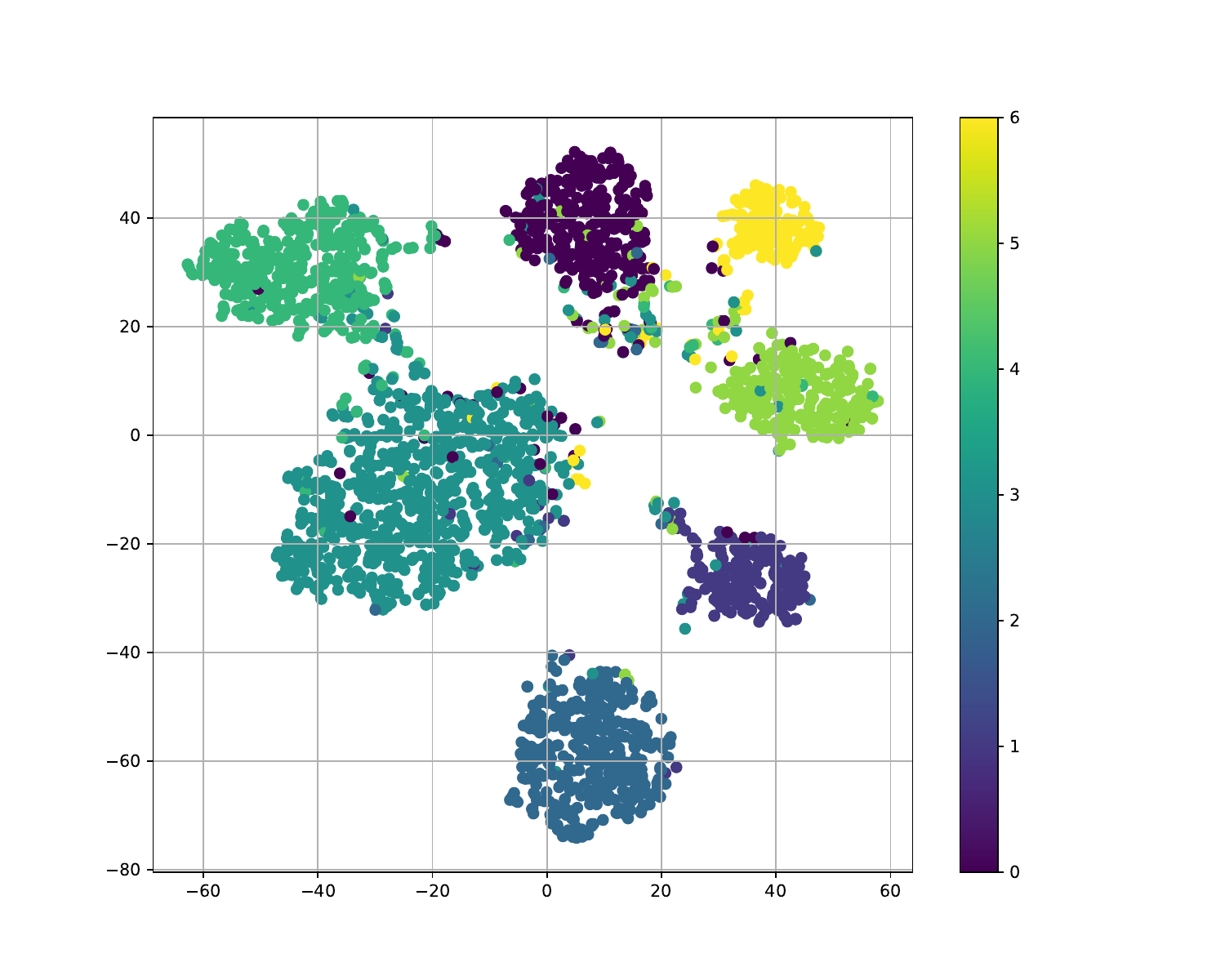}
    \caption{Horner}
    \label{figa:cora_tsne_hornerconv}
\end{subfigure}
\hfill
\begin{subfigure}[b]{0.19\textwidth}
    \centering
    \includegraphics[width=1.1\textwidth]{figs/tsne/cora/DecoupledVar_JacobiConv.pdf}
    \caption{Jacobi}
    \label{figa:cora_tsne_jacobiconv}
\end{subfigure}
\hfill
\begin{subfigure}[b]{0.19\textwidth}
    \centering
    \includegraphics[width=1.1\textwidth]{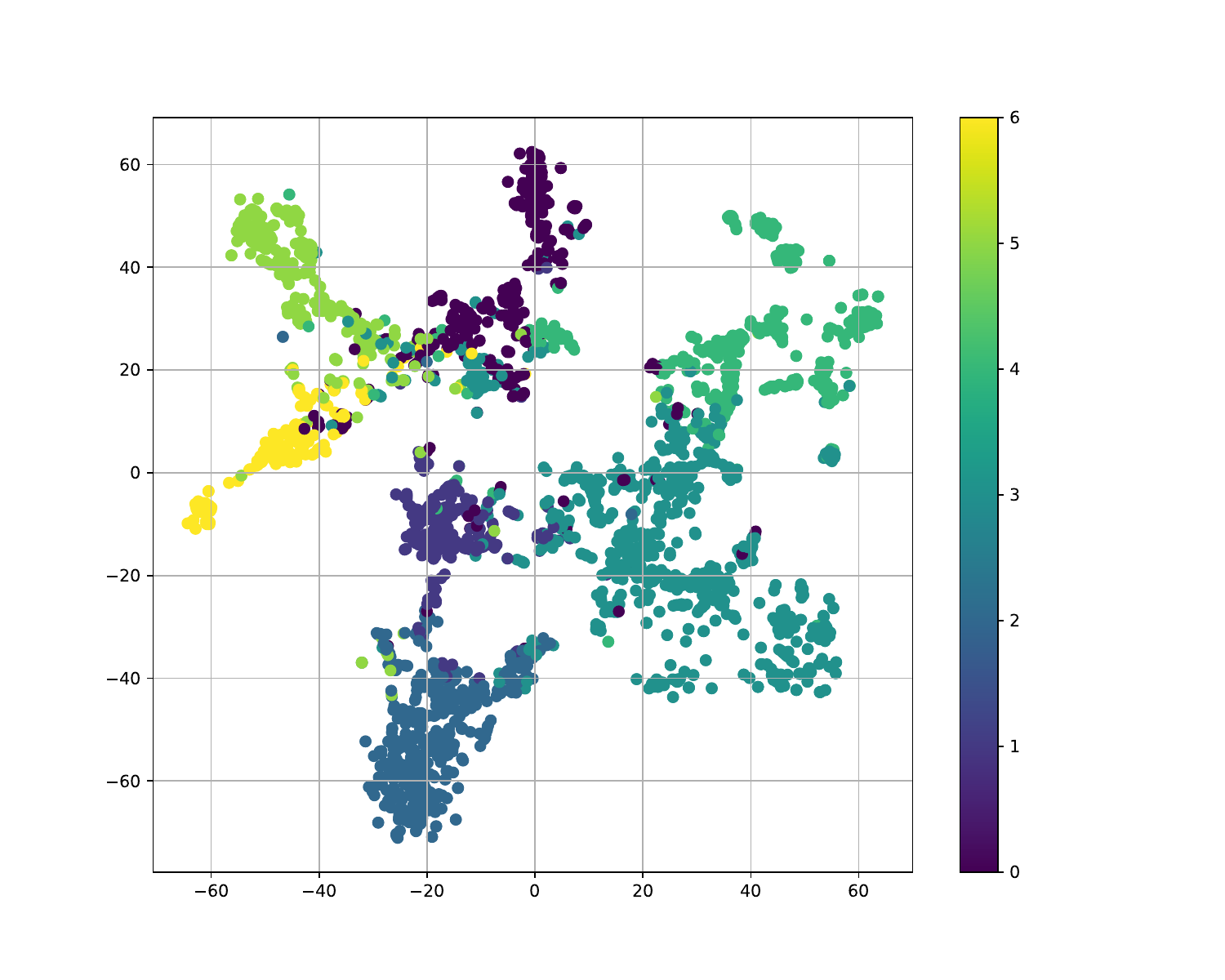}
    \caption{Legendre}
    \label{figa:cora_tsne_legendreconv}
\end{subfigure}
\hfill
\begin{subfigure}[b]{0.19\textwidth}
    \centering
    \includegraphics[width=1.1\textwidth]{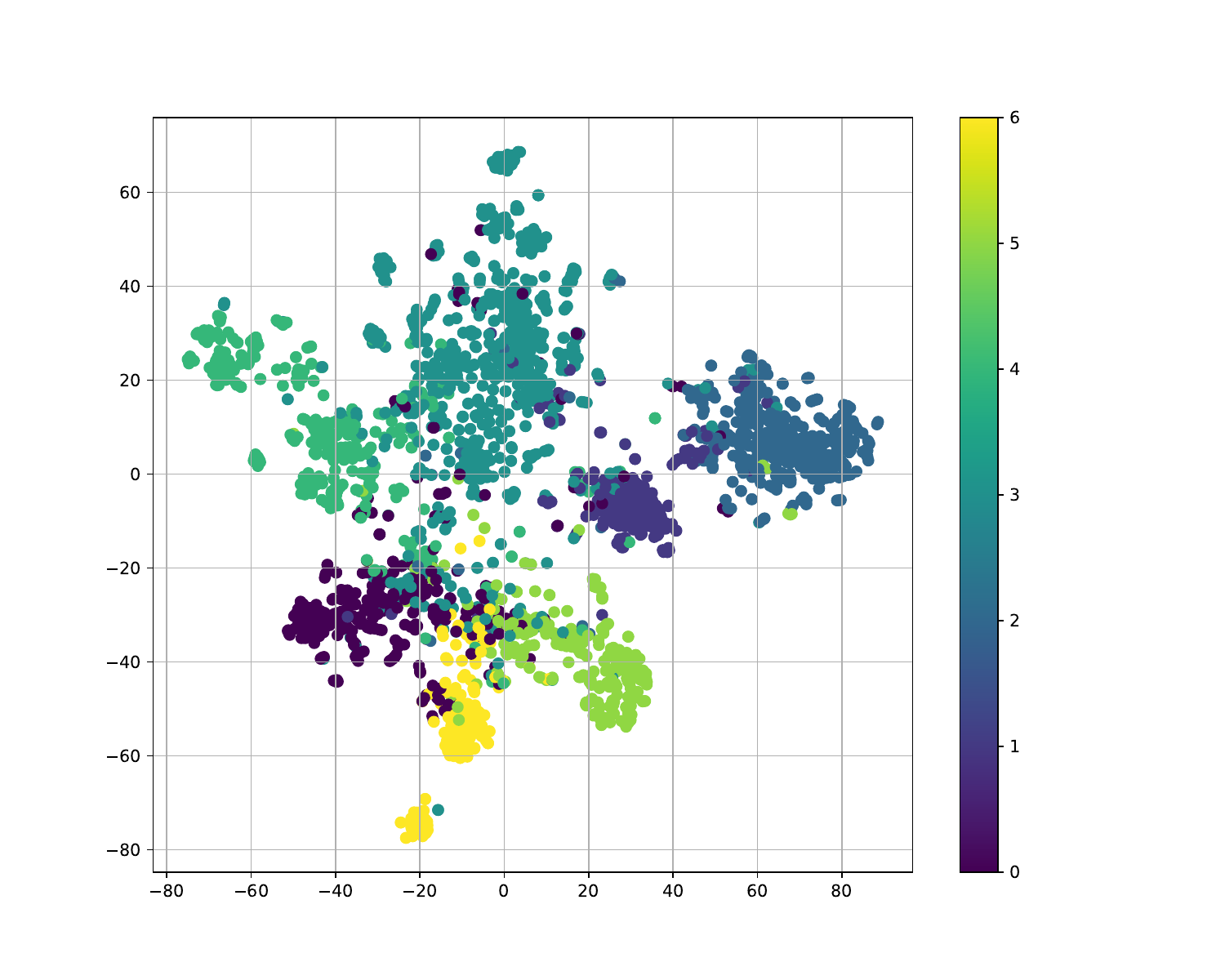}
    \caption{OptBasis}
    \label{figa:cora_tsne_optbasisconv}
\end{subfigure}

\caption{Clusters of different filters on \ds{CORA}.}
\label{fig:tsne_cora}

\centering
\begin{subfigure}[b]{0.19\textwidth}
    \centering
    \includegraphics[width=1.1\textwidth]{figs/tsne/chameleon/DecoupledFixed_AdjConv-appr.pdf}
    \caption{PPR}
    \label{figa:chameleon_tsne_appr}
\end{subfigure}
\hfill
\begin{subfigure}[b]{0.19\textwidth}
    \centering
    \includegraphics[width=1.1\textwidth]{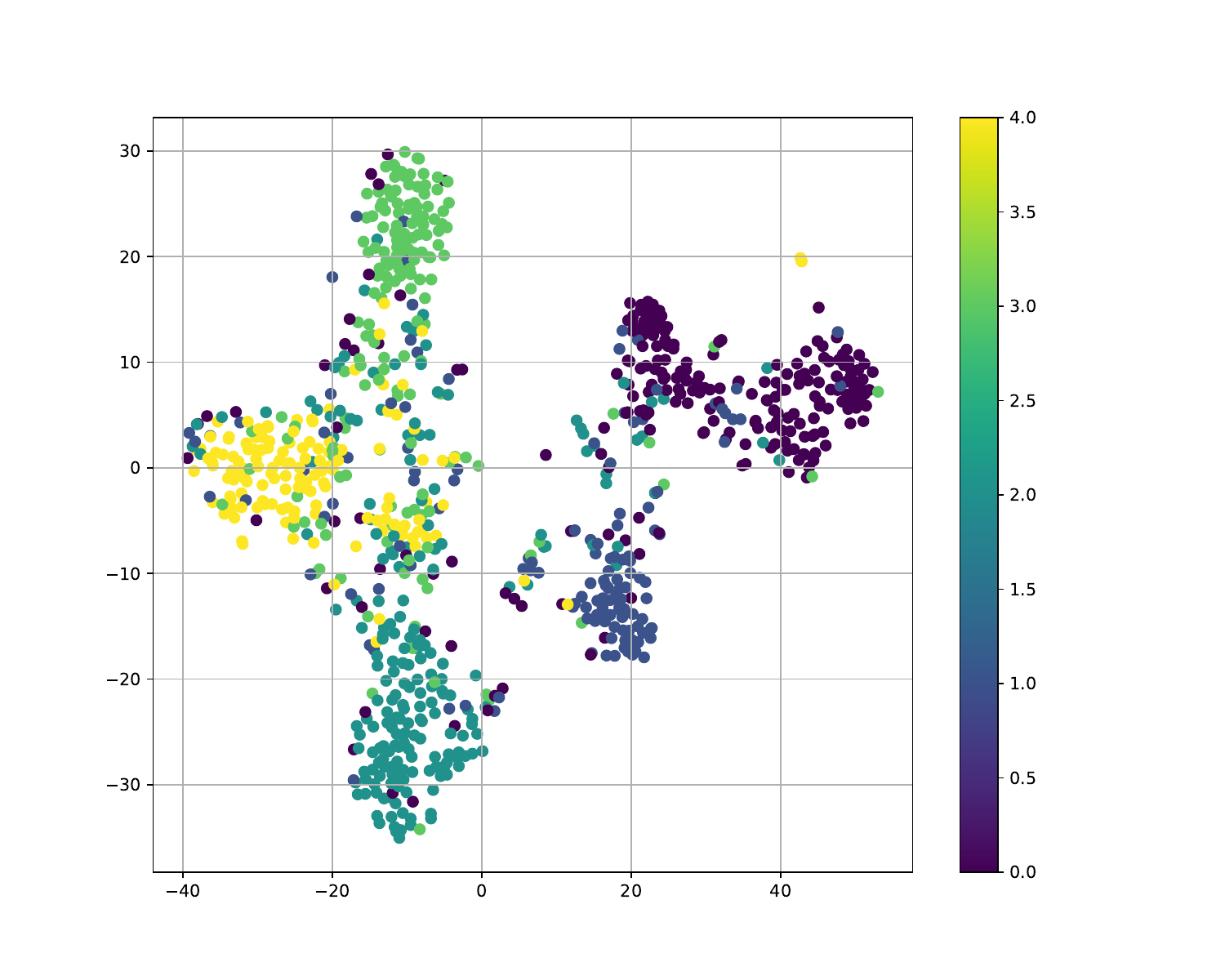}
    \caption{Gaussian}
    \label{figa:chameleon_tsne_gaussian}
\end{subfigure}
\hfill
\begin{subfigure}[b]{0.19\textwidth}
    \centering
    \includegraphics[width=1.1\textwidth]{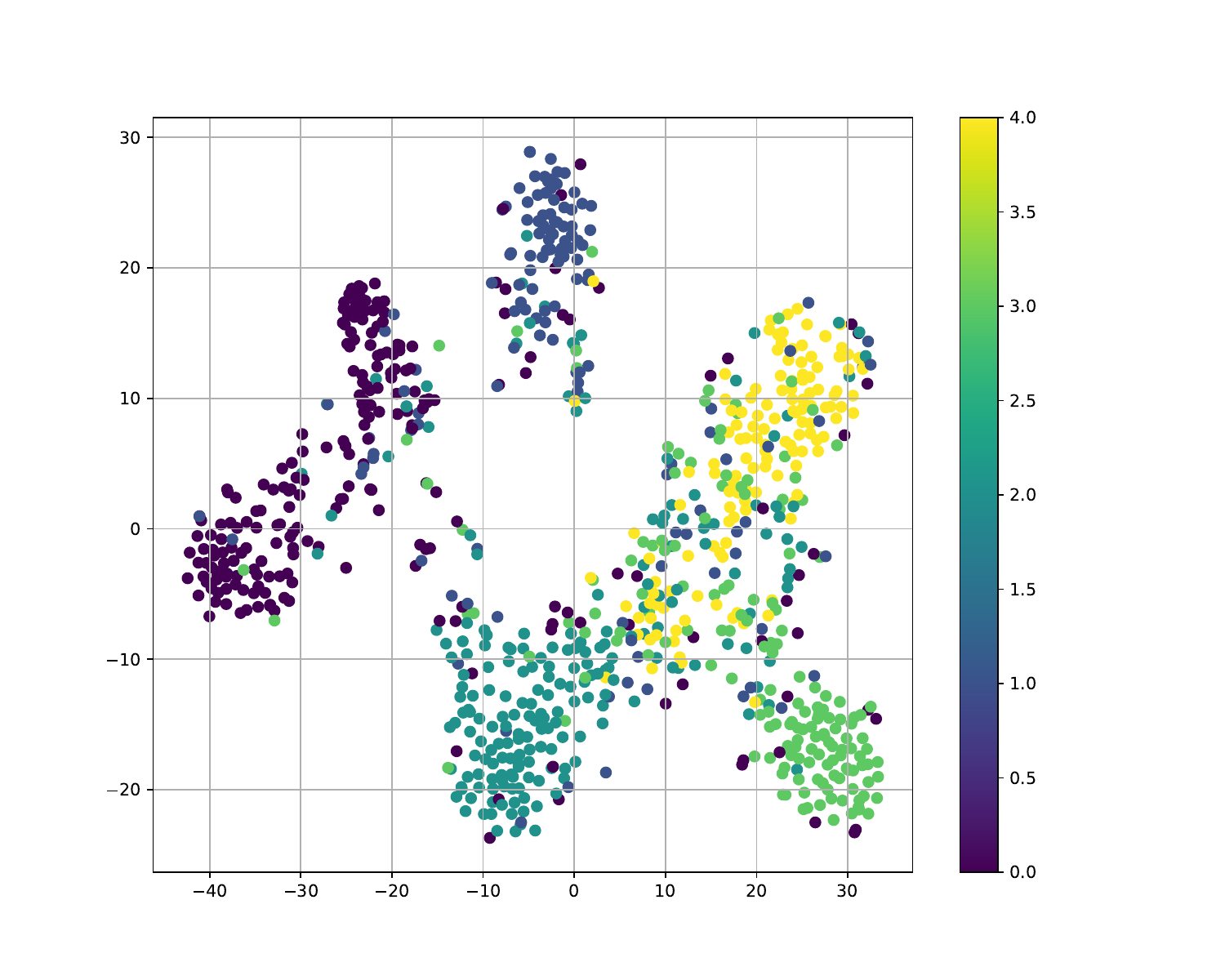}
    \caption{HK}
    \label{figa:chameleon_tsne_hk}
\end{subfigure}
\hfill
\begin{subfigure}[b]{0.19\textwidth}
    \centering
    \includegraphics[width=1.1\textwidth]{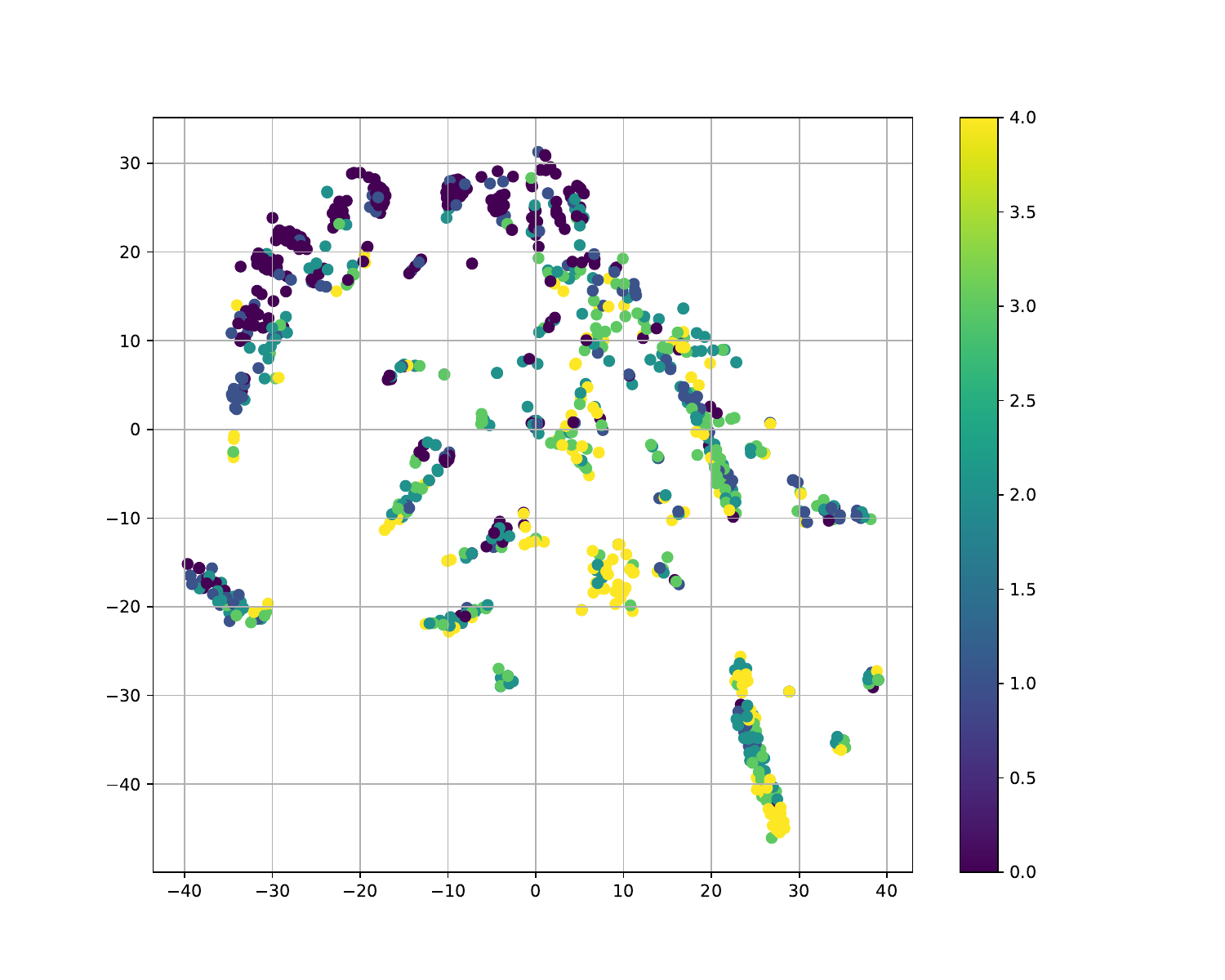}
    \caption{Impulse}
    \label{figa:chameleon_tsne_impulse}
\end{subfigure}
\hfill
\begin{subfigure}[b]{0.19\textwidth}
    \centering
    \includegraphics[width=1.1\textwidth]{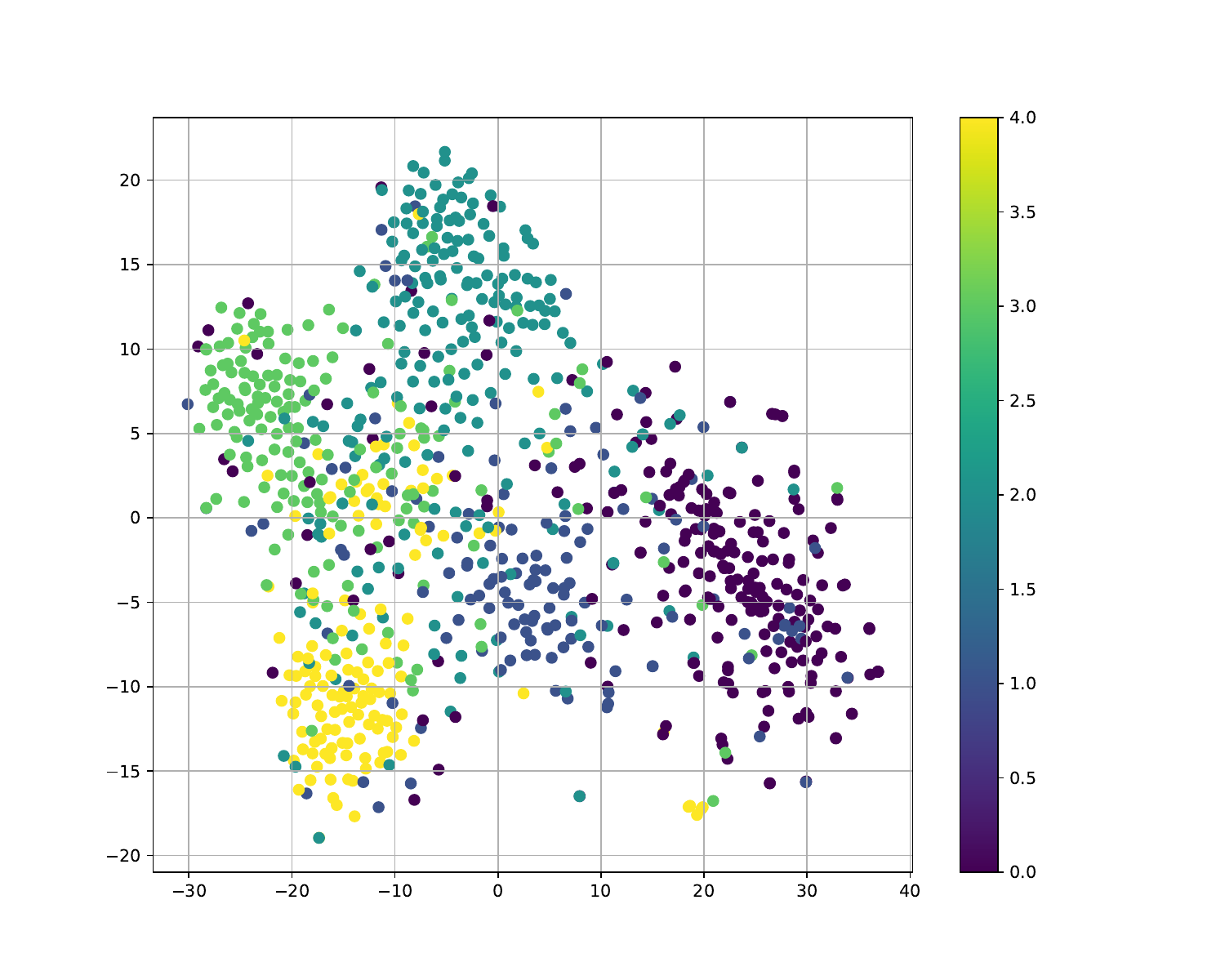}
    \caption{Mono}
    \label{figa:chameleon_tsne_mono}
\end{subfigure}
\begin{subfigure}[b]{0.19\textwidth}
    \centering
    \includegraphics[width=1.1\textwidth]{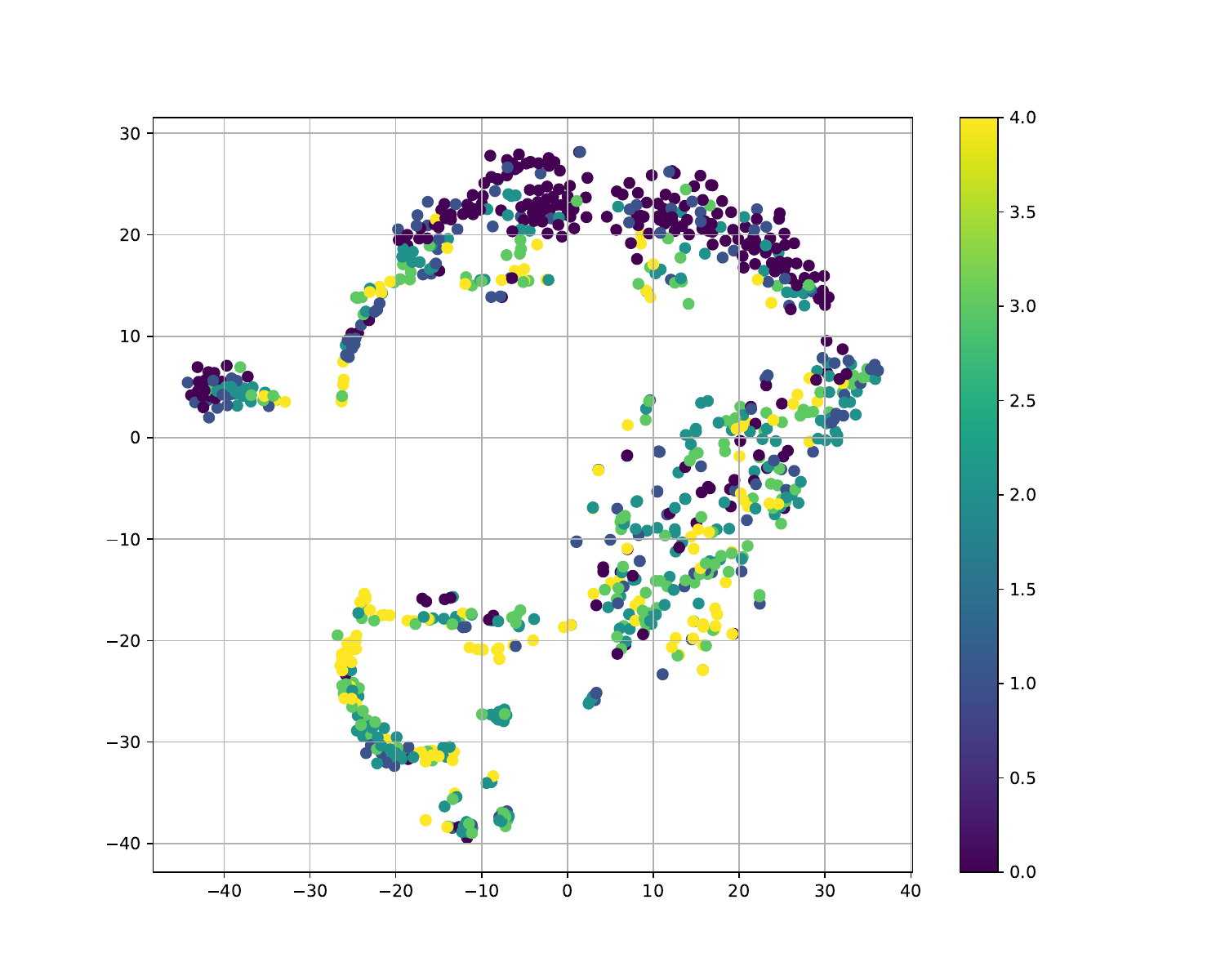}
    \caption{Var-Mono}
    \label{figa:chameleon_tsne_adjconv}
\end{subfigure}
\hfill
\begin{subfigure}[b]{0.19\textwidth}
    \centering
    \includegraphics[width=1.1\textwidth]{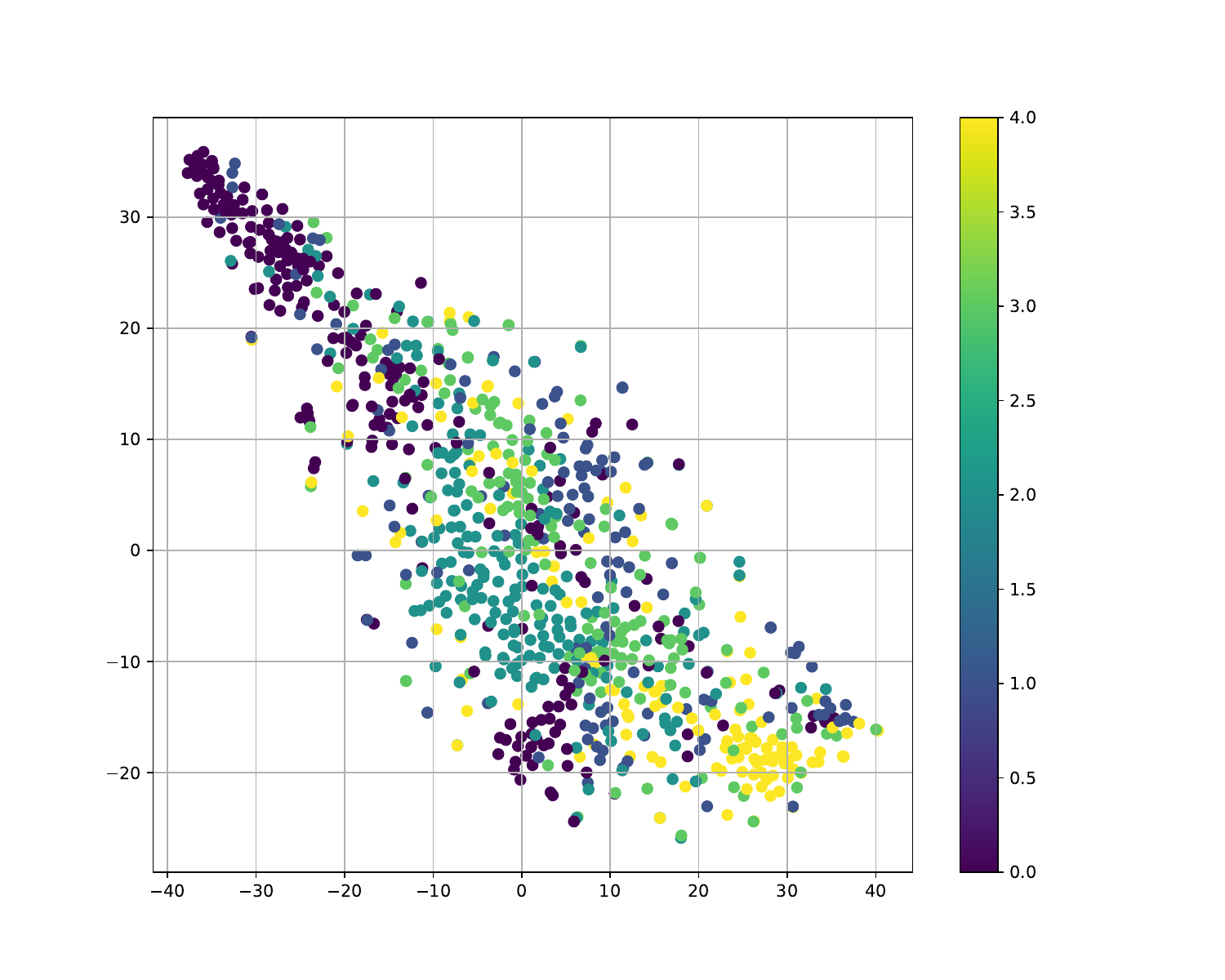}
    \caption{Bern}
    \label{figa:chameleon_tsne_bernconv}
\end{subfigure}
\hfill
\begin{subfigure}[b]{0.19\textwidth}
    \centering
    \includegraphics[width=1.1\textwidth]{figs/tsne/chameleon/DecoupledVar_ChebConv.pdf}
    \caption{Cheb}
    \label{figa:chameleon_tsne_chebconv}
\end{subfigure}
\hfill
\begin{subfigure}[b]{0.19\textwidth}
    \centering
    \includegraphics[width=1.1\textwidth]{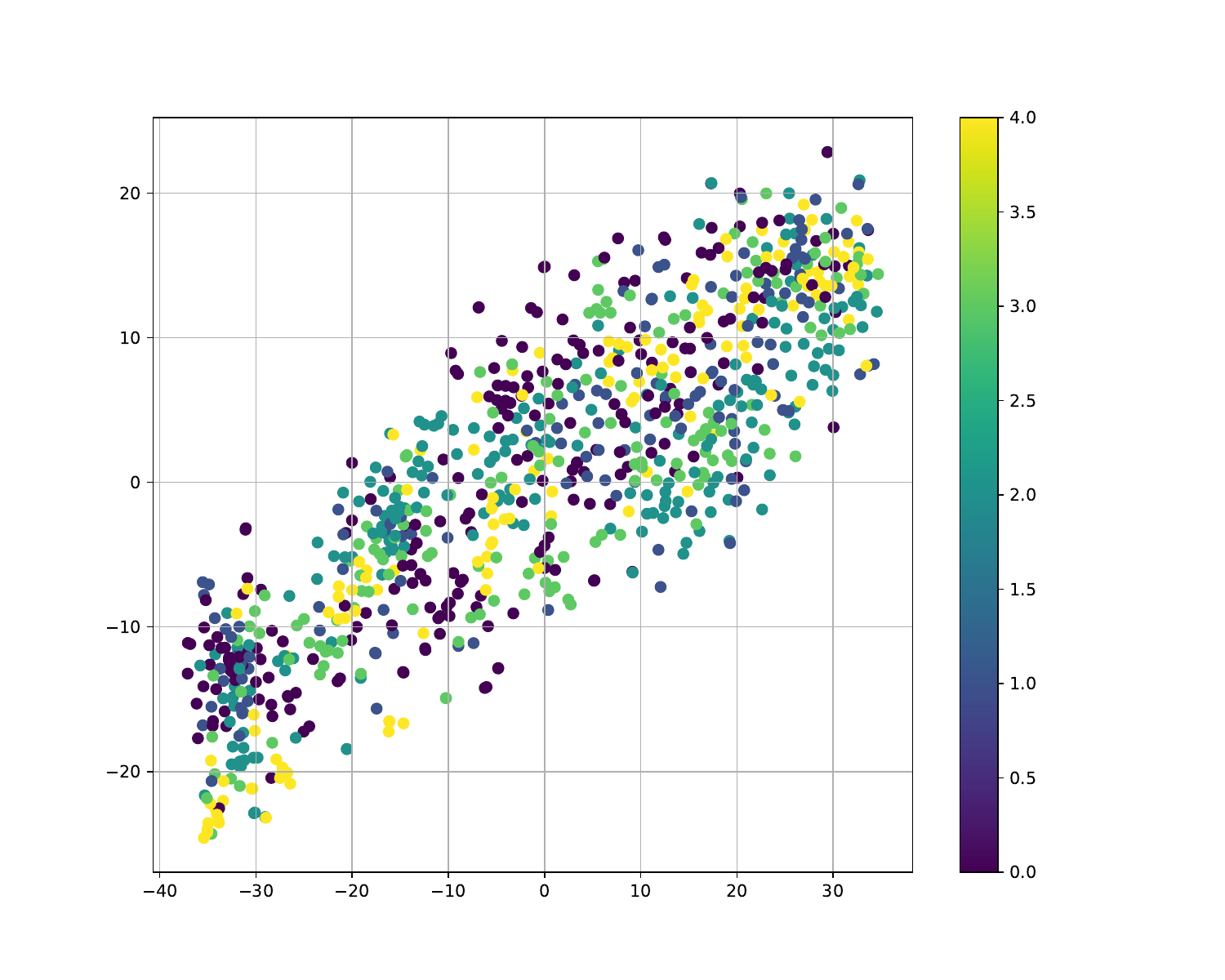}
    \caption{ChebInterp}
    \label{figa:chameleon_tsne_chebiiconv}
\end{subfigure}
\hfill
\begin{subfigure}[b]{0.19\textwidth}
    \centering
    \includegraphics[width=1.1\textwidth]{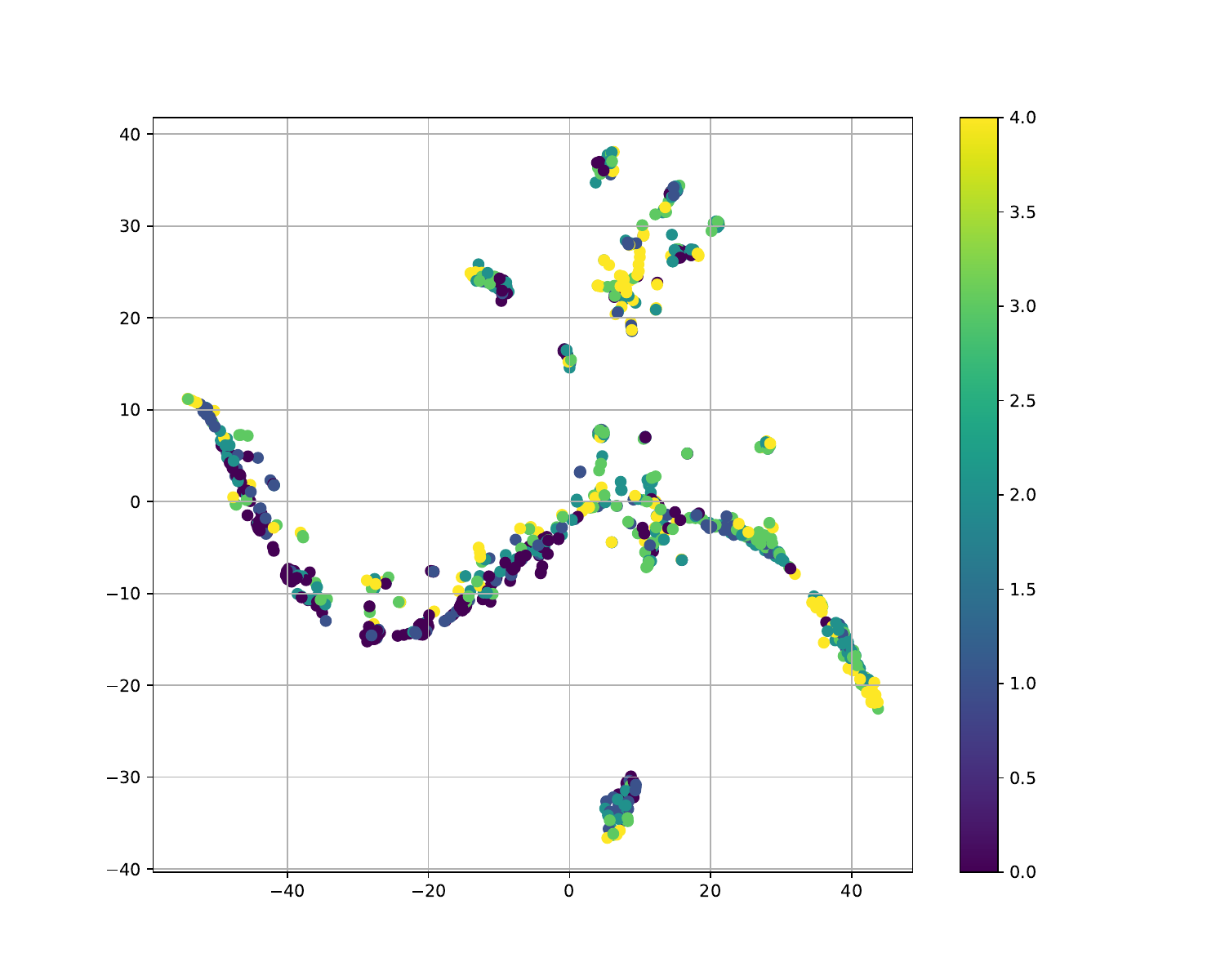}
    \caption{Clenshaw}
    \label{figa:chameleon_tsne_clenshawconv}
\end{subfigure}
\begin{subfigure}[b]{0.19\textwidth}
    \centering
    \includegraphics[width=1.1\textwidth]{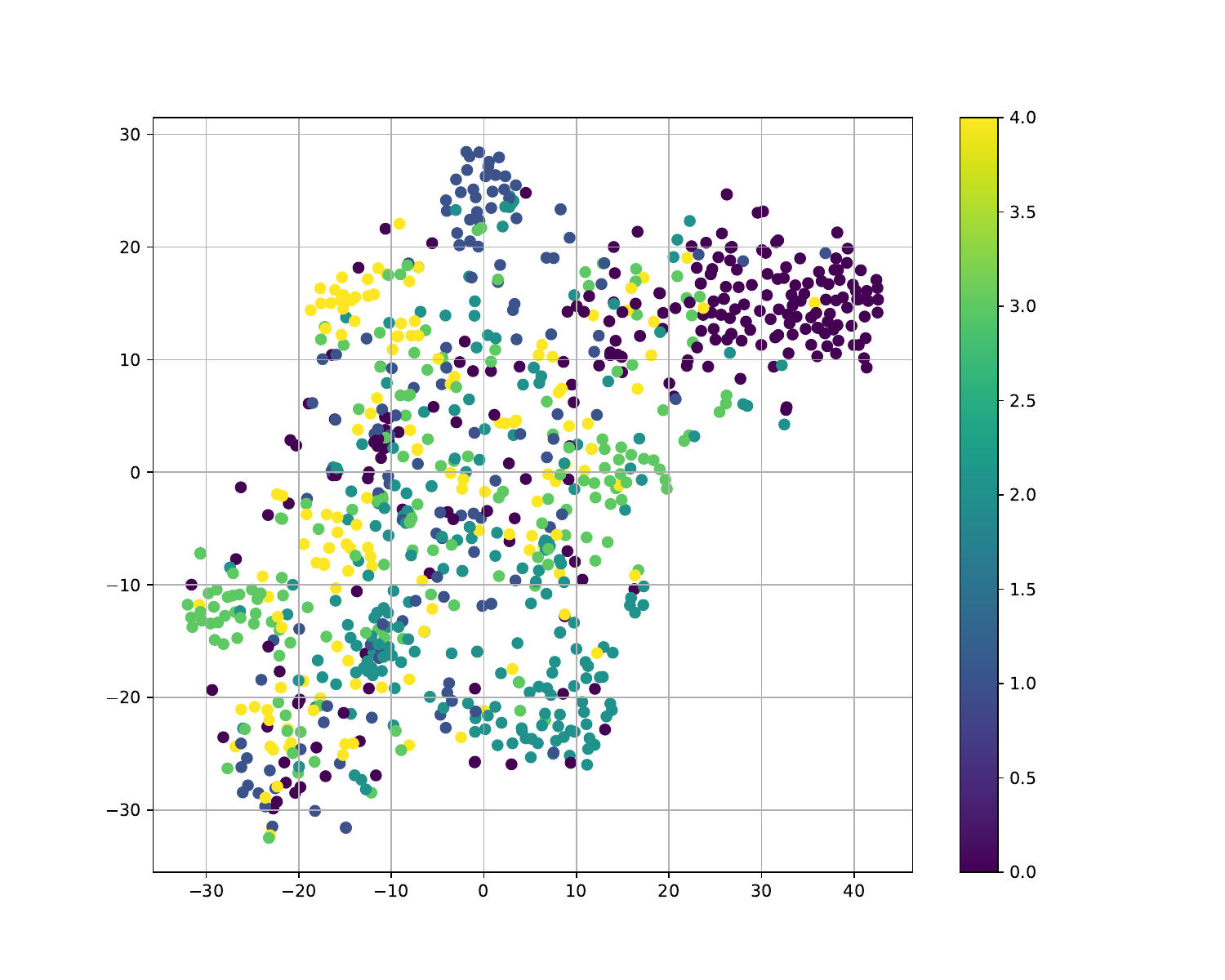}
    \caption{Favard}
    \label{figa:chameleon_tsne_favardconv}
\end{subfigure}
\hfill
\begin{subfigure}[b]{0.19\textwidth}
    \centering
    \includegraphics[width=1.1\textwidth]{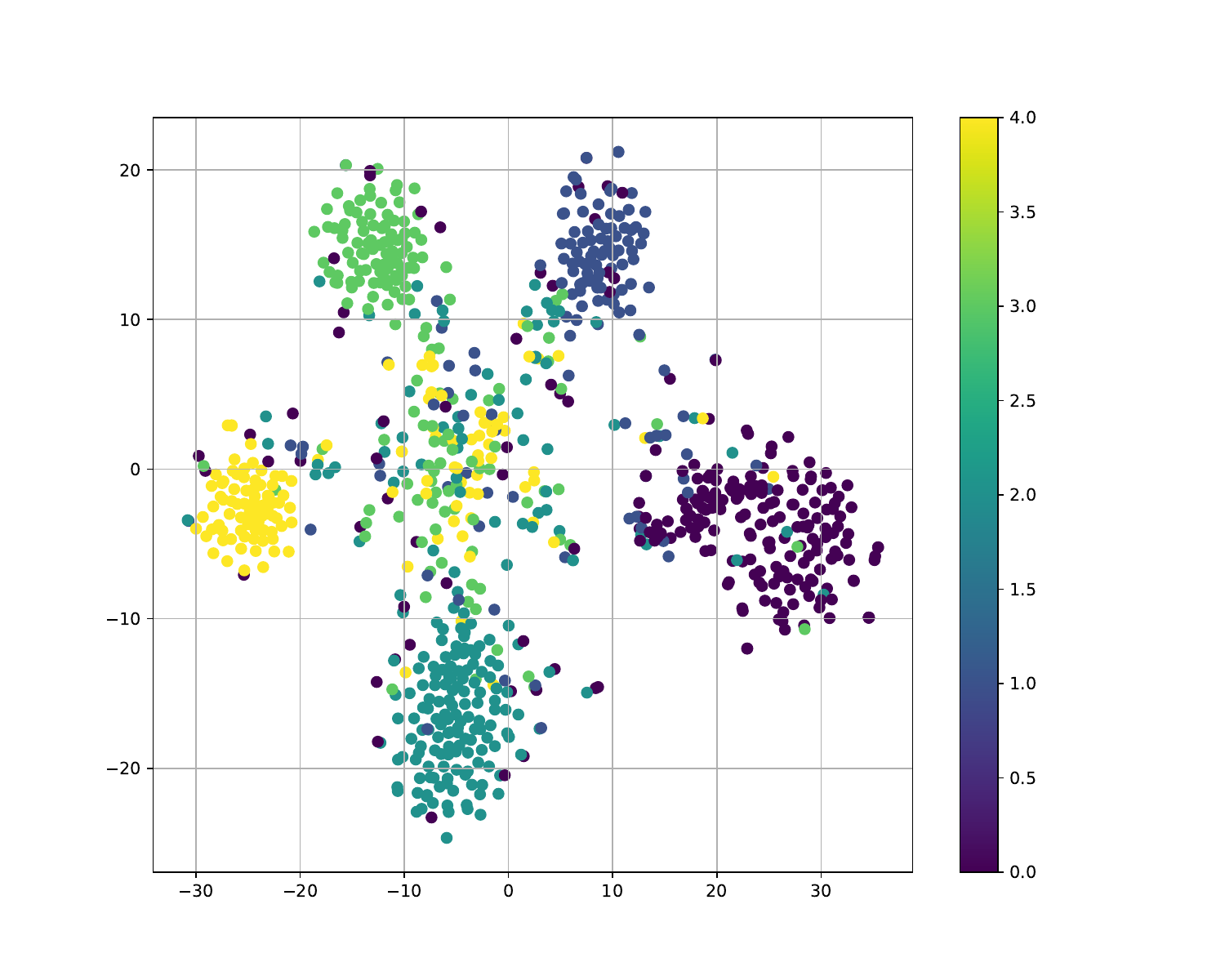}
    \caption{Horner}
    \label{figa:chameleon_tsne_hornerconv}
\end{subfigure}
\hfill
\begin{subfigure}[b]{0.19\textwidth}
    \centering
    \includegraphics[width=1.1\textwidth]{figs/tsne/chameleon/DecoupledVar_JacobiConv.pdf}
    \caption{Jacobi}
    \label{figa:chameleon_tsne_jacobiconv}
\end{subfigure}
\hfill
\begin{subfigure}[b]{0.19\textwidth}
    \centering
    \includegraphics[width=1.1\textwidth]{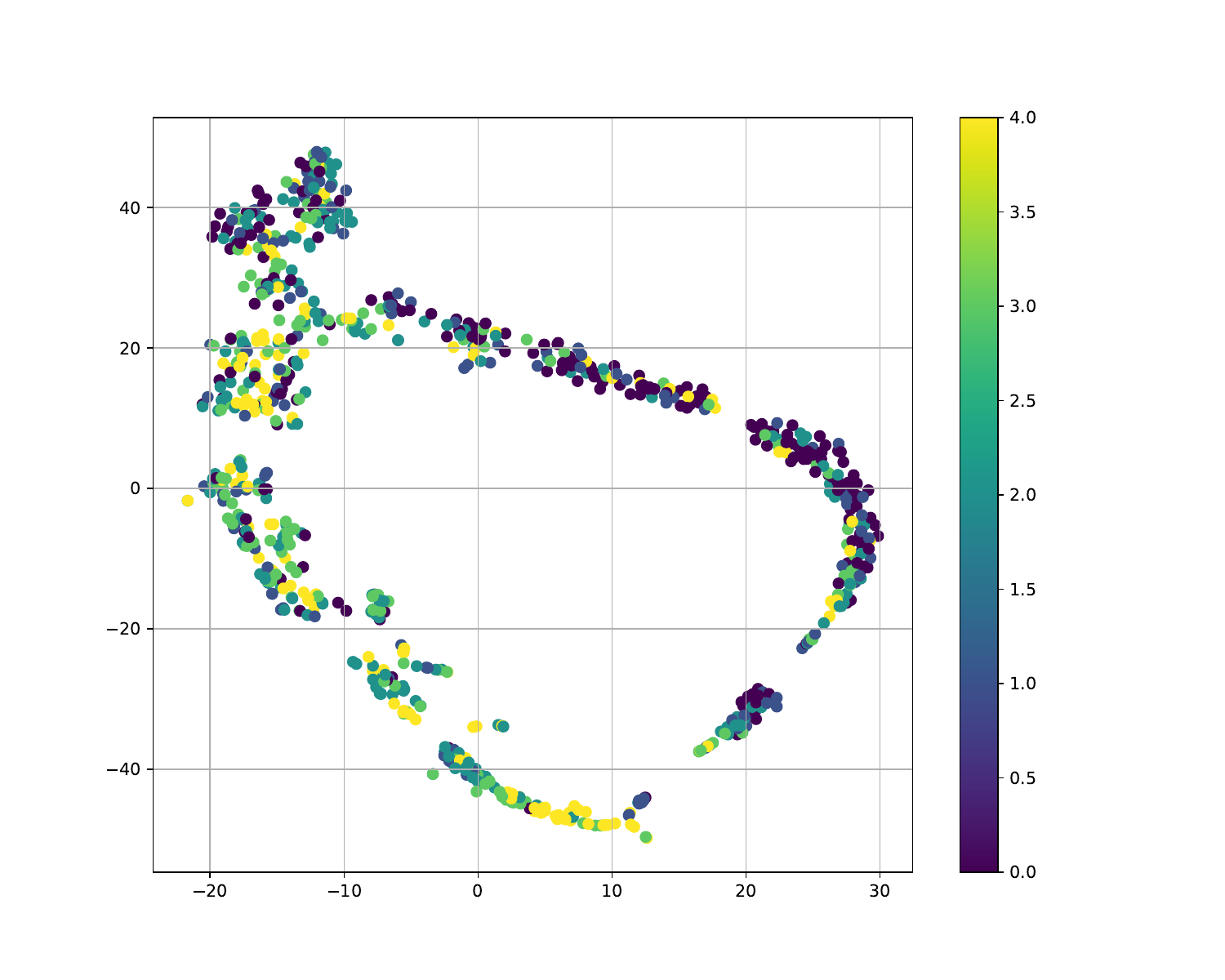}
    \caption{Legendre}
    \label{figa:chameleon_tsne_legendreconv}
\end{subfigure}
\hfill
\begin{subfigure}[b]{0.19\textwidth}
    \centering
    \includegraphics[width=1.1\textwidth]{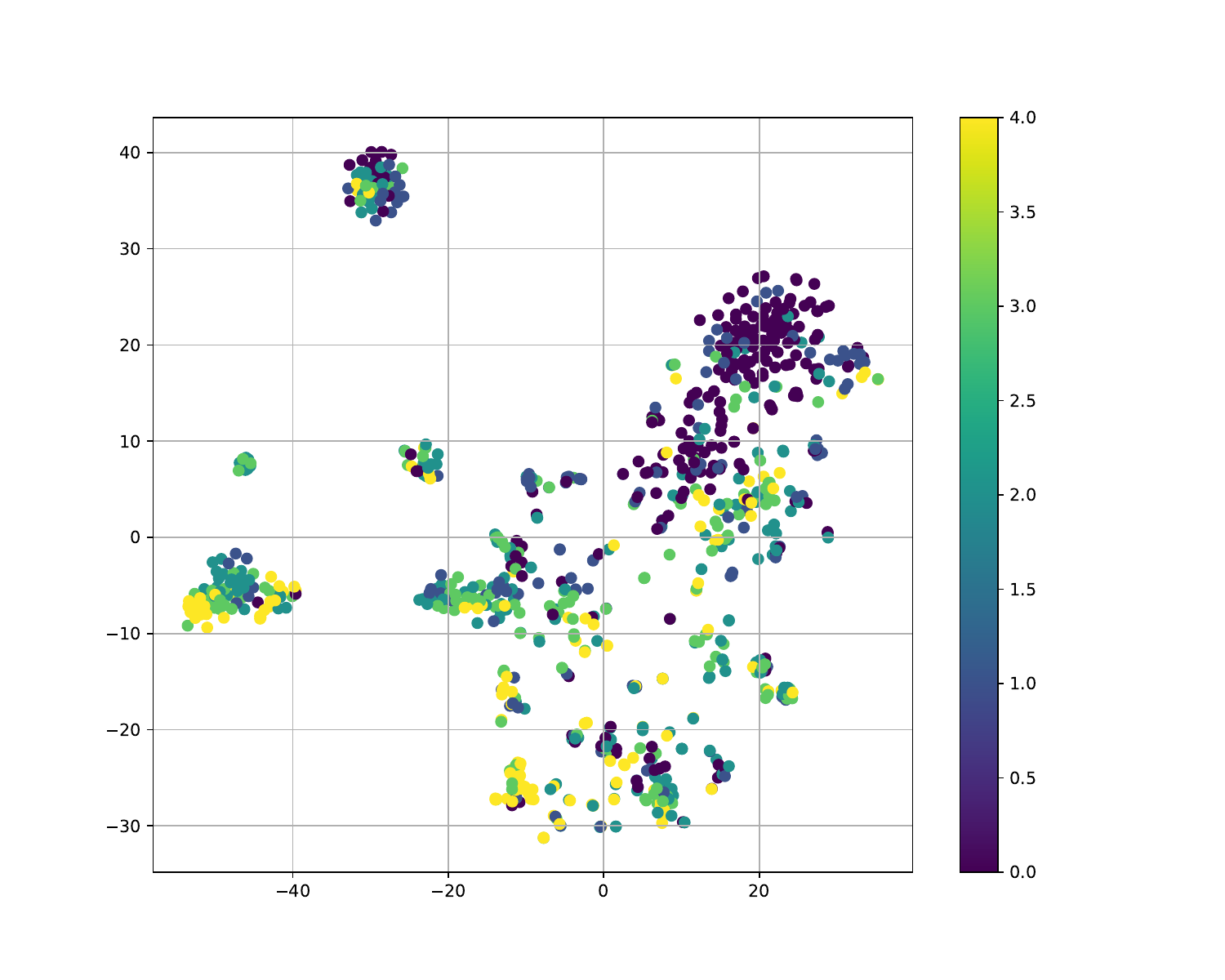}
    \caption{OptBasis}
    \label{figa:chameleon_tsne_optbasisconv}
\end{subfigure}

\caption{Clusters of different filters on \ds{CHAMELEON}.}
\label{figa:tsne_chameleon}
\end{figure*}

\clearpage

\begin{figure*}[!t]
\subsection{Degree-wise Accuracy Gap}
\label{seca:extra_deg_best}
    \includegraphics[height=1.8in]{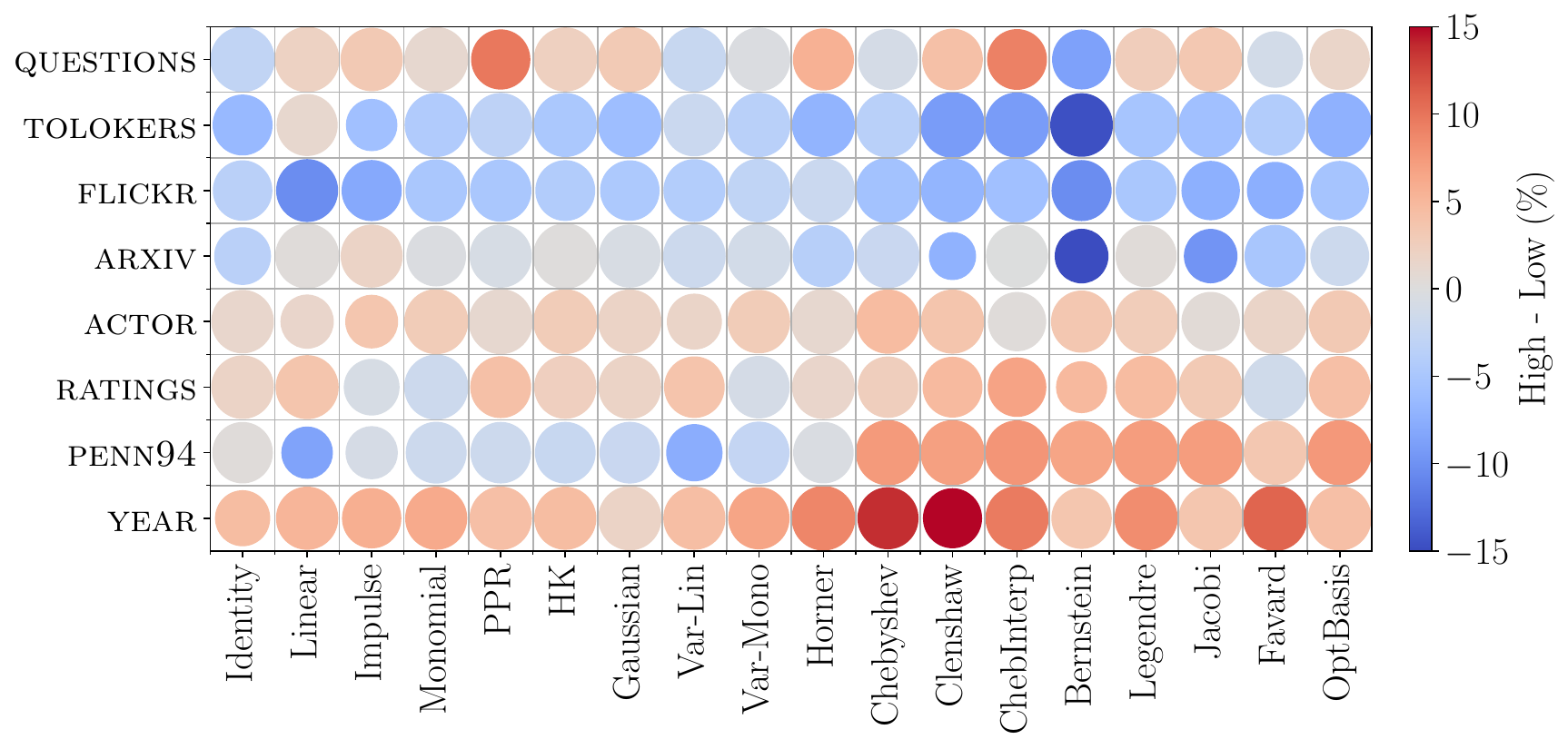}
\caption{Accuracy gap between high- and low-degree nodes on remaining datasets. } 
\label{fig:deg_best2}
\end{figure*}

\begin{figure*}[!t]
\subsection{Effect of Graph Normalization on Degree-wise Performance}
\label{seca:extra_deg}
    \centering
    \vspace{14pt}
    \subcaptionbox{\ds{cora}\label{ffiga:degng_cora}}%
    [0.24\linewidth]{\includegraphics[height=1.05in]{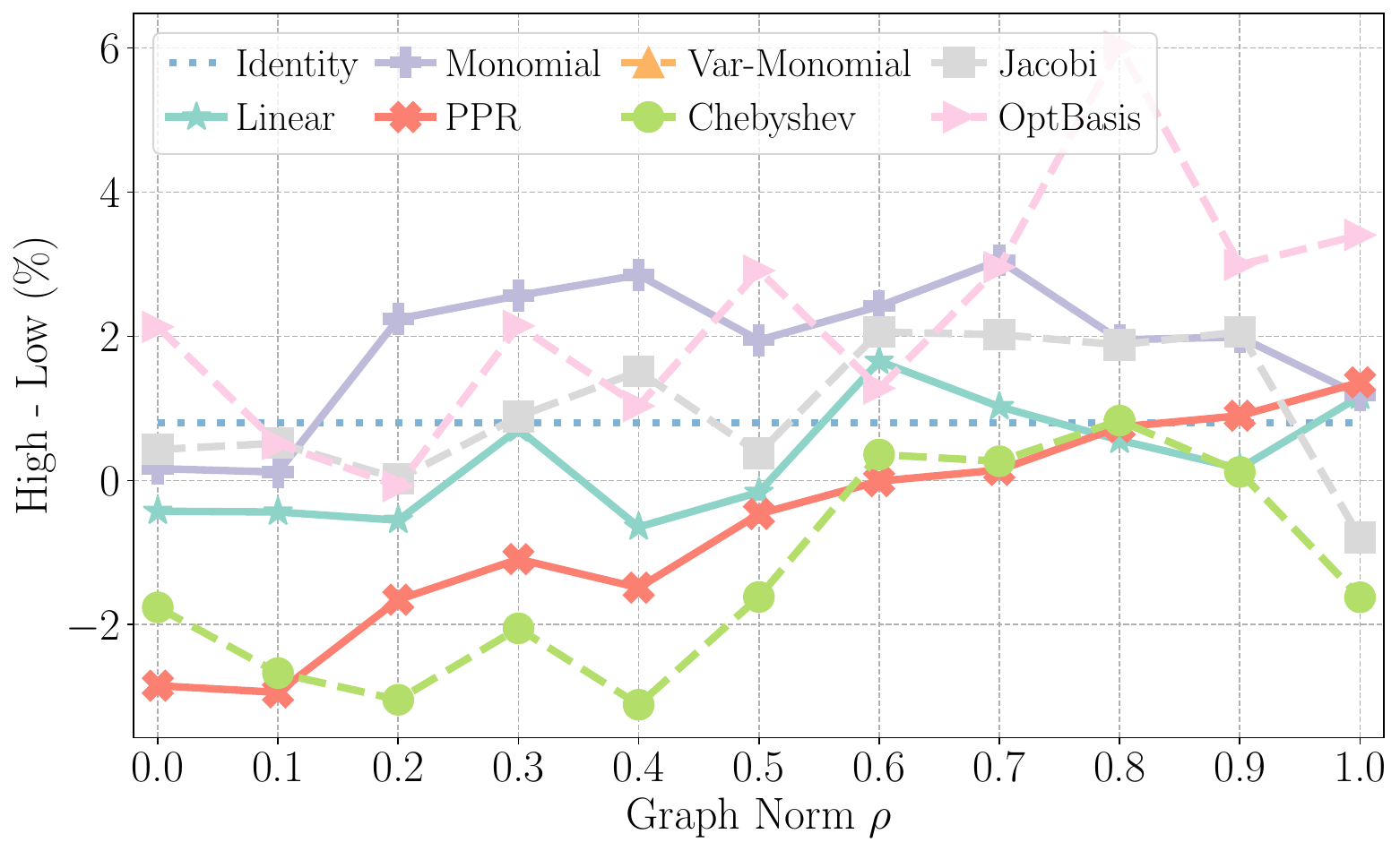}}
    \hfil
    \subcaptionbox{\ds{citeseer}\label{ffiga:degng_citeseer}}%
    [0.24\linewidth]{\includegraphics[height=1.05in]{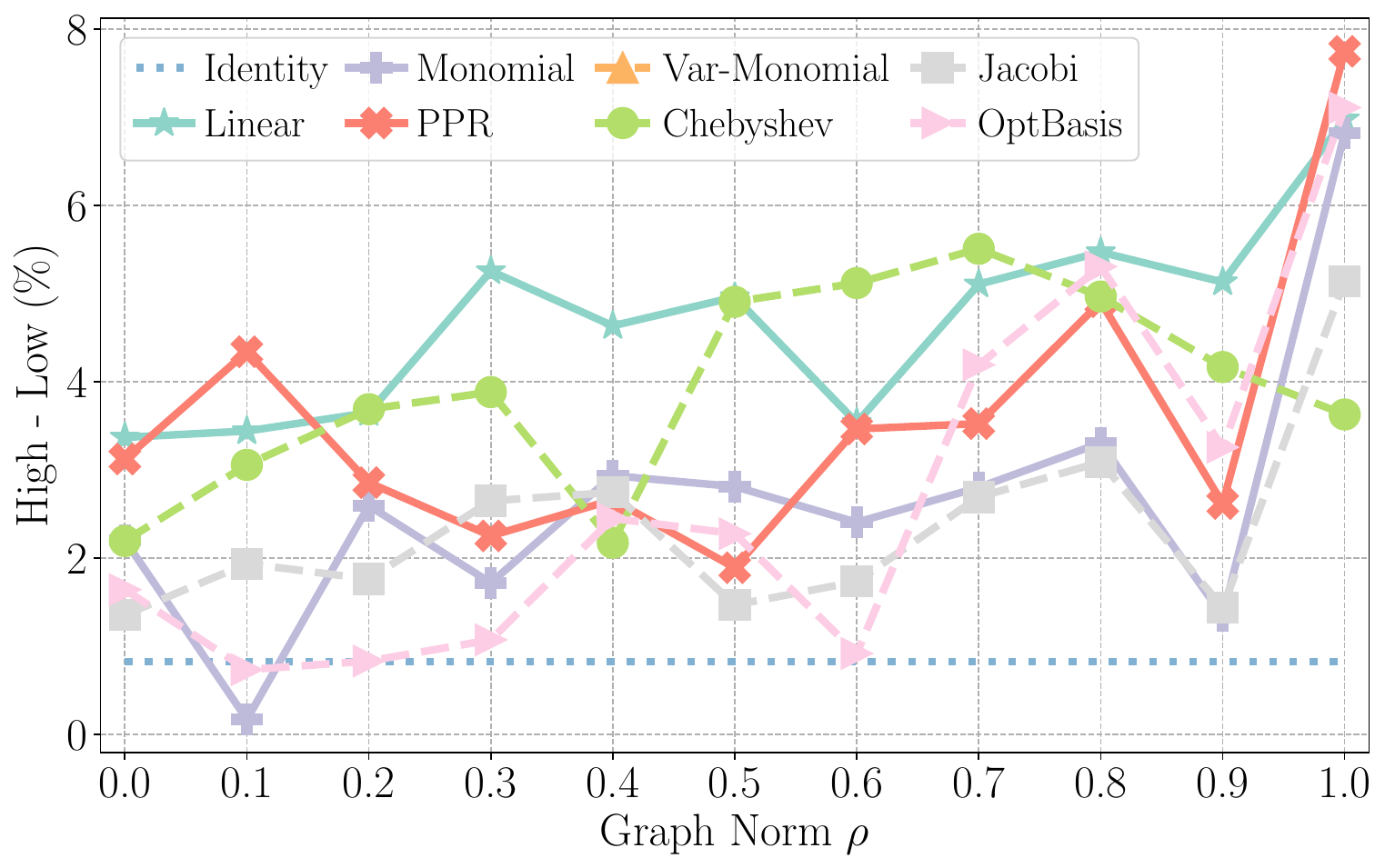}}
    \hfil
    \subcaptionbox{\ds{pubmed}\label{ffiga:degng_pubmed}}%
    [0.24\linewidth]{\includegraphics[height=1.05in]{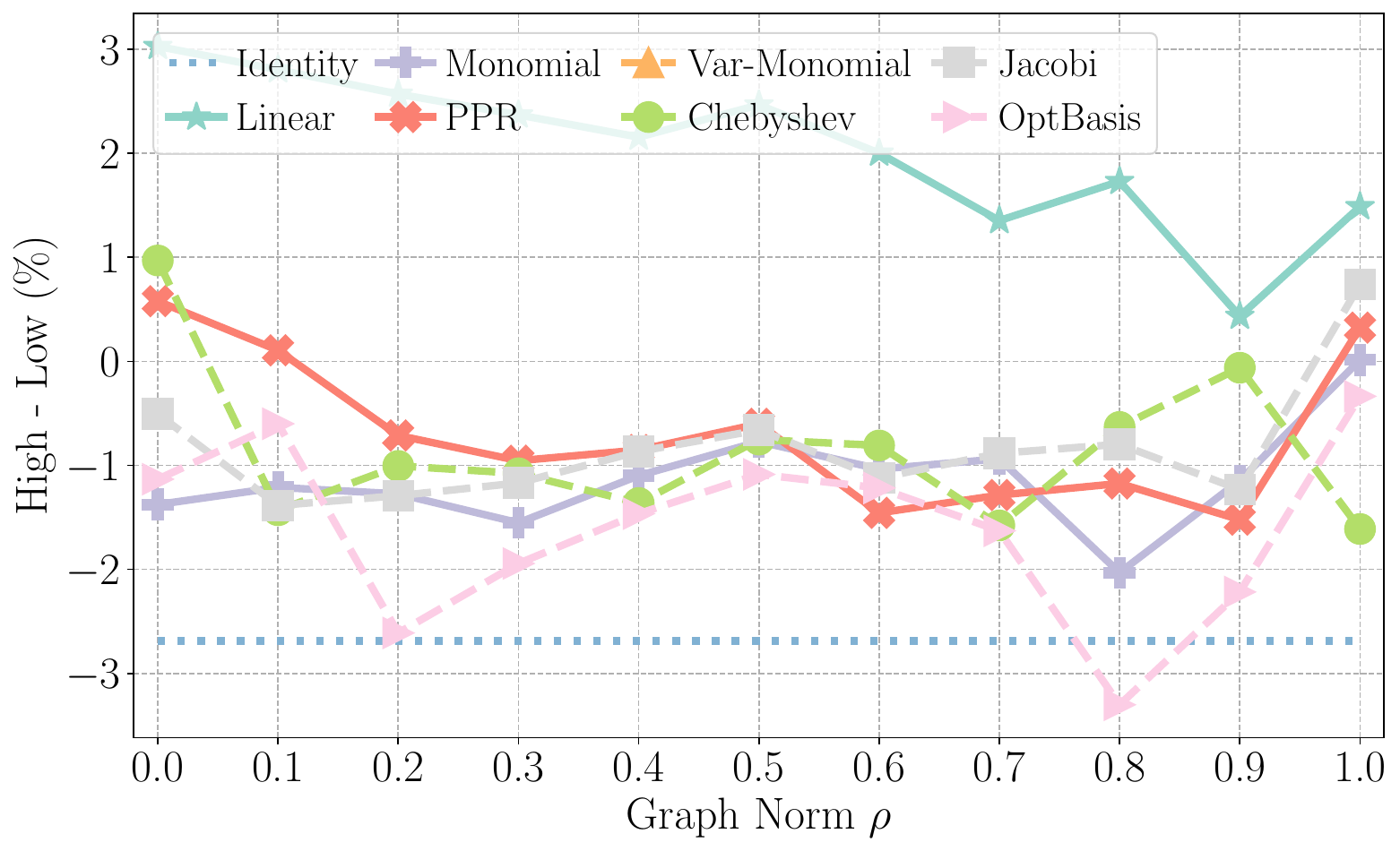}}
    \hfil
    \subcaptionbox{\ds{flickr}\label{ffiga:degng_flickr}}%
    [0.24\linewidth]{\includegraphics[height=1.05in]{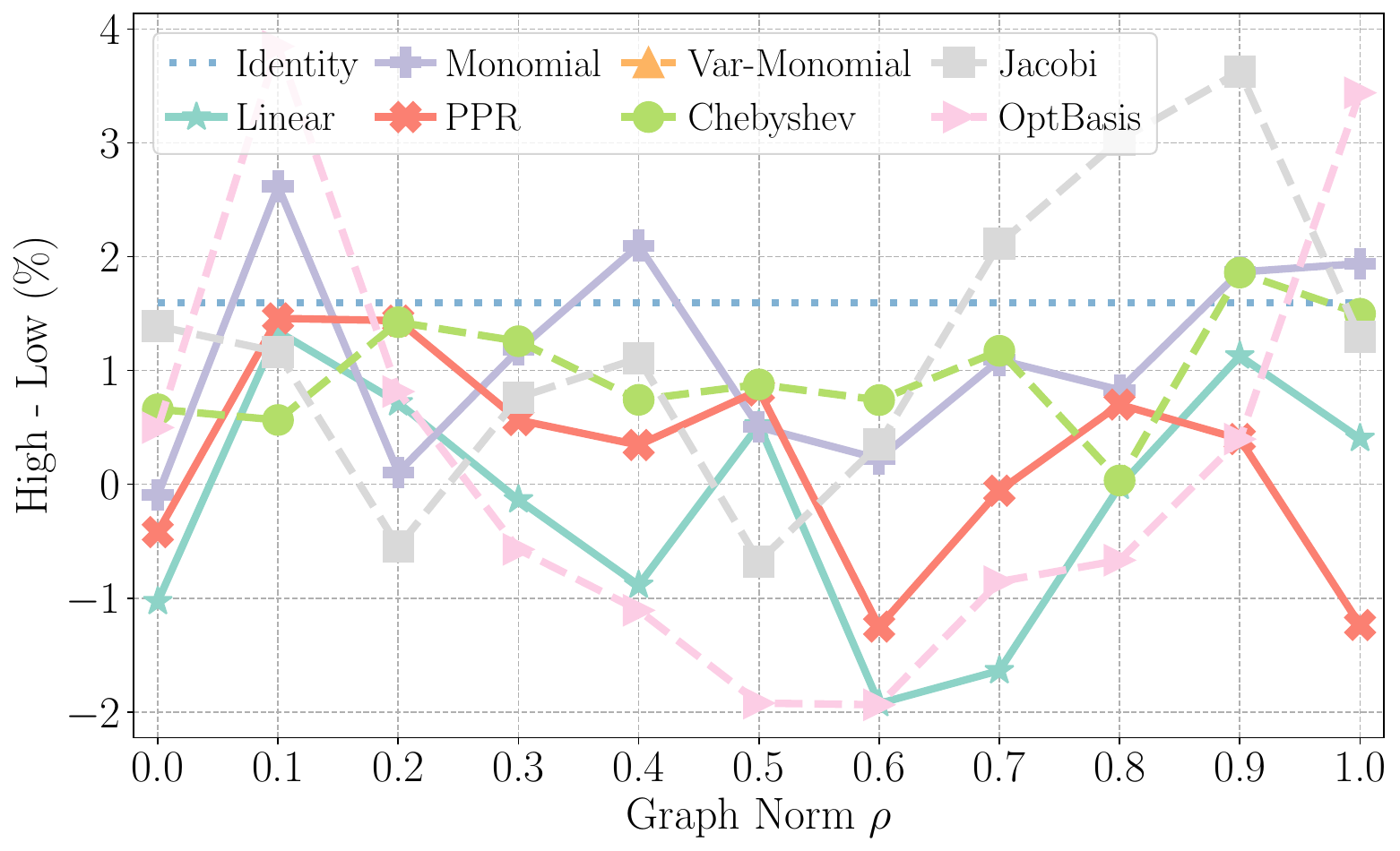}}
    \\ \vspace{8pt}
    \subcaptionbox{\ds{squirrel}\label{ffiga:degng_squirrel_filtered}}%
    [0.24\linewidth]{\includegraphics[height=1.05in]{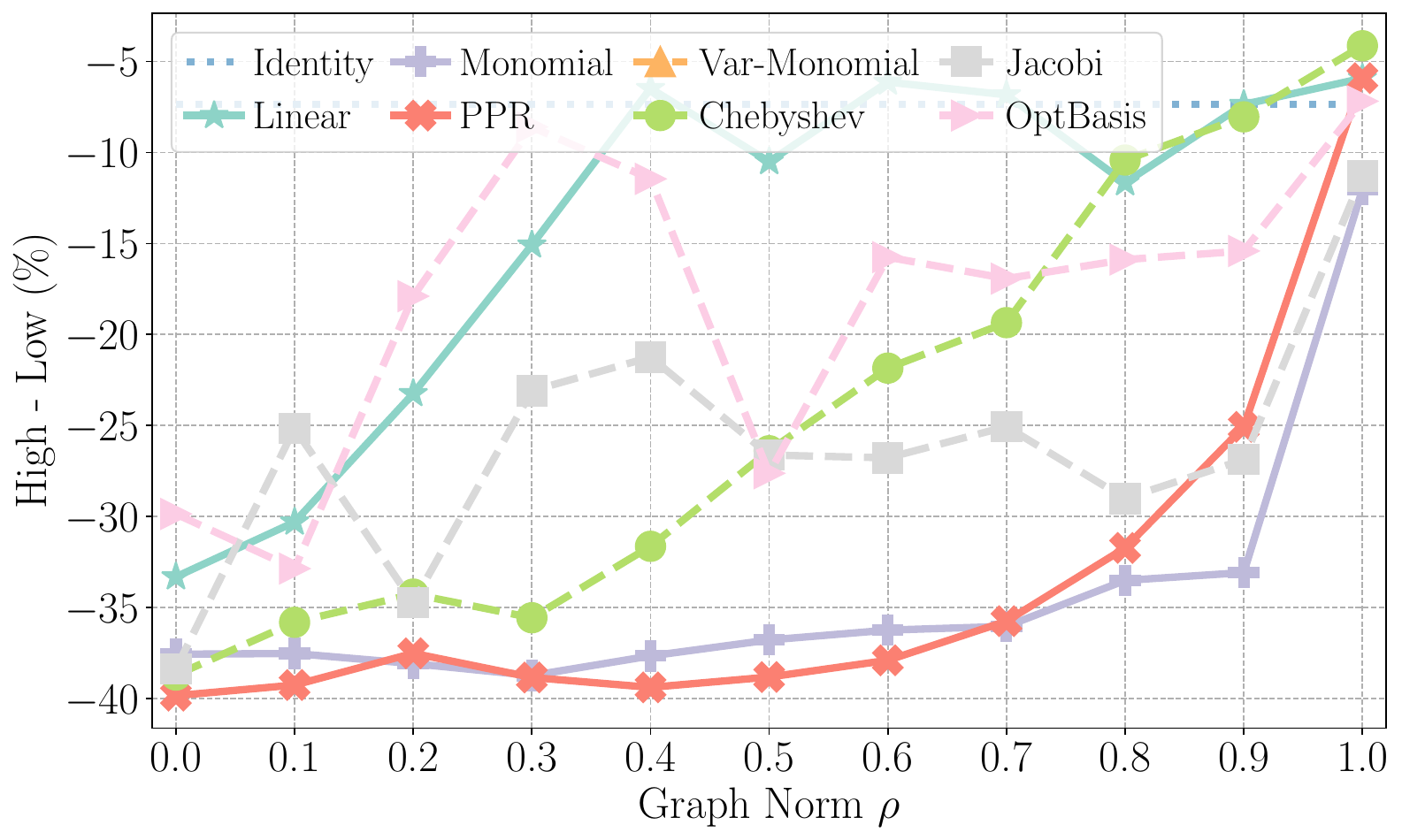}}
    \hfil
    \subcaptionbox{\ds{chameleon}\label{ffiga:degng_chameleon_filtered}}%
    [0.24\linewidth]{\includegraphics[height=1.05in]{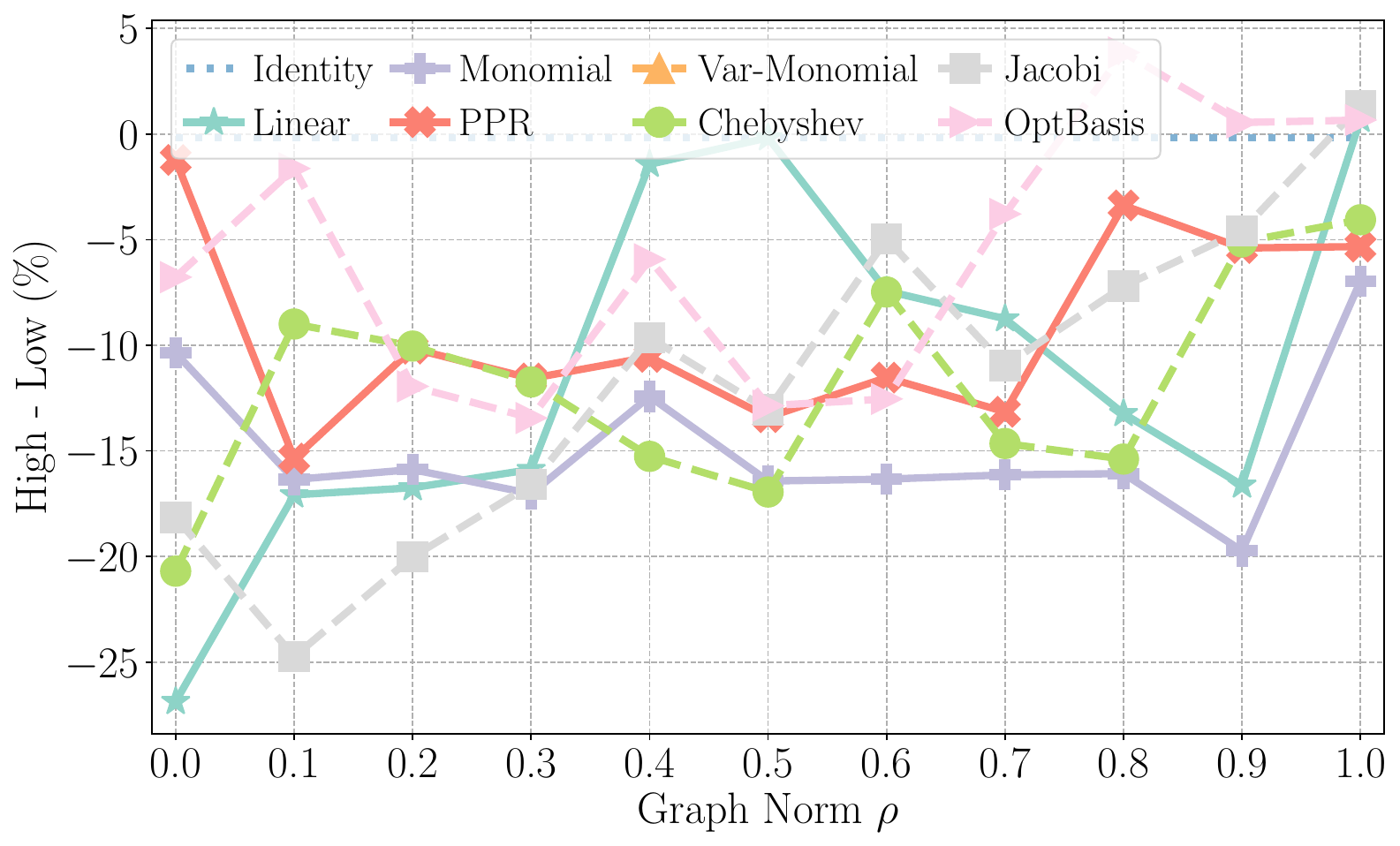}}
    \hfil
    \subcaptionbox{\ds{actor}\label{ffiga:degng_actor}}%
    [0.24\linewidth]{\includegraphics[height=1.05in]{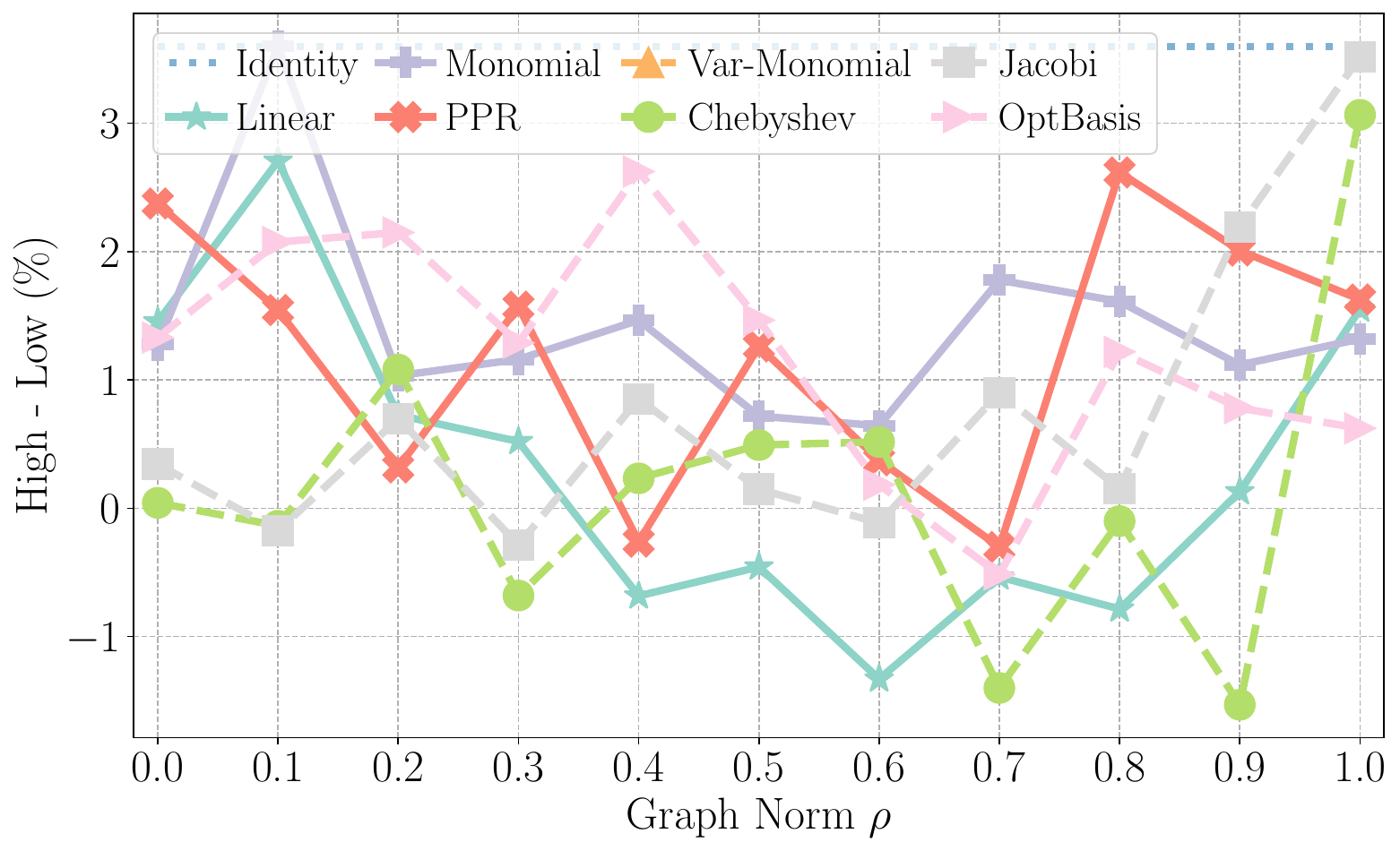}}
    \hfil
    \subcaptionbox{\ds{roman}\label{ffiga:degng_roman_empire}}%
    [0.24\linewidth]{\includegraphics[height=1.05in]{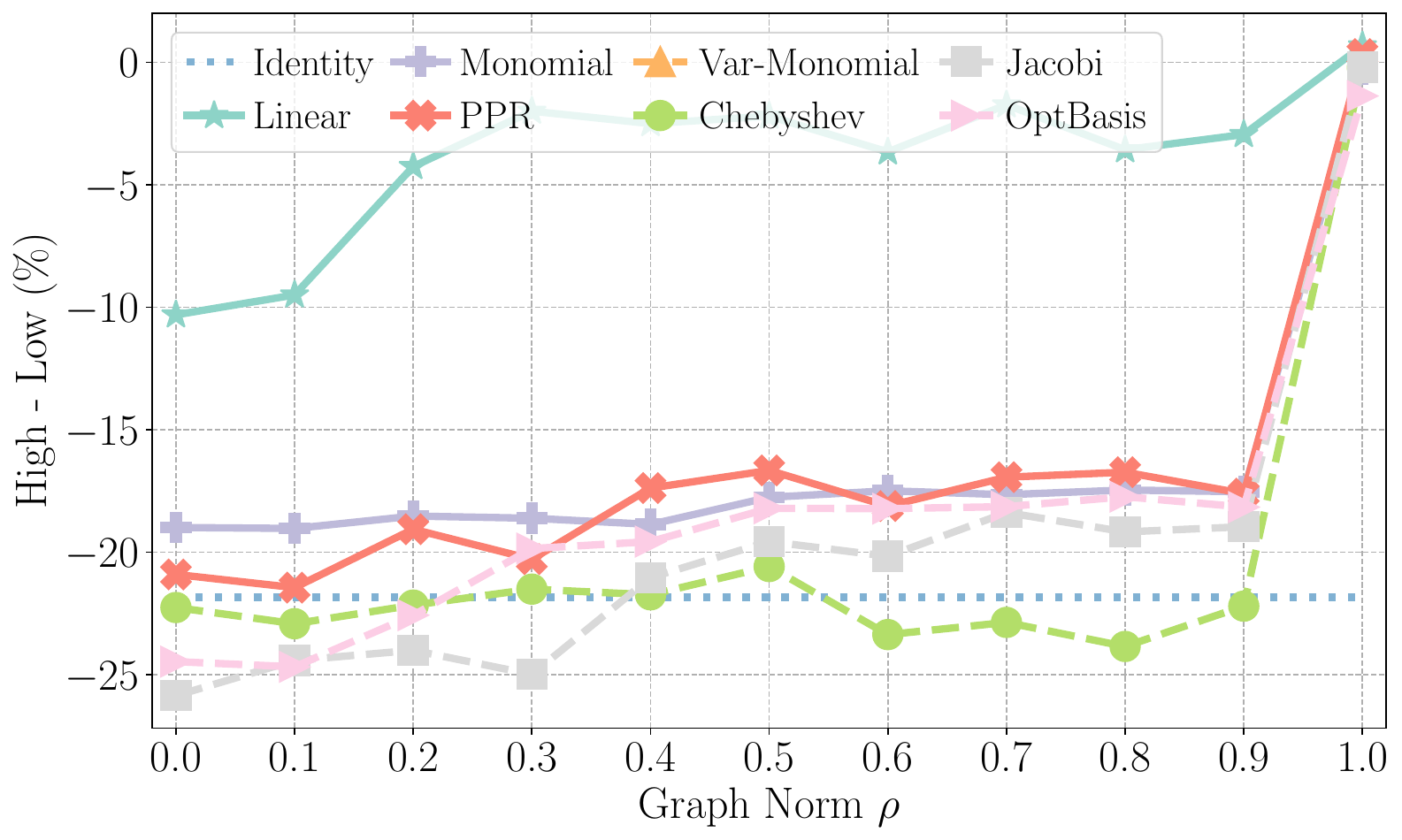}}
    \caption{Effect of \textit{graph normalization} $\rho$ on the accuracy difference between high- and low-degree nodes of selected full-batch fixed and variable filters on 4 homophilous and 4 heterophilous datasets. }
  \label{figa:degng}
\end{figure*}

\begin{figure*}[!t]
\subsection{Efficiency on Different Hardware}
\label{seca:plat_full}
\vspace{14pt}
    \centering
    \subcaptionbox{\ds{arxiv}\label{ffiga:plat_ogbn-arxiv_full}}%
    [0.31\linewidth]{\includegraphics[height=1.45in]{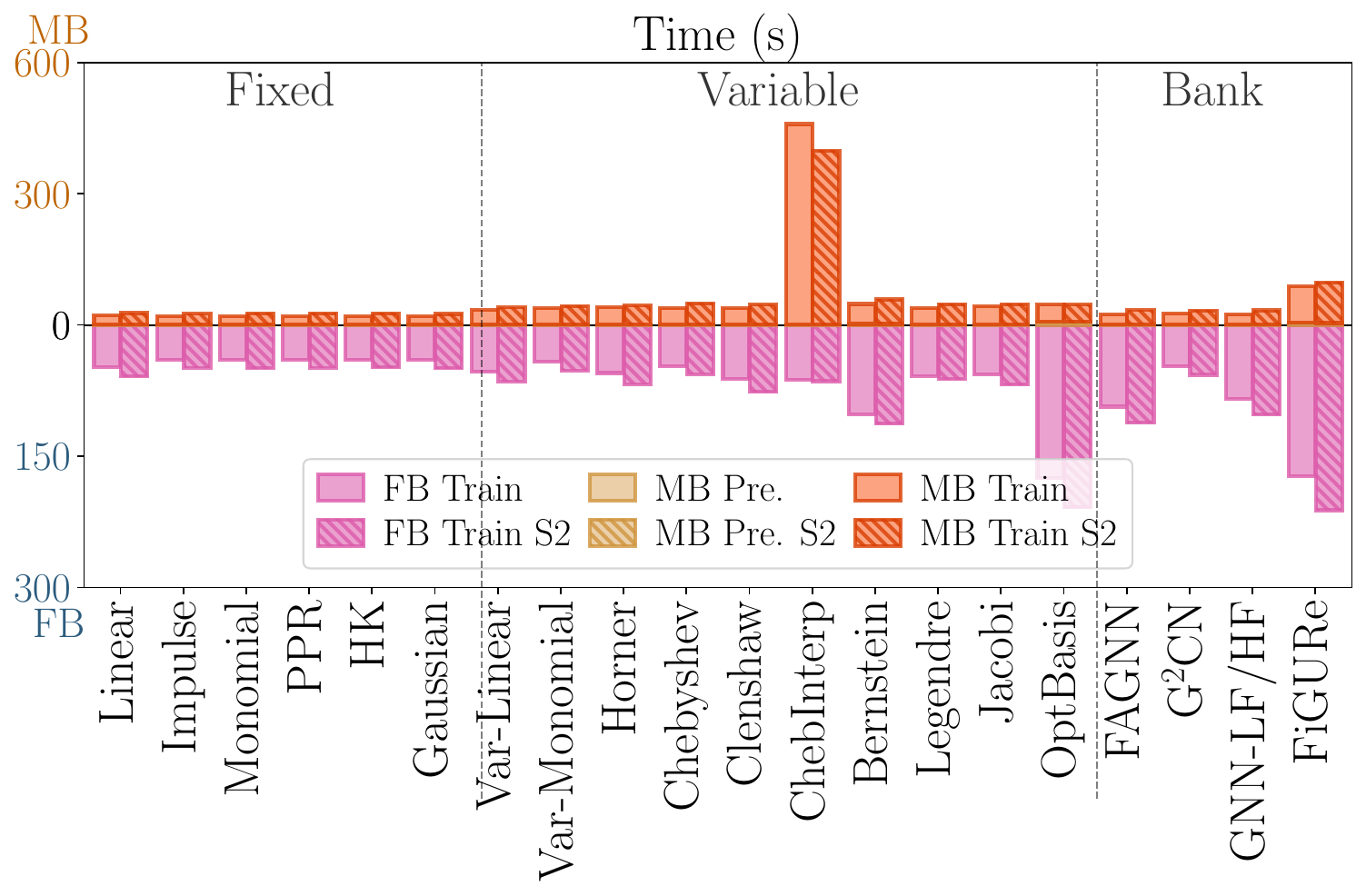}}
    \hfil
    \subcaptionbox{\ds{genius}\label{ffiga:plat_genius_full}}%
    [0.31\linewidth]{\includegraphics[height=1.45in]{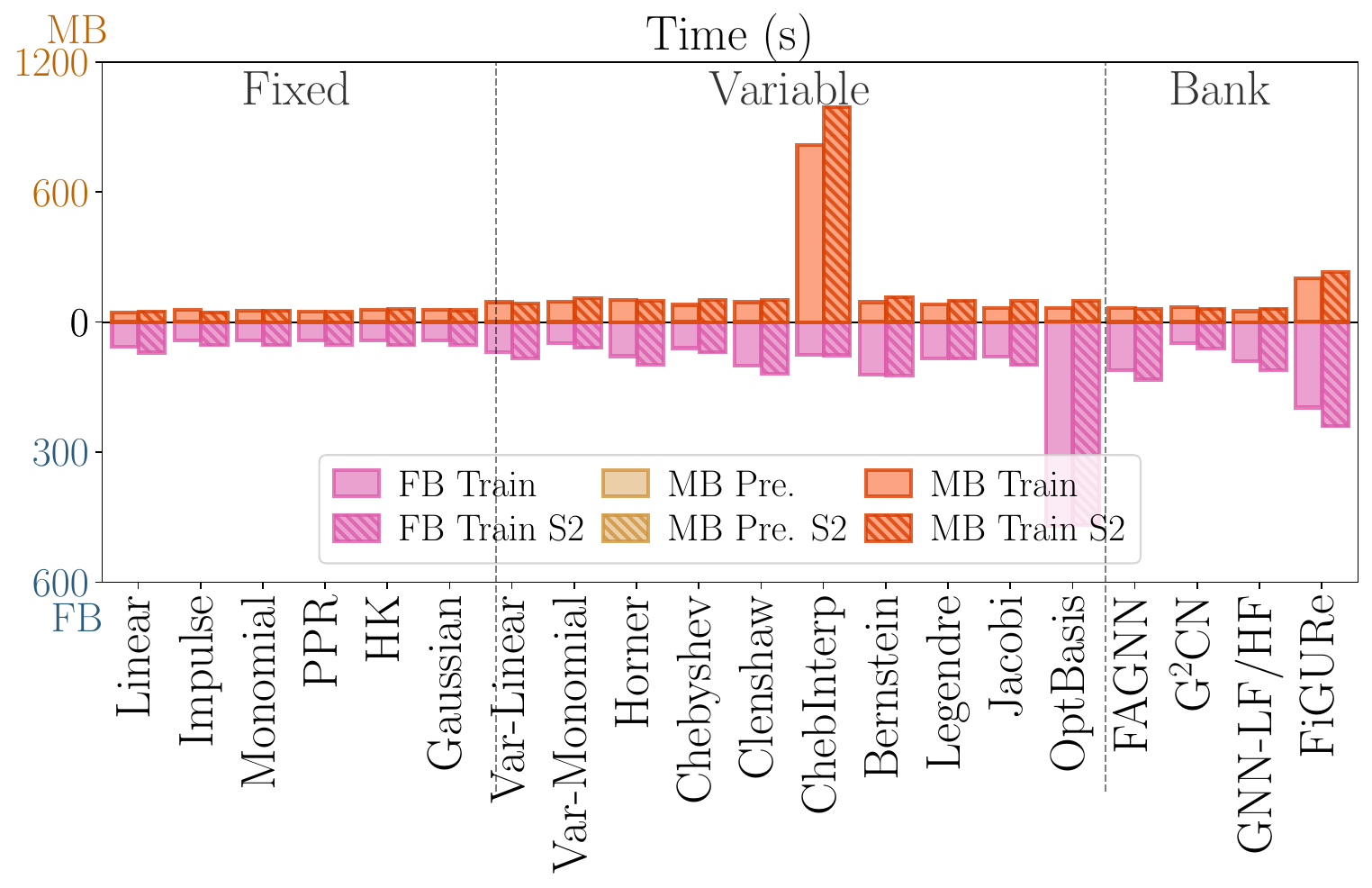}}
    \hfil
    \subcaptionbox{\ds{pokec}\label{ffiga:plat_pokec_full}}%
    [0.31\linewidth]{\includegraphics[height=1.45in]{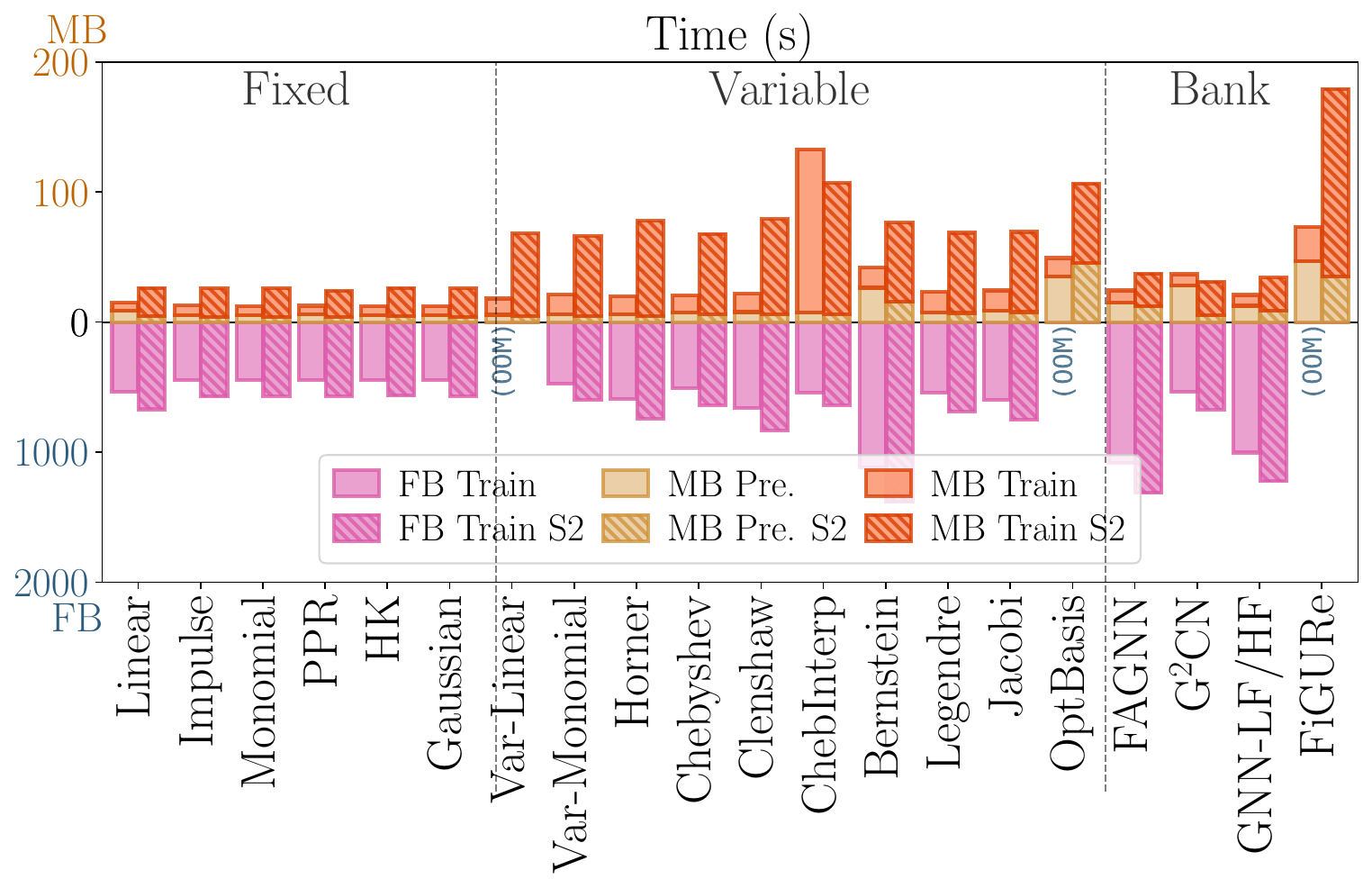}}
    \caption{\rra{Time efficiency comparison of FB (lower axis) and MB (upper axis) training with different hardware.}}
  \label{figa:plat_full}
\end{figure*}

\end{document}